\documentclass[11pt]{article}
\ifdefined\pdfsuppressptexinfo\pdfsuppressptexinfo=-1\fi
\newif\ifbridgeReview
\bridgeReviewfalse

\usepackage[T1]{fontenc}
\usepackage[utf8]{inputenc}
\usepackage{lmodern}
\usepackage{iftex}
\ifPDFTeX
  \usepackage[activate={true,nocompatibility},final]{microtype}
\else
  \usepackage[protrusion=true,final]{microtype}
\fi
\usepackage[margin=1in]{geometry}
\usepackage{amsmath}
\usepackage{amssymb,amsfonts}
\usepackage{mathtools}      
\usepackage{bm}             
\usepackage{nicefrac}
\allowdisplaybreaks         

\usepackage{amsthm}

\usepackage{graphicx}
\usepackage{placeins}
\usepackage{float} 
\graphicspath{{figures/}{./}}                       
\usepackage{booktabs}                               
\usepackage{multirow,makecell,array,longtable}
\usepackage[font=small,labelfont=bf]{caption}       
\usepackage{subcaption}                             
\usepackage{algorithm}
\usepackage{algpseudocode}                          

\usepackage{enumitem}
\setlist[itemize]{leftmargin=2.2em,itemsep=2pt,topsep=2pt}
\setlist[enumerate]{leftmargin=2.2em,itemsep=2pt,topsep=2pt}

\usepackage{xcolor}
\definecolor{LinkColor}{rgb}{0.10,0.40,0.75}        
\definecolor{CiteColor}{rgb}{0.70,0.25,0.20}        
\definecolor{UrlColor} {rgb}{0.20,0.50,0.50}        

\usepackage{tikz}
\usetikzlibrary{positioning,calc,arrows.meta}

\usepackage[round,sort&compress]{natbib}

\usepackage{xurl}
\usepackage{hyperref}
\hypersetup{
  colorlinks=true,
  linkcolor=LinkColor,
  citecolor=CiteColor,
  urlcolor=UrlColor,
  breaklinks=true,
  bookmarksnumbered=true,
}
\usepackage{bookmark}                               
\usepackage[capitalise,noabbrev,sort&compress]{cleveref}  

\numberwithin{equation}{section}

\newcommand{\nctx}{n_{\rm ctx}}
\newcommand{\rctx}{d_{\rm width}}
\newcommand{\Nobs}{N_{\rm obs}}
\newcommand{\iclrVspace}[1]{%
  \ifbridgeReview
    \vspace{#1}%
  \fi
}

\newtheorem{theorem}{Theorem}[section]
\newtheorem{proposition}[theorem]{Proposition}
\newtheorem{corollary}[theorem]{Corollary}
\newtheorem{lemma}[theorem]{Lemma}
\theoremstyle{definition}
\newtheorem{definition}[theorem]{Definition}
\theoremstyle{remark}

\newcommand{\R}{\mathbb R}
\newcommand{\E}{\mathbb E}

\newcommand{\norm}[1]{\lVert#1\rVert}
\newcommand{\ip}[2]{\langle#1,#2\rangle}

\newcommand{\op}{\mathrm{op}}
\newcommand{\softmax}{\operatorname{softmax}}

\newcommand{\diag}{\operatorname{diag}}

\title{What Pretraining and Midtraining Make Learnable from Rewards?}
\author{Chiwun Yang\thanks{City University of Hong Kong. \href{mailto:christiannyang37@gmail.com}{\texttt{christiannyang37@gmail.com}}.}
\and Xiaoyu Li\thanks{University of New South Wales. \href{mailto:7.xiaoyu.li@gmail.com}{\texttt{7.xiaoyu.li@gmail.com}}.}}
\date{}
\newif\ifbridgeMeasured
\IfFileExists{experiments/generated/generated_experiments.tex}{\bridgeMeasuredtrue}{\bridgeMeasuredfalse}
\newcommand{\sgn}{\operatorname{sign}}

\newcommand{\cmin}{c_{\min}}
\newcommand{\Btar}{B_{\star}}

\hypersetup{pdftitle={What Pretraining and Midtraining Make Learnable from Rewards?}}
\ifbridgeReview\hypersetup{pdfauthor={}}\else\hypersetup{pdfauthor={Chiwun Yang and Xiaoyu Li}}\fi
\begin{document}
\maketitle
\begin{abstract}
\ifbridgeReview\input{arxiv/sections/iclr9_abstract}\elseA reward can identify a correct answer while leaving the computation needed for new inputs undetermined. We study how task-independent prediction supplies the missing information and turns it into computation that rewards can use. In sequential state computation and contextual memory, we characterize exactly which mechanism--rule worlds agree on every training reward yet demand opposite held-out answers. Source observations expose the mechanism before its effects cancel in training rewards. Finite sampled Adam realizes this division of labor from random initialization in the same parameters: the sequential learner acquires execution and selects the task binding in two reward updates, reaching held-out success at least $7/8$; the contextual learner acquires retrieval and an input-dependent task rule within explicit finite budgets, reaching $19/20$, while randomly initialized frozen attention stays near chance. Across eight worlds, source midtraining from pretrained Qwen2.5 checkpoints supplies executable state predictions and Memory responses sensitive to stored values. With task-independent first-operation supervision, Sequential models reach 82.61\% success against 44.15\% for a private-random source control, with a 67.57-point task-selectivity advantage. Memory replay retains 92.21\% original/value-flip pair accuracy during reward adaptation, compared with 0\% under pure reward training. An independent eight-world Memory confirmation reaches 75.32\% task success versus 49.86\% after matched alternative-retrieval training. GSM8K and HotpotQA separate accuracy at reward entry, subsequent gain and final performance. The results connect three steps in learning from rewards: identifying a mechanism, acquiring its computation and learning its task-specific use.
\fi
\end{abstract}
\ifbridgeReview
  \input{arxiv/sections/iclr9_intro}
  \input{arxiv/sections/iclr9_related}
  \input{arxiv/sections/iclr9_setup}
  \input{arxiv/sections/iclr9_sequential}
  \input{arxiv/sections/iclr9_contextual}
  \input{arxiv/sections/iclr9_experiments}
\else
  \section{Introduction}
Reinforcement learning can substantially improve reasoning in pretrained language models, as demonstrated by DeepSeek-R1 \citep{DeepSeekR12025}. What makes a model ready to learn from rewards? Earlier prediction stages shape downstream performance: domain- and task-adaptive training improve classification \citep{Gururangan2020}, and agent-native midtraining strengthens subsequent software-engineering training \citep{Zeng2026DaVinci}. Learning-dynamics analyses connect optimization to generalization in transformers \citep{ChiwunYang2025Scaling}; prefix-learning theory gives a complementary account of adaptation through trainable context \citep{Liang2025Prefix}. To choose training data and interpret these gains, we need to understand what each stage contributes. Prediction may supply information absent from rewards, build a computation that uses it, or already teach part of the task. These possibilities lead to different explanations of the same final score. We ask which information and computations make a new task rule learnable from rewards.

A terminal reward evaluates an answer without directly observing its intermediate mechanism. Consider a learner that composes unknown operations on a hidden state. Two transition rules can agree on every training reward yet demand opposite held-out answers: their difference cancels along training sequences and reappears on test sequences. More reward queries restricted to that support cannot distinguish them; observing individual transitions can. We study such \emph{source prediction} during initial pretraining or \emph{midtraining} of an existing model, followed by reward adaptation of the same parameters (\cref{fig:source-reward-bridge}). A \emph{world} pairs a reusable mechanism with a task rule; our theoretical source laws reveal the former while leaving the latter unspecified. Source prediction trains composition or retrieval; reward training then learns a task-specific use of that computation.
\ifbridgeReview\begin{figure}[!t]\else\begin{figure}[t]\fi
\centering
\begin{tikzpicture}[x=.995cm,y=1cm,>=Stealth,
 box/.style={draw=black!45,rounded corners=2pt,align=center,inner sep=4pt,font=\fontsize{9}{10.5}\selectfont},
 flow/.style={->,thick,draw=black!65}]
\node[box,fill=blue!5,text width=3.15cm,minimum height=1.48cm] (source) at (1.68,0)
 {\textbf{Source observations}\\$s_0\xrightarrow{u_1}s_1\xrightarrow{u_2}s_2$\\$(k_1,v_1)\quad(k_2,v_2)$\\\textit{Task rule hidden}};
\node[box,fill=blue!9,text width=3.45cm,minimum height=1.48cm] (shared) at (6.55,0)
 {\textbf{Reusable computation}\\Compose: $u_2\circ u_1$\\Retrieve: $q=k_2\;\Rightarrow\;v_2$\\\textit{One shared checkpoint}};
\node[box,fill=green!6,text width=2.85cm] (a) at (11.82,.58)
 {Reward A $\to$ task-A policy};
\node[box,fill=green!6,text width=2.85cm] (b) at (11.82,-.58)
 {Reward B $\to$ task-B policy};
\draw[flow] (source.east) -- (shared.west);
\node[font=\fontsize{9}{10.5}\selectfont,align=center] at (4.05,1.12) {Source prediction};
\draw[flow] (shared.east) -- (a.west);
\draw[flow] (shared.east) -- (b.west);
\node[font=\fontsize{9}{10.5}\selectfont,align=center] at (11.82,-1.62)
 {\textbf{Held-out evaluation}\\New combinations or memories};
\node[font=\fontsize{9}{10.5}\selectfont,align=center] at (5.65,-1.40)
 {Measure: information $\to$ acquired computation $\to$ task selection};
\end{tikzpicture}
\caption{Source prediction prepares a shared computation for reward adaptation. States $s_i$ and operations $u_i$ illustrate composition; keys $k_i$, values $v_i$ and query $q$ illustrate retrieval. Rewards select task-specific use.}
\iclrVspace{-3mm}
\label{fig:source-reward-bridge}
\input{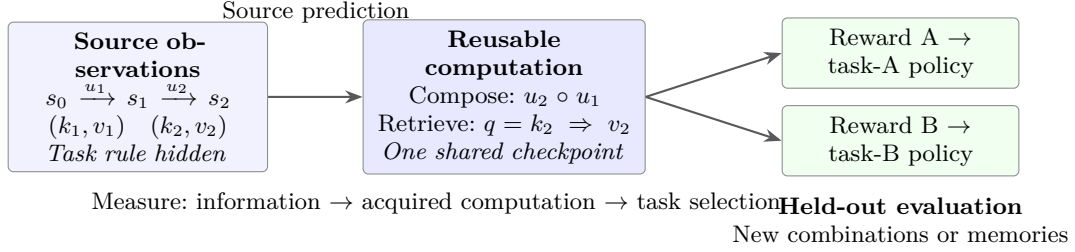}
\end{figure}

\paragraph{Exactly what rewards leave open.}
Prior theory explains how RL exploits rare task demonstrations in prediction data \citep{Tsilivis2025}; controlled experiments also show RL composing pretrained primitives into reusable strategies \citep{Abdulsalam2026}. We characterize what source observations must add when training rewards leave the mechanism ambiguous. In our Sequential family, a two-parameter family of transformations preserves all training rewards, while half reverse every held-out answer (\cref{thm:quotient}). Local source observations distinguish these possibilities. Contextual memory gives a second exact characterization: after finitely many numerical source records, unobserved contexts impose an exact error floor that additional rewards cannot lower (\cref{thm:context-information}).

\paragraph{A finite learning path through the same parameters.}
From specified random initializations, we construct finite sampled Adam paths through both stages with high-probability guarantees. Sequential source prediction teaches a transformer to execute transitions on its own generated prefixes; two REINFORCE updates using $2B_R$ terminal verifications then select the initial operation, giving held-out success at least $7/8$ (\cref{thm:neural}). In contextual memory, source Adam acquires retrieval and reward Adam learns an input-dependent rule, reaching at least $19/20$ within explicit budgets; freezing randomly initialized query/key attention under the same schedule keeps success at most $1/64$ above chance (\cref{ctx:thm:main}). Widths, batches, step sizes and budgets are chosen without test information.

\paragraph{Measuring acquisition and task-specific use.}
Our Qwen2.5 experiments apply source midtraining to pretrained checkpoints and measure acquisition, retention and task-specific use separately. Across eight matched worlds, correct-source models acquire Sequential execution and Memory retrieval; action-target Sequential training reaches 82.61\% success versus 44.15\% for a private-random source control (\cref{fig:v5-acquisition-retention,fig:v5-sequential}). Source-checkpoint probes reveal the computation acquired before rewards; paired reward branches test its use for opposite tasks. Memory replay preserves retrieval; an independent eight-world confirmation reaches 75.32\% task success versus 49.86\% with alternative retrieval (\cref{app:memory-confirmation}). An independent single-world comparison shows correct-source preparation enabling rapid reward learning from zero task success at reward entry (\cref{fig:parameter-behavior}). The measured parameter slice in \cref{fig:parameter-plane} complements these actual-checkpoint curves by displaying source loss and task success across a common plane. On GSM8K and HotpotQA, accuracy at reward entry, reward-stage gain and final accuracy reveal how preparation changes subsequent learning.

{\bf Contributions.}
\begin{itemize}
    \item Exact information characterizations: which mechanisms training rewards leave indistinguishable, and how source observations resolve the ambiguity (\cref{sec:results,sec:retrieval}).
    \item Finite sampled-Adam paths that turn source observations into execution or attention and then into task learning, with a frozen-attention comparison isolating the role of acquired retrieval (\cref{thm:neural,ctx:thm:main}). 
    \item Measurements that distinguish acquiring a computation, retaining it and learning its task-specific use in Qwen2.5, together with the contribution of each training stage to public-task accuracy (\cref{sec:experiments}). \Cref{sec:related} connects the results to prior work; \cref{sec:discussion} develops their implications.
\end{itemize}

  \section{Related Work}\label{sec:related}\label{app:more-related}
We organize prior work around three questions: what information each training stage supplies, how optimization turns that information into computation, and how rewards use the acquired computation.

\subsection{Information supplied by observations and feedback}
\paragraph{Identification with different observation channels.}
Learning an unknown mechanism depends on which queries and observations are permitted. Membership and equivalence queries support regular-language identification \citep{Angluin1987}; related methods learn weighted automata from queries and counterexamples \citep{Weiss2019}. Bounded-width branching programs and statistical-query learning of semiautomata provide further settings in which the observation model determines learnability \citep{Ergun1995,Giapitzakis2025}. Pairwise measurements with side information illustrate a complementary distinction: edge observations constrain relative labels, while additional vertex information helps recover the labels themselves \citep{Foster2018}. In attention inversion, \citet{Deng2023Unmasking} study input recovery from supplied attention weights and outputs. The specified observations likewise determine which reconstruction problem the learner faces. Our information results identify exactly which transition and task transformations leave all training rewards unchanged, and which source observations distinguish their held-out predictions. The resulting obstruction concerns indistinguishable observations even with unlimited reward queries; it is different from an optimizer's difficulty in finding an informative trajectory.

\paragraph{What a verifier supplies.}
The feedback interface matters beyond a binary reward's numerical range. Learning a chain-of-thought verifier online uses an expert label stream and has distinct soundness and completeness objectives \citep{OnlineVerifier2026}. Machine teaching can also choose among demonstrations and other feedback modalities \citep{Larian2026}. These settings supply different information from the terminal correctness feedback in our sequential task. In the contextual binary-action task, correctness together with the chosen action does reveal the binary label, so supervised adaptation is a valid alternative there. We account for this difference when comparing learning routes.

\paragraph{Specifications, safety, and external structure.}
Reach--avoid control specifies dynamics, target sets, and constraints \citep{Fisac2015}. Shielding uses a supplied specification and model information to restrict actions, including under partial observability \citep{Alshiekh2018,Carr2023}; almost-sure reachability in partially observable Markov decision processes (POMDPs) studies the corresponding structural conditions \citep{Junges2021}. Automata-conditioned reinforcement learning likewise uses an explicit automaton interface \citep{Yalcinkaya2025}. These comparisons identify what is supplied to a learner or controller; our sequential result concerns task utility and public-coordinate violations within a specified finite grammar.

Other reliability objectives require different observations and success events. Noninterference compares executions with different secret inputs \citep{Volpano1996}; selective classification couples conditional error with coverage \citep{Geifman2017}; randomized smoothing certifies predictions within a specified perturbation set \citep{CohenSmooth2019}. Safe exploration also distinguishes zero-violation guarantees under suitable structural or startup information from bounds on accumulated constraint violation \citep{Turchetta2016,Amani2021,WeiSafe2024}. Copyright-protection objectives can instead constrain optimization itself: \citet{Chu2024Copyright} formulate a softmax-regression method for avoiding protected outputs. These objectives each specify their own success event and verification requirement, distinct from terminal task success.

\subsection{Learning across source and adaptation stages}
\paragraph{Intermediate training and task information.}
Supervised training on intermediate labeled-data tasks (STILTs) places a supervised intermediate task between language-model pretraining and downstream adaptation \citep{Phang2019}. Domain- and task-adaptive pretraining continue prediction on unlabeled text from a relevant distribution \citep{Gururangan2020}. Agent-native midtraining instead uses trajectories preserving agent workflows, including observations from executed tools and tests \citep{Zeng2026DaVinci}. Their common use of an intermediate stage does not make their information interfaces identical: supervised examples may reveal the downstream choice, whereas our source laws leave the task binding or contextual decision rule unspecified. The public-task experiments use domain adaptation as an empirical comparison; the controlled tasks separate source mechanism information from downstream task information.

\paragraph{From source computation to reward learning.}
\citet[Section 4, Theorems 1--2 and Appendices D.3--D.4]{Tsilivis2025} analyze next-token prediction followed by reinforcement learning on mixtures of short and long demonstrations of a single parity task. Their autoregressive linear model is trained with finite source stochastic gradient descent and self-taught reasoner (STaR) rounds. Their finite learning theorem uses a reward that checks the complete chain of thought, and explains how post-training increases the probability of long, correct computations. In our sequential construction, source observations describe local transitions without revealing the downstream task binding, and feedback checks the terminal answer. This changes both the missing information and what the feedback stage must learn.

\citet[Sections 3.2--3.3 and 4.3--4.4]{WeiKim2026} study a related world-model-to-reward sequence for graph backtracking. They establish a pretraining convergence result for a linear-softmax bigram model, then analyze post-training from a model assumed arbitrarily close to the world law. Their edge-state bigram analysis uses an outcome reward with a length penalty. Its gradient-flow and sign-policy-gradient-flow analysis compares outcome-reward learning with shortest-path supervised fine-tuning (SFT) through inference-time efficiency. Our sequential theorem follows finite sampled updates across both stages, from random initialization, and evaluates held-out operation words.

Reusable modules and compositional curricula provide further accounts of what stages contribute \citep{Kong2026,Rajaraman2026}. Empirically, \citet{Abdulsalam2026} study primitive rewrite pretraining followed by final-answer reinforcement learning (RL) and trace how valid composed procedures develop. Their distinction between exposure to primitives and usable procedures is particularly relevant to our computation interventions. Staged training can also change where a computation occurs: \citet{Internalize2026} analyze a structured curriculum for internalizing explicit chain-of-thought steps. Our question is the acquisition and downstream use of reusable computation across the two stages.

\subsection{Acquisition and reuse of representations}
\paragraph{Transferable prediction structure.}
Language-modeling performance can imply useful downstream features when the source and target tasks satisfy an appropriate relationship \citep{Saunshi2021}. Using linear attention and preconditioned projected proximal updates for sparse contextual bigrams, \citet{Ren2024Transfer} show that a correlated pretrained transition matrix supplies a signal that allows their learning algorithm to bypass its initial sample-intensive stage. Our contextual result instead derives the source-trained attention and head-factor distribution and follows it into learning a new reward-dependent rule with sampled Adam, including its two moment states. Another adaptation route trains an auxiliary prefix: \citet{Liang2025Prefix} establish convergence in a stylized neural-tangent-kernel setting and approximate ultra-long prefix attention using a compact parameterization. Representation reuse also depends on later updates: full fine-tuning can distort pretrained features and worsen out-of-distribution performance \citep{Kumar2022}. We therefore track both source acquisition and retention of the information needed during feedback.

\paragraph{Representing and learning sequential computation.}
Graph algorithms, parallel computation, and automata shortcuts characterize what transformer architectures can express and how depth or intermediate computation changes their resources \citep{Sanford2024,SanfordDepth2024,LiuAutomata2022}. \citet{HuangCoT2025} analyze gradient-based learning and length generalization in structured state-tracking tasks. These results motivate explicit operation and history interfaces. Our sequential source theorem additionally establishes execution on generated prefixes outside the source trajectories, which is what makes the learned local computation available during reward rollouts.

\paragraph{Optimization, generalization and numerical resources.}
\citet{ChiwunYang2025Scaling} connects transformer learning dynamics to generalization through a kernel approximation and separate optimization and statistical regimes. For binary-weight networks, \citet{Daliri2025OneBit} analyze width-dependent kernel behavior, convergence and the discrepancy from full-precision networks. These analyses make the model, training dynamics and resource scaling explicit. Our finite-training results track a further object: a source-acquired computation whose role changes when a new reward objective is introduced.

\paragraph{Learning attention and memory.}
Margin-based token selection, induction-head dynamics, and latent causal structure provide different accounts of attention acquisition \citep{Tarzanagh2024,Chen2024Induction,NichaniCausal2024}. Random-feature, sparse-parity, regular-language, mean-field, and Bayesian copy-head analyses study complementary representations and training regimes \citep{Fu2023,Han2026,Huang2025Regular,Huan2026MeanField,Herty2026MeanField,Lavie2026Copy}. \citet{Ke2025Curse} identify asymmetric feature learning in attention-based time-series forecasting, including difficulties on sign-inconsistent inputs. Sparse-attention approximation asks a complementary question: \citet{Deng2024Sparse} analyze when retaining large attention entries approximates the full attention output. Normalization can change gradient scales at initialization \citep{Xiong2020}. In our contextual model, the relevant interaction occurs across stages: attention pooling changes correlation and signal scale, and the normalization family allows useful reward learning with different directions of attention-contrast change.

Memory is not a single observation interface. Associative-memory analyses study parametric factual storage \citep{NichaniRecall2024}; contextual recall combines pretraining knowledge with implicit attribute inference after fine-tuning \citep{Vasudeva2026}. Statistical in-context learning and abstract-symbol reasoning address other forms of adaptation and generalization \citep{Bai2023,BoixAdsera2023}. Titans changes writable neural memory, while Verifiable Memory trains an agent's operations over long-term memory, active context, and episodic history using multiple verifier signals \citep{Titans2025,VerifiableMemory2026}. At inference time, ParallelComp uses parallel chunk processing and KV-cache eviction to extend the context handled by a fixed language model \citep{Xiong2025ParallelComp}. This changes access to contextual information through compression. Our contextual theorem concerns retrieval and a context--cue rule in a supplied numerical interface, and follows their acquisition through both training stages.

\subsection{Reward learning, empirical comparisons, and capability interfaces}
\paragraph{What rewards change.}
Outcome-reward learning can acquire a graph traversal strategy when the training distribution has sufficient mass on simple instances \citep{RanMilo2026}. That analysis follows expected-gradient flow of attention matrices with fixed values and prescribed initial structure, and also gives a discrete extension. Implicit curricula and learned search have also been analyzed through RL dynamics \citep{Huang2026,Yang2026}, while \citet{Lyu2026Boolean} compare RL and supervised Boolean-policy learning with a supplied dependency mask and process rewards. \citet{Zhou2025} give a reference-policy-dependent resource analysis for Kullback--Leibler (KL)-regularized distributional value learning. These works clarify different entry conditions and learning objectives. Our paired constructions derive source-acquired computation and then study how feedback uses it, alongside the information that feedback alone cannot identify.

Empirical comparisons require the same distinctions. \citet{Yue2025} report improved small-budget success after RL without an expansion of large-budget coverage in their tested settings. Our equivalence result is of a different kind: it identifies held-out targets that rewards on the training distribution cannot reliably distinguish across reward-equivalent worlds. Self-correction training supplies a sequence of attempts and rewards \citep{SCoRe2024}; thought-action models study when initial policies make internal computation useful \citep{HannaThought2025}; logit-bias adaptation studies a different route for using a fixed model \citep{Cohen2026}. Goal-conditioned control also depends on whether success means arrival at a deadline, first arrival, or occupancy \citep{ControlMax2026}. Our experiments distinguish access to successful trajectories, acquisition of the required computation, and reward-dependent adaptation, rather than treating any one of these measurements as all three.

\paragraph{Task-specific computation and verification.}
Several capability families require additional resources that are explicit in their original settings. SATNet incorporates a differentiable satisfiability solver \citep{SATNet2019}; DeepCoder uses learned guidance for program search \citep{DeepCoder2016}; LeanDojo provides an executable theorem-proving and retrieval environment \citep{LeanDojo2023}. Toolformer and ReAct study tool calls and environment interaction \citep{Toolformer2023,ReAct2022}. These are distinct sources of computation and feedback from our finite transition and contextual tasks. Plan-verification theory, rule-based inference attacks, and attention interventions similarly specify particular symbolic or architectural interfaces \citep{PlanVerification2026,Logicbreaks2024,AttentionBoosting2025}.

Instruction-hierarchy training and external enforcement address different parts of agent reliability \citep{InstructionHierarchy2024,IHChallenge2026,CaMeL2025}. Transfer across theory-of-mind tasks must also be measured rather than inferred from in-distribution success \citep{TheoryMind2025}.

  \section{Framework and Preliminaries}\label{sec:setup}
\paragraph{Two stages, two kinds of information.}
A world is a fixed but unknown pair of a reusable mechanism and a task rule. The mechanism maps inputs to intermediate computations; the rule specifies which use of that computation earns a reward. The learner first observes task-independent \emph{source records}, whose distribution we call the source law, and trains a predictor. It then adapts the same parameters $w$ with terminal correctness rewards on training prompts. Evaluation uses held-out prompts and no further rewards. In the sequential family below, the mechanism is a transition table $\theta$ and the task rule is an initial operation $\eta$; in the contextual family of \cref{sec:retrieval}, they are a retrieval map and a context--cue decision rule.

\paragraph{Information and choice order.}
The world and the training/test split are fixed before data sampling; the learner and initialization are chosen independently of the world. Source observations reveal samples of the mechanism but carry no information about the task rule. Each feedback query supplies a fresh training prompt and scores one generated candidate; the learner cannot choose or repeat prompts except where a theorem states otherwise. A symbolic learner may choose its candidate, whereas the neural learner samples it from its current policy. Neither receives test feedback or intermediate correctness labels. Throughout, $\R$ denotes the reals, iid abbreviates independent and identically distributed, and $\E$, $\Pr$ are taken over the randomness specified locally.

\paragraph{Adam across the two stages.}
The neural theorems use bias-corrected Adam \citep{KingmaBa2015}. At stage-local step $t$, let $g_t$ be the minibatch gradient of the source loss (descent) or of the sampled reward score (ascent). For coordinate $j$,
\begin{align}\label{eq:adam-update}
 m_{t,j}&=\beta_1m_{t-1,j}+(1-\beta_1)g_{t,j},&
 v_{t,j}&=\beta_2v_{t-1,j}+(1-\beta_2)g_{t,j}^2,\\
 \widehat m_{t,j}&=m_{t,j}/(1-\beta_1^t),&
 \widehat v_{t,j}&=v_{t,j}/(1-\beta_2^t),\\
 w_{t+1,j}&=w_{t,j}\mp\gamma_{t,j}
 \frac{\widehat m_{t,j}}{\sqrt{\widehat v_{t,j}}+\epsilon_j},&&
\end{align}
with decay coefficients $\beta_1,\beta_2$, coordinate step $\gamma_{t,j}>0$ and offset $\epsilon_j>0$; the sign is $-$ for source descent and $+$ for reward ascent. Moments start at zero. At the stage boundary every parameter is retained, both moments reset, and the step counter restarts. Each theorem fixes its source objective, stopping rule and blockwise steps and offsets. The SGD variant is in \cref{app:sgd-result}.

\paragraph{Sequential states and operations.}
Let $\mathbb F_2=\{0,1\}$ with arithmetic modulo two. Fix $r\ge1$, put $Z=\mathbb F_2^{2r+1}$, $m=|Z|=2^{2r+1}$ and $q=2m$. States and controls form disjoint $q$-symbol alphabets $\mathcal S$ and $\mathcal U$, indexed by $(s,z)\in\mathbb F_2\times Z$ and $(b,u)\in\mathbb F_2\times Z$; with PAD and EOS the vocabulary $\mathcal V$ has $2q+2$ symbols. An unknown array $\theta\in\mathbb F_2^{m\times m}$ defines the transition
\begin{equation}\label{eq:mechanism}
 \phi_{(b,u)}(s,z)=(s+b+\theta_{u,z},\ z+u).
\end{equation}
We call $z$ and $u$ the \emph{public} coordinates: they follow a known update and are readable from any token through a known index map. The state bit $s$ is the \emph{private} coordinate: its evolution depends on the $m^2$ unknown entries of $\theta$. Every learner, neural or symbolic, has the index map and the grammar; neural token embeddings are initialized independently of them. Each operation is a permutation, and uniform averaging over controls sends any state to the uniform law $\Pi$.

Write $u=(x,y,t)$ with $x,y\in\mathbb F_2^r$, $t\in\mathbb F_2$, and define the tag
 $\ell(u)=t+x\cdot y\in\mathbb F_2$,
which is balanced under a uniform control. Let $H$ be the number of executed operations and $L=H-1$ the suffix length. A sequence of controls is a \emph{word}. For a displayed suffix $U_2,\ldots,U_H$ with $U_i=(b_i,u_i)$, training prompts are uniform conditional on $\sum_{i=2}^H\ell(u_i)=0$ and test prompts on the disjoint class with sum one; the split is fixed before source data, initialization and feedback. Equivalently,
\begin{equation}
 C_{\rm suffix}=(-1)^{\sum_{i=2}^H\ell(u_i)},\qquad
 \E_\pm=\E[\,\cdot\mid C_{\rm suffix}=\pm1].
\end{equation}
For prompt functions these expectations include the uniform initial state and suffix; for source functions they also include the independent first control and transition noise. $\E_0$ denotes the source law without tag conditioning.

\paragraph{Source records and prediction.}
A source record has $N=2H+2$ symbols,
\begin{equation}\label{eq:record}
 S_0,U_2,\ldots,U_H,U_1,S_1,\ldots,S_H,\mathrm{EOS},
\end{equation}
with $S_0$ uniform, $U_1$ independent uniform, a training-tag suffix, and states evolving in execution order $U_1,\ldots,U_H$ under the kernel $T_U=aP_U+(1-a)\Pi$, where $P_U$ is the permutation kernel of $\phi_U$ and $0<a\le1$ is the probability of applying the mechanism rather than a uniform reset. Records are independent, unfiltered by downstream success and carry no success labels. The \emph{task binding} $\eta=(b_\eta,u_\eta)$, the unknown initial control that attaches the task to the mechanism, has no effect on the source law; its token may occur in records, but its downstream role is hidden.

The source objective averages next-token cross-entropy (CE) over every position of every record in a minibatch. Random controls and noisy states give a nonzero conditional entropy; the constructions use $a=1-c/H$ with reset scale $c=1/64$ for Adam ($0<c\le1/64$ for SGD) and bound the excess risk above this entropy. All prefix tokens are retained in the $N$-token context, left-padded.

\paragraph{Terminal feedback and success measurements.}
A prompt is $X=(S_0,U_2,\ldots,U_H)$ with target
\begin{equation}
 y^*(X)=\phi_{U_H}\circ\cdots\circ\phi_{U_2}\circ\phi_\eta(S_0).
\end{equation}
The learner generates $A,Y_1,\ldots,Y_H,\mathrm{EOS}$ in $n=H+2$ decisions, where $A$ is the chosen first control and $Y_i$ the $i$th state. The verifier returns one exactly when $A$ is a control, every $Y_i$ is a state, EOS is placed correctly, the public coordinates follow $A$ and then the displayed suffix, and $Y_H=y^*(X)$. It does not require $A=\eta$ or a faithful private history: an incorrect state-bit history that repairs itself is rewarded.

For parameters $w$ and a prompt $X$, under free generation from the induced policy, let $E(w,X)$ be the probability of grammar-respecting endpoint success, $S(w,X)$ the probability of endpoint success obtained with the correct first control $A=\eta$ (task utility), $M(w,X)$ the probability that $A$ is a control and $Y_1=\phi_\eta(S_0)$, and $V(w,X)$ the probability of a grammar or public-coordinate violation. Task utility implies endpoint success; an incomplete execution has zero utility. The memory metric $M$ is a delayed-retention task: prompts differing only in the state bit of $S_0$ share a long suffix but have opposite first-state targets, so the model must carry the initial information through the prefix until it generates $Y_1$.

\paragraph{Learning resources and probability statements.}
The resource vector
\[
 B=(N_{\rm pre},Q_{\rm verifier},N_{\rm generated},N_{\rm updates})
\]
counts observed source tokens, training verifier queries (abbreviated $Q$), generated tokens and optimizer updates. The verifier is the function that returns the terminal reward; one call evaluates $H$ private transition entries and $H$ public checks, so the private table is an oracle resource. High-probability bounds hold for each fixed world $(\theta,\eta)$ over initialization, source data and feedback sampling, simultaneously for every test prompt on that event.

  \section{Information and a finite neural learning path}\label{sec:results}
\subsection{What training feedback cannot identify}
The first question concerns every possible reward, before any optimizer is chosen. Once $u_\eta$ is known, call two worlds \emph{reward-equivalent} if every grammar-valid training candidate receives the same reward in both.

\begin{theorem}[Mechanisms with identical training rewards; informal version of Theorem~\ref{thm:quotient-formal}]\label{thm:quotient}
Let $H\ge3$ and fix $u_\eta$. Two worlds are reward-equivalent if and only if, for some $\kappa,d\in\mathbb F_2$,
\begin{equation}\label{eq:gauge}
 \theta'_{u,z}=\theta_{u,z}+\kappa+d\,\ell(u),\qquad
 b'_\eta=b_\eta+(H\bmod2)\,\kappa+d\,\ell(u_\eta).
\end{equation}
Their targets differ on every test prompt exactly when $d=1$.
\end{theorem}

\paragraph{A three-step example.}
Take $r=1$, $H=3$, $\theta_{u,z}=0$, and $\eta=(0,(0,0,0))$. Compare it with $\theta'_{u,z}=\ell(u)$ and the same binding, the case $\kappa=0,d=1$ in \eqref{eq:gauge}. Starting with private bit zero and using controls with $b=0$, a training suffix with tags $(1,1)$ leaves the original private bit at zero throughout, while the changed mechanism produces private bits $0,1,0$ after the three operations. The terminal target is identical. A test suffix with tags $(1,0)$ instead produces $0,1,1$ under the changed mechanism, so the terminal targets differ. Public coordinates agree throughout. Source observations expose the intermediate distinction: on the same input with $s=0$ and a tag-one control, the probability that the next private bit is one is $(1-a)/2$ in the original source law and $(1+a)/2$ in the changed law. Thus training rewards conceal a difference that a single-step source distribution reveals.

The theorem shows that these tag shifts and the constant shift $\kappa$ exhaust the ambiguity. Its proof compares short paths padded with zero controls, forcing all tag-zero differences to agree and all tag-one differences to agree (\cref{app:quotient}, including the verifier that accepts self-repairing private histories).

Two reward-equivalent worlds with $d=1$ demand opposite answers on every test prompt, so any learner that sees only training rewards has average test success at most $1/2$ on that pair, after any number of queries. Source records break the tie: they expose single transitions before their effects cancel along a training word. This is the information that source observations add. \Cref{thm:resource} gives the symbolic, query-optimal learner that uses it at logarithmic horizon; \cref{thm:neural} below gives the neural one.

\subsection{A complete neural learner in the same family}
\Cref{thm:quotient} identifies what source observations must supply. We now show that a trainable network acquires the reusable execution from them and that rewards then teach the missing binding, along one finite optimization path.

For a full padded prefix with token embedding $e_l$ at lag $l$, the network uses $R$ causal attention heads and a bilinear feed-forward block:
\begin{align}\label{eq:network}
 \alpha_h(l)&=\softmax_l\!\left(\frac{b_h(l)}{t_{\rm att}}
                   +\frac{\ip{Q_he_0}{K_he_l}}{\sqrt{d_h}}\right),\\
 z_{\rm att}&=e_0+R^{-1/2}\sum_hO_h\sum_l\alpha_h(l)V_he_l,\label{eq:network-attention}\\
 f_w(X)&=W\left[z_{\rm att}+U\big((Bz_{\rm att}+b)\odot(Cz_{\rm att}+c)\big)+b_{\rm out}\right].\label{eq:network-output}
\end{align}
Position $l=0$ is the current token and $h$ indexes heads; $b_h(l)$ is a learned positional bias, $t_{\rm att}$ the attention temperature, $d_h$ the query/key width, and $Q_h,K_h,V_h,O_h$ the query, key, value and output maps. The residual attention feature $z_{\rm att}$ feeds a feed-forward/readout block with matrices $U,B,C,W$ and biases $b,c,b_{\rm out}$; $\odot$ is elementwise multiplication and $f_w$ the logit vector. Every matrix, embedding and bias trains. Widths, head count, temperature and Gaussian scales are prescribed independently of $\theta,\eta$ (\cref{seqadam:app:adam-joint,seqadam:app:foundations}). Writing $u_{\rm ff}$ for the bracketed feature in \eqref{eq:network-output}, a second member of the class replaces it by $g\odot\sqrt2u_{\rm ff}/\sqrt{1+\norm{u_{\rm ff}}^2/12}$ with a trainable gain $g$ initialized to ones; \cref{seqadam:app:norm-main} handles the changed function and derivative.

The source objective is the average CE over all targets plus $\lambda\norm w^2/2$ with $\lambda>0$, a coupled $L_2$ penalty on every parameter. Adam runs with $\beta_1=0.9$, $\beta_2=0.999$, bias correction and separate positive readout and backbone rates; the source stage stops when the fresh-batch head gradient falls below a prescribed norm threshold, a criterion computed from raw source data alone. At feedback entry the regularizer is removed, every parameter is retained and both moments reset. Two REINFORCE policy-gradient updates with Adam then use fresh on-policy trajectories, terminal reward and baseline zero; the moments of the first update enter the second, control-output rows use a faster rate, and zero-reward trajectories, EOS scores and malformed outputs all stay in the batch. Let $\varepsilon_{\rm logit}>0$ be the prescribed uniform logit-approximation tolerance (denoted $h$ in \eqref{seqadam:eq:adam-h}).

\begin{theorem}[Finite source learning and sampled Adam improvement; informal version of Theorem~\ref{thm:neural-formal}]\label{thm:neural}
Fix $q=2^{2r+2}$ with $r\ge1$, $c=1/64$ and $\delta\in(0,1)$. Run the source algorithm on either network member in \eqref{eq:network}--\eqref{eq:network-output}, with the horizon, dimensions and training choices of \cref{seqadam:app:adam-joint}, then take exactly two feedback Adam updates, each on $B_R$ fresh sampled trajectories. For every fixed world, with probability at least $1-\delta$, source stopping succeeds and, simultaneously for every held-out prompt $X$,
\begin{align}
 \min\{E(w_f,X),M(w_f,X),S(w_f,X)\}&\ge7/8,\\
 \max\{E(w_s,X),M(w_s,X),S(w_s,X)\}&\le1/q+2n\varepsilon_{\rm logit},\\
 \min_{J\in\{E,M,S\}}\big[J(w_f,X)-J(w_s,X)\big]&\ge3/4,\\
 V(w_f,X)&\le9c/8+1/128,
\end{align}
Here $w_s$ is the source stopping state, $w_f$ the second feedback iterate and $n=H+2$. Feedback uses $Q=2B_R$ queries; the complete source-token and update counts are in \cref{thm:neural-formal}.
\end{theorem}

\paragraph{Reading the theorem.}
The dimensions, precisions, batch sizes, update bound and inverse rates are polynomial in $q,H,\delta^{-1}$ up to logarithms (\cref{seqadam:app:adam-joint}); feedback uses no checkpoint selection. The three metrics move together: a source model already executes, but its success is of order $1/q$ because it guesses the binding; two reward updates raise every metric by at least $3/4$ while keeping format violations at the reset-noise level.

\paragraph{Source loss must control the complete policy.}
Accurate conditional transition predictions do not by themselves make a model emit a state token. In the canonical coefficient class of \cref{seqadam:adam-src-canonical}, an unregularized population near-minimizer can keep exact conditional transitions and an arbitrarily small source gradient yet emit EOS with probability close to one on a chosen suffix (\cref{seqadam:prop:format-obstruction}). Low source loss alone therefore does not deliver a usable executor; the regularizer does. In the canonical class the ridge objective has a unique optimum, which an invariant-gradient calculation identifies as class/PAD terms plus two state--operation interactions. The source excess decomposition controls transition errors and token-class masses (control, state, EOS, PAD) together, including the opposing EOS term; signed contraction bounds the effect of conditioning on the training word class; strong convexity carries the optimum to the random feature coordinates; and ridge bounds every readout coordinate along the Adam trajectory. A momentum argument along the full trajectory, with fresh-record concentration, then forces the observable stopping criterion within a finite budget.

\paragraph{How that computation changes feedback updates.}
The source model executes any supplied operation but chooses the binding nearly uniformly; its task success is of order $1/q$ because the source law leaves the binding unspecified. Forcing a candidate first control makes the model generate its learned transition sequence from there, so the reward stage chooses among computations that already exist: the correct binding earns reward near one, the opposite private bit earns a small reward from self-repairing histories, and wrong public coordinates earn zero. The choice is global and transfers to held-out suffixes because the same learned transitions execute them. Adam's denominator is correlated with the sampled gradient and depends on the previous update. We prove a bound on the feature operator that is uniform over all bounded second-moment histories; Gaussian reference dynamics (the population updates around which the sampled path is controlled) then reaches correct-binding probability $15/16$ in two updates, with a small first step and a larger second step retaining both histories. A comparison of the full sampled path with its stopped reference absorbs sampling, non-choice scores, EOS, wrong-token-class events and all backbone motion into the batch-size requirement of \cref{thm:neural}.

\paragraph{An SGD learning path.}
The same family also admits finite source minibatch SGD followed by sampled policy-gradient updates, with the same $7/8$ final success and $3/4$ improvement for all three metrics, using unregularized source training, a predetermined source endpoint and separately chosen horizons, block rates and budgets (\cref{thm:neural-sgd}; proofs in \cref{app:source,app:foundations,app:fb-main}).

In the sequential model, execution is learned through rich initial features while every backbone block trains at a small positive rate, and feedback selects one global initial operation. The contextual construction next brings the representation itself into the learning problem: attention forms during source training and carries a task rule that varies with the input.

\paragraph{From exact ambiguity to finite exposure.}
Initial source observations and inherited checkpoint weights can be treated jointly as an observation $O$ about the world. A reward transcript can only resolve ambiguity that its allowed prompts expose. For the known two-world gauge experiment in \cref{prop:rare-pair}, with initial-law overlap $\alpha$ and odd training prompts appearing independently with probability $\varepsilon$, the optimal equal-prior endpoint error is $\alpha(1-\varepsilon)^Q/2$. The full raw-source formula is in \cref{cor:rare-source}. These are information bounds for the specified pair; the finite neural path above supplies a separate optimization guarantee. \Cref{app:learning-routes} compares it with quotient recovery from few source records and table recovery followed by binding search.

  \section{Learning the representation that rewards use}\label{sec:retrieval}
The sequential model shows how learned transitions support a new global binding. The harder case is a task rule that varies with the input, because rewards can only teach such a rule through a representation that exposes the relevant input. A contextual memory task separates the two ingredients: a hidden retrieval rule, which source training must acquire, and a decision made with the retrieved information, which only rewards can teach. We characterize what each observation channel identifies, then follow finite Adam updates from random factors through retrieval acquisition to reward-based task learning.

\subsection{A contextual memory task}\label{ctx:sec:experiment}

Let ${\nctx}\ge1$ be the number of contexts, let the content dimension $p\ge17$ be odd, and put $k=p-2$. A fixed unknown world consists of
$
 U=\diag(U_1,\ldots,U_{\nctx}), U_\ell\in\{I_2,P_{\rm swap}\},
  s\in\{\pm1\}^{\nctx},  \tau\in\{\pm1\}^k.
$
Here $I_2$ is the two-dimensional identity, $P_{\rm swap}$ exchanges the two slots, and $k$ is the cue dimension. The map $U_\ell$ specifies which of two slots a query selects; $s_\ell$ is a contextual sign; and $\tau$ specifies a Boolean cue function shared across contexts. The world is fixed before initialization and training data are drawn.

The width ${\rctx}$ counts factor rows. There are ${\rctx}$ query/key rows $q_i,k_i\in\R^{2{\nctx}}$ and ${\rctx}$ head rows $a_i,b_i\in\R^{{\nctx}+p}$, $o_i\in\R$. Every coordinate is initially an independent standard Gaussian. Define the attention and head coefficient matrices (called cores below):
\begin{equation}\label{ctx:eq:cores}
 M=\frac1{\rctx}\sum_{i=1}^{\rctx} q_i k_i^\top,\qquad
 C=\frac1{\rctx}\sum_{i=1}^{\rctx} o_i a_i b_i^\top.
\end{equation}
For query $x$, keys $\kappa_1,\kappa_2$ and values $V_1,V_2$, the network computes
\begin{equation}\label{ctx:eq:model}
 \pi_j=\frac{e^{x^\top M\kappa_j}}{\sum_{j^\prime=1}^2e^{x^\top M\kappa_{j^\prime}}},\quad
 h=\sum_{j=1}^2\pi_j V_j,\quad
 g=h\sqrt{\psi(\norm h^2)},\quad f=g^\top Cg.
\end{equation}
The weights $\pi_j$ pool the memory values into $h$, $g$ is the normalized pooled representation, and $f$ is a scalar logit. The binary policy takes action $+1$ with probability $\sigma(f)=(1+e^{-f})^{-1}$. Its readout is a trainable bilinear feed-forward block, since $f={\rctx}^{-1}\sum_i o_i(a_i^\top g)(b_i^\top g)$. The head receives only the pooled values. All ${\rctx}(6{\nctx}+2p+1)$ supplied parameters are trainable and retained across both stages.

The normalizer is any fixed finite mixture
\begin{equation}\label{ctx:eq:class}
 \psi(u)=\sum_{j=1}^{m_{\rm norm}}\omega_j(u+\zeta_j)^{-\alpha_j},
 \quad \omega_j>0,\quad\sum_j\omega_j=1,\quad
 0\le\alpha_j\le1,\quad0\le\zeta_j\le Z_{\max}<\infty.
\end{equation}
Here $m_{\rm norm}$ counts mixture components, $\omega_j$ are their weights, $\zeta_j$ their offsets, $\alpha_j$ their powers, and $Z_{\max}$ bounds the offsets; $u$ is a squared norm. The mixture constants specify the architecture and remain fixed. Define $g=0$ and its selected derivative as zero at an exceptional zero pool; the proof events avoid this point. The class takes context features and numerical query/key/value inputs, and varies normalization at nonvanishing scale while retaining the attention and factorized head.

Here $O(2)$ is the group of two-dimensional orthogonal matrices, $e_i$ is a standard basis vector, $\otimes$ is the tensor product, $0_{\nctx}$ is a length-${\nctx}$ zero vector, and $N(0,I_p)$ is the standard $p$-dimensional Gaussian. The sign map $\sgn$ codes the future bit as $-1$ or $+1$.
A source record draws a uniform context $\ell$, an independent uniformly random orthogonal frame $R\in O(2)$, and a uniform selected slot $J\in\{1,2\}$. It supplies
\[
 \kappa_j=e_\ell\otimes Re_j,\qquad x=U\kappa_J,
 \qquad V_j=(0_{\nctx},G_j),\qquad G_1,G_2\stackrel{\rm iid}{\sim}N(0,I_p),
\]
followed by the raw future bit $y=\sgn(G_{J,1}G_{J,2})$. Source matching is the correct-slot probability $\pi_J$; its uniform bounds range over the source context, frame and query. Only the future-bit position contributes the cross-entropy $\log(1+e^{-yf})$. The source law is independent of $s,\tau$ and unfiltered by downstream success, so next-bit prediction teaches retrieval without touching the downstream Boolean rule.

A training prompt draws independent uniform $\ell$, query index $q\in\{1,2\}$ and cue $z\in\{\pm1\}^k$. Keys are $e_\ell\otimes e_1,e_\ell\otimes e_2$, the query is $e_\ell\otimes e_q$, and values are
 $(e_\ell,0,0,z),\qquad(e_\ell,0,0,-z).$
The selected slot $j_\star\in\{1,2\}$ is defined by $U_\ell e_{j_\star}=e_q$. Its cue is $(-1)^{j_\star-1}z$, so the correct action is $s_\ell(-1)^{j_\star-1}\sgn(\tau^\top z)$. The policy samples an action $A$, receives only the correctness reward $R_{\rm rew}\in\{0,1\}$, and uses $R_{\rm rew}\nabla\log\Pr(A)$ with baseline zero. Every sample, including zero reward, remains in the minibatch denominator.

Let $\cmin$ be the attention-contrast threshold, $m_{\rm mem}$ the finite amplitude resolution and $\rho_{\rm mem}$ the smaller test amplitude. Set
\[
 \cmin=\frac1{\sqrt k},\quad m_{\rm mem}=\lceil\log_2(2\sqrt k)\rceil,
 \quad \rho_{\rm mem}=1-2^{-m_{\rm mem}}.
\]
Test values have cues $u_1z,u_2z$, with independent uniform $u_1\in\{\pm1\}$ and $u_2\in\{\pm\rho_{\rm mem}\}$. All four amplitude pairs differ from the training pair $(1,-1)$. Their label is
\begin{equation}\label{ctx:eq:testlabel}
 y_{\rm test}=s_\ell\sgn(u_{\rm selected})\sgn(\tau^\top z).
\end{equation}
The two memories share a cue direction. Testing keeps the context identities and changes their memory configurations. Let $J_{\rm test}$ denote average success of one sampled action. For an individual test input the corresponding success is $\sigma(y_{\rm test}f)$.

\subsection{What the two observation channels identify}
Write $\xi_\ell=+1$ for the identity orientation and $\xi_\ell=-1$ for the swap, and put $v_1=1,v_2=-1$. Training labels factor as
$
 y_{\rm train}=s_\ell \xi_\ell v_q\sgn(\tau^\top z).
$
Rewards identify the products $s_\ell\xi_\ell$ and the cue vector $\tau$ up to a common sign, and nothing more; source geometry identifies each orientation $\xi_\ell$. This factorization yields a complete information characterization for the very task that the attention model learns.

\begin{theorem}[Information supplied by contextual source observations; informal version of Theorem~\ref{thm:context-information-formal}]\label{thm:context-information}\label{ctx:prop:information}
Two worlds are reward-equivalent on every training input if and only if
\begin{equation}\label{eq:context-equivalence}
 \tau'=\varepsilon\tau,\qquad s'_\ell \xi'_\ell=\varepsilon s_\ell \xi_\ell
 \quad(\ell=1,\ldots,{\nctx}),\qquad \varepsilon\in\{\pm1\}.
\end{equation}
The ${\nctx}$ orientations may vary independently; the global sign leaves all predictions unchanged. Flipping an orientation reverses exactly the equal-sign test-amplitude labels in that context.

For the exact numerical source, let ${\Nobs}$ count contextual source records, and let $\mathcal R_{\Nobs}$ be the infimum, over all world-independent learners and finite fixed iid feedback-query budgets, of worst-world expected test error after ${\Nobs}$ iid source records. Then
\begin{equation}\label{eq:context-risk}
 \mathcal R_{\Nobs}=\frac14(1-1/{\nctx})^{\Nobs}.
\end{equation}
For ${\nctx}=1$, the factor is one at ${\Nobs}=0$ and zero at ${\Nobs}\ge1$. With source observations alone, the minimax error is $1/2$ for every ${\Nobs}$.
\end{theorem}

The first Fourier coefficients of the cue function identify $\tau$ up to sign, which gives necessity in \eqref{eq:context-equivalence}; one source record identifies the orientation of its context almost surely, which gives \eqref{eq:context-risk}: a test context missed by all ${\Nobs}$ records, an event of probability $(1-1/{\nctx})^{\Nobs}$, keeps its orientation hidden. The orientation changes labels on the equal-sign half of test configurations, where indistinguishability forces error $1/2$, giving $\tfrac12\cdot\tfrac12=\tfrac14$. A finite feedback-label table approaches this floor; \cref{thm:context-information-formal} gives the explicit query count and tolerance (proof in \cref{app:context-information}).

The theorem separates two needs that a task score merges. Source observations remove the orientation ambiguity that rewards preserve; rewards identify a task rule absent from the source law. The next result shows how ordinary training acquires the representation through which that rule becomes learnable. (The optimization result also covers a finite coordinate alphabet; \eqref{eq:context-risk} uses exact numerical observations.)

\subsection{Finite acquisition and feedback learning}\label{ctx:sec:result}

We use bias-corrected Adam with first- and second-moment decay parameters $0\le\beta_1<1$, $0<\beta_2<1$, and $\beta_1^2<\beta_2$. Source minibatch descent starts with zero moments. At its fixed endpoint all parameters are retained, both moments are reset once, and feedback minibatch ascent begins.

The dimension-normalized coordinates are those in \eqref{ctx:eq:cores}. Physical query/key row parameters are $Q_i=q_i/\sqrt {\rctx},K_i=k_i/\sqrt {\rctx}$; physical head rows are $a_i,b_i,o_i$. A normalized step $\eta_t$ and positive denominator offset $\epsilon$ are implemented using query/key step $\eta_t/\sqrt {\rctx}$ and offset $\epsilon/\sqrt {\rctx}$, and head step $\eta_t$ and offset $\epsilon/{\rctx}$. Thus every normalized coordinate has the same nonzero rate and offset. These physical step and offset choices implement the common normalized update.

\begin{theorem}[Acquisition, contextual transfer and intervention; informal version of Theorem~\ref{ctx:thm:main-formal}]\label{ctx:thm:main}
Fix the task dimensions, an architecture \eqref{ctx:eq:class}, Adam parameters $(\beta_1,\beta_2)$ as above, a world $(U,s,\tau)$ and $\delta\in(0,1)$. Choose width, batches, steps, offsets and input precision as in \cref{ctx:sec:budgets,ctx:sec:alphabet}; they are explicit and finite. With probability at least $1-\delta$ over initialization and all training samples, jointly:
\begin{enumerate}
\item \emph{Acquisition.} Initial source matching is at most $0.51$; after source training it is at least $0.9$ uniformly over contexts, frames and queries, and greedy source-bit accuracy exceeds $0.9$.
\item \emph{Transfer.} At the final feedback iterate, sampled success is at least $19/20$ on every held-out context, query, cue and amplitude pair, an improvement of at least $2/5$ over the source checkpoint.
\item \emph{Intervention.} The same schedules and budgets with the initial query/key parameters frozen in both stages, and every head factor trained, end with sampled test success at most $1/2+1/64$; the paired gap exceeds $2/5$.
\end{enumerate}
The conclusions hold for the finite coordinate alphabet of \cref{ctx:sec:alphabet} with supplied numerical decoding and real-valued optimizer states. At fixed ${\nctx},p,Z_{\max},m_{\rm norm}$ the budget formulas may be shared across mixture members, with probability evaluated for each fixed member.
\end{theorem}

The learned policy is input-dependent: reversing the cue sign reverses the correct action at the same context and query, and a constant action has success exactly $1/2$. The learned coefficient shares $\tau$ across contexts and combines it with the contextual sign $s_\ell$. A single shared step and offset per stage carries the matching statistic from at most $0.51$ to at least $0.9$. The frozen-attention run is a complete training run under the same schedule; it keeps the small random initial attention and whatever the trained head can amplify from it, and remains within $1/64$ of chance. Acquired attention exposes the task information that makes the reward stage effective.

Throughout, a \emph{reference} is the deterministic population dynamics around which the sampled Adam path is controlled. The reference source state has attention core $M=\beta U$ with $\beta\ge0$ and correct-slot attention weight (matching) $\sigma(\beta)$.
Pooling two Gaussian values changes both signal correlation and scale, which can compete under normalization. Let $L(t,\beta)$ be the source cross-entropy at the symmetric reference whose logit is $t\,h_1h_2\psi(\|h\|^2)$, where $h_1,h_2$ are the two content coordinates that determine the source bit and $t$ is the coefficient on their product. The Gaussian identity in \cref{ctx:eq:drift} gives
\begin{equation}\label{ctx:eq:mainidentity}
 \partial_\beta L(t,\beta)<0\qquad(t>0,\ 0\le\beta<\infty).
\end{equation}
Increasing matching therefore improves the source objective throughout the normalization family once the head has acquired positive parity alignment. The head's initial source direction creates that alignment, and a uniform drift bound carries attention to reliable matching. The appendix derives the identity by separating conditional label uncertainty from the correctly signed loss.

The source stage also controls the distribution of the trained factors: inactive context and cue factors retain their Gaussian reference law, independently of the evolved source-active coordinates, while the output factor stays symmetric. The proof bounds the deviation of the trained state from this law. This retained distribution supplies the starting state of the feedback argument.

The block entry $C_{\ell i}$ below pairs context coordinate $\ell$ with cue coordinate $i$ (full-vector index ${\nctx}+2+i$); $C_{i\ell}$ denotes the transposed block entry. Let $B$ denote the scalar strength of the learned context--cue interaction. Under the feedback reference law, its coefficients take the form
$
 C_{\ell i}+C_{i\ell}=B s_\ell\tau_i.
$
The pooled training cue is a signed multiple $c z$, where $c=\tanh(\lambda/2)$ is the learned attention contrast and $\lambda$ is the contrast coefficient of the reference attention block (\cref{ctx:sec:feedbackproof}). The policy's population success becomes
\[
 J(B,c)=\E\sigma\!\left(Bw(c)|\tau^\top z|\right),
 \qquad w(c)=c\psi(1+kc^2).
\]
Here $w(c)$ is the effective cue-signal scale. Uniform retrieval gives $c=0$ and erases the cue in training. Acquired retrieval gives a positive direction for $B$ under the specified coordinatewise update. For all class members, $c\ge\cmin$ is a lower boundary preserved by the reference dynamics, because
\[
 \frac{d}{dc}\frac{c}{(1+kc^2+\zeta)^\alpha}
 =\frac{1+\zeta+(1-2\alpha)kc^2}{(1+kc^2+\zeta)^{\alpha+1}}
 \ge0\quad\text{at }c=\cmin.
\]
Here $\alpha,\zeta$ denote a single mixture component's power and offset. This is the structural condition that survives the class variation. When $\alpha=1,\zeta=0$, further sharpening above this boundary can reduce $w$; useful feedback is therefore compatible with decreasing attention contrast. Every held-out selected cue still has a positive pooled signed amplitude, so a sufficiently large learned $B$ gives a uniform test margin.

We compare sampled Adam with a reference evolving each row by $f_\epsilon(v)=v/(|v|+\epsilon)$ applied to a scalar coordinate $v$ of its current scaled population gradient. The proof controls sampled, bias-corrected Adam around it using a deterministic coordinate bound
\[
 \left|\frac{\widehat m_t}{\sqrt{\widehat v_t}+\epsilon}\right|
 \le K_A:=\frac{1-\beta_1}{\sqrt{(1-\beta_2)(1-\beta_1^2/\beta_2)}}.
\]
Here $\widehat m_t,\widehat v_t$ are Adam's bias-corrected first and second gradient moments, and $K_A$ bounds the normalized update. History comparisons retain old momentum and the correlation between each sample gradient and its denominator. Because the head moment map is cubic, its Gaussian rows do not admit a uniform sensitivity bound. We solve the rowwise sensitivity inequality first and only then average, using a finite exponential Gaussian moment. This yields an explicit joint path bound for the trained and frozen branches. Source errors persist across the switch; only optimizer moments reset.

\subsection{Training resources}
For one run, let $B_{\rm mb}$ be the per-update batch size and $T_s,T_f$ the source and feedback update counts. Write $N_s=B_{\rm mb}T_s$ for source records and $N_f=B_{\rm mb}T_f$ for sampled feedback actions. The resource coordinates count source records, verifier queries, sampled actions and optimizer updates:
\[
 (N_{\rm source\ records},Q_{\rm verifier},N_{\rm actions},N_{\rm updates})
 =(N_s,N_f,N_f,T_s+T_f).
\]
Each source record contributes one predicted bit and the numerical coordinates counted at the precision in \cref{ctx:sec:alphabet}; the learned and frozen runs receive the same budget. Together with the minimax error in \cref{thm:context-information}, these budgets give both the information available for learning and a finite algorithm that uses it.

\paragraph{Attention contrast and useful reward signal.}
Retrieval exposes the task cue, but its effective update scale also depends on normalization. For the member $\psi(u)=u^{-1}$, $w(c)=c/(1+kc^2)$ increases up to $c=1/\sqrt{k}$ and decreases above it. \Cref{prop:contrast-scale} derives this dependence within the stated normalization model. The finite theorem combines a preserved sufficient contrast bound with control of the sampled-path error.

  \section{Experiments}\label{sec:experiments}
We measure source acquisition, reward-task selection and retention separately on Qwen2.5 Base models \citep{qwen25report}. The intervention is full-parameter source midtraining from pretrained checkpoints before terminal-reward adaptation. Greedy evaluation uses one highest-probability completion per input; reward training samples at temperature one in the original experiments, with calibrated sampling in the independent Memory confirmation (\cref{app:memory-confirmation}). A world--training-seed pair is one independent unit. We average A/B branches within world before computing 95\% Student-$t$ intervals; effects use paired world differences. Methods and complete comparisons are in \cref{app:v5-results,app:memory-confirmation}; earlier experimental designs are described separately (\cref{app:v4-results,app:additional-comparisons}).

\subsection{What computation does source training acquire?}
Eight fresh worlds per module compare correct source with a matched structural control, using Qwen2.5-0.5B. A task-independent format stage precedes source training, with no later format stage. Sequential observes $2^{22}$ source tokens; its control preserves public states and randomizes private bits. We cross source structure with two objectives: predict states and END/EOS, or additionally predict the first operation already present in each record. This uniformly sampled operation is independent of the downstream task binding. The records and observed-token budgets are unchanged; the added operation positions increase the supervised targets. Memory observes $2^{24}$ tokens with an all-token objective, comparing correct with wrong retrieval on matched inputs.

With the first operation supplied at evaluation, action-objective Sequential source models generate 87.74\% strictly complete trajectories versus 1.93\% for the private-random control (\cref{fig:v5-acquisition-retention}a). Correct-source Memory first-field accuracy is 99.76\%, versus 51.95\% under wrong retrieval; joint original/value-flip accuracy is 99.41\% versus 5.81\%. The latter probe requires both answers to be correct, exposing input-sensitive source predictions without supplying a downstream task rule. These tests distinguish mechanism-dependent responses from output syntax or a constant bit.

\begin{figure}[!htb]\centering
\includegraphics[width=\linewidth]{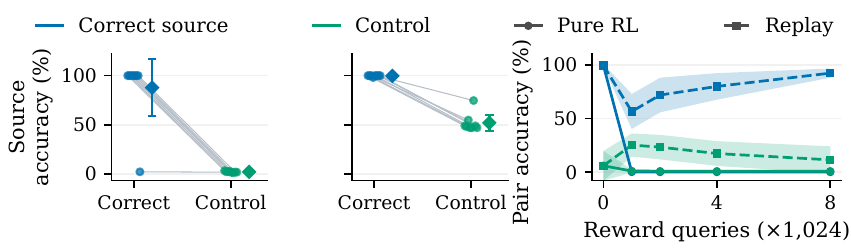}
\input{arxiv/experiments/generated/v5_main_acquisition_labels.tex}
\caption{Acquiring and retaining source computation. (a) Strict trajectories on 512 source probes per world with first operation supplied, using the action objective. (b) Correct first field on 256 Memory records. Points and gray lines show paired worlds; diamonds give means and 95\% intervals. (c) Both answers correct on 256 original/value-flip pairs during reward training, for correct/wrong source with pure RL or replay. A/B branches are averaged within each of eight worlds; bands are untruncated 95\% intervals. Replay adds 131,072 source targets and 128 CE updates per branch.}
\label{fig:v5-acquisition-retention}\end{figure}

\subsection{Do rewards select and retain the computation?}
The original source checkpoints branch into opposite tasks A/B for 8,192 reward queries, with fresh Adam moments and a fixed learning rate $3\times10^{-6}$. Both verifiers score each output on the same 1,024 held-out evaluation records per world. If $J_{XY}$ is success of the task-$X$ branch under verifier $Y$, task selectivity is $\Delta_{\rm task}=\{(J_{AA}-J_{AB})+(J_{BB}-J_{BA})\}/2$. Shared ancestry fixes what the two branches acquired; their cross-scored outputs measure whether different rewards select different task behavior.

\paragraph{Sequential: source computation supports task selection.}
With first-operation supervision, correct-source success rises from 4.83\% to 82.61\%, compared with a control endpoint of 44.15\% (\cref{fig:v5-sequential}a). The paired final-score advantage is 38.46 pp, with 95\% interval $[19.78,57.15]$. The predeclared task-selectivity advantage is 67.57 pp, interval $[37.04,98.10]$, with two-test Holm-adjusted exact $p=.015625$ (\cref{fig:v5-sequential-detail}b). All sixteen correct-source and all sixteen control branches make 128 nonzero reward updates. Thus the source comparison is made between actively adapting models with the same reward budget.

The first-operation target trains an output position needed to begin a free rollout; it supplies no preferred task binding. Both action-objective routes can therefore initiate reward-bearing trajectories, while their source records differ in private-state structure. The resulting selectivity contrast tests the contribution of that structure under the action objective. Comparing action and state objectives is a separate, predeclared secondary analysis: the source-by-objective selectivity interaction is 42.59 pp with $t$ interval $[0.82,84.36]$, but exact $p=.08594$ (\cref{tab:v5-secondary}). The exact test leaves this secondary interaction unresolved.

\begin{figure}[!htb]\centering
\includegraphics[width=\linewidth]{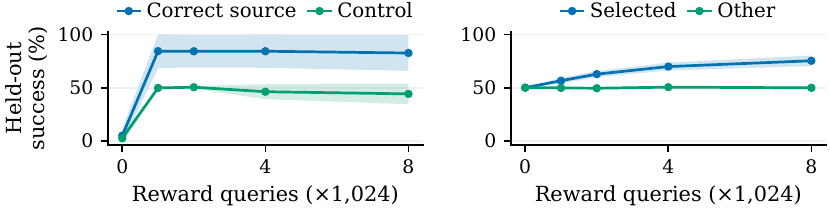}
\input{arxiv/experiments/generated/source_task_learning_labels.tex}
\caption{Source preparation and reward learning. (a) Sequential action-objective routes, worlds 500--507. (b) Independent Memory confirmation, worlds 508--515: selected versus alternative retrieval after a shared correct-source start. Success uses 1,024 fixed held-out records; A/B are averaged within each world, with 95\% intervals across eight worlds. Source costs differ between panels (\cref{app:v5-results,app:memory-confirmation}).}
\label{fig:v5-sequential}\end{figure}

\paragraph{Memory: source replay preserves acquired behavior.}
From the same source checkpoints, Memory compares pure reward training with a source CE step after each reward update. Replay uses 1,024 targets from that route's own previously observed source records, including wrong labels for the wrong-retrieval control. Both kinds of update share the branch's Adam moments and learning rate. After 8,192 queries, correct-source replay retains 92.21\% joint original/flip-pair accuracy versus 0\% under pure RL (\cref{fig:v5-acquisition-retention}c); first-field accuracy is 96.24\% versus 49.66\%. All branches perform 128 nonzero RL updates; replay additionally uses 131,072 source targets in 128 CE updates.

Correct-source final task accuracy is numerically higher than wrong retrieval by 4.40 pp under pure RL and 5.24 pp with replay, with paired intervals $[-11.38,20.18]$ and $[-8.10,18.58]$. The predeclared correct-minus-control task-selectivity effect under replay is 0.67 pp, interval $[-1.22,2.57]$ (\cref{tab:v5-primary,tab:v5-main-endpoints}). The source probes establish that replay preserves responses to stored values; the task intervals span zero. Measuring both distinguishes retaining the source computation from selecting it for the reward task.

\paragraph{Memory: task-relevant retrieval supports reward learning.}
In an independent confirmation on eight new Memory worlds, task-relevant retrieval reaches 75.32\% success versus 49.86\% for alternative retrieval, from A/B-averaged entry scores of 50.00\% (\cref{fig:v5-sequential}b). The paired accuracy advantage is 25.46 pp (95\% CI 20.10--30.81), positive in every world. The predeclared selectivity advantage is 50.92 pp (CI 40.20--61.63; exact $p=.0078125$), a separate single primary test. Both routes inherit the same correct $2^{24}$-token source checkpoint, then receive matched training on public-projection and retrieval-vector targets. Reward adaptation uses calibrated sampling, vector/syntax replay without supervising source answer signs, and guarded steps. This comparison shows task-relevant retrieval supporting reward learning under this recipe; full methods, added source costs and all 32 endpoints appear in \cref{app:memory-confirmation}.

\paragraph{Stage order and computation retention.}
Format learning and computation retention are distinct: an earlier study found that learning termination could coincide with a loss of source-probe accuracy (\cref{fig:acquisition-format,app:v2-results}). This observation motivates format preparation before source acquisition. The resulting sequence lets probes distinguish what source training acquires, what adaptation retains, and which task the rewards select. Initialization and budget comparisons appear in \cref{app:additional-comparisons}.

\ifbridgeReview
\subsection{How does source training change the subsequent reward trajectory?}
\label{sec:parameter-plane}
A separate Sequential world (508, Task A) compares direct RL, correct source and private-random source from one format checkpoint. Source routes use $2^{22}$ tokens with first-operation supervision; all routes use 8,192 reward queries and 128 nonzero RL updates. On 1,024 held-out inputs, correct-source success rises from 0\% to 100\% within 1,024 queries and remains there; direct RL and the private-random route finish at 51.76\% and 51.27\% (\cref{fig:parameter-behavior}). This independent world complements the eight-world A/B tests with a direct-RL comparison: source preparation enables rapid reward learning without already solving the task at entry. The original trajectories measure how preparation changes subsequent learning, while the parameter slice shows where those trajectories lie.

The common parameter plane (\cref{fig:parameter-plane}) retains 91.96\% of aggregate squared displacement. Its SVD basis is fitted jointly to source and reward checkpoints from all three routes. Its 81 grid models are evaluated on 256 task inputs and 128 source records. Task success uses strict output scoring; source cross-entropy is in nats per target. Every grid model uses the same evaluation records. Correct and private-random source move in different projected directions; direct RL stays nearer the origin in this projection. Colors score the slice models, while actual learning is read from the original-checkpoint curves: projected and original models can differ. Construction, residuals and reconstruction measurements are in \cref{app:parameter-plane}.
\iclrVspace{-3mm}
\else
\subsection{How does source training change the subsequent reward trajectory?}
\label{sec:parameter-plane}
A separate, prespecified Sequential world (508, Task A) compares three Qwen2.5-0.5B routes from the same format checkpoint: direct RL, correct source followed by RL, and public-preserving/private-random source followed by RL. The source routes each observe $2^{22}$ tokens with first-operation supervision; all routes use 8,192 reward queries. Original-checkpoint evaluation on 1,024 fixed held-out inputs shows correct-source success rising from 0\% at reward entry to 100\% after 1,024 queries, remaining there through 8,192 queries. Direct RL finishes at 51.76\% and the private-random route at 51.27\%; all three make 128 nonzero RL updates (\cref{fig:parameter-behavior}). Thus, in this world, correct source prepares a model for rapid reward learning without already solving the downstream task at reward entry. This independently specified world complements the eight-world A/B task-selection experiment with a direct-RL comparison.

\Cref{fig:parameter-plane} evaluates a common two-dimensional parameter slice at 81 grid models: Task A success on 256 held-out inputs and source cross-entropy on 128 correct-source records. Correct and private-random source move the parameters in different projected directions, while direct RL remains closer to the shared origin in this projection. The plane retains 91.96\% of aggregate squared parameter displacement. Its colors describe the evaluated slice models; projected checkpoints can have different task scores from their originals, so training performance is read from the actual-checkpoint curves in \cref{fig:parameter-behavior}. This separates two views of preparation: the measured slice displays how source loss and task success vary across parameters, and the original trajectories show how each starting point responds to rewards. \Cref{app:parameter-plane} gives the plane construction, route-wise residuals and paired reconstruction measurements.
\fi

\paragraph{An additional source stage and final public-task performance.}
Five paired Qwen2.5-1.5B seeds compare direct RL, domain-source training followed by RL, and shuffled-source training followed by RL. Source data are raw OpenWebMath documents \citep{OpenWebMath2023} for GSM8K arithmetic \citep{GSM8K2021}, or title-disjoint Wikipedia passages for HotpotQA multi-hop questions \citep{HotpotQA2018}. Source training receives no downstream answer or supporting-fact labels; it uses $2^{22}$ prediction tokens, and reward adaptation uses 8,192 queries. The shuffled control permutes tokens within 64-token document blocks. This comparison asks how continued prediction changes reward-entry performance and subsequent reward gains.

For these public-task comparisons, let $J_{a,Q}$ be route $a$'s greedy success after $Q$ reward queries. Its within-seed gain and paired difference from direct RL are
\begin{equation}\label{eq:supplement-gains}
G_a=J_{a,8192}-J_{a,0},\qquad D=G_{\mathrm{source}}-G_{\mathrm{direct}}.
\end{equation}

On identical before/after sets of 1,319 GSM8K or 1,024 HotpotQA questions, GSM8K domain-source training moves feedback-entry accuracy from 8.57\% to 2.32\%, and source+RL then reaches 62.05\% against 62.32\% for direct RL: the source route's paired reward-stage gain is $5.97\pm3.66$ points larger. HotpotQA source training raises the starting score from 23.93\% to 35.10\%, with final scores of 45.59\% and 47.60\%: here the source route's paired gain is $13.18\pm1.50$ points smaller (\cref{tab:public-summary}; full figure and learning curves in \cref{fig:revision-public,fig:revision-public-curves}). Final accuracy is entry accuracy plus reward gain, and a source stage changes the starting point and the subsequent gain in different directions, so none of the three quantities can stand in for the others. A $2^{20}$-token source route on the same seeds reaches 61.76\% on GSM8K and 46.23\% on HotpotQA; the seed-paired endpoint differences from the $2^{22}$ route are $-0.29\pm5.64$ and $0.64\pm1.12$ points (\cref{tab:v2-public-dose,tab:v2-public-contrasts}).
\begin{table}[!htb]
\centering
\begingroup
\newcommand{\pairvalue}[2]{\ensuremath{#1}\,{\scriptsize\ensuremath{\pm#2}}}
\captionsetup{justification=raggedright,singlelinecheck=false}
\small\setlength{\tabcolsep}{2.15pt}\renewcommand{\arraystretch}{1.12}
\begin{minipage}[t]{0.48363483\linewidth}
\vspace{0pt}\raggedright
\caption{Task accuracy (\%) in the original eight-world comparison (500--507), after 8,192 queries. Means $\pm$ 95\% Student-$t$ half-widths after A/B averaging, on 1,024 records; $\Delta$ is paired correct minus control (pp).}
\label{tab:v5-main-endpoints}
\end{minipage}\hspace{.02\linewidth}%
\begin{minipage}[t]{0.49636517\linewidth}
\vspace{0pt}\raggedright
\caption{Public-task accuracy (\%) at entry and after 8,192 queries; gain is final minus entry (pp). Means $\pm$ 95\% Student-$t$ half-widths over five paired seeds, on 1,319 GSM8K / 1,024 HotpotQA records. None denotes direct RL; its fixed-checkpoint entries have $\pm0$.}
\label{tab:public-summary}
\end{minipage}%
\par\noindent%
\begin{minipage}[t]{0.48363483\linewidth}
\vspace{0pt}\raggedright
\begin{tabular*}{\linewidth}{@{\extracolsep{\fill}}lrrr@{}}\toprule
Setting & Correct & Control & $\Delta$ \\\midrule
\multicolumn{4}{@{}l}{\textit{Sequential}} \\
State & \pairvalue{27.99}{30.48} & \pairvalue{12.41}{15.57} & \pairvalue{+15.58}{19.42} \\
Action & \pairvalue{82.61}{16.80} & \pairvalue{44.15}{9.71} & \pairvalue{+38.46}{18.69} \\
\midrule
\multicolumn{4}{@{}l}{\textit{Memory}} \\
Pure RL & \pairvalue{48.16}{3.61} & \pairvalue{43.76}{14.75} & \pairvalue{+4.40}{15.78} \\
Replay & \pairvalue{47.54}{3.44} & \pairvalue{42.30}{11.84} & \pairvalue{+5.24}{13.34} \\
\bottomrule\end{tabular*}
\end{minipage}\hspace{.02\linewidth}%
\begin{minipage}[t]{0.49636517\linewidth}
\vspace{0pt}\raggedright
\begin{tabular*}{\linewidth}{@{\extracolsep{\fill}}lrrr@{}}\toprule
Source & Entry & Gain & Final \\\midrule
\multicolumn{4}{@{}l}{\textit{GSM8K}} \\
None & \pairvalue{8.57}{0} & \pairvalue{53.75}{4.14} & \pairvalue{62.32}{4.14} \\
Domain & \pairvalue{2.32}{0.39} & \pairvalue{59.73}{3.16} & \pairvalue{62.05}{2.93} \\
Shuffled & \pairvalue{8.58}{1.31} & \pairvalue{41.12}{28.34} & \pairvalue{49.70}{29.29} \\
\midrule
\multicolumn{4}{@{}l}{\textit{HotpotQA}} \\
None & \pairvalue{23.93}{0} & \pairvalue{23.67}{1.02} & \pairvalue{47.60}{1.02} \\
Domain & \pairvalue{35.10}{0.98} & \pairvalue{10.49}{1.74} & \pairvalue{45.59}{0.86} \\
Shuffled & \pairvalue{5.62}{0.96} & \pairvalue{38.83}{1.91} & \pairvalue{44.45}{1.13} \\
\bottomrule\end{tabular*}
\end{minipage}%
\endgroup
\iclrVspace{-3mm}
\end{table}
\iclrVspace{-5mm}

\paragraph{Public controls and transfer.}
Public answer-only SFT is weaker than RL on GSM8K (12.36\%) and stronger on HotpotQA (53.74\%) (\cref{tab:refinement-public-sft-greedy}). Secondary measurements show that the routes differ in answer style, not only in correctness: GSM8K source routes generate longer solutions and the shuffled route's wide interval coincides with format failures ($17.8\pm47.5\%$) and truncated outputs ($17.7\pm47.6\%$), while HotpotQA answers are short and well formed on every route (\cref{tab:public-secondary}); the HotpotQA direct-RL lead over the domain source, about two points, holds on both bridge and comparison questions (\cref{tab:public-question-types}). Transfer depends on the evaluation task: after GSM8K training, GSM-Symbolic success is 53.0\% for source+RL against 54.5\% for direct RL, and BIG-Bench Hard (BBH) logic and object-tracking results depend on the training task and route (\cref{tab:public-transfer,tab:refinement-controlled-bbh}). Public-task best-of-16 selection measures candidate coverage under test-time verification, a separate resource from reward training (\cref{tab:public-transfer,app:additional-comparisons}).

  \FloatBarrier
  \section{Discussion}\label{sec:discussion}\label{ctx:sec:comparisons}
\paragraph{Rewards select among computations that exist.}
The information results and training constructions explain complementary parts of the same mechanism. Source observations distinguish mechanisms that training rewards leave indistinguishable. The Adam constructions turn this information into execution or retrieval, then use sampled rewards to learn its task-specific use: two updates in the sequential construction and an explicit finite budget in the contextual one. The acquired computation applies to held-out inputs, allowing the task rule to transfer. This gives a concrete question to ask of a reward-stage gain: which acquired computation supports the new task rule? The measurements in \cref{sec:experiments} address each part of this question.

\paragraph{Three interventions, three questions.}
The no-source bound in contextual memory concerns world-independent learners with the specified observations. The frozen-attention run holds the training schedule fixed and freezes the initial random query/key factors in both stages while training the head. Direct language-model RL omits additional source midtraining and starts from inherited pretrained weights. Each intervention isolates one question: information access, representation learning, or continued source training. \Cref{app:learning-routes,tab:routes} compare the neural path, which reaches the binding through the model's own sampled trajectories at $2B_R$ verifier calls, with an explicit controller that needs at most $q-1$ verifications, and with supervised and test-time alternatives.

\paragraph{Initial information and the stage of source prediction.}
The information results specify what the learner observes. An inherited checkpoint that already reveals the hidden mechanism can distinguish worlds even when their training rewards coincide; the no-source bounds assume a world-independent learner and initialization. Likewise, the source-only bound uses source laws that leave the downstream rule unspecified. The finite Adam constructions start from the stated random distributions, whose initial geometry lets the proofs establish head alignment and control the training trajectory. The Qwen experiments study continued source prediction from inherited weights, making source midtraining the intervention. The constructions and experiments thus study acquisition from two distinct starting points: specified random factors and inherited language-model weights. On public tasks, domain-related source text without downstream labels is a different information interface from the rule-independent source laws of the controlled families.

\paragraph{From acquired computation to task use.}
Source-checkpoint probes, retention measurements and paired reward branches separate acquired computation from its task-specific use. The historical output decompositions illustrate this distinction: a valid public path or output format can coexist with chance-level private-task behavior (\cref{app:historical-evaluation,app:memory-diagnostic}). The original eight-world Sequential comparison shows reward-task selection after execution is acquired; its Memory comparison shows that replay preserves intervention-sensitive retrieval (\cref{app:v5-results}). The independent Memory confirmation connects retrieval preparation to task learning: under a matched adaptation recipe, task-relevant retrieval reaches 75.32\% success versus 49.86\% for alternative retrieval (\cref{app:memory-confirmation}). Both confirmation routes inherit correct source information and receive the same added training budget. The original comparison uses a two-test Holm family; the confirmation has a separate single primary test and additional source preparation. Together, these experiments show how stage-specific measurements connect the computation acquired from source data to its retention and reward-dependent use.

\paragraph{Entry accuracy, reward gain and final accuracy.}
Final accuracy combines two effects of source training: accuracy at reward entry and gain during reward adaptation. Our public-task comparisons show that these effects can move in different directions. Reporting all three scores locates the improvement in the training sequence; the controlled-task probes identify the computation available at each stage. This connects public-task evaluation to the paper's central question: which computations does source prediction make available for rewards to use?

\fi
\FloatBarrier
\ifbridgeReview\input{arxiv/sections/iclr9_conclusion}\else\section{Conclusion}
Pretraining and midtraining can make a task learnable by revealing a mechanism that training rewards leave ambiguous and building the computation needed to use it. Our finite families connect this information gain to execution, retrieval and task learning along sampled-Adam paths from random initialization. With first-operation supervision, Sequential experiments demonstrate reward-task selection; Memory experiments connect retrieval retention to task use. Together with public-task comparisons, these results show why understanding learning from rewards requires tracking both the computation a model brings to the reward stage and how that stage uses it.
\fi
\ifbridgeReview\clearpage\input{arxiv/sections/iclr_statements}\fi
\bibliographystyle{plainnat}
\bibliography{arxiv/refs}
\clearpage
\appendix
\begin{center}\LARGE\bf Appendix\end{center}
\section{Experimental details and supplementary results}
\label{app:experiments}
World/training seeds are the independent units throughout; shared source trajectories, evaluation prompts and checkpoints are repeated measurements within those units. Item-level results, configurations and analysis code accompany the paper.

\subsection{Output supervision and source replay}\label{app:v5-results}
We use eight fresh worlds (seeds 500--507) for each module, Qwen2.5-0.5B, and a shared task-independent format stage followed by source training and reward adaptation. The format stage has 128 updates for Sequential and 32 for Memory, at learning rate $10^{-5}$. Source training also uses $10^{-5}$: Sequential receives $2^{22}$ observed tokens in 512 updates, and discrete Memory receives $2^{24}$ in 2,048 updates. There is no format stage after source training.

Sequential crosses correct versus public-preserving/private-randomized source with state/END/EOS targets versus first-operation/state/END/EOS targets. The new target is an already observed random operation, independent of the private task binding; initial-state and suffix inputs are not prediction targets. Memory crosses correct versus wrong-retrieval source with pure reward adaptation versus source replay. Both retrieval routes use the all-token source objective. Replay samples only complete records previously observed by that route, with an independent random stream, and supplies exactly 1,024 source prediction targets after each reward update. It shares the branch's Adam state and learning rate, $3\times10^{-6}$, with reward training. Each replay branch thus uses an additional 131,072 source targets and 128 cross-entropy updates.

Each source checkpoint initializes two reward-trained branches, one for task A and one for task B. Relative to A, B flips the Sequential binding's private bit, or negates all Memory context signs; mechanism and prompt distribution remain fixed. Every branch receives 8,192 verifier queries in 128 batches of 64, at fixed learning rate $3\times10^{-6}$. We score unmodified generated outputs with the strict verifier. At 0, 1,024, 2,048, 4,096 and 8,192 queries, all routes and both task-trained models use the same fixed 1,024 held-out evaluation records within a world. Source probes are repeated at these points. There are 16 format stages, 48 source stages and 128 reward branches; shared ancestors do not create additional independent samples. Across both modules the reward-training budget is 1,048,576 queries.

The two primary paired effects compare correct and matched-control task selectivity under first-operation supervision for Sequential and under source replay for Memory. We first average the task-A and task-B contributions within each world, then report the eight world-level paired differences, a 95\% Student-$t$ interval, and a two-sided exact sign-flip test. Exact sign-flip inference assumes sign symmetry of paired world effects under the null. Holm correction applies jointly to these two primary tests. The target-objective and replay interactions are secondary comparisons. These samples and this test family are separate from the earlier eight-world acquisition study.

\paragraph{Resources, implementation and evaluation units.}
 Sequential state-only and action-plus-state objectives share observed source records, but adding operation targets increases optimized target positions. Observed tokens, optimized prediction targets and optimizer calls are therefore distinct resources. Memory pure/replay A/B branches reuse the same route-specific source ancestor. Replay uses that ancestor's recorded labels, excludes partial boundary and held-out records, and preserves the reward-prompt random stream. A replay branch has 128 RL calls and 128 source CE calls, versus 128 RL calls for pure training. Both modules use BF16 model evaluation with FP32 master parameters and Adam moments in this eight-world comparison. Sampled completions use the generation backend; response log probabilities for updates are recomputed by the trainer. The constructions establish finite learning paths in the specified architectures and schedules; these Qwen experiments measure acquisition and reward-dependent use in pretrained language models.

The 1,024 evaluation items in a world are fixed across routes, tasks and checkpoints. Sampling can repeat a prompt: Sequential worlds 501 and 502 each contain 1,023 distinct prompt texts among 1,024 item IDs; the remaining worlds contain 1,024. These repeats are identical across paired routes and are retained with their original sample multiplicities. The independent statistical unit is the world, not an item, A/B branch or checkpoint. Source probes use 512 Sequential records with first operation supplied, and 256 Memory records with matched relevant-value interventions. These probes measure execution of the source computation on held-out records.

\paragraph{Primary and secondary effects.}
\Cref{tab:v5-primary} reports the complete predeclared two-test family; \cref{tab:v5-secondary} reports every predeclared secondary contrast. The Sequential primary effect compares correct and private-random source within the action objective. The source-by-objective interaction measures how adding action supervision changes the source-control selectivity advantage. Its Student-$t$ interval is positive, while the unadjusted exact sign-flip test gives $p=.08594$ over the eight world effects.
\begin{table}[!htb]\centering
\caption{Predeclared source-control effects on task selectivity (percentage points), eight paired worlds per module. Two-sided exact sign-flip tests; Holm adjustment over both primary tests.}\label{tab:v5-primary}
\small
\begin{tabular}{lrrrr}\toprule
Module & Effect & 95\% CI & Exact $p$ & Holm $p$ \\
\midrule
Sequential & 67.57 & [37.04, 98.10] & 0.007812 & 0.015625 \\
Memory & 0.67 & [-1.22, 2.57] & 0.640625 & 0.640625 \\
\bottomrule\end{tabular}
\end{table}

\begin{table}[!htb]\centering
\caption{All predeclared secondary contrasts at 8,192 queries (percentage points). Eight paired world differences; 95\% Student-$t$ intervals and unadjusted exact sign-flip tests. }\label{tab:v5-secondary}
\small
\begin{tabular}{lllrrr}\toprule
Module & Comparison & Metric & Effect & 95\% CI & Exact $p$ \\
\midrule
Seq. & Correct: action--state & Select. & 43.55 & [1.47, 85.64] & 0.08594 \\
Seq. & Correct: action--state & Success & 54.62 & [23.75, 85.50] & 0.01562 \\
Seq. & Control: action--state & Select. & 0.96 & [-0.09, 2.02] & 0.04688 \\
Seq. & Control: action--state & Success & 31.74 & [13.15, 50.32] & 0.01562 \\
Seq. & Source interaction & Select. & 42.59 & [0.82, 84.36] & 0.08594 \\
Seq. & Source interaction & Success & 22.88 & [-1.44, 47.20] & 0.08594 \\
Mem. & Correct: replay--pure & Select. & -0.06 & [-0.52, 0.40] & 1.00000 \\
Mem. & Correct: replay--pure & Success & -0.62 & [-5.54, 4.31] & 0.62500 \\
Mem. & Control: replay--pure & Select. & -0.59 & [-2.45, 1.27] & 1.00000 \\
Mem. & Control: replay--pure & Success & -1.45 & [-6.13, 3.22] & 0.56250 \\
Mem. & Source interaction & Select. & 0.52 & [-1.40, 2.45] & 0.87500 \\
Mem. & Source interaction & Success & 0.84 & [-6.20, 7.87] & 0.68750 \\
\bottomrule\end{tabular}
\end{table}

\begin{table}[!htb]\centering
\caption{Original eight-world task success: reward entry, within-world reward gain, and final score at 8,192 queries. A/B branches are averaged within each world; means and 95\% Student-$t$ half-widths over eight worlds. Entry and final are percentages; gain is percentage points.}\label{tab:v5-endpoints}
\small
\begin{tabular}{llrrr}\toprule
Module & Route & Entry & Gain & Final \\
\midrule
Sequential & Correct/state & $0.20\,{\pm 0.46}$ & $27.80\,{\pm 30.37}$ & $27.99\,{\pm 30.48}$ \\
Sequential & Control/state & $0.00\,{\pm 0.00}$ & $12.41\,{\pm 15.57}$ & $12.41\,{\pm 15.57}$ \\
Sequential & Correct/action & $4.83\,{\pm 6.72}$ & $77.78\,{\pm 17.72}$ & $82.61\,{\pm 16.80}$ \\
Sequential & Control/action & $2.32\,{\pm 3.39}$ & $41.83\,{\pm 11.12}$ & $44.15\,{\pm 9.71}$ \\
Memory & Correct/pure & $49.70\,{\pm 0.71}$ & $-1.54\,{\pm 3.80}$ & $48.16\,{\pm 3.61}$ \\
Memory & Control/pure & $41.41\,{\pm 11.93}$ & $2.35\,{\pm 9.53}$ & $43.76\,{\pm 14.75}$ \\
Memory & Correct/replay & $49.70\,{\pm 0.71}$ & $-2.16\,{\pm 2.82}$ & $47.54\,{\pm 3.44}$ \\
Memory & Control/replay & $41.41\,{\pm 11.93}$ & $0.90\,{\pm 5.18}$ & $42.30\,{\pm 11.84}$ \\
\bottomrule\end{tabular}
\end{table}

\paragraph{Absolute success and task selection.}
Correct-source Sequential action models finish at 78.09\% and 87.13\% on tasks A/B, versus 38.39\% and 49.90\% under private-random source. Averaging A/B within world gives 82.61\% versus 44.15\%, a paired difference of 38.46 pp (95\% interval $[19.78,57.15]$). These absolute-success summaries complement the task-selectivity primary test. Here the source advantage is measured against the private-random source control.

Correct-source Memory replay retains the acquired source response: original/flip-pair accuracy is 92.21\%, compared with 0\% under pure RL, with a paired retention-difference interval of $[87.93,96.50]$ pp. First-field accuracy is 96.24\% versus 49.66\%. Correct source exceeds wrong retrieval in final task accuracy by 4.40 pp under pure RL (95\% interval $[-11.38,20.18]$) and 5.24 pp with replay ($[-8.10,18.58]$). These endpoint intervals and the primary task-selectivity result leave the task advantage unresolved at this budget. Replay preserves the measured source computation, while its task-specific use remains a separate outcome.

\begin{figure}[!htb]\centering
\includegraphics[width=\linewidth]{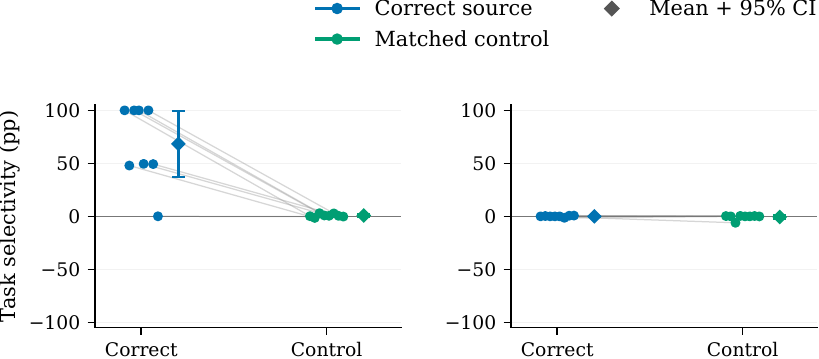}
\caption{Correct and matched-control task selectivity at 8,192 reward queries: (a) Sequential action objective; (b) Memory source replay. Small dots are eight worlds, gray lines pair worlds, and diamonds give route means with untruncated 95\% intervals. Primary inference uses paired differences, not overlap of route intervals.}\label{fig:v5-primary}
\end{figure}

\begin{figure}[!htb]\centering
\includegraphics[width=\linewidth]{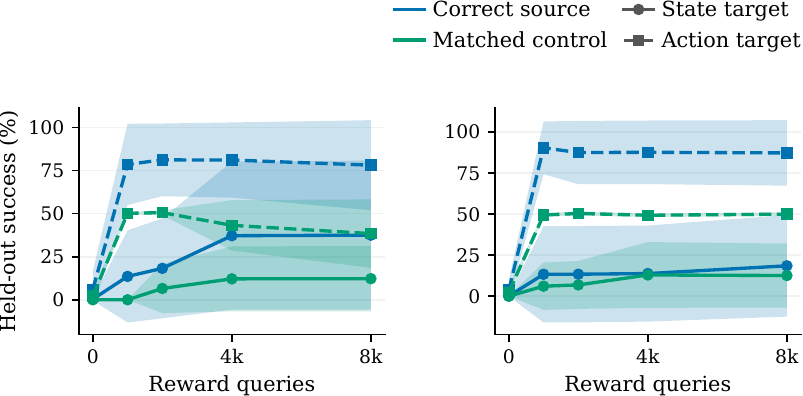}
\caption{Sequential learning curves for (a) task A and (b) task B, with correct/private-random source crossed with state/action-plus-state targets. Each point averages eight worlds on 1,024 fixed evaluation items per world; bands are untruncated 95\% Student-$t$ intervals.}\label{fig:v5-seq-ab}
\end{figure}

\begin{figure}[!htb]\centering
\includegraphics[width=\linewidth]{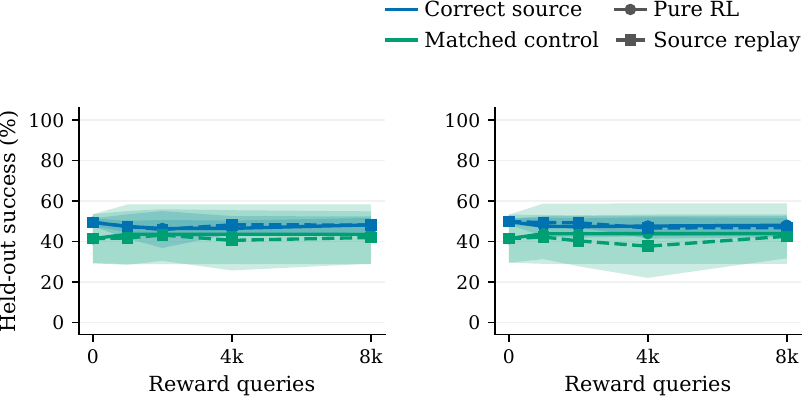}
\caption{Memory learning curves for (a) task A and (b) task B. Correct/wrong retrieval is crossed with pure reward training/source replay; the same ancestors and evaluation items are paired. Means and untruncated 95\% intervals over eight worlds.}\label{fig:v5-memory-ab}
\end{figure}

\begin{figure}[!htb]\centering
\includegraphics[width=\linewidth]{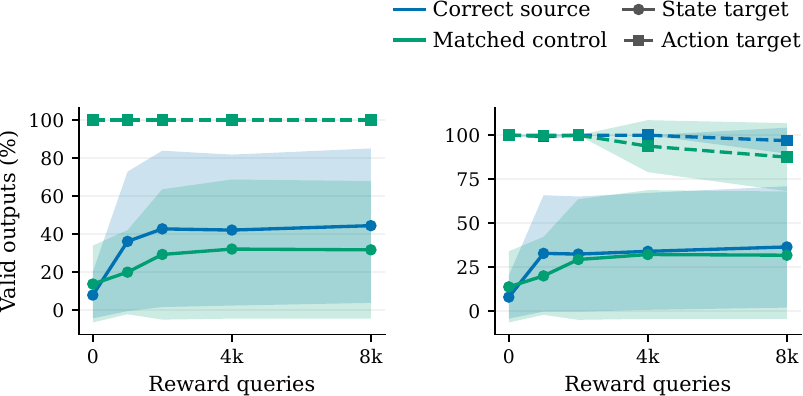}
\caption{(a) Legal first-operation output and (b) complete syntax on the fixed held-out reward inputs. Four source routes, A/B averaged within world, with 95\% intervals across eight worlds. Correct syntax and task correctness are separate metrics.}\label{fig:v5-interface}
\end{figure}

\begin{figure}[!htb]\centering
\includegraphics[width=\linewidth]{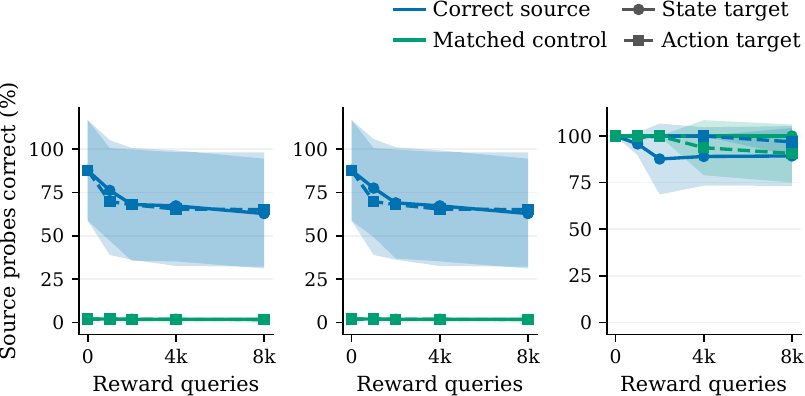}
\caption{(a) Strict complete source trajectory, (b) required state-prefix content and (c) termination, on 512 fixed source probes per world with the first operation supplied. A/B are averaged within world; bands are 95\% intervals over eight worlds.}\label{fig:v5-seq-retention}
\end{figure}

\begin{figure}[!htb]\centering
\includegraphics[width=\linewidth]{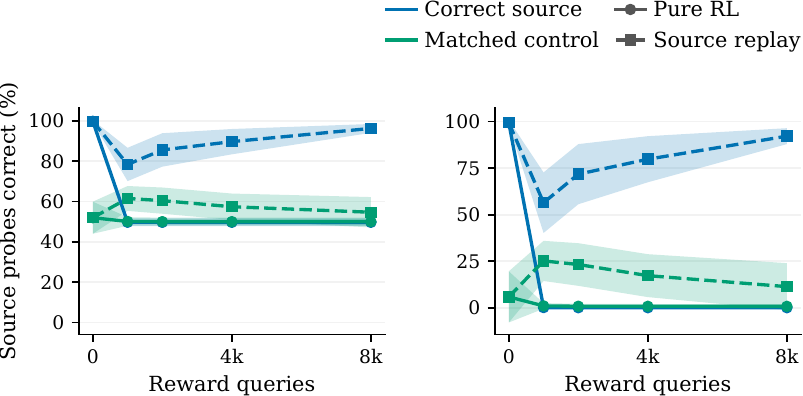}
\caption{(a) Correct first field on 256 source records and (b) joint correctness on 256 original/relevant-flip pairs. A/B are averaged within each of eight worlds; bands are untruncated 95\% intervals. Pure/replay branches share ancestors; replay contributes 131,072 additional source targets.}\label{fig:v5-memory-retention}
\end{figure}

\begin{figure}[!htb]\centering
\includegraphics[width=\linewidth]{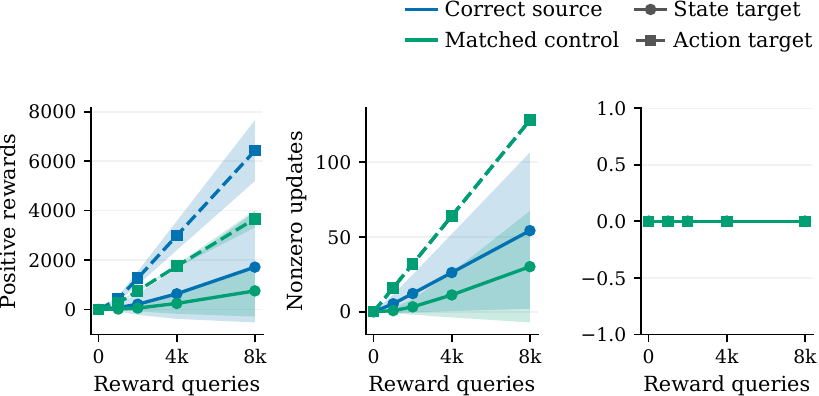}
\caption{Sequential training diagnostics: (a) positive training rewards, (b) nonzero RL updates and (c) nonzero source CE updates (zero for these routes). Cumulative counts average A/B within world, then eight worlds, with 95\% intervals. Logged calls are distinguished from measured nonzero updates.}\label{fig:v5-seq-updates}
\end{figure}

\begin{figure}[!htb]\centering
\includegraphics[width=\linewidth]{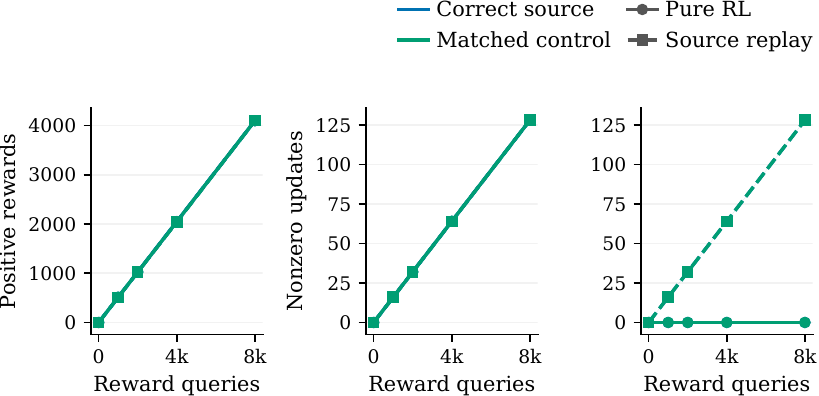}
\caption{Memory training diagnostics: (a) positive training rewards, (b) nonzero RL updates and (c) nonzero source CE updates. A/B are averaged within world, with 95\% intervals over eight worlds. CE calls do not increment the verifier-query count.}\label{fig:v5-memory-updates}
\end{figure}

\FloatBarrier

\begin{figure}[!htb]\centering
\includegraphics[width=\linewidth]{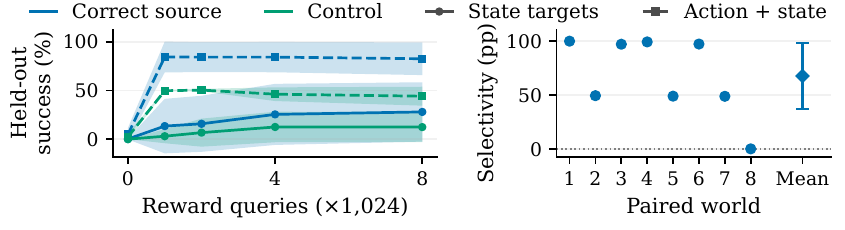}
\\[-1pt]\makebox[\linewidth]{\makebox[.55\linewidth]{(a) Reward learning}\makebox[.45\linewidth]{(b) Source-selectivity effect}}
\caption{Complete original Sequential comparison. (a) State-only and action-plus-state source objectives, correct and private-random source, on 1,024 fixed held-out records. (b) Action-objective source-selectivity effects for all eight worlds, with their mean and 95\% interval. A/B are averaged within world. World-level results correspond to \cref{tab:v5-primary,tab:v5-secondary}.}
\label{fig:v5-sequential-detail}\end{figure}

\FloatBarrier
\subsection{Independent Memory confirmation: retrieval that rewards can use}
\label{app:memory-confirmation}
We test task-relevant retrieval on eight new worlds, 508--515, using Qwen2.5-0.5B and the discrete Memory representation. Each world has four reward branches: selected/other retrieval crossed with opposite tasks A/B. The two interventions start from the same correct-source checkpoint and receive matched additional training; A/B then branch from the same intervention endpoint. The experiment asks whether preparing the relevant retrieval computation supports reward-task selection under a fixed adaptation recipe.

\paragraph{Source interventions.}
After format preparation and $2^{24}$ correct-source tokens, each route observes 131,072 additional records in 2,048 batches of 64. The selected route predicts the sign of the sum of the 15 content coordinates in the task-relevant memory slot; the other route predicts the same public projection of the other slot. Prompts are identical across routes. These targets depend on the retrieval mechanism, without the downstream contextual signs or cue-rule coefficients. Both routes therefore inherit correct source information; the comparison changes which retrieval computation the additional stage trains.

A second source stage uses 65,536 record exposures in 1,024 batches of 64, adding a retrieval-vector objective to next-token CE. For context $c\in\{1,\ldots,4\}$, the target $z\in\mathbb R^{60}$ places the retrieved 15-coordinate value in block $c$ and sets the other blocks to zero. The route is inferred from observed source input/bit pairs. With final-prefix hidden state $h$, a fixed, data-independent $60\times896$ orthonormal-row projection $P$ (NumPy seed 970506) predicts $P(h/\operatorname{RMS}(h))$; the auxiliary loss is $\|P(h/\operatorname{RMS}(h))-z\|^2/15$, averaged over records and added to CE with coefficient one. Here $\operatorname{RMS}(h)=\sqrt{\max(\|h\|^2/d,10^{-12})}$. The projection is a training buffer, absent from generated answers. These stages train all model parameters at nominal learning rate $10^{-5}$; the vector stage uses the step safeguard below.

\paragraph{Reward adaptation and representation retention.}
Each branch uses 8,192 fresh on-policy reward queries: 128 batches of 64, REINFORCE baseline $1/2$, and nominal Adam learning rate $3\times10^{-6}$. Sampling and scoring use the same full-vocabulary autoregressive policy. After each reward update, 256 previously observed prefixes supply the same retrieval-vector loss and a syntax objective. Each prefix is paired with both supplied signs; CE supervises only the three public suffix tokens after each sign. Thus this step uses 1,536 CE targets and 15,360 vector components, without CE supervision of the answer sign. Adam moments are reset at stage boundaries and shared by reward and retention updates within each branch.

At the intervention endpoint, temperature is calibrated using 256 unlabeled task-training prompts to mean conditional sign entropy 0.5 bit, then fixed throughout adaptation. If the original next-token probabilities of the two sign tokens are $p_+,p_-$, write $m=p_++p_-$ and $d=\log(p_+/p_-)$. The processed probabilities are
\[
 p'_+=m\,\sigma(d/T),\qquad p'_-=m\,[1-\sigma(d/T)],
\]
with all other token probabilities unchanged. Only the first sign position is transformed. The temperatures range from 3.7381 to 4.8753; training samples, sequence log probabilities and greedy evaluation use the same processing. Calibration uses no reward labels or held-out examples.

For each guarded Adam proposal, the displacement is scaled by the first accepted value in $\{1,10^{-1},10^{-2},10^{-3},10^{-4},0\}$. Acceptance requires finite values, loss at most its pre-update value plus $10^{-7}$, and mean first-position full-vocabulary $D_{\rm KL}(\pi_{\rm old}\|\pi_{\rm new})\le .01$ on the update prompts. The guard scales parameter displacement; Adam's updated moments are retained. Every branch makes 128 actual reward updates; 121--128 of its 128 retention attempts change parameters. These settings define the complete recipe used for both source interventions.

\paragraph{Budgets.}
The shared warm start uses 16,777,216 observed source tokens. Per intervention, the public-projection stage adds approximately 20.56--20.60 million observed tokens in 131,072 records; the vector stage adds 65,536 record exposures. Per reward branch, retention adds 196,608 CE targets and 1,966,080 vector components in 128 attempts, separately from 8,192 reward queries. The 32 branches use 262,144 new training queries in total. These costs are additional to, and separately recorded from, the original experiment in \cref{app:v5-results}. The contrast evaluates task-relevant versus alternative retrieval under the complete recipe; individual components are not separately ablated.

\paragraph{Evaluation and paired inference.}
Each branch is evaluated greedily at 0, 1,024, 2,048, 4,096 and 8,192 reward queries on 1,024 fixed held-out records and 256 separate training-distribution records. Both task verifiers score each output. A/B scores are averaged within world before taking world means or differences; the eight worlds are the independent units. The single predeclared primary effect is selected-minus-other task selectivity $\Delta_{\rm task}$ as defined in \cref{sec:experiments}. Its exact two-sided sign-flip test enumerates all $2^8$ sign assignments and remains separate from the earlier two-test Holm family. Intervals use Student's $t$ with seven degrees of freedom.

Selected retrieval reaches 75.32\% held-out success, compared with 49.86\% for other retrieval; both routes enter reward training at 50.00\% after averaging A/B within world. The paired accuracy difference is 25.46 pp (95\% CI 20.10--30.81). All eight world differences are positive. The primary selectivity difference is 50.92 pp (95\% CI 40.20--61.63), exact $p=2/256=.0078125$. All 32 branches have 100\% format accuracy on their final evaluation records. With valid binary answers and opposite task labels, $J_{AB}=1-J_{AA}$ and $J_{BA}=1-J_{BB}$, so the selectivity difference is exactly twice the accuracy difference. \Cref{fig:memory-confirmation-detail,tab:memory-confirmation-worlds} display every world and both task branches.

\begin{figure}[!htb]\centering
\includegraphics[width=\linewidth]{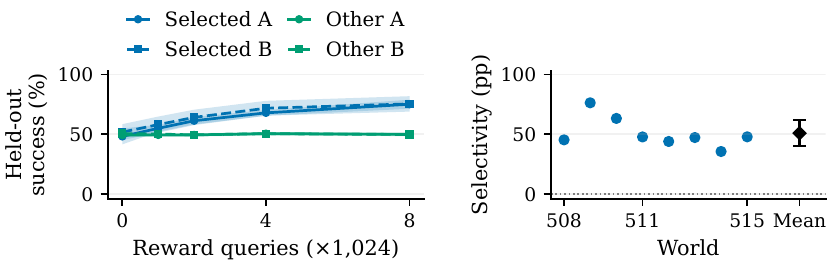}
\input{arxiv/experiments/generated/memory_confirmation_detail_labels.tex}
\caption{Independent Memory confirmation. (a) Separate task-A and task-B success curves for selected and other retrieval; means and 95\% Student-$t$ bands over eight worlds, using 1,024 held-out records per branch. (b) Each point is a world's selected-minus-other task-selectivity effect after averaging A/B; black diamond and error bar give the mean and 95\% interval. All fixed evaluation points and all worlds are retained.}
\label{fig:memory-confirmation-detail}\end{figure}
\begin{table}[!htb]\centering\small
\caption{Independent Memory confirmation: all 32 branch endpoints after 8,192 reward queries. Success is a percentage of 1,024 fixed held-out records per branch. Differences first average A/B within world; all eight worlds are included.}\label{tab:memory-confirmation-worlds}
\begin{tabular*}{\linewidth}{@{\extracolsep{\fill}}rrrrrrr@{}}\toprule
World & Selected A & Selected B & Other A & Other B & Accuracy $\Delta$ & Selectivity $\Delta$\\\midrule
508 & 73.34 & 75.10 & 52.44 & 50.68 & 22.66 & 45.31\\
509 & 83.69 & 91.31 & 49.32 & 49.32 & 38.18 & 76.37\\
510 & 80.66 & 81.84 & 49.80 & 49.41 & 31.64 & 63.28\\
511 & 70.90 & 73.05 & 47.56 & 48.63 & 23.88 & 47.75\\
512 & 73.54 & 71.09 & 50.10 & 50.59 & 21.97 & 43.95\\
513 & 73.83 & 72.27 & 48.83 & 50.00 & 23.63 & 47.27\\
514 & 73.83 & 66.31 & 52.54 & 52.05 & 17.77 & 35.55\\
515 & 71.68 & 72.66 & 48.73 & 47.75 & 23.93 & 47.85\\
\bottomrule\end{tabular*}
\end{table}

\FloatBarrier
\subsection{Independent Sequential parameter-plane measurement}
\label{app:parameter-plane}
This descriptive comparison uses the prespecified world 508 and Task A, separately from the formal worlds 500--507. Three routes share a Qwen2.5-0.5B checkpoint after 128 task-independent format updates. Direct RL skips added source training. Correct source and the public-preserving/private-random control each use $2^{22}$ observed tokens and 512 source updates, predicting the record's first operation, subsequent states and END/EOS. The input initial state and suffix operations are not supervised targets under this objective. Each reward branch uses learning rate $3\times10^{-6}$, batch size 64 and 128 updates, totaling 8,192 verifier queries. The three branches make 128 nonzero updates each and receive 3,874, 7,668 and 3,786 positive training rewards, respectively. Additional source cost is separate from the common inherited checkpoint and equal reward budgets.

\paragraph{Actual learning and evaluation units.}
\Cref{fig:parameter-behavior} reports greedy strict success on 1,024 fixed held-out records, balanced between suffix lengths four and six, at 0, 1,024, 2,048, 4,096 and 8,192 reward queries. Source-endpoint task success is distinct from a source-computation probe: the correct-source model has 0\% task success before RL and 100\% after 1,024 queries. Final advantages are 48.24 percentage points over direct RL and 48.73 over the private-random source control. The bands use 2,000 item-bootstrap resamples stratified by suffix length, with shared indices across routes and checkpoints. These intervals describe item uncertainty conditional on the prespecified world and trained run; they collapse at all-correct or all-incorrect endpoints. This single-world comparison is analyzed separately from the eight-world tests.

\begin{figure}[!htbp]
\centering
\includegraphics[width=\linewidth]{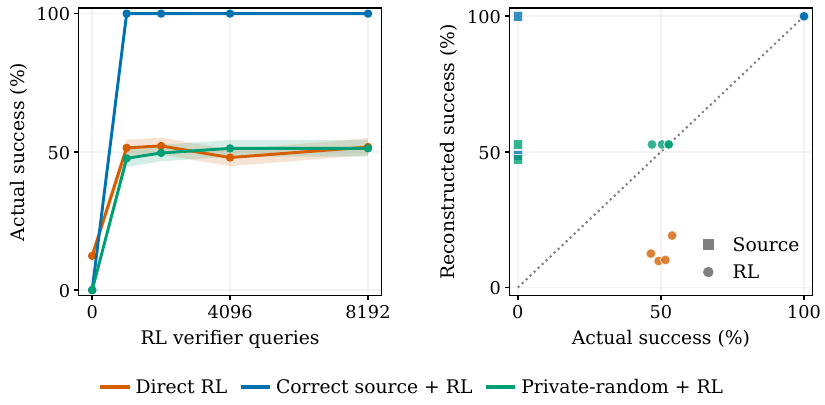}
\caption{Actual reward learning and projection fidelity in world 508. Left: original-checkpoint success on 1,024 fixed records; bands are conditional 95\% item-bootstrap intervals. Right: original versus reconstructed checkpoint success on the same 256 slice-evaluation inputs. Squares mark source checkpoints and circles mark RL checkpoints; overlapping points are retained. The diagonal denotes equal scores. The source routes receive $2^{22}$ additional tokens before RL; direct RL skips this stage. All three routes share format preparation and use 8,192 reward queries.}
\label{fig:parameter-behavior}
\end{figure}

\paragraph{One common plane and complete measurement grid.}
Let $\vartheta_0$ denote the shared format checkpoint. We use 17 distinct checkpoints: the origin, both source routes at $2^{20}$ and $2^{22}$ tokens, and all three reward routes at 1,024, 2,048, 4,096 and 8,192 queries. Shared source-to-reward boundaries use the same weights and are counted once. An uncentered rank-two SVD of displacements $\vartheta_i-\vartheta_0$ across all these states gives orthonormal directions $v_1,v_2$. A checkpoint is projected to coordinates $\alpha_{ij}=\langle\vartheta_i-\vartheta_0,v_j\rangle$ and reconstructed as $\widetilde\vartheta_i=\vartheta_0+\alpha_{i1}v_1+\alpha_{i2}v_2$. Grid models use the same affine plane and are cast to the evaluation model's dtype.

We evaluate the 81 points of the fixed $9\times9$ grid, all 17 original checkpoints and their 17 reconstructions. Every measurement uses the same 256 held-out inputs (128 of each suffix length) and 128 fixed correct-source records. Source loss is the summed negative log-likelihood over 3,856 supervised first-operation/state/END/EOS targets divided by that target count, in nats per target. The colored cells in \cref{fig:parameter-plane} are measured grid values; only the white source-loss contours interpolate between grid values.

\paragraph{Geometric coverage and behavioral fidelity.}
The plane retains 91.96\% of $\sum_i\|\vartheta_i-\vartheta_0\|^2$. For the direct-RL endpoint alone, the relative residual $\|\vartheta_i-\widetilde\vartheta_i\|/\|\vartheta_i-\vartheta_0\|$ is 89.63\%, leaving only 19.67\% of its squared displacement in the plane. Aggregate coverage thus masks route-wise reconstruction loss. On the common 256 inputs, actual/reconstructed success is 0.00\%/100.00\% at the correct-source endpoint, 0.00\%/52.73\% at the private-random source endpoint, and 53.91\%/19.14\% at the direct-RL endpoint. The two source-plus-RL endpoints agree at 100.00\% and 52.73\%, respectively. These denominators differ from the full 1,024-record learning curves.

Because the basis includes later RL checkpoints, reconstructing a source checkpoint introduces directions learned during reward adaptation. The grid describes the behavior of reconstructed models in this shared plane; the original-checkpoint curves measure the actual training behavior. Together they relate the measured slice to learning on task A, while the paired A/B experiments measure task selectivity.

\FloatBarrier
\subsection{Eight-world source acquisition and reward adaptation}\label{app:v4-results}
This comparison separates learning a source computation from using it under a new reward. Each module uses eight independently generated worlds (seeds 300--307), two source routes and two reward tasks A/B per route. Sequential uses its original representation, comparing the correct source with a source that preserves every public state and randomizes the private bits. Memory uses discrete signed-permutation keys and $\pm1$ values, comparing correct retrieval with wrong retrieval under the same all-token objective. Inputs and source exposure are matched within each world; A/B branches share the same source checkpoint.

\paragraph{Training and evaluation.}
All runs use Qwen2.5-0.5B Base and full-parameter Adam. A task-independent format stage precedes source training: 128 updates for Sequential and 32 for Memory, batch 64, learning rate $10^{-5}$. Sequential then observes $2^{22}$ source prediction tokens in 512 updates, optimizing state tokens and the existing END/EOS targets; Memory observes $2^{24}$ tokens in 2,048 updates, optimizing all tokens. Source learning rate is $10^{-5}$ for both. No subsequent format stage is applied. Each source checkpoint branches directly into tasks A/B, with fresh Adam moments, fixed learning rate $3\times10^{-6}$ and 8,192 verifier queries (128 batches of 64 sampled completions). The strict reward and output parser are unchanged. Greedy evaluation uses 1,024 fixed held-out prompts at 0, 1,024, 2,048, 4,096 and 8,192 queries; both task verifiers score every output. All 64 reward branches finish the full budget, totaling 524,288 training queries.

\paragraph{Source computation.}
The source-stage diagnostics in \cref{fig:v4-acquisition} use 512 Sequential prompts per world, with the first operation given and horizons five and seven equally represented. A complete answer contains all correct states and END in exactly the required positions. Memory uses 256 original source records and 256 corresponding relevant-value interventions per world; a pair is correct only when both first-field answers are correct. These are source-distribution probes without the downstream task binding. Across eight worlds, correct minus control differences are $87.50$ percentage points for Sequential complete trajectories (95\% paired $t$ interval $[62.89,112.11]$), $44.48$ for Memory original records ($[31.92,57.05]$), and $88.28$ for Memory intervention pairs ($[60.57,115.99]$). These source diagnostics are distinct from the two primary reward-task comparisons below. The $t$ intervals are untruncated; their endpoints can exceed the bounded metric range.

\begin{figure}[!htb]\centering
\includegraphics[width=\linewidth]{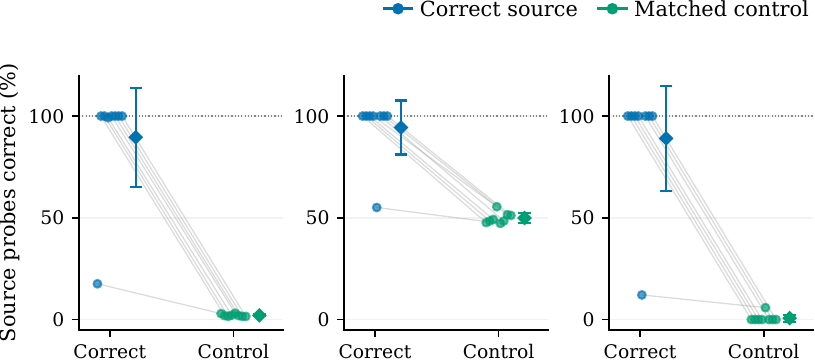}
\caption{Historical source acquisition, worlds 300--307. (a) Sequential complete trajectories on 512 probes with first operation supplied; (b) Memory first-field correctness and (c) original/value-flip pair correctness, each on 256 records or pairs. Correct source and matched structural controls are paired within world; points are worlds, diamonds means with untruncated 95\% Student-$t$ intervals. This protocol is separate from worlds 500--507 in \cref{app:v5-results}.}\label{fig:v4-acquisition}
\end{figure}

\paragraph{Task selection.}
For each route and world, selectivity is $\tfrac12[(J_{AA}-J_{AB})+(J_{BB}-J_{BA})]$, where $J_{XY}$ is the success of the task-$X$ branch under verifier $Y$. The two primary effects subtract the matched-control selectivity from correct-source selectivity at 8,192 queries. We average eight world-level differences, use Student-$t$ intervals with seven degrees of freedom, enumerate all $2^8$ sign flips for two-sided tests, and apply Holm correction to these two hypotheses. This two-test family is separate from the earlier comparisons. Neither interval excludes zero (\cref{tab:v4-effects,fig:v4-selectivity}). Sequential world effects are $[0,0,50,0,-1.123,0,0,-1.172]$ percentage points, so its positive mean is concentrated in one world. Full learning curves appear in \cref{fig:v4-feedback}.
\begin{table}[!htbp]\centering\small
\caption{Correct-source minus matched-control task selectivity after 8,192 reward queries, eight worlds per module. Effects and 95\% paired Student-$t$ intervals are in percentage points; $p$ is the exact two-sided sign-flip value and $p_H$ applies Holm correction to the two comparisons.}
\label{tab:v4-effects}
\begin{tabular}{lrrrr}\toprule
Module & Effect & 95\% interval & $p$ & $p_H$ \\\midrule
Sequential & $5.96$ & $[-8.92, 20.85]$ & $1.000$ & $1.000$ \\
Memory & $-0.87$ & $[-2.10, 0.35]$ & $0.125$ & $0.250$ \\
\bottomrule\end{tabular}\end{table}

\begin{figure}[t]\centering
\makebox[\linewidth]{\makebox[.5\linewidth]{(a) Sequential}\makebox[.5\linewidth]{(b) Memory}}\\[-2pt]
\includegraphics[width=\linewidth]{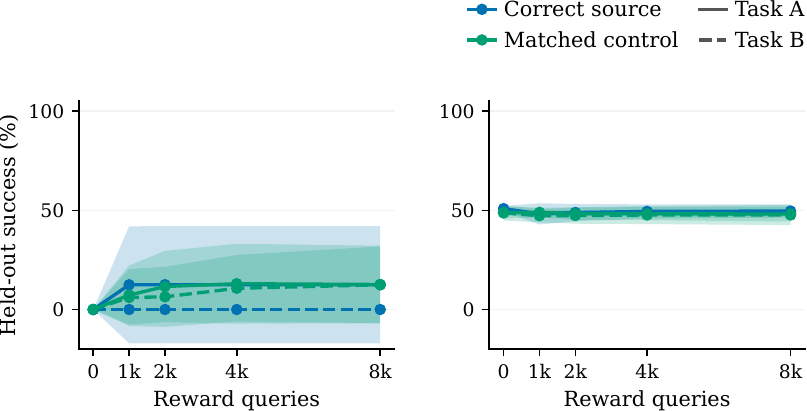}
\caption{Held-out success during reward adaptation, eight worlds per route. Colors distinguish correct source and the matched control (random private states for Sequential; wrong retrieval for Memory); solid and dashed lines denote tasks A and B. Markers are world means on 1,024 prompts per world at each query budget, with 95\% Student-$t$ bands. Training samples at temperature one; evaluation is greedy. Intervals are shown without clipping to the success-rate range.}
\label{fig:v4-feedback}\end{figure}

\begin{figure}[t]\centering
\makebox[\linewidth]{\makebox[.5\linewidth]{(a) Sequential}\makebox[.5\linewidth]{(b) Memory}}\\[-2pt]
\includegraphics[width=\linewidth]{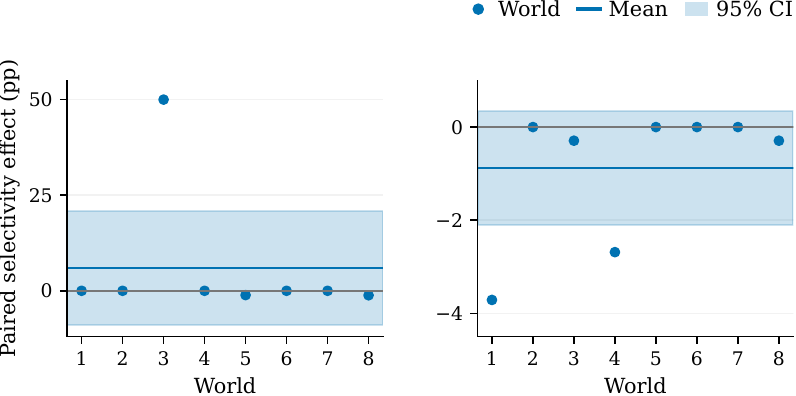}
\caption{Correct-source minus matched-control task selectivity at 8,192 queries. Each dot is one complete world-level A/B contrast; worlds 1--8 correspond to seeds 300--307. The blue line is their mean, the shaded band its 95\% Student-$t$ interval, and the gray line marks zero. Units are percentage points. Both comparisons include all eight worlds; vertical scales differ between panels.}
\label{fig:v4-selectivity}\end{figure}

\paragraph{Computation during reward adaptation.}
\Cref{fig:v4-retention} repeats the fixed source probes through the reward stage. A/B probe scores are averaged within each world before forming the eight-world mean and interval. Correct-source Sequential complete trajectories change from 89.60\% to 88.94\%, with END placement at 100\% throughout. Twenty-six of the 32 Sequential reward branches receive no positive reward and make no parameter updates, so most branches retain the source checkpoint unchanged. Correct-source Memory first-field accuracy changes from 94.38\% to 50.34\% and relevant-pair accuracy from 89.01\% to 0\%, while its source-probe public-format accuracy stays at 100\%. All 32 Memory branches make 128 nonzero updates. Thus computation acquired on source inputs and its expression after reward adaptation are separately observable properties.

\begin{figure}[t]\centering
\makebox[\linewidth]{\makebox[.333\linewidth]{(a) Sequential trajectory}\makebox[.333\linewidth]{(b) Memory record}\makebox[.333\linewidth]{(c) Memory flip pair}}\\[-2pt]
\includegraphics[width=\linewidth]{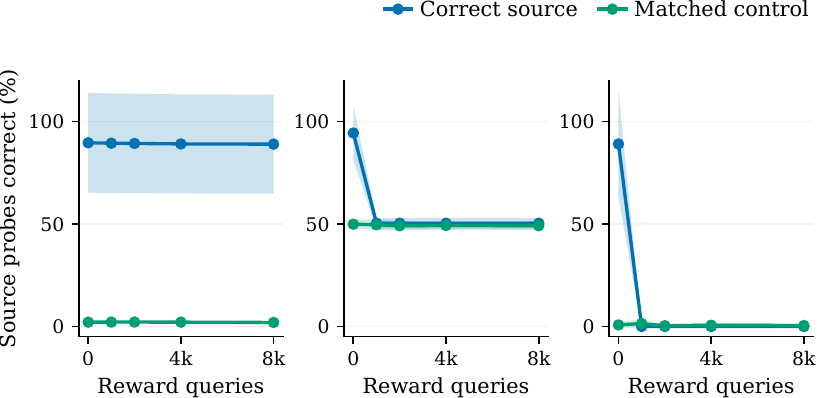}
\caption{Source-computation probes through reward adaptation, using the same metrics and matched controls as \cref{fig:v4-acquisition}. Lines join the five measured query budgets; shaded bands are 95\% Student-$t$ intervals over eight worlds after averaging A/B within each world. Shared ancestors are not counted as additional independent samples. Sequential probes supply the first operation; Memory pairs require both the original and relevant-value-flipped bit to be correct.}
\label{fig:v4-retention}\end{figure}
\FloatBarrier

The following subsections describe the earlier controlled-task and public-task comparisons; the eight-world source-acquisition design above has its own source objectives and budgets.

\subsection{Earlier execution, task-selection and interface comparisons}\label{app:historical-interpretation}
These earlier comparisons identify which part of learning changes with source preparation. They use their own objectives, budgets and checkpoints, described below.

\paragraph{Execution and task choice.}
The endpoint decomposition separates a valid public trajectory from its private answer, while opposite-reward branches test task-specific selection (\cref{fig:revision-sequential,app:historical-evaluation,app:task-choice}). This distinguishes learning to execute from learning which execution the task requires.

\begin{figure}[!htbp]\centering
\includegraphics[width=\linewidth]{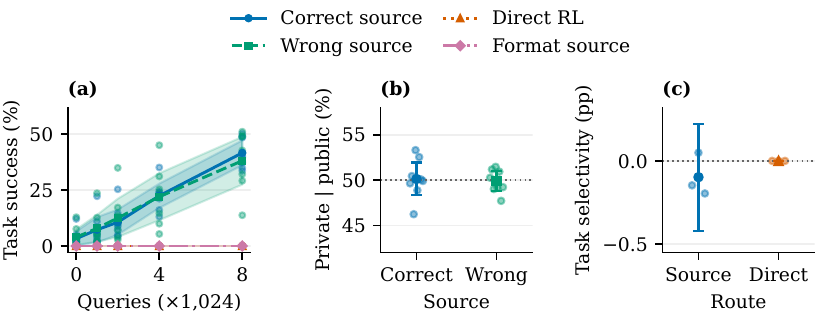}
\caption{Historical Sequential execution and private-task adaptation, using $2^{20}$ source prediction tokens. (a) Greedy success on 1,024 held-out prompts per world: eight worlds for correct source, wrong source and direct RL, and three for format control. (b) Final private-bit accuracy among outputs with valid public execution and the correct public endpoint, computed within each world before averaging; direct and format routes have no eligible outputs. (c) Shared-source A/B branches in three worlds: paired task selectivity $\Delta_{\rm task}$ in percentage points (pp). Small translucent dots show individual worlds; connected markers in (a) and large markers in (b,c) show world means. Shading and error bars are 95\% Student-$t$ intervals across worlds; horizontal jitter separates overlapping points. Dotted lines mark 50\% private accuracy in (b) and zero task selectivity in (c).}
\label{fig:revision-sequential}\label{fig:revision-task-choice}\end{figure}

\paragraph{Acquisition and stage order.}
Memory source probes test the retrieved-value rule directly (\cref{app:memory-diagnostic}). The separate two-world study measures state content, termination and value-flip responses before and after format training (\cref{app:v2-results}). Together these diagnostics locate the computation available at each stage.

\paragraph{Initialization, budgets and learning routes.}
\Cref{app:additional-comparisons} compares source dose, inherited versus random initialization, model scale, continued prediction, answer supervision and test-time selection. \Cref{app:historical-evaluation} gives the output-level decomposition and sampled-decoding comparison, with the full denominators and uncertainty.

\subsection{Task definitions and observation permissions}
For sequential execution, a state is $(s,z)\in\{0,1\}\times\mathbb F_2^3$, and an operation $(b,u)$ maps it to $(s\oplus b\oplus\theta(u,z),z\oplus u)$. Here $\oplus$ is addition modulo two. Writing $u=(t,x,y)$, let $a(u)=t\oplus xy$. Displayed suffixes have length $L\in\{4,6\}$; the actual execution horizon is $H=L+1$, including the hidden initial operation. Training and validation draw independently from suffixes with $\bigoplus_j a(u_j)=0$; held-out suffixes have parity one. The paired mechanism replaces $\theta(u,z)$ by $\theta(u,z)\oplus a(u)$ and compensates the hidden initial binding. The terminal training-reward function is unchanged. Evaluation has 512 prompts at each length.

A source record serializes the initial state, the operation suffix, the independently uniform first operation, and the resulting states. At each transition a reset occurs with probability $(1/64)/H$, replacing the state uniformly. Every source token is a next-token prediction target, including random fields and EOS. In textual outputs, END is an explicit field terminator, distinct from the tokenizer's end-of-sequence marker. The verifier checks output syntax, the public $z$ trajectory, and the final state; correct intermediate answer bits are not required. Source data do not contain the downstream hidden binding.

The contextual memory experiment uses four hidden orientations and a 15-dimensional shared cue rule, as in the contextual model. Source records contain random keys, a transformed query, Gaussian values, and the raw future bit, with no task-dependent signs. Target memory amplitudes separate training from four equally weighted test configurations $(\pm1,\pm0.875)$. Equal-sign and opposite-sign subsets are reported separately. Language-model inputs use textual KEY, QUERY, VALUE, and BIT fields, with numeric values rendered to eight significant digits. The language-model objective predicts all serialized fields; the contextual theorem trains on the future-bit position of typed numerical input.

The information panel uses eight paired worlds and nested source record counts $0,1,2,4,8,16,32,64$. The contextual curve $\tfrac14(\tfrac34)^N$ is the analytic expected error floor; the measured coverage statistic is the number of contexts actually observed. The paired-world construction is verified by enumeration.

\subsection{Models, training and evaluation}
The models are Qwen2.5-0.5B Base and Qwen2.5-1.5B Base, with all parameters trainable; \cref{tab:experiment-design} lists the comparisons and their independent seeds. Source training uses 8,192 prediction targets per update, with saved states at $2^{18},2^{19},2^{20}$ targets. Feedback uses one fresh batch of 64 sampled completions per REINFORCE update, for 8,192 verifications in total. Its loss is the negative batch mean of reward times the sum of sampled completion log probabilities, including EOS, with baseline zero; zero-reward batches still advance Adam moments. There is no reward filtering, Kullback--Leibler (KL) penalty, weight decay or gradient clipping. Parameters and Adam moments are stored in bfloat16 with 32-bit residual, normalization and probability computations; sampled completions are produced by SGLang, with trainer-side response log probabilities recomputed for the update. Contexts are limited to 2,048 tokens and continuations to 128 new tokens. Sampling uses temperature one without top-$k$ or top-$p$ truncation. Feedback begins from the source parameters with reset moments.

Learning rates are selected on two pilot worlds disjoint from the evaluated seeds, by source validation cross-entropy over $\{10^{-5},3\cdot10^{-5},10^{-4}\}$ and then by validation feedback on the allowed training support over $\{10^{-6},3\cdot10^{-6},10^{-5}\}$, with smaller rates breaking ties. The selected rates (\cref{tab:revision-recipes}) are shared by all routes within a task, model and initialization. Evaluation occurs at $0,1024,2048,4096,8192$ training queries. Each checkpoint has 1,024 fixed paired test prompts, evaluated with both greedy and sampled decoding; raw outputs, success, public-path and format errors, and truncation are saved. BBH uses every item from the two three-object tasks, a zero-shot option-only prompt, greedy decoding, and a strict option parser at the base, source, and final reward checkpoints.

\begin{table}[!htbp]
\centering\small
\caption{Experimental comparisons. Source-only measurements reuse source checkpoints; the fixed external base model contributes one checkpoint.}
\label{tab:experiment-design}
\begin{tabular}{@{}p{.51\linewidth}p{.40\linewidth}@{}}
\toprule
Comparison & Independent seeds and scope\\
\midrule
Direct RL, correct source, paired wrong source & 8 sequential and 5 contextual, 0.5B\\
Random initialization and 1.5B transfer & 3 sequential seeds per three-arm panel\\
Source dose & 5 sequential seeds, 0.5B, shared source\\
Alternative learning routes & 3 sequential LM seeds\\
Alternate downstream task choice & 3 sequential seeds, shared source states\\
Format-only source then RL & 3 sequential seeds\\
BBH checkpoint evaluation & Base, source, feedback; 3 source/feedback seeds\\
\bottomrule
\end{tabular}
\end{table}
\begin{table}[!htbp]\centering\small\setlength{\tabcolsep}{4pt}
\caption{Adam learning rates used in the historical controlled and public-task comparisons, selected on pilot worlds by source validation cross-entropy and training-support validation feedback. Source and reward stages reset optimizer moments.}
\label{tab:revision-recipes}
\begin{tabular}{lrr}
\toprule
Task / model / initialization & Source rate & Reward rate \\
\midrule
Contextual memory / 0.5B / pretrained & $3\times10^{-5}$ & $10^{-5}$ \\
\midrule
Sequential / 0.5B / pretrained & $3\times10^{-5}$ & $10^{-6}$ \\
Sequential / 0.5B / random & $10^{-4}$ & $10^{-6}$ \\
Sequential / 1.5B / pretrained & $3\times10^{-5}$ & $10^{-6}$ \\
\midrule
GSM8K / 1.5B / pretrained & $10^{-5}$ & $10^{-5}$ \\
HotpotQA / 1.5B / pretrained & $3\times10^{-5}$ & $10^{-5}$ \\
\bottomrule
\end{tabular}
\end{table}

\subsection{Alternative routes and controls}\label{app:supplement}
Continued source prediction consumes the next $2^{20}$ targets with the source optimizer state continued. Interleaved source updates add one 1,024-target source cross-entropy update after each reward update, drawing only from observed source records. Answer-only SFT generates its own prefix and supervises just the terminal answer and END; it receives no expert intermediate states, and answer labels are counted separately from verifications. Best-of-$N$ reuses nested sets of $N=1,2,4,8$ candidates from a fixed source model with test-time verification. The classical comparator estimates transitions from the same complete observed source records and searches the hidden binding with allowed training feedback. Source balancing changes sampling over initial public contexts at the same token budget.

\paragraph{Format exposure without a shared transition rule.}\label{app:format-control}
The format control uses three sequential Qwen2.5-0.5B worlds. It reuses the source input law: the initial state, suffix, independent first control, vocabulary, field positions, sequence lengths and END marker. Every output-state token is instead an independent uniform element of the legal state space. The generator reads neither the target transition table nor the downstream binding. Thus it preserves the format and the state marginal distribution while removing any persistent state--operation relation. The mechanism-sensitive contrast is $G_{\mathrm{source}}-G_{\mathrm{format}}$.

\paragraph{Shared-source, opposite-task feedback branches.}\label{app:task-choice}
For three sequential worlds, write the hidden first operation as $\eta=(b,u)$. Task A uses the original $\eta$ and task B uses $(b\oplus1,u)$, with identical $\theta$. The source generator does not read $\eta$ and the displayed prompts omit it. Every public coordinate is unchanged, while the terminal private bit flips under the same suffix, so the two success sets are disjoint. Task-B direct RL and correct-source RL branches are paired with the task-A branches; correct-source branches share the exact source checkpoint. Both tasks' scoring functions are applied offline to the same outputs at zero and final queries; \cref{tab:revision-choice} reports the cross-scored rates and \cref{tab:revision-choice-denominators} the exact eligible counts $C=N_A+N_B$ behind them.

The paired selectivity is $-0.10$ percentage points with interval $[-0.42,0.22]$ (\cref{fig:revision-task-choice,tab:revision-choice,tab:revision-choice-denominators}). Together with the conditional private accuracy in \cref{app:historical-evaluation}, this identifies shared public execution without a measurable preference for the branch's reward task.

\paragraph{Memory-rule acquisition probes.}\label{app:memory-diagnostic}
Per-update answer frequencies locate input-independent behavior, and training label frequencies distinguish it from a majority-label task. Two matched source runs, with the same source token budget, learning rate and Adam schedule as the historical continuous-numeric Memory runs, are probed before and after source training: 32 fixed held-out source records each generate three input domains and three value interventions, giving 288 observations per checkpoint. The domains use the original random key frames, basis keys, or basis keys with discrete nuisance coordinates; the first two content coordinates retain the source parity rule. Negating the selected slot's first coordinate reverses the future bit, whereas the same intervention on the unselected slot preserves it. For every input we record the log-likelihoods of the two valid bit completions (including the common terminator), their difference, two-candidate accuracy, and loss by source field. A correct relevant-value response together with invariance to irrelevant values indicates rule acquisition.

\begin{table}[!htbp]\centering\small
\caption{Source-rule diagnostics from two matched memory source runs. Intervention triples share an original record. Fractions report correct cases over the displayed denominator; CE denotes cross-entropy.}
\label{tab:memory-source-diagnostics}
\begin{tabular}{lrr}\toprule
Measurement & Run A & Run B \\\midrule
Two-candidate completion accuracy & 46/96 & 49/96 \\
Original records correct & 14/32 & 17/32 \\
Prediction reverses on relevant-value flip & 1/32 & 0/32 \\
Original and relevant-flip pair both correct & 0/32 & 0/32 \\
\midrule
Query-field CE, before & 0.2045 & 0.2114 \\
Query-field CE, after & 0.0430 & 0.0397 \\
\midrule
Bit-field CE per token, after & 0.3483 & 0.3869 \\
\bottomrule\end{tabular}
\end{table}
Each group of 96 cases consists of 32 records and their relevant and irrelevant interventions. The answer field contains two predicted tokens; normalizing the two completion likelihoods gives binary cross-entropies of 0.6949 and 0.7705. The lower query-field loss shows learning of predictable source content, but the pair test shows that these models do not reliably express the composed retrieval-and-sign-product response. After feedback, all five correct-source runs produce a constant valid action, with 49.18\% greedy correctness on balanced task labels; wrong-source runs share this endpoint (\cref{tab:revision-memory}). Comparing source and feedback checkpoints separates predictable-field learning from input-dependent task responses.

\subsection{Uncertainty}
Means and 95\% Student-$t$ intervals are computed across world/training seed means, with individual seed values retained; paired differences use the same seeds. The sampled sequential correct-source-versus-direct and correct-source-versus-wrong-source endpoint contrasts additionally use exact two-sided sign-flip tests with Holm correction at 0.05 (\cref{tab:sampled-robustness}). The sign-flip tests assume sign symmetry of paired effects under the null. Holm adjustment uses a fixed family size of three, with $p=1$ assigned to the remaining slot; the adjusted values for the two reported contrasts are 0.0234 and 0.672. Secondary comparisons report effect sizes and intervals. Separately, 5,000 paired item-bootstrap resamples quantify uncertainty conditional on the trained models.

\subsection{Public-task adaptation and feedback}\label{app:public-experiments}
Public-task source midtraining uses domain-related text without downstream labels; it does not instantiate the exact rule-independence assumption of the theoretical source laws. Reported budgets cover additional source tokens and reward queries, excluding inherited pretraining. The public-task panel uses five paired Qwen2.5-1.5B seeds for each of GSM8K and HotpotQA, with three reward routes: direct, domain-source and shuffled-source learning. Each source trajectory predicts $2^{22}$ tokens and retains $2^{20}$ and $2^{22}$ checkpoints. Shuffling acts within each document's 64-token blocks, preserving the target-token multiset. Source and feedback learning rates use the same grids and selection rule as the controlled experiments, with two pilot seeds per task; selection uses source validation cross-entropy and training-distribution validation rewards.

The numeric answer parser compares final GSM8K answers; HotpotQA uses normalized exact match for reward and official answer F1 as an additional measurement. Mathematical continuations have at most 512 new tokens, and question-answering continuations at most 256. HotpotQA prompts are admitted by tokenized length at most 4,096; all 7,405 development candidates qualify, and a fixed 1,024-question subset is evaluated. All 1,319 GSM8K test questions are evaluated. Source article titles are disjoint from downstream training and evaluation article titles, and question and document identities, dataset versions and tokenization are fixed before training.

Reward queries are evaluated at $0,1024,2048,4096,8192$. The three intermediate points use 512 paired prompts; the pre-RL and final evaluations use the full 1,319 or 1,024 items, and learning curves use the common 512 prompts at every checkpoint. Each task also has three answer-only SFT seeds initialized from the corresponding domain-source state, supervising only the answer and EOS with 8,192 answer labels and the same update batch. Fixed-model best-of-16 uses the same 256 test prompts for base and domain-source models. Public transfer uses 100 GSM-Symbolic templates with ten fixed variants each, and every item in the two three-object BBH tasks; symbolic variants are repeated observations of a template. Public results show every seed, means and 95\% Student-$t$ intervals over seed-level outcomes and paired gains; conditional item-bootstrap intervals preserve item pairing, and GSM-Symbolic resamples templates. \Cref{fig:revision-public} shows the entry scores, endpoints and paired gain differences summarized in \cref{tab:public-summary}, and \cref{fig:revision-public-curves} the greedy learning curves on the common 512 prompts, which locate where along the 8,192 queries each route's gain accrues.
\begin{figure}[!htbp]\centering
\includegraphics[width=\linewidth]{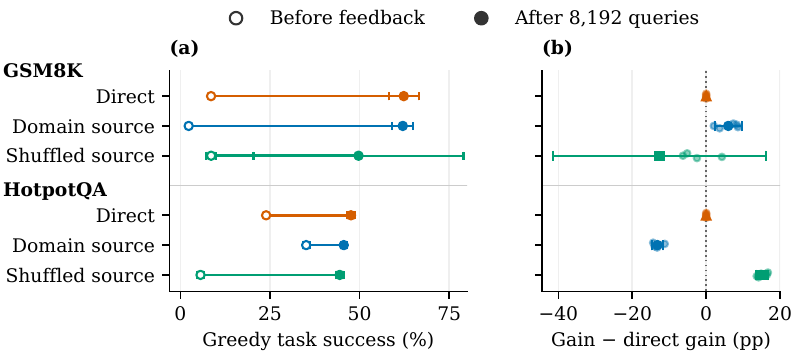}
\caption{Public-task starting scores and reward gains, five paired training seeds. (a) Open and filled markers show mean greedy success before and after RL; segments connect the same route. (b) Within-seed reward gain minus direct RL's gain, in percentage points: small translucent dots are seeds, large markers are means, and the dotted line is zero. Horizontal bars are 95\% Student-$t$ intervals. Each seed and route uses the same full endpoint sets: 1,319 GSM8K and 1,024 HotpotQA prompts. Direct starting scores evaluate one fixed pretrained checkpoint.}
\label{fig:revision-public}\end{figure}

\IfFileExists{experiments/generated/generated_public_appendix.tex}{\subsection{Secondary public-task metrics and external transfer}
Success, numeric correctness and F1 are higher-is-better; format errors and truncation are lower-is-better. Uncertainty describes training-seed variation.
\begingroup\small\setlength{\tabcolsep}{3pt}
\begin{longtable}{llllr}
\caption{Public-task secondary greedy measurements. Percentages apply to correctness, F1 and error rates; generated length is tokens. RL uses five seeds and answer-only SFT three. F1 is answer token-overlap F1; format and truncation rates use all generated answers. Entries are seed means with 95\% Student-$t$ half-widths.}\label{tab:public-secondary}\\
\toprule
Task & Method & Decoding & Metric & Mean and 95\% CI \\
\midrule\endfirsthead
\caption[]{Public-task secondary greedy measurements. Percentages apply to correctness, F1 and error rates; generated length is tokens. (continued)}\\
\toprule
Task & Method & Decoding & Metric & Mean and 95\% CI \\
\midrule\endhead
\midrule\multicolumn{5}{r}{Continued on next page}\\\endfoot
\bottomrule\endlastfoot
GSM8K & Direct RL & greedy & numeric correct & $62.3\,{\textstyle\pm 4.1}$ \\*
GSM8K & Direct RL & greedy & Format error & $0.5\,{\textstyle\pm 0.2}$ \\*
GSM8K & Direct RL & greedy & Length (tokens) & $108.1\,{\textstyle\pm 8.8}$ \\*
GSM8K & Direct RL & greedy & truncated & $0.3\,{\textstyle\pm 0.1}$ \\
\midrule
GSM8K & Correct source & greedy & numeric correct & $62.0\,{\textstyle\pm 2.9}$ \\*
GSM8K & Correct source & greedy & Format error & $2.2\,{\textstyle\pm 4.2}$ \\*
GSM8K & Correct source & greedy & Length (tokens) & $149.8\,{\textstyle\pm 101.5}$ \\*
GSM8K & Correct source & greedy & truncated & $1.6\,{\textstyle\pm 3.4}$ \\
\midrule
GSM8K & Shuffled source & greedy & numeric correct & $49.7\,{\textstyle\pm 29.3}$ \\*
GSM8K & Shuffled source & greedy & Format error & $17.8\,{\textstyle\pm 47.5}$ \\*
GSM8K & Shuffled source & greedy & Length (tokens) & $177.8\,{\textstyle\pm 210.9}$ \\*
GSM8K & Shuffled source & greedy & truncated & $17.7\,{\textstyle\pm 47.6}$ \\
\midrule
GSM8K & Answer SFT & greedy & numeric correct & $12.4\,{\textstyle\pm 0.4}$ \\*
GSM8K & Answer SFT & greedy & Format error & $0.0\,{\textstyle\pm 0.0}$ \\*
GSM8K & Answer SFT & greedy & Length (tokens) & $4.5\,{\textstyle\pm 0.0}$ \\*
GSM8K & Answer SFT & greedy & truncated & $0.0\,{\textstyle\pm 0.0}$ \\
\midrule
HotpotQA & Direct RL & greedy & Answer F1 & $60.4\,{\textstyle\pm 1.6}$ \\*
HotpotQA & Direct RL & greedy & Format error & $0.0\,{\textstyle\pm 0.0}$ \\*
HotpotQA & Direct RL & greedy & Length (tokens) & $5.3\,{\textstyle\pm 0.3}$ \\*
HotpotQA & Direct RL & greedy & truncated & $0.0\,{\textstyle\pm 0.0}$ \\
\midrule
HotpotQA & Correct source & greedy & Answer F1 & $58.4\,{\textstyle\pm 1.1}$ \\*
HotpotQA & Correct source & greedy & Format error & $0.0\,{\textstyle\pm 0.0}$ \\*
HotpotQA & Correct source & greedy & Length (tokens) & $5.6\,{\textstyle\pm 0.2}$ \\*
HotpotQA & Correct source & greedy & truncated & $0.0\,{\textstyle\pm 0.0}$ \\
\midrule
HotpotQA & Shuffled source & greedy & Answer F1 & $57.4\,{\textstyle\pm 1.4}$ \\*
HotpotQA & Shuffled source & greedy & Format error & $0.0\,{\textstyle\pm 0.0}$ \\*
HotpotQA & Shuffled source & greedy & Length (tokens) & $5.5\,{\textstyle\pm 0.4}$ \\*
HotpotQA & Shuffled source & greedy & truncated & $0.0\,{\textstyle\pm 0.0}$ \\
\midrule
HotpotQA & Answer SFT & greedy & Answer F1 & $69.0\,{\textstyle\pm 0.5}$ \\*
HotpotQA & Answer SFT & greedy & Format error & $0.0\,{\textstyle\pm 0.0}$ \\*
HotpotQA & Answer SFT & greedy & Length (tokens) & $4.9\,{\textstyle\pm 0.2}$ \\*
HotpotQA & Answer SFT & greedy & truncated & $0.0\,{\textstyle\pm 0.0}$ \\
\end{longtable}
\endgroup

\begingroup\small\setlength{\tabcolsep}{3pt}
\begin{longtable}{llllr}
\caption{HotpotQA greedy results within each indicated question-type subset. Success is normalized exact match; rates are percentages with seed-level 95\% Student-$t$ half-widths.}\label{tab:public-question-types}\\
\toprule
Method & Decoding & Type & Metric & Mean and 95\% CI \\
\midrule\endfirsthead
\caption[]{HotpotQA greedy results within each indicated question-type subset. Success is normalized exact match; rates are percentages with seed-level 95\% Student-$t$ half-widths. (continued)}\\
\toprule
Method & Decoding & Type & Metric & Mean and 95\% CI \\
\midrule\endhead
\midrule\multicolumn{5}{r}{Continued on next page}\\\endfoot
\bottomrule\endlastfoot
Direct RL & greedy & bridge & Answer F1 & $60.5\,{\textstyle\pm 1.1}$ \\*
Direct RL & greedy & bridge & success & $46.2\,{\textstyle\pm 0.4}$ \\*
Direct RL & greedy & comparison & Answer F1 & $60.2\,{\textstyle\pm 3.5}$ \\*
Direct RL & greedy & comparison & success & $52.9\,{\textstyle\pm 3.7}$ \\
\midrule
Correct source & greedy & bridge & Answer F1 & $58.3\,{\textstyle\pm 1.1}$ \\*
Correct source & greedy & bridge & success & $44.2\,{\textstyle\pm 0.9}$ \\*
Correct source & greedy & comparison & Answer F1 & $58.9\,{\textstyle\pm 2.4}$ \\*
Correct source & greedy & comparison & success & $51.2\,{\textstyle\pm 2.6}$ \\
\midrule
Shuffled source & greedy & bridge & Answer F1 & $57.3\,{\textstyle\pm 1.3}$ \\*
Shuffled source & greedy & bridge & success & $43.0\,{\textstyle\pm 1.0}$ \\*
Shuffled source & greedy & comparison & Answer F1 & $57.7\,{\textstyle\pm 2.4}$ \\*
Shuffled source & greedy & comparison & success & $50.0\,{\textstyle\pm 2.4}$ \\
\midrule
Answer SFT & greedy & bridge & Answer F1 & $70.0\,{\textstyle\pm 1.1}$ \\*
Answer SFT & greedy & bridge & success & $52.9\,{\textstyle\pm 2.2}$ \\*
Answer SFT & greedy & comparison & Answer F1 & $65.0\,{\textstyle\pm 3.2}$ \\*
Answer SFT & greedy & comparison & success & $56.9\,{\textstyle\pm 1.8}$ \\
\end{longtable}
\endgroup

\begingroup\small\setlength{\tabcolsep}{3pt}
\begin{longtable}{llllrr}
\caption{External-transfer and verification-selection results. Bo16 denotes best-of-16 selection with test-time verification. $\pm$ gives training-seed 95\% Student-$t$ half-widths; brackets give item/template-bootstrap intervals for a fixed model. BBH logic/tracking are three-object logical deduction/object tracking.}\label{tab:public-transfer}\\
\toprule
Training & Method & Evaluation & Checkpoint & Seeds & Success (\%) \\
\midrule\endfirsthead
\caption[]{External-transfer and verification-selection results. Bo16 denotes best-of-16 selection with test-time verification. $\pm$ gives training-seed 95\% Student-$t$ half-widths; brackets give item/template-bootstrap intervals for a fixed model. BBH logic/tracking are three-object logical deduction/object tracking. (continued)}\\
\toprule
Training & Method & Evaluation & Checkpoint & Seeds & Success (\%) \\
\midrule\endhead
\midrule\multicolumn{6}{r}{Continued on next page}\\\endfoot
\bottomrule\endlastfoot
GSM8K & Base/Bo16 & GSM8K & base & 1 & $44.5\,{\textstyle[38.3,50.8]}^\dagger$ \\
\midrule
GSM8K & Correct source/Bo16 & GSM8K & source & 5 & $22.1\,{\textstyle\pm 4.9}$ \\
\midrule
GSM8K & Direct RL & BBH logic & feedback & 5 & $40.1\,{\textstyle\pm 1.6}$ \\*
GSM8K & Direct RL & GSM-Symbolic & feedback & 5 & $54.5\,{\textstyle\pm 3.2}$ \\*
GSM8K & Direct RL & BBH tracking & feedback & 5 & $11.4\,{\textstyle\pm 17.6}$ \\
\midrule
GSM8K & Correct source & BBH logic & feedback & 5 & $30.9\,{\textstyle\pm 21.0}$ \\*
GSM8K & Correct source & GSM-Symbolic & feedback & 5 & $53.0\,{\textstyle\pm 3.5}$ \\*
GSM8K & Correct source & BBH tracking & feedback & 5 & $13.2\,{\textstyle\pm 22.3}$ \\*
GSM8K & Correct source & BBH logic & source & 5 & $40.7\,{\textstyle\pm 1.2}$ \\*
GSM8K & Correct source & GSM-Symbolic & source & 5 & $2.1\,{\textstyle\pm 0.4}$ \\*
GSM8K & Correct source & BBH tracking & source & 5 & $30.8\,{\textstyle\pm 3.3}$ \\
\midrule
GSM8K & Shuffled source & BBH logic & feedback & 5 & $27.1\,{\textstyle\pm 21.3}$ \\*
GSM8K & Shuffled source & GSM-Symbolic & feedback & 5 & $43.6\,{\textstyle\pm 26.5}$ \\*
GSM8K & Shuffled source & BBH tracking & feedback & 5 & $11.4\,{\textstyle\pm 18.5}$ \\*
GSM8K & Shuffled source & BBH logic & source & 5 & $39.5\,{\textstyle\pm 2.9}$ \\*
GSM8K & Shuffled source & GSM-Symbolic & source & 5 & $6.3\,{\textstyle\pm 0.8}$ \\*
GSM8K & Shuffled source & BBH tracking & source & 5 & $32.8\,{\textstyle\pm 2.6}$ \\
\midrule
HotpotQA & Base/Bo16 & HotpotQA & base & 1 & $45.7\,{\textstyle[39.5,52.0]}^\dagger$ \\
\midrule
HotpotQA & Correct source/Bo16 & HotpotQA & source & 5 & $50.7\,{\textstyle\pm 1.3}$ \\
\midrule
HotpotQA & Direct RL & BBH logic & feedback & 5 & $41.5\,{\textstyle\pm 1.2}$ \\*
HotpotQA & Direct RL & BBH tracking & feedback & 5 & $35.5\,{\textstyle\pm 0.8}$ \\
\midrule
HotpotQA & Correct source & BBH logic & feedback & 5 & $41.0\,{\textstyle\pm 1.6}$ \\*
HotpotQA & Correct source & BBH tracking & feedback & 5 & $36.1\,{\textstyle\pm 1.1}$ \\*
HotpotQA & Correct source & BBH logic & source & 5 & $39.0\,{\textstyle\pm 1.8}$ \\*
HotpotQA & Correct source & BBH tracking & source & 5 & $38.2\,{\textstyle\pm 1.3}$ \\
\midrule
HotpotQA & Shuffled source & BBH logic & feedback & 5 & $26.8\,{\textstyle\pm 10.7}$ \\*
HotpotQA & Shuffled source & BBH tracking & feedback & 5 & $34.0\,{\textstyle\pm 2.8}$ \\*
HotpotQA & Shuffled source & BBH logic & source & 5 & $5.8\,{\textstyle\pm 2.3}$ \\*
HotpotQA & Shuffled source & BBH tracking & source & 5 & $35.2\,{\textstyle\pm 2.0}$ \\
\midrule
Shared & Base & GSM-Symbolic & base & 1 & $5.8\,{\textstyle[3.5,8.5]}$ \\*
Shared & Base & BBH logic & base & 1 & $41.6\,{\textstyle[35.6,48.0]}$ \\*
Shared & Base & BBH tracking & base & 1 & $34.4\,{\textstyle[28.4,40.4]}$ \\
\end{longtable}
\endgroup

A dagger marks a fixed-base best-of-16 estimate. Bracketed base intervals bootstrap prompts (or GSM-Symbolic templates) conditional on the one fixed model; GSM-Symbolic uses equal template means over ten variants per template.
}{}

\subsection{Endpoint decomposition and complete controlled-task results}\label{app:historical-evaluation}\label{app:sampled}
For the sequential decomposition, let $N$ be all evaluated prompts, $P$ syntactically valid outputs with a consistent public path, $C$ those outputs that also have the correct terminal public coordinate, and $K$ those in $C$ with the correct private bit. We report $P/N$, $C/N$ and $K/C$; the last quantity is undefined when $C=0$. Confidence intervals use per-seed ratios, and exact counts are shown. Cross-scoring yields disjoint task-verifier counts $N_A+N_B=C$. Memory validity is the fraction producing a legal bit and terminator; its conditional correctness uses only valid completions. All gains are computed from the two feedback evaluations on identical prompts. Greedy fixed-base scores repeat across training seeds because they evaluate one inherited checkpoint.

\paragraph{Public execution and private-task accuracy.}
Sequential evaluation uses held-out suffixes of lengths four and six. Each source route receives $2^{20}$ prediction tokens, then 8,192 reward queries: 128 updates of 64 sampled completions. Eight paired worlds compare correct source, coherently wrong source and direct RL. The wrong source is a consistent source law with the same public transition rule but a private transition table different from the target mechanism. Correct-source models obtain 242--584 positive training rewards per world, whereas direct RL and a format control that randomizes every output state obtain none. Consistent source structure, not exposure to the output format, is what gives rewards successful trajectories to reinforce.

Correct-source greedy success rises from 3.30\% to 41.67\% and wrong-source success from 3.94\% to 38.15\% (\cref{tab:revision-gain-contrasts}). The endpoint decomposition, which splits greedy success into a valid public path, a correct public endpoint and a correct private bit, says what this shared improvement is made of. Correct-source outputs have valid public paths on 96.22\% of prompts and correct public endpoints on 83.25\%, while private-bit accuracy conditional on correct public execution is 50.12\% for the correct source and 49.87\% for the wrong source (\cref{tab:revision-decomposition,fig:revision-sequential}). The gain is acquired public execution. The error partition over all prompts (\cref{tab:revision-errors-greedy}) identifies the private bit as the largest correct-source error category, 41.58\% of all prompts, against 12.96\% for a wrong public endpoint and under 4\% for malformed or path errors, so the decomposition locates the remaining error in private-state computation and its task-specific use. A classical transition-table learner given the same source records and training feedback reaches 100\% in three tested worlds (\cref{tab:controls-greedy}), so the observations carry the private information, and the decomposition measures how much of it a neural route has turned into computation.

\paragraph{Format and reward-bearing trajectories.}
In the historical Sequential direct-RL runs, all 1,024 greedy outputs per world are malformed. The format-source control instead produces syntactically valid answers on 97.75\% of prompts on average, but none has a consistent public trajectory (\cref{tab:revision-errors-greedy}). Thus a zero public-path rate can reflect either invalid syntax or invalid execution. Historical Memory direct-RL outputs likewise contain no valid bit-and-terminator answer. Across these eight Sequential direct, three format-source and five Memory direct runs, each full 8,192-query training budget contains zero positive rewards and all 128 reward-loss values are zero. The zero endpoints therefore accompany an absence of reward-weighted learning signal. Tables reporting decomposition and cross-task counts reuse these outputs; they are not additional independent runs.

\begin{table}[!htbp]\centering\small\setlength{\tabcolsep}{4pt}
\caption{Greedy evaluation on identical before/after prompts. Start and end are success percentages; gain is percentage points. Entries are means with 95\% Student-$t$ half-widths across paired seeds. Start is the feedback-entry evaluation on the same prompts; direct public-task starts evaluate one fixed base checkpoint.}
\label{tab:revision_endpoints}
\begin{tabular}{lrrrr}
\toprule
Training route & Seeds & Before (\%) & After (\%) & Gain (pp) \\
\midrule
\multicolumn{5}{l}{\textbf{Sequential}}\\[2pt]
Correct source & 8 & $3.30\,{\textstyle\pm 3.27}$ & $41.67\,{\textstyle\pm 5.56}$ & $38.38\,{\textstyle\pm 5.21}$ \\
Direct RL & 8 & $0.00\,{\textstyle\pm 0.00}$ & $0.00\,{\textstyle\pm 0.00}$ & $0.00\,{\textstyle\pm 0.00}$ \\
Wrong source & 8 & $3.94\,{\textstyle\pm 3.48}$ & $38.15\,{\textstyle\pm 10.68}$ & $34.20\,{\textstyle\pm 9.87}$ \\
Format source & 3 & $0.00\,{\textstyle\pm 0.00}$ & $0.00\,{\textstyle\pm 0.00}$ & $0.00\,{\textstyle\pm 0.00}$ \\
\midrule
\multicolumn{5}{l}{\textbf{Memory}}\\[2pt]
Correct source & 5 & $51.02\,{\textstyle\pm 2.53}$ & $49.18\,{\textstyle\pm 1.43}$ & $-1.84\,{\textstyle\pm 3.15}$ \\
Direct RL & 5 & $0.00\,{\textstyle\pm 0.00}$ & $0.00\,{\textstyle\pm 0.00}$ & $0.00\,{\textstyle\pm 0.00}$ \\
Wrong source & 5 & $49.90\,{\textstyle\pm 1.46}$ & $49.18\,{\textstyle\pm 1.43}$ & $-0.72\,{\textstyle\pm 1.35}$ \\
\midrule
\multicolumn{5}{l}{\textbf{GSM8K}}\\[2pt]
Direct RL & 5 & $8.57\,{\textstyle\pm 0.00}$ & $62.32\,{\textstyle\pm 4.14}$ & $53.75\,{\textstyle\pm 4.14}$ \\
Domain source & 5 & $2.32\,{\textstyle\pm 0.39}$ & $62.05\,{\textstyle\pm 2.93}$ & $59.73\,{\textstyle\pm 3.16}$ \\
Shuffled source & 5 & $8.58\,{\textstyle\pm 1.31}$ & $49.70\,{\textstyle\pm 29.29}$ & $41.12\,{\textstyle\pm 28.34}$ \\
\midrule
\multicolumn{5}{l}{\textbf{HotpotQA}}\\[2pt]
Direct RL & 5 & $23.93\,{\textstyle\pm 0.00}$ & $47.60\,{\textstyle\pm 1.02}$ & $23.67\,{\textstyle\pm 1.02}$ \\
Domain source & 5 & $35.10\,{\textstyle\pm 0.98}$ & $45.59\,{\textstyle\pm 0.86}$ & $10.49\,{\textstyle\pm 1.74}$ \\
Shuffled source & 5 & $5.62\,{\textstyle\pm 0.96}$ & $44.45\,{\textstyle\pm 1.13}$ & $38.83\,{\textstyle\pm 1.91}$ \\
\bottomrule
\end{tabular}
\end{table}

\begin{table}[H]\centering\small
\caption{Sequential source dose on five matched worlds. Greedy success after 8,192 reward queries; values are percentages with 95\% Student-$t$ half-widths. Source checkpoints share their training trajectories.}
\label{tab:story-dose}
\begin{tabular}{lrr}\toprule
Source prediction tokens & Worlds & Final success (\%)\\\midrule
$2^{18}$ & 5 & $0.00\,{\scriptstyle\pm 0.00}$ \\
$2^{19}$ & 5 & $0.00\,{\scriptstyle\pm 0.00}$ \\
$2^{20}$ & 5 & $41.99\,{\scriptstyle\pm 8.92}$ \\
\bottomrule\end{tabular}\end{table}

\begin{table}[!htbp]\centering\small\setlength{\tabcolsep}{4pt}
\caption{Greedy sequential endpoints for initialization, scale and alternative-route controls after 8,192 reward queries. Entries are seed means with 95\% Student-$t$ half-widths; the classical learner uses the same source records and permitted training feedback.}
\label{tab:controls-greedy}
\begin{tabular}{llrr}
\toprule
Setting & Route & Worlds & Success (\%) \\
\midrule
\multirow{3}{*}{0.5B, pretrained, $2^{20}$ source tokens} & Balanced source & 3 & $39.71\,{\textstyle\pm 32.80}$ \\
 & Interleaved source updates & 3 & $38.57\,{\textstyle\pm 24.05}$ \\
 & Classical table learner & 3 & $100.00\,{\textstyle\pm 0.00}$ \\
\midrule
\multirow{3}{*}{0.5B, random initialization, $2^{20}$ source tokens} & Correct source & 3 & $0.00\,{\textstyle\pm 0.00}$ \\
 & Wrong source & 3 & $0.00\,{\textstyle\pm 0.00}$ \\
 & Direct RL & 3 & $0.00\,{\textstyle\pm 0.00}$ \\
\midrule
\multirow{3}{*}{1.5B, pretrained, $2^{20}$ source tokens} & Correct source & 3 & $26.43\,{\textstyle\pm 35.44}$ \\
 & Wrong source & 3 & $22.27\,{\textstyle\pm 30.76}$ \\
 & Direct RL & 3 & $0.00\,{\textstyle\pm 0.00}$ \\
\bottomrule
\end{tabular}
\end{table}

\begin{table}[!htbp]\centering\small\setlength{\tabcolsep}{4pt}
\caption{Sequential greedy endpoints. Public path requires valid syntax and a consistent public trajectory. Correct public additionally requires the target terminal public coordinate. Private accuracy conditions on this latter event; -- denotes a zero denominator in every seed. The first two rates use all prompts; the last is the within-seed conditional ratio. Entries are seed means and 95\% Student-$t$ half-widths.}
\label{tab:revision-decomposition}
\begin{tabular}{lrrrr}
\toprule
Route & Seeds & Public path (\%) & Correct public (\%) & Private $\mid$ public (\%) \\
\midrule
Correct source & 8 & $96.22\,{\textstyle\pm 8.56}$ & $83.25\,{\textstyle\pm 11.38}$ & $50.12\,{\textstyle\pm 1.81}$ \\
Wrong source & 8 & $98.13\,{\textstyle\pm 2.87}$ & $76.77\,{\textstyle\pm 21.79}$ & $49.87\,{\textstyle\pm 1.09}$ \\
\midrule
Direct RL & 8 & $0.00\,{\textstyle\pm 0.00}$ & $0.00\,{\textstyle\pm 0.00}$ & -- \\
Format source & 3 & $0.00\,{\textstyle\pm 0.00}$ & $0.00\,{\textstyle\pm 0.00}$ & -- \\
\bottomrule
\end{tabular}
\end{table}

\begin{table}[!htbp]\centering\small\setlength{\tabcolsep}{4pt}
\caption{Exact denominators for the sequential greedy decomposition. $N$ is all prompts, $P$ valid public paths, $C$ correct public endpoints, and $K$ correct private bits among $C$. The conditional accuracy is $K/C$ separately in each seed; $C=0$ is missing.}
\label{tab:revision-denominators}
\begin{tabular}{lrrrrr}
\toprule
Route & Seed & $N$ & $P$ & $C$ & $K$ \\
\midrule
Correct source & 0 & 1024 & 1012 & 695 & 345 \\
Correct source & 1 & 1024 & 1024 & 1021 & 515 \\
Correct source & 2 & 1024 & 1024 & 824 & 381 \\
Correct source & 3 & 1024 & 1024 & 773 & 412 \\
Correct source & 4 & 1024 & 1024 & 1017 & 497 \\
Correct source & 5 & 1024 & 1024 & 771 & 405 \\
Correct source & 6 & 1024 & 726 & 715 & 358 \\
Correct source & 7 & 1024 & 1024 & 1004 & 501 \\
\midrule
Wrong source & 0 & 1024 & 957 & 772 & 388 \\
Wrong source & 1 & 1024 & 1024 & 1024 & 524 \\
Wrong source & 2 & 1024 & 1024 & 610 & 299 \\
Wrong source & 3 & 1024 & 1022 & 274 & 141 \\
Wrong source & 4 & 1024 & 1024 & 1021 & 502 \\
Wrong source & 5 & 1024 & 1024 & 646 & 329 \\
Wrong source & 6 & 1024 & 940 & 918 & 438 \\
Wrong source & 7 & 1024 & 1024 & 1024 & 504 \\
\midrule
Direct RL & 0 & 1024 & 0 & 0 & 0 \\
Direct RL & 1 & 1024 & 0 & 0 & 0 \\
Direct RL & 2 & 1024 & 0 & 0 & 0 \\
Direct RL & 3 & 1024 & 0 & 0 & 0 \\
Direct RL & 4 & 1024 & 0 & 0 & 0 \\
Direct RL & 5 & 1024 & 0 & 0 & 0 \\
Direct RL & 6 & 1024 & 0 & 0 & 0 \\
Direct RL & 7 & 1024 & 0 & 0 & 0 \\
\midrule
Format source & 3 & 1024 & 0 & 0 & 0 \\
Format source & 4 & 1024 & 0 & 0 & 0 \\
Format source & 5 & 1024 & 0 & 0 & 0 \\
\bottomrule
\end{tabular}
\end{table}

\begin{table}[!htbp]\centering\small\setlength{\tabcolsep}{4pt}
\caption{Greedy cross-scoring of the same saved outputs, three paired worlds. A/B success sets are disjoint. The conditional A score uses the union of the two sets as denominator; it is undefined for direct runs with no eligible outputs. Entries are seed means with 95\% Student-$t$ half-widths.}
\label{tab:revision-choice}
\begin{tabular}{llrrr}
\toprule
Route & Reward task & Test A (\%) & Test B (\%) & A $\mid$ A or B (\%) \\
\midrule
Correct source & A & $42.77\,{\textstyle\pm 12.42}$ & $40.59\,{\textstyle\pm 21.93}$ & $51.57\,{\textstyle\pm 5.88}$ \\
Correct source & B & $41.83\,{\textstyle\pm 14.49}$ & $39.45\,{\textstyle\pm 22.79}$ & $51.73\,{\textstyle\pm 6.34}$ \\
\midrule
Direct RL & A & $0.00\,{\textstyle\pm 0.00}$ & $0.00\,{\textstyle\pm 0.00}$ & -- \\
Direct RL & B & $0.00\,{\textstyle\pm 0.00}$ & $0.00\,{\textstyle\pm 0.00}$ & -- \\
\bottomrule
\end{tabular}
\end{table}
\begin{table}[!htbp]\centering\small\setlength{\tabcolsep}{4pt}
\caption{Exact cross-task greedy counts. $C=N_A+N_B$ is the common eligible denominator: valid public execution with the correct public endpoint. Each world uses 1,024 prompts shared across its A/B branches and scorers; task B changes only the required private bit.}
\label{tab:revision-choice-denominators}
\begin{tabular}{lrrrrr}
\toprule
Route & Seed & $N$ & $C$ & $N_A$ & $N_B$ \\
\midrule
Correct source & 3 & 1024 & 773 & 412 & 361 \\
Correct source & 4 & 1024 & 1017 & 497 & 520 \\
Correct source & 5 & 1024 & 771 & 405 & 366 \\
\midrule
Source / task B & 3 & 1024 & 824 & 439 & 385 \\
Source / task B & 4 & 1024 & 988 & 482 & 506 \\
Source / task B & 5 & 1024 & 685 & 364 & 321 \\
\midrule
Direct RL & 3 & 1024 & 0 & 0 & 0 \\
Direct RL & 4 & 1024 & 0 & 0 & 0 \\
Direct RL & 5 & 1024 & 0 & 0 & 0 \\
\midrule
Direct / task B & 3 & 1024 & 0 & 0 & 0 \\
Direct / task B & 4 & 1024 & 0 & 0 & 0 \\
Direct / task B & 5 & 1024 & 0 & 0 & 0 \\
\bottomrule
\end{tabular}
\end{table}

\begin{table}[!htbp]\centering\small\setlength{\tabcolsep}{4pt}
\caption{Contextual greedy endpoints. Correctness conditional on a valid bit is undefined if no valid answer is produced. Constant counts runs producing a single valid action on all 1,024 prompts. A balanced input-independent bit rule has 50\% population accuracy. Entries are seed means with 95\% Student-$t$ half-widths.}
\label{tab:revision-memory}
\begin{tabular}{lrrrrr}
\toprule
Route & Seeds & Valid (\%) & Correct (\%) & Correct $\mid$ valid (\%) & Constant runs \\
\midrule
Correct source & 5 & $100.00\,{\textstyle\pm 0.00}$ & $49.18\,{\textstyle\pm 1.43}$ & $49.18\,{\textstyle\pm 1.43}$ & 5 \\
Wrong source & 5 & $100.00\,{\textstyle\pm 0.00}$ & $49.18\,{\textstyle\pm 1.43}$ & $49.18\,{\textstyle\pm 1.43}$ & 5 \\
Direct RL & 5 & $0.00\,{\textstyle\pm 0.00}$ & $0.00\,{\textstyle\pm 0.00}$ & -- & 0 \\
\bottomrule
\end{tabular}
\end{table}

\begin{table}[!htbp]\centering\small\setlength{\tabcolsep}{4pt}
\caption{Sequential greedy endpoint error partition (percent of all prompts). Malformed, invalid public path, wrong public endpoint, wrong private bit and success are mutually exclusive and exhaustive. Private-bit failures here use all prompts as denominator; conditional accuracy uses only correct-public outputs. Entries are seed means with 95\% Student-$t$ half-widths.}
\label{tab:revision-errors-greedy}
\begin{tabular}{lrrrrr}
\toprule
Route & Malformed & Path error & Public error & Private error & Success \\
\midrule
Correct source & $3.72\,{\textstyle\pm 8.58}$ & $0.06\,{\textstyle\pm 0.14}$ & $12.96\,{\textstyle\pm 11.00}$ & $41.58\,{\textstyle\pm 6.16}$ & $41.67\,{\textstyle\pm 5.56}$ \\
Wrong source & $1.81\,{\textstyle\pm 2.83}$ & $0.06\,{\textstyle\pm 0.10}$ & $21.36\,{\textstyle\pm 22.37}$ & $38.62\,{\textstyle\pm 11.18}$ & $38.15\,{\textstyle\pm 10.68}$ \\
Direct RL & $100.00\,{\textstyle\pm 0.00}$ & $0.00\,{\textstyle\pm 0.00}$ & $0.00\,{\textstyle\pm 0.00}$ & $0.00\,{\textstyle\pm 0.00}$ & $0.00\,{\textstyle\pm 0.00}$ \\
Format source & $2.25\,{\textstyle\pm 5.25}$ & $97.75\,{\textstyle\pm 5.25}$ & $0.00\,{\textstyle\pm 0.00}$ & $0.00\,{\textstyle\pm 0.00}$ & $0.00\,{\textstyle\pm 0.00}$ \\
\bottomrule
\end{tabular}
\end{table}

\begin{table}[!htbp]\centering\small\setlength{\tabcolsep}{4pt}
\caption{Paired differences of reward-stage gains: correct source minus the indicated route. Only shared seeds enter each contrast; entries are seed-paired mean differences in percentage points with 95\% Student-$t$ half-widths.}
\label{tab:revision-gain-contrasts}
\begin{tabular}{lllrr}
\toprule
Task & Evaluation & Comparator & Seeds & Difference (pp) \\
\midrule
sequential & greedy & Direct RL & 8 & $38.38\,{\textstyle\pm 5.21}$ \\
sequential & greedy & Wrong source & 8 & $4.17\,{\textstyle\pm 8.95}$ \\
sequential & greedy & Format source & 3 & $41.96\,{\textstyle\pm 10.16}$ \\
\midrule
contextual & greedy & Direct RL & 5 & $-1.84\,{\textstyle\pm 3.15}$ \\
contextual & greedy & Wrong source & 5 & $-1.11\,{\textstyle\pm 2.06}$ \\
\midrule
GSM8K & greedy & Direct RL & 5 & $5.97\,{\textstyle\pm 3.66}$ \\
GSM8K & greedy & Shuffled source & 5 & $18.61\,{\textstyle\pm 30.16}$ \\
\midrule
HotpotQA & greedy & Direct RL & 5 & $-13.18\,{\textstyle\pm 1.50}$ \\
HotpotQA & greedy & Shuffled source & 5 & $-28.34\,{\textstyle\pm 2.42}$ \\
\bottomrule
\end{tabular}
\end{table}

\paragraph{Sampled decoding.}
Under sampled decoding of the same checkpoints (one sampled answer per prompt), source-trained sequential models also outperform direct RL, by $5.48\pm1.68$ points at the endpoint, while the correct-source reward-stage gain is $0.31\pm0.55$ percentage points against $38.38\pm5.21$ under greedy decoding. The output decompositions in this subsection use greedy decoding (\cref{fig:revision-sampled,tab:sampled-robustness}).
\begin{table}[!htbp]\centering\small\setlength{\tabcolsep}{4pt}
\caption{Sampled-decoding robustness check on the same checkpoints and prompts as the greedy results (one sampled answer per prompt at temperature one). Start and end are success percentages; gain is percentage points; entries are seed means with 95\% Student-$t$ half-widths. The two sequential endpoint contrasts use exact two-sided sign-flip tests; adjusted $p$-values use a fixed three-hypothesis Holm family, of which the two measured endpoint contrasts are shown and the third hypothesis is entered conservatively at $p=1$.}
\label{tab:sampled-robustness}
\begin{tabular}{lrrrr}
\toprule
Training route & Seeds & Before (\%) & After (\%) & Gain (pp) \\
\midrule
\multicolumn{5}{l}{\textbf{Sequential}}\\[2pt]
Correct source & 8 & $5.18\,{\textstyle\pm 1.71}$ & $5.48\,{\textstyle\pm 1.68}$ & $0.31\,{\textstyle\pm 0.55}$ \\
Direct RL & 8 & $0.00\,{\textstyle\pm 0.00}$ & $0.00\,{\textstyle\pm 0.00}$ & $0.00\,{\textstyle\pm 0.00}$ \\
Wrong source & 8 & $5.16\,{\textstyle\pm 1.59}$ & $5.73\,{\textstyle\pm 1.72}$ & $0.56\,{\textstyle\pm 0.77}$ \\
Format source & 3 & $0.00\,{\textstyle\pm 0.00}$ & $0.00\,{\textstyle\pm 0.00}$ & $0.00\,{\textstyle\pm 0.00}$ \\
\midrule
\multicolumn{5}{l}{\textbf{Memory}}\\[2pt]
Correct source & 5 & $48.26\,{\textstyle\pm 1.16}$ & $49.18\,{\textstyle\pm 1.43}$ & $0.92\,{\textstyle\pm 1.41}$ \\
Direct RL & 5 & $0.00\,{\textstyle\pm 0.00}$ & $0.00\,{\textstyle\pm 0.00}$ & $0.00\,{\textstyle\pm 0.00}$ \\
Wrong source & 5 & $48.16\,{\textstyle\pm 2.66}$ & $49.18\,{\textstyle\pm 1.43}$ & $1.02\,{\textstyle\pm 2.39}$ \\
\midrule
\multicolumn{5}{l}{\textbf{GSM8K}}\\[2pt]
Direct RL & 5 & $5.13\,{\textstyle\pm 0.22}$ & $45.75\,{\textstyle\pm 3.96}$ & $40.62\,{\textstyle\pm 4.04}$ \\
Domain source & 5 & $1.62\,{\textstyle\pm 0.25}$ & $45.17\,{\textstyle\pm 5.19}$ & $43.55\,{\textstyle\pm 4.99}$ \\
Shuffled source & 5 & $2.68\,{\textstyle\pm 0.91}$ & $40.89\,{\textstyle\pm 13.87}$ & $38.21\,{\textstyle\pm 13.87}$ \\
\midrule
\multicolumn{5}{l}{\textbf{HotpotQA}}\\[2pt]
Direct RL & 5 & $13.59\,{\textstyle\pm 0.90}$ & $45.27\,{\textstyle\pm 1.02}$ & $31.68\,{\textstyle\pm 0.83}$ \\
Domain source & 5 & $19.55\,{\textstyle\pm 1.97}$ & $43.71\,{\textstyle\pm 1.48}$ & $24.16\,{\textstyle\pm 1.82}$ \\
Shuffled source & 5 & $2.99\,{\textstyle\pm 0.71}$ & $42.23\,{\textstyle\pm 1.05}$ & $39.24\,{\textstyle\pm 1.13}$ \\
\midrule
\multicolumn{5}{l}{\textbf{Sequential endpoint contrasts} (difference in pp, $p$, $p_{\rm Holm}$)}\\[2pt]
\multicolumn{2}{l}{Correct source $-$ Direct RL} & \multicolumn{3}{r}{$5.48\,{\textstyle\pm 1.68}$;\quad $p=0.0078$;\quad $p_{\rm Holm}=0.0234$} \\
\multicolumn{2}{l}{Correct source $-$ Wrong source} & \multicolumn{3}{r}{$-0.24\,{\textstyle\pm 0.51}$;\quad $p=0.336$;\quad $p_{\rm Holm}=0.672$} \\
\bottomrule
\end{tabular}
\end{table}

\begin{figure}[!htbp]\centering
\includegraphics[width=\linewidth]{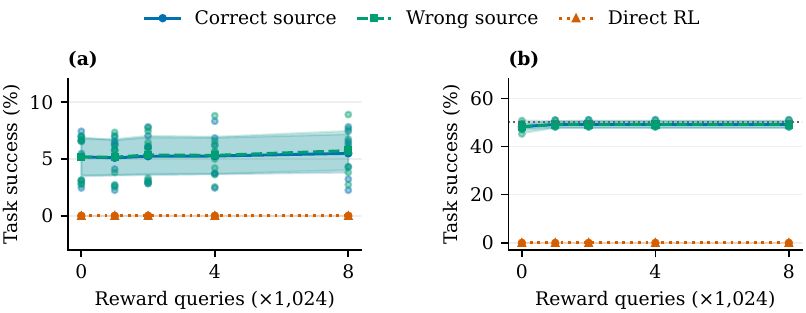}
\caption{Sampled evaluation of (a) Sequential, eight worlds, and (b) Memory, five worlds. Each checkpoint uses one answer per prompt at temperature one on 1,024 prompts per world. Curves and large markers show world means, small translucent dots individual worlds, and shaded bands 95\% Student-$t$ intervals. The dotted 50\% line in (b) marks binary chance. \Cref{fig:revision-sequential,tab:revision_endpoints,tab:revision-decomposition} report greedy evaluation of these historical checkpoints.}
\label{fig:revision-sampled}\end{figure}
\begin{figure}[!htbp]\centering
\includegraphics[width=\linewidth]{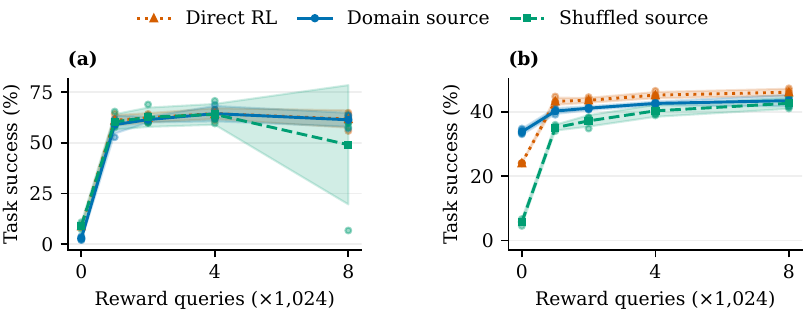}
\caption{Greedy learning curves for (a) GSM8K and (b) HotpotQA on the same 512 prompts at every checkpoint, five training seeds. Curves and large markers are seed means, small translucent dots are individual seeds, and bands are 95\% Student-$t$ intervals. The full endpoint comparison in \cref{fig:revision-public,tab:public-summary} uses the larger matched evaluation sets.}
\label{fig:revision-public-curves}\end{figure}
\begin{figure}[!htbp]\centering
\includegraphics[width=\linewidth]{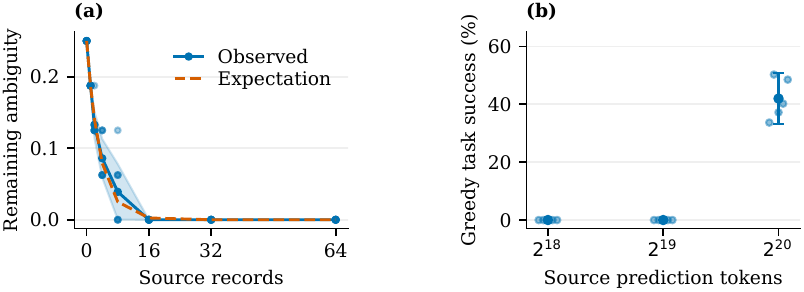}
\caption{Two effects of source observations on distinct horizontal scales. (a) Remaining contextual ambiguity versus the number $N$ of source records in eight worlds; the dashed curve is the analytic expectation $\tfrac14(\tfrac34)^N$. (b) Greedy Sequential success after 8,192 reward queries versus source prediction tokens, five matched worlds. Solid curves and large markers are world means; small translucent dots are individual worlds. Shading in (a) and error bars in (b) are 95\% Student-$t$ intervals.}
\label{fig:revision-information}\label{fig:revision-source-budget}\end{figure}

\subsection{Alternative routes, selection and external transfer}\label{app:recovered-coverage}\label{app:additional-comparisons}
\paragraph{Source dose and access to reward.}
The budget comparison evaluates the $2^{18}$, $2^{19}$ and $2^{20}$ checkpoints of five source trajectories. At $2^{18}$ and $2^{19}$ source tokens, no feedback run receives a positive reward; at $2^{20}$ every run does, and greedy success reaches 41.99\% (\cref{tab:story-dose,fig:revision-information}). Access to successful trajectories therefore switches on within a factor of two in source budget.

\paragraph{Initialization, model scale and adaptation.}
In the three-world comparison at $2^{20}$ source tokens, no random-initialized route attains sequential success. The contrast with pretrained initialization shows that inherited weights matter under this budget and protocol. At 1.5B parameters, correct-source and wrong-source routes reach 26.43\% and 22.27\% while direct RL stays at zero; source sampling balanced over initial public contexts, at the same token budget, reaches $39.71\pm32.80\%$ over three worlds, with substantial between-world uncertainty (\cref{tab:controls-greedy}). Continued prediction and answer-only supervised fine-tuning (SFT) do not reproduce the reward-stage improvement, whereas interleaving source updates with RL retains it at 38.57\% (\cref{tab:refinement-alternatives-greedy,tab:controls-greedy}). Fixed-source best-of-$N$ selection raises success from 4.95\% at $N=1$ to 26.73\% at $N=8$ by spending test-time verifier queries instead of training queries (\cref{tab:refinement-controlled-bestof}). 
Source-only, doubled-source and answer-only SFT scores on the same three worlds are 3.55\%, 1.40\% and zero (\cref{tab:refinement-alternatives-greedy}); their separate source and answer-label costs are listed with the endpoints.

Public SFT yields 12.36\% on GSM8K and 53.74\% on HotpotQA over three seeds (\cref{tab:refinement-public-sft-greedy}). Transfer to GSM-Symbolic gives 53.0\% after source+RL versus 54.5\% after direct RL; BBH logic and object-tracking results vary by task and route (\cref{tab:public-transfer}). Controlled sequential checkpoints change little on the same external tasks (\cref{tab:refinement-controlled-bbh}). On separate 256-question sets, best-of-16 selection scores 22.1\% after domain source versus 44.5\% for the fixed base on GSM8K, and 50.7\% versus 45.7\% on HotpotQA (\cref{tab:public-transfer}).
\begin{table}[H]\centering\small
\setlength{\tabcolsep}{5pt}\renewcommand{\arraystretch}{1.12}
\caption{Greedy sequential source-only and alternative-adaptation endpoints.}\label{tab:refinement-alternatives-greedy}
\begin{tabular}{@{}lrrrr@{}}
\toprule
Route & Seeds & Source tokens & Answer labels & Success (\%)\\
\midrule
Source only, all worlds & 8 & 1,048,576 & 0 & $3.30\,{\textstyle\pm 3.27}$\\
\midrule
Source only, paired worlds & 3 & 1,048,576 & 0 & $3.55\,{\textstyle\pm 3.76}$\\
Continue source prediction & 3 & 2,097,152 & 0 & $1.40\,{\textstyle\pm 3.40}$\\
Answer-only adaptation & 3 & 1,048,576 & 8,192 & $0.00\,{\textstyle\pm 0.00}$\\
\bottomrule
\end{tabular}
\par\smallskip\parbox{\linewidth}{\footnotesize All evaluations use 1,024 identical prompts within a world. The paired-world rows use three matched worlds. All routes have zero reward-training queries. Answer-only adaptation additionally uses its recorded answer-prediction targets, separately from source observations. Intervals are 95\% Student-$t$ intervals across seed means.}
\end{table}

\begin{table}[H]\centering\small
\setlength{\tabcolsep}{5pt}\renewcommand{\arraystretch}{1.12}
\caption{Verification-selected sampling from the fixed sequential source checkpoint. Uncertainty is the 95\% Student-$t$ half-width across training seeds.}\label{tab:refinement-controlled-bestof}
\begin{tabular}{@{}rrrrr@{}}
\toprule
Candidates & Seeds & Prompts & Test queries/seed & Success (\%)\\
\midrule
1 & 3 & 1,024 & 1,024 & $4.95\,{\textstyle\pm 6.29}$\\
2 & 3 & 1,024 & 2,048 & $8.72\,{\textstyle\pm 9.86}$\\
4 & 3 & 1,024 & 4,096 & $15.92\,{\textstyle\pm 16.01}$\\
8 & 3 & 1,024 & 8,192 & $26.73\,{\textstyle\pm 24.57}$\\
\bottomrule
\end{tabular}
\par\smallskip\parbox{\linewidth}{\footnotesize Each row selects a correct answer, when present, among the first $N$ sampled candidates. All rows use nested prefixes of the same eight candidates per prompt, with no reward-stage parameter updates. Intervals summarize three seed-level means.}
\end{table}

\begin{table}[H]\centering\small
\setlength{\tabcolsep}{5pt}\renewcommand{\arraystretch}{1.12}
\caption{Greedy external transfer after controlled sequential training of Qwen2.5-0.5B.}\label{tab:refinement-controlled-bbh}
\begin{tabular}{@{}llrr@{}}
\toprule
Checkpoint & Independent units & Prompts & Success (\%)\\
\midrule
\multicolumn{4}{@{}l}{\textbf{Logical deduction} (250 prompts)}\\
Base checkpoint & 1 fixed model & 250 & $30.40\,[24.80,36.00]$\\
\midrule
After source prediction & 3 training seeds & 250 & $34.40\,{\textstyle\pm 2.63}$\\
\midrule
After source + reward & 3 training seeds & 250 & $34.27\,{\textstyle\pm 3.19}$\\
\midrule
\multicolumn{4}{@{}l}{\textbf{Object tracking} (250 prompts)}\\
\midrule
Base checkpoint & 1 fixed model & 250 & $21.20\,[16.40,26.40]$\\
\midrule
After source prediction & 3 training seeds & 250 & $28.13\,{\textstyle\pm 14.10}$\\
\midrule
After source + reward & 3 training seeds & 250 & $27.60\,{\textstyle\pm 12.69}$\\
\bottomrule
\end{tabular}
\par\smallskip\parbox{\linewidth}{\footnotesize Source and feedback uncertainty is a 95\% Student-$t$ interval over three training seeds. Base brackets give the 95\% item-bootstrap interval for the fixed pretrained model.}
\end{table}

\begin{table}[H]\centering\small
\setlength{\tabcolsep}{5pt}\renewcommand{\arraystretch}{1.12}
\caption{Greedy answer-only supervised adaptation on public tasks.}\label{tab:refinement-public-sft-greedy}
\begin{tabular}{@{}lrrrr@{}}
\toprule
Task & Seeds & Prompts/seed & Answer labels/seed & Success (\%)\\
\midrule
GSM8K & 3 & 1,319 & 8,192 & $12.36\,{\textstyle\pm 0.38}$\\
HotpotQA & 3 & 1,024 & 8,192 & $53.74\,{\textstyle\pm 1.38}$\\
\bottomrule
\end{tabular}
\par\smallskip\parbox{\linewidth}{\footnotesize Each Qwen2.5-1.5B run starts from its domain-source checkpoint and receives 8,192 terminal-answer labels, with no reward-training queries. Source training uses $2^{22}$ prediction tokens. Success is exact answer correctness; intervals are 95\% Student-$t$ intervals over three training seeds, not prompt intervals.}
\end{table}

\subsection{Source-acquisition probes and public source dose}\label{app:v2-results}
This subsection gives the two-world format-stage study discussed in \cref{sec:experiments} and the $2^{20}$-token public source-dose comparison.

\paragraph{Public source-dose comparison.}
The $2^{20}$-token source route uses Qwen2.5-1.5B, the same source learning rates as the $2^{22}$ route and 8,192 training verifications per feedback run. Within each seed, the direct, $2^{20}$ and $2^{22}$ routes share identical evaluation prompts and the same ordered sequence of 8,192 training prompts. \Cref{tab:v2-public-dose,tab:v2-public-contrasts} report start, endpoint, gain and seed-paired contrasts.
\begin{table}[H]\centering\small
\caption{Public source-dose endpoints. Five paired training seeds; the same 1,319 GSM8K or 1,024 HotpotQA prompts at both endpoints and across routes. Rates are percentages; gains are percentage points. Uncertainty is a seed-level 95\% Student-$t$ half-width. Direct starting scores evaluate one fixed pretrained checkpoint.}
\label{tab:v2-public-dose}
\begin{tabular}{lrrr}\toprule
Source tokens & Before RL (\%) & After RL (\%) & Gain (pp)\\\midrule
\multicolumn{4}{l}{\textit{GSM8K; 8,192 reward queries}}\\
None (direct) & $8.57\,\text{\footnotesize$\pm 0.00$}$ & $62.32\,\text{\footnotesize$\pm 4.14$}$ & $53.75\,\text{\footnotesize$\pm 4.14$}$\\
$2^{20}$ & $3.11\,\text{\footnotesize$\pm 0.57$}$ & $61.76\,\text{\footnotesize$\pm 3.15$}$ & $58.65\,\text{\footnotesize$\pm 3.67$}$\\
$2^{22}$ & $2.32\,\text{\footnotesize$\pm 0.39$}$ & $62.05\,\text{\footnotesize$\pm 2.93$}$ & $59.73\,\text{\footnotesize$\pm 3.16$}$\\
\midrule
\multicolumn{4}{l}{\textit{HotpotQA; 8,192 reward queries}}\\
None (direct) & $23.93\,\text{\footnotesize$\pm 0.00$}$ & $47.60\,\text{\footnotesize$\pm 1.02$}$ & $23.67\,\text{\footnotesize$\pm 1.02$}$\\
$2^{20}$ & $35.55\,\text{\footnotesize$\pm 1.44$}$ & $46.23\,\text{\footnotesize$\pm 1.28$}$ & $10.68\,\text{\footnotesize$\pm 2.61$}$\\
$2^{22}$ & $35.10\,\text{\footnotesize$\pm 0.98$}$ & $45.59\,\text{\footnotesize$\pm 0.86$}$ & $10.49\,\text{\footnotesize$\pm 1.74$}$\\
\bottomrule
\end{tabular}\end{table}

\begin{table}[H]\centering\small
\caption{Seed-paired public source-dose contrasts, in percentage points (pp). Each effect is calculated within seed before forming the five-seed mean and 95\% Student-$t$ half-width.}
\label{tab:v2-public-contrasts}
\begin{tabular}{lrr}\toprule
Contrast & Endpoint difference (pp) & Gain difference (pp)\\\midrule
\multicolumn{3}{l}{\textit{GSM8K}}\\
$2^{20}$ minus direct & $-0.56\,\text{\footnotesize$\pm 4.87$}$ & $4.90\,\text{\footnotesize$\pm 5.28$}$\\
$2^{22}$ minus direct & $-0.27\,\text{\footnotesize$\pm 3.63$}$ & $5.97\,\text{\footnotesize$\pm 3.66$}$\\
$2^{20}$ minus $2^{22}$ & $-0.29\,\text{\footnotesize$\pm 5.64$}$ & $-1.08\,\text{\footnotesize$\pm 6.38$}$\\
\midrule
\multicolumn{3}{l}{\textit{HotpotQA}}\\
$2^{20}$ minus direct & $-1.37\,\text{\footnotesize$\pm 0.37$}$ & $-12.99\,\text{\footnotesize$\pm 1.67$}$\\
$2^{22}$ minus direct & $-2.01\,\text{\footnotesize$\pm 0.88$}$ & $-13.18\,\text{\footnotesize$\pm 1.50$}$\\
$2^{20}$ minus $2^{22}$ & $0.64\,\text{\footnotesize$\pm 1.12$}$ & $0.20\,\text{\footnotesize$\pm 1.82$}$\\
\bottomrule
\end{tabular}\end{table}

\begin{figure}[t]\centering
\includegraphics[width=\linewidth]{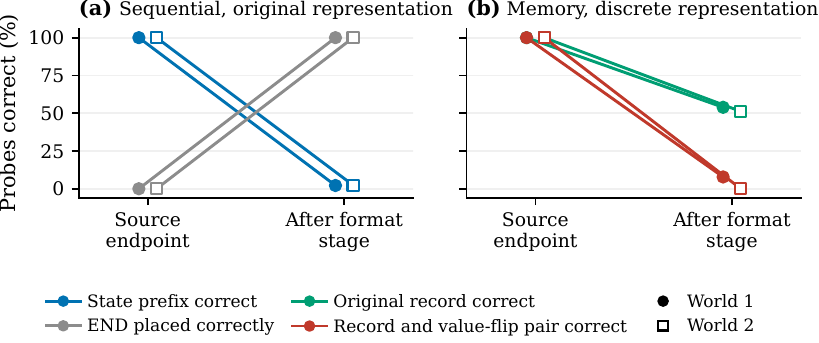}
\caption{Acquired computation before and after the answer-format stage, two worlds at $2^{24}$ source tokens. (a) Sequential, original representation with state targets: correct first-$H$ state prefixes and correct END placement, out of 512 probes with the first operation supplied. (b) Memory, discrete representation with the all-token objective: correct first-bit answers on original records and on both members of record/value-flip pairs, each out of 256. Circles and squares are the two worlds, slightly offset where they coincide. Colors identify metrics; lines join the same world and metric across stages.}
\label{fig:acquisition-format}\end{figure}

\paragraph{Acquisition design.}
Two source worlds per representation compare learning rates $10^{-5}$ and $3\times10^{-5}$ at $2^{22}$ prediction tokens, and the better rate per representation is trained to $2^{24}$ tokens with matched objective controls. Sequential compares state targets with private-bit-weighted state targets (private weight four, other state targets weight one). Memory compares all-token prediction with future-bit prediction, the latter excluding END and end-of-sequence (EOS) from its source loss. Each source update observes 8,192 prediction-token positions. Alongside the original representations, an alternative Sequential representation adds an explicit length boundary and a local-to-complete-trajectory curriculum, and an alternative Memory representation uses discrete keys and values; alternative tasks are reported separately from the original continuous-numeric Memory task.

An answer-format stage precedes and follows source training. Its loss supervises complete public-format answers, including termination, on records whose private fields are independent random labels, so it reveals neither the target mechanism nor the task binding. Sequential uses 128 format updates per stage and Memory 32, with batch 64 and learning rate $10^{-5}$; stage transitions reset Adam moments.

\paragraph{Separating state content from termination.}
Each Sequential endpoint has 512 source-validation prompts, split equally between horizons $H=5$ and $H=7$, with the random first operation supplied and no task binding. The complete-answer metric requires exactly the expected states followed by END; a content diagnostic instead checks the first $H$ generated states without modifying the output. Seven of the eight original-representation $2^{22}$ endpoints already generate at least 511/512 correct state prefixes. At $2^{24}$, all four original source models and three of four curriculum models generate every state prefix correctly, continuing to the 128-token limit rather than emitting END. After the format stage, every $2^{24}$ Sequential model places END correctly on all 512 prompts, while state-prefix counts fall to 6--14/512 (\cref{tab:v2-pilot-counts}). The two metrics separate termination, learned in the format stage, from state-prefix accuracy, which falls sharply after this stage.

\begin{figure}[!htbp]\centering
\includegraphics[width=\linewidth]{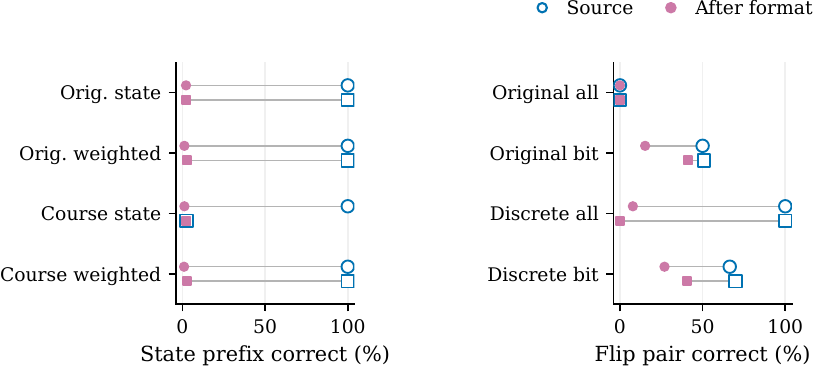}
\caption{Two-world acquisition probes before and after the format stage at $2^{24}$ source tokens. Left: exact Sequential state prefixes, excluding termination, on 512 probes. Right: both members correct in each of 256 Memory original/relevant-flip pairs. Open blue markers are source endpoints; filled purple markers follow the format stage. Circles and squares denote the two worlds, and links pair the same world and representation. Legend abbreviations: ``Orig.'' is the original representation and ``Course'' the Sequential curriculum representation; ``state'' uses state-token targets and ``weighted'' upweights private state targets; ``all'' and ``bit'' are the Memory all-token and future-bit objectives. The Sequential diagnostic uses source-validation support with the first operation given. \Cref{fig:acquisition-format} shows the original-representation Sequential state-target and discrete Memory all-token conditions, adding termination and first-bit accuracy to the two diagnostics shown here.}
\label{fig:v2-stage}\end{figure}

\paragraph{Memory interventions and the format interface.}
Each Memory world has 256 original source records and paired relevant-value flips; a successful pair requires the correct first bit on both inputs. At $2^{24}$ tokens, original-representation all-token source accuracy averages 49.61\% with no successful pairs, and future-bit accuracy averages 75.20\% with 50.39\% successful pairs. Discrete all-token source models answer all originals and relevant pairs correctly and already satisfy the public format checks. After the format stage, these means fall to 52.54\% and 3.91\%; discrete future-bit means change from 85.35\%/68.16\% to 64.45\%/33.79\%. For a binary private field predicted with probability $p$, independent uniform labels give expected cross-entropy $-\tfrac12\log p-\tfrac12\log(1-p)$, minimized at $p=1/2$: supervising random private content pulls the model toward an input-independent private prediction, which is consistent with the observed loss of intervention sensitivity after the format stage and motivates placing format preparation before source acquisition and excluding such random-label supervision afterwards.

{\small\setlength{\tabcolsep}{4pt}\renewcommand{\arraystretch}{0.92}
\begin{longtable}{llrrrr}
\caption{Acquisition-probe counts at $2^{24}$ source tokens. Sequential columns count correct first-$H$ state prefixes out of 512; Memory columns count correct original bits and relevant-flip pairs out of 256. Each cell shows before/after format. Before the format stage no Sequential model emits END; after it, every model places END correctly on all 512 prompts.}\label{tab:v2-pilot-counts}\\
\toprule Representation & Objective & World & Content before & Content after & Pair before/after\\\midrule\endfirsthead
\caption[]{Acquisition-probe counts (continued).}\\
\toprule Representation & Objective & World & Content before & Content after & Pair before/after\\\midrule\endhead
\multicolumn{6}{l}{\textit{Sequential; content /512}}\\
Original & State & 1 & 512 & 11 & --\\
Original & State & 2 & 512 & 12 & --\\
Original & Weighted & 1 & 512 & 6 & --\\
Original & Weighted & 2 & 512 & 14 & --\\
Curriculum & State & 1 & 512 & 6 & --\\
Curriculum & State & 2 & 12 & 12 & --\\
Curriculum & Weighted & 1 & 512 & 5 & --\\
Curriculum & Weighted & 2 & 512 & 14 & --\\
\midrule
\multicolumn{6}{l}{\textit{Memory; content and pairs /256}}\\
Original & All-token & 1 & 130 & 130 & 0/0\\
Original & All-token & 2 & 124 & 132 & 0/0\\
Original & Future-bit & 1 & 192 & 145 & 128/39\\
Original & Future-bit & 2 & 193 & 182 & 130/105\\
Discrete & All-token & 1 & 256 & 138 & 256/20\\
Discrete & All-token & 2 & 256 & 131 & 256/0\\
Discrete & Future-bit & 1 & 220 & 152 & 170/69\\
Discrete & Future-bit & 2 & 217 & 178 & 179/104\\
\bottomrule
\end{longtable}
}

\FloatBarrier
\section{Reading map and theorem interfaces}\label{app:complete-statements}
The information results first specify what observations can distinguish. The optimization results then construct finite training paths under their stated architectures and schedules. Complete statements appear next to their proofs. Shared network facts are proved once; each optimizer checks its own parameter choices and failure allocations.

\begin{center}\small
\begin{tabular}{@{}ll@{}}\toprule
Question & Complete statement and proof \\\midrule
Sequential reward equivalence & \cref{thm:quotient-formal,app:quotient}\\
Contextual source information & \cref{thm:context-information-formal,app:context-information}\\
Shared initialization and network bounds & \cref{app:shared-foundations}\\
Sequential finite Adam path & \cref{thm:neural-formal,seqadam:app:adam-joint}\\
Contextual finite Adam path & \cref{ctx:thm:main-formal,ctx:sec:budgets}\\
SGD alternative & \cref{thm:neural-sgd,app:sgd-path}\\
Information and computation costs & \cref{app:learning-routes}\\
Experimental protocols and results & \cref{app:experiments}\\\bottomrule
\end{tabular}
\end{center}
Here $q=2m$ counts sequential states and controls, $H$ is the operation horizon, and $N=2H+2$ counts tokens in a source record. Within shared network bounds, $N$ is the padded context length, $R$ the attention-head count, $k_{\rm head}=d/R$ its width, and $D_{\rm feat}$ the feature dimension. Optimizer sections explicitly instantiate these quantities; their local rate and radius symbols do not change the task binding. Contextual dimensions are $n_{\rm ctx}$ and $k=p-2$. A source record, prediction target, verifier query and optimizer update are distinct resources.

\section{Information supplied by source and reward observations}\label{app:information}
In either family a world fixes a reusable mechanism and a downstream rule.
Initial information $O$ includes any source sample and inherited weights; reward adaptation produces a transcript of prompts, sampled outputs and terminal feedback. The decision is an output on a held-out input. In Sequential the mechanism is $\theta$, the rule is $\eta$, and source reveals noisy transitions; in Contextual they are $U$ and $(s,\tau)$, and source reveals selected-value predictions. Reward equivalence concerns the complete permitted feedback interface, while learnability by the specified network additionally requires the finite optimization arguments below. World-independent initialization is part of each no-source lower-bound experiment; arbitrary pretrained weights must instead be counted in $O$.
\subsection{Mechanisms with identical training rewards}
Use the sequential state space, operation tags, training and held-out supports, and terminal verifier of \cref{sec:setup}. A world fixes the transition table $\theta$ and initial operation $\eta=(b_\eta,u_\eta)$. All sums below are in $\mathbb F_2$.

\begin{theorem}[Mechanisms with identical training rewards]\label{thm:quotient-formal}
Let $H\ge3$ and fix $u_\eta$. Two worlds are reward-equivalent if and only if, for some $\kappa,d\in\mathbb F_2$,
\begin{equation}\label{eq:gauge-formal}
 \theta'_{u,z}=\theta_{u,z}+\kappa+d\,\ell(u),\qquad
 b'_\eta=b_\eta+(H\bmod2)\,\kappa+d\,\ell(u_\eta).
\end{equation}
Their targets differ on every test prompt exactly when $d=1$.
\end{theorem}

The proof is given in \cref{app:quotient}.

\subsection{Exact identification from source and terminal observations}\label{app:statistical}
This appendix proves the statistical results. Throughout, $L=H-1$ is the suffix length, $m=|Z|$, and all additions of bits or coordinates are in $\mathbb F_2$. Matrix norms and inner products are over the reals. The matrix arguments are analysis of a binary-equation learner; they do not grant it an additional oracle.

The iid-prompt restriction makes recovery quantitative. A symbolic learner first tests public binding coordinates with random endpoint bits. Once that coordinate is identified, one terminal reply reveals one exact binary equation in the unknown transition table and binary binding coordinate. After solving these equations, the learner uses its stored source records to recover $\kappa,d$.

Canonical execution starts with $\eta$ and applies the true transition at every step; $\delta$ below is the allowed failure probability.

\begin{theorem}[A matched low-source query regime]\label[theorem]{thm:resource}
Let $L=H-1\ge\lceil2^{16}\ln(2m)\rceil$ and $0<\delta<1$. The explicit fresh-iid learner with budgets \eqref{eq:rank-budget} and \eqref{eq:source-budget} has, for each fixed world, probability at least $1-\delta$ of correct canonical execution on every test prompt. Its endpoint success, target first-state retention and task utility then equal one. It uses
\begin{align}
 N_{\rm pre}&\le32a^{-2}\ln(6/\delta)+2H+2,\\
 Q&=O\!\left(m^2+\left(m+\frac{m^2}{L}\right)\ln(m/\delta)
                       +m\ln(1/\delta)\right),
\end{align}
with $nQ$ generated decisions, $HQ$ prompt tokens and zero gradient updates.
At fixed $\delta=1/16$, fixed $c\in(0,1/64]$, $a=1-c/H$ and
$L=\lceil2^{16}\ln(2m)\rceil$, this is $O(\log m)$ raw source tokens and $O(m^2)$ queries.
For $m\ge512$, any learner with $n_{\rm rec}$ source records and worst-case expected endpoint risk at most $1/16$ must satisfy
\begin{equation}\label{eq:main-lower}
 Q\ge\frac5{896}m^2-n_{\rm rec}H-1,
\end{equation}
even when granted the binding and adaptive chosen training prompts. Hence the query coordinate is minimax-optimal in this fixed-confidence, logarithmic-source-budget and logarithmic-horizon regime.
\end{theorem}

The low-error structure is the key to the upper bound. For the binary entrywise difference $f$ between two transition tables, form a signed transition matrix $A(z,z')=(-1)^{f(z'+z,z)}/m$, indexed by public states $z,z'$ and its tagged counterpart. Their powers describe the probability of agreement along a random suffix. Writing $\varepsilon$ for training-endpoint disagreement probability, near-perfect agreement forces one matrix close to a signed rank-one form. After sign rounding, only $O(\varepsilon m^2/L)$ transition defects and $O(\varepsilon m)$ vertex-sign choices remain at error $\varepsilon$. The resulting small hypothesis count makes $O(m^2)$ iid equations sufficient when $L=\Theta(\log m)$. A separate short-witness argument rules out arbitrarily small nonzero error before applying this rounding estimate.

The binding prefix has its own sampling law. Retaining its factor throughout the suffix approximation controls accuracy on this distinguished row. Finally, bounded source sign statistics recover the two residual bits. The source is independent of the feedback calculation because it is stored but not used during that calculation. The lower bound packs $2^{\Theta(m^2)}$ separated mechanisms and counts at most $n_{\rm rec}H+Q$ information bits. The proofs and complete computation/storage charges are in \cref{app:statistical}.

The logarithmic source-token charge includes the length of a mandatory whole record. At fixed confidence and fixed noise level, decoding the two ambiguity parameters needs only a constant-order number of informative transitions. Thus record length and the information needed for the two residual bits make distinct contributions to the token count.

\begin{lemma}[The complete two-bit kernel]\label[lemma]{lem:seq-quotient}\label[lemma]{app:quotient}
In the sequential model of \cref{sec:setup}, fix $u_\eta$ and let $H\ge3$. The two worlds have equal endpoints on all training prompts if and only if their difference has the form \eqref{eq:gauge-formal}. Such a difference changes every test endpoint by the bit $d$. Equality of these endpoints is equivalent to equality of the full terminal verification oracle.
\end{lemma}
\begin{proof}
Fix the public binding coordinate $u_\eta$. Write $f(u,z)=\theta'_{u,z}+\theta_{u,z}$ and $\beta=b'_\eta+b_\eta$. At suffix-start coordinate $z$, equality of all training endpoints is equivalent to
\begin{equation}\label{eq:quotient-path}
 \beta+f(u_\eta,z+u_\eta)+\sum_{j=1}^{L}f(v_j,z_{j-1})=0,
 \qquad z_j=z_{j-1}+v_j,\quad \sum_j\ell(v_j)=0.
\end{equation}
Consequently every admissible length-$L$ path starting at $z$ has the same cost. Its all-zero path has cost $Lf(0,z)$. For $\ell(u)=0$, compare $(u,0,\ldots,0)$ with $(0,u,0,\ldots,0)$. This gives $f(0,z)=f(0,z+u)$. The set $\ell^{-1}(0)$ spans $Z$: it contains the coordinate vectors in the $x$ and $y$ components and $t+x_1+y_1$, whose tag is zero. Thus $f(0,z)=\kappa$ is constant. Comparing a single tag-zero control followed by zeros with the all-zero path gives $f(u,z)=\kappa$ for every tag-zero control.

For tag-one controls $u,v$, the path $(u,v,0,\ldots,0)$ has tag zero. Comparing with the all-zero path gives $f(u,z)=f(v,z+u)$. At any fixed vertex all tag-one outgoing costs therefore agree; call their value $a(z)$. The same equality gives $a(z)=a(z+u)$ for each tag-one $u$. The tag-one controls contain $t,t+x_i,t+y_i$ and span $Z$, so $a(z)=\lambda$ is constant. Hence
\begin{equation}
 f(u,z)=\kappa+d\ell(u),\qquad d=\kappa+\lambda.
\end{equation}
Substitution into \eqref{eq:quotient-path} forces $\beta=H\kappa+d\ell(u_\eta)$. Conversely these differences cancel on every tag-zero suffix and produce the difference $d$ on every tag-one suffix. This argument requires only $L\ge2$.

Endpoint equality is equivalent to equality of the full binary verification oracle once $u_\eta$ agrees. Invalid grammar or public-coordinate paths are rejected in both worlds. On any given prompt one can instead choose a valid public-coordinate path and either final state bit. Exactly the correct endpoint is accepted, independently of the intermediate state bits. These two candidates distinguish unequal endpoints. This proves the claimed equivalence for the actual verifier, including repaired trajectories.
\end{proof}

\begin{proof}[Proof of \cref{thm:quotient-formal}]
With $L=H-1\ge2$, \cref{lem:seq-quotient} gives both directions of \eqref{eq:gauge-formal} and equivalence for the actual verifier. On every test prompt the remaining endpoint difference is $d$, so all test targets differ exactly when $d=1$.
\end{proof}

\begin{lemma}[Binding-coordinate identification and exact equations]\label[lemma]{lem:seq-binary-equations}\label[lemma]{app:equations}
For $0<\delta<1$, $Q_{\rm id}=m\lceil\log_2(3/\delta)\rceil$ fresh iid queries identify $u_\eta$ with failure probability at most $\delta/3$. Conditional on successful identification, each subsequent fresh prompt and binary reply yield the exact iid binary equation \eqref{eq:binary-row} in $m^2+1$ unknown bits.
\end{lemma}
\begin{proof}
First identify $u_\eta$. For each of its $m$ possible values, use at most
$k=\lceil\log_2(3/\delta)\rceil$ fresh iid training prompts. Output a grammar-compliant candidate with the required public coordinates for that coordinate and a fresh fair final state bit. A wrong coordinate never succeeds; the correct coordinate succeeds with probability $1/2$ independently on each trial. Stop at the first success, or declare failure after the fixed budget. The failure probability is at most $2^{-k}\le\delta/3$, and the query cost is at most $Q_{\rm id}=mk$.

Thereafter choose $A=(0,u_\eta)$ and a compliant trajectory with final state bit zero. Its binary reward reveals the true endpoint bit: reward one means bit zero, and reward zero means bit one. Subtract the known initial state bit and the displayed control bits. The remaining label is
\begin{equation}\label{eq:binary-row}
 b_\eta+\theta_{u_\eta,z_0}
       +\sum_{i=2}^{H}\theta_{u_i,z_{i-1}},
 \qquad z_1=z_0+u_\eta,\quad z_i=z_{i-1}+u_i.
\end{equation}
Each observation is a known binary row in $m^2+1$ unknown bits. The rows are iid because prompts remain fresh. Gaussian elimination returns any consistent representative. It remains to show that a controlled number of these particular path rows eliminates every difference outside the two-bit kernel. Counting the number of unknown bits alone does not establish that assertion.
\end{proof}

\begin{lemma}[Tagged convolution norm]\label[lemma]{lem:seq-tagged-norm}\label[lemma]{app:coarse}
For the sequential tag $\ell(t,x,y)=t+x\cdot y$, the matrix $M_\ell(z,w)=m^{-1}(-1)^{\ell(w+z)}$ has operator norm $\rho=2^{-r}\le1/2$.
\end{lemma}
\begin{proof}
Define the signed convolution
\begin{equation}
 M_\ell(z,w)=m^{-1}(-1)^{\ell(w+z)}.
\end{equation}
At Walsh frequency $(\alpha,\beta,\tau)$ its eigenvalue is the average of
$(-1)^{t+x\cdot y+\alpha\cdot x+\beta\cdot y+\tau t}$.
It vanishes when $\tau=0$. When $\tau=1$, the sum over $y$ forces $x=\beta$ and leaves $2^{-r}(-1)^{\alpha\cdot\beta}$. Thus $\norm{M_\ell}=\rho=2^{-r}\le1/2$.
\end{proof}

\begin{lemma}[Short witness or vertex potential]\label[lemma]{lem:seq-short-witness}
Every binary transition cost $f$ either has two length-nine control blocks with the same start, end and tag but opposite costs, or admits \eqref{eq:potential} for a vertex potential $h:Z\to\mathbb F_2$ and bits $\kappa,d$.
\end{lemma}
\begin{proof}
For any cost $f$, either two length-nine blocks have the same start, end and tag but opposite costs, or
\begin{equation}\label{eq:potential}
 f(u,z)=\kappa+h(z)+h(z+u)+d\ell(u).
\end{equation}
To see this, comparisons $(0,u)$ versus $(u,0)$ detect nonconstant self-loop costs. Without such a witness, subtract their constant $\kappa$. Comparisons $(u,u)$ versus $(0,0)$ then detect unequal forward and reverse costs. If those also agree, use the undirected lifted graph on $(z,e)\in Z\times\mathbb F_2$, whose $u$-edge leads to $(z+u,e+\ell(u))$. A direct edge gives any public displacement; the three-edge loop $(x_1,y_1,x_1+y_1)$ has zero displacement and odd tag. The graph therefore has diameter at most four. A breadth-first spanning tree has depth at most four. A non-tree edge violating a potential produces a fundamental closed walk of length at most nine with odd cost. Pad this walk and the zero walk to length nine. If no such violation occurs, the graph has a potential $G(z,e)$. Since costs do not depend on $e$, the difference $G(z,e+1)+G(z,e)$ is constant along edges, hence is a constant $d$. Writing $G(z,e)=h(z)+de$ proves \eqref{eq:potential}. All earlier short witnesses can likewise be padded because their endpoints agree.
\end{proof}

\begin{lemma}[Coarse separation of nonkernel differences]\label[lemma]{lem:seq-coarse-separation}
For $L\ge10$, every binary difference hypothesis outside the kernel in \cref{lem:seq-quotient} has training-endpoint disagreement probability at least $m^{-10}$.
\end{lemma}
\begin{proof}
Use the alternatives in \cref{lem:seq-short-witness}. Place either witness block at the beginning of the suffix and fix its public starting vertex. At least one suffix control remains when $L\ge10$. Conditioning the entire suffix tag to zero leaves the probability of each fixed block and starting vertex exactly $m^{-10}$. Couple the remaining controls identically. The endpoint-difference parities of the paired words disagree, so at least one is wrong. The error is at least $m^{-10}$; the two block probabilities already pay for the pairing, with no additional factor $1/2$.

For \eqref{eq:potential}, the full endpoint difference is
\begin{equation}
 \beta+H\kappa+h(z_0)+h(z_H)+d\ell(u_\eta).
\end{equation}
The conditional final public-coordinate law is
$m^{-1}+M_\ell^L(z_0+u_\eta,z_H)$. Every entry is at least
$m^{-1}-2^{-rL}\ge(2m)^{-1}$ for $L\ge10$. A nonconstant $h$ therefore gives error at least $(2m)^{-1}$. A constant $h$ gives a kernel difference or error one. In particular every nonkernel hypothesis has training disagreement at least $m^{-10}$.
\end{proof}

\begin{lemma}[Low-error count and refined separation]\label[lemma]{lem:seq-low-count}\label[lemma]{app:counting}
Assume $L\ge\lceil2^{16}\ln(2m)\rceil$. For $0<\varepsilon\le1/800$, the number of difference hypotheses with error at most $\varepsilon$ is bounded by
\begin{equation}\label{eq:low-count}
 16\exp\!\left\{160\varepsilon\left(m+\frac{m^2}{L}\right)\ln(m+1)\right\}.
\end{equation}
Moreover every nonkernel difference has error at least
\begin{equation}\label{eq:epsilon-star}
 \varepsilon_* =\min\left\{\frac{L}{24m^2},\frac1{52m},\frac1{800}\right\}.
\end{equation}
\end{lemma}
\begin{proof}
Here is the quantitative argument. Put
$A(z,w)=(-1)^{f(w+z,z)}/m$ and $A_c(z,w)=(-1)^{\ell(w+z)}A(z,w)$.
Both have Frobenius norm one. Keep the binding factor
$v(z)=(-1)^{\beta+f(u_\eta,z+u_\eta)}$. Then
\begin{equation}\label{eq:signed-agreement}
 1-2\varepsilon=m^{-1}\ip{v}{(A^L+A_c^L)\mathbf1}.
\end{equation}
Select the branch of larger operator norm $a_0$. It satisfies
$a_0^L\ge(1-2\varepsilon)/2\ge1/4$, so
$\delta_A=1-a_0^2\le2\ln4/L$. Its best rank-one singular approximation has squared residual $\delta_A$. Round its left and right singular vectors to signs $h,k$. Each incorrectly rounded matrix entry contributes at least $1/m^2$ to this residual. The rounded matrix
$P(z,w)=h(z)k(w)/m$ thus differs in at most $\delta_A m^2$ signs and in Frobenius norm by at most $2\sqrt{\delta_A}$.
The complementary tagged branch has norm at most
$\rho+2\sqrt{\delta_A}\le5/8$.
Consequently
$a_0^L\ge1-2\varepsilon-(5/8)^L$.
For a nonkernel difference the preceding coarse bound gives
$\varepsilon\ge m^{-10}$, whereas $(5/8)^L\le m^{-20}$.
Taking logarithms yields
\begin{equation}\label{eq:round-defects}
 \delta_A\le12\varepsilon/L,
 \qquad \#\{f\ne f_0\}\le12\varepsilon m^2/L,
\end{equation}
where $f_0$ is the rounded separable cost, with the chosen tag branch restored.
Zero singular-vector coordinates cause no problem: any wrong sign still has the stated residual cost. Kernel differences may be counted separately below the coarse separation.

Retain the \emph{original} factor $v$ and replace only the suffix costs by $f_0$.
Before tag conditioning every suffix edge is uniform on $m^2$ pairs; conditioning costs at most a factor two. A union bound over $L$ edges and \eqref{eq:round-defects} bounds the change of suffix parity by $24\varepsilon$. The auxiliary predictor therefore has error at most $25\varepsilon$. This replacement does not require a bound on the distinguished binding row.

Write $g=kh$, with pointwise multiplication. The exact rank-one identity is
\begin{equation}
 P^L\mathbf1=h\,\overline{kh}^{\,L-1}\bar k.
\end{equation}
The complementary signed matrix of $P$ has norm $\rho$. Thus auxiliary agreement is at most
$|\bar k|\,|\bar g|^{L-1}+\rho^L$.
Since $\rho^L\le\varepsilon$, the elementary logarithmic bounds from agreement at least $1-50\varepsilon$ imply
\begin{equation}
 \operatorname{minority}(k)\le26\varepsilon m,
 \qquad \operatorname{minority}(g)\le102\varepsilon m/L.
\end{equation}
Choose the branch, the two global signs and $\beta$ in at most $16$ ways. The remaining choices are bounded by
\begin{align}
16&\left[\sum_{i\le\lfloor26\varepsilon m\rfloor}\binom{m}{i}\right]
\left[\sum_{j\le\lfloor102\varepsilon m/L\rfloor}\binom{m}{j}\right]
\left[\sum_{d\le\lfloor12\varepsilon m^2/L\rfloor}\binom{m^2}{d}\right].
\end{align}
Using $\sum_{j\le t}\binom{n}{j}\le(n+1)^t$ for integral $t\ge0$ gives \eqref{eq:low-count}; the coefficient $160$ is conservative. If $\varepsilon<\varepsilon_*$, all three integer radii are zero. Then $k,g$ are constant, there are no defects, and $f$ is a constant plus a tag. Its binding bit gives either the kernel or error one. This proves \eqref{eq:epsilon-star}.

\end{proof}

\begin{lemma}[Finite iid schedule for coset recovery]\label[lemma]{lem:seq-rank-schedule}\label[lemma]{app:schedules}
Under the horizon assumption of \cref{lem:seq-low-count}, the iid binary-row learner of \cref{lem:seq-binary-equations} with $Q_{\rm rank}$ in \eqref{eq:rank-budget} recovers the correct observable coset with conditional failure probability at most $\gamma=\delta/3$.
\end{lemma}
\begin{proof}
Let $D=m+m^2/L$, $B=2+\lceil\log_2(52m^2)\rceil$ and $\gamma=\delta/3$. Use
\begin{equation}\label{eq:rank-budget}
Q_{\rm rank}=\left\lceil\max\left\{
\begin{aligned}
&800\big[(m^2+1)\ln2+\ln(2/\gamma)\big],\\
&640D\ln(m+1)+\varepsilon_*^{-1}\ln(32B/\gamma)
\end{aligned}\right\}\right\rceil.
\end{equation}
The $2^{m^2+1}$ hypotheses of error at least $1/800$, including equality, survive with total probability at most $\gamma/2$. Partition the remaining interval $[\varepsilon_*,1/800)$ into at most $B$ dyadic bins with lower endpoints $e_j$, truncating the final bin. Equation~\eqref{eq:low-count} bounds each bin's size by $16\exp[320e_jD\ln(m+1)]$; survival is at most $\exp(-Q_{\rm rank}e_j)$. Equation~\eqref{eq:rank-budget} makes the contribution at most $\gamma/(2B)$ per bin. If $\varepsilon_*=1/800$, the low-error interval is empty. Thus elimination recovers the correct coset with failure at most $\gamma$.
\end{proof}

\begin{lemma}[Recovering both gauges from retained source records]\label[lemma]{lem:seq-source-gauges}
Let $a>0$, $L=H-1$, and suppose the coordinate/rank procedure has recovered the correct coset without using the independently stored source records. With $n_{\rm rec}$ in \eqref{eq:source-budget}, both residual gauge bits are recovered with conditional failure at most $\delta/3$. Correcting the representative then gives the true binding bit and transition table, hence canonical execution and the target first state on every test prompt.
\end{lemma}
\begin{proof}
Store raw source records before feedback but do not use them in the coordinate/rank procedure. Conditional on a fixed world and the feedback representative $\theta^0$, raw source transitions remain independent of that procedure. For an observed transition from $(s,z)$ under $(b,u)$ to a state with state bit $s'$, set
\begin{equation}
 R=s'+s+b+\theta^0_{u,z},\qquad
 X_g=\mathbf1\{\ell(u)=g\}(-1)^R,\quad g\in\{0,1\}.
\end{equation}
On coset recovery, $R=\kappa+d\ell(u)$ in the true-transition component, and it is a fair bit in the reset component, even when the public coordinate is wrong. For a sequentially uniform control,
$\E[X_g\mid\text{past}]=(a/2)(-1)^{\kappa+dg}$.
Use the first $H-1$ execution transitions in each record. The control $U_1$ is independent uniform, and $U_2,\ldots,U_{H-1}$ are a strict subset of the tag-conditioned suffix and remain sequentially uniform. Marginalize the unused final control. This is an execution-order proof filtration, not a claim that the raw serialization reveals variables in that order.

For $n_{\rm rec}$ records, bounded martingale differences give a two-sign error probability at most
$2\exp[-n_{\rm rec}(H-1)a^2/8]$. Taking
\begin{equation}\label{eq:source-budget}
 n_{\rm rec}=\left\lceil\frac{8a^{-2}\ln(6/\delta)}{L}\right\rceil
\end{equation}
makes this at most $\delta/3$. Both gauges, including the endpoint-irrelevant common gauge, are recovered. Correcting $\theta^0$ and the binding bit therefore gives the true first-state target and the full canonical execution on every test prompt. Coordinate, rank and source failure allowances sum to $\delta$.
\end{proof}

\begin{lemma}[Complete low-source learner and its resources]\label[lemma]{prop:seq-low-source-resources}
Combining \cref{lem:seq-binary-equations,lem:seq-rank-schedule,lem:seq-source-gauges} with their three failure allocations gives simultaneous canonical execution on every test prompt with probability at least $1-\delta$. Its source tokens are at most $32a^{-2}\ln(6/\delta)+2H+2$, and its query, generation, computation and storage costs are the quantities below.
\end{lemma}
\begin{proof}
The three failure allowances sum to $\delta$, and on their common success event all test executions are correct by \cref{lem:seq-source-gauges}. The raw token cost is
\begin{equation}
 N_{\rm pre}=(2H+2)n_{\rm rec}
 \le32a^{-2}\ln(6/\delta)+2H+2.
\end{equation}
The query cost is $Q=Q_{\rm id}+Q_{\rm rank}$, generation costs $(H+2)Q$ decisions, and prompts contribute $HQ$ tokens. There are zero gradient updates and no checkpoint or test selection. Dense binary elimination costs conservatively
$O(Q(H\log q+m^4)+N_{\rm pre}\log q)$ bit operations, with
$O(m^4+H\log q+N_{\rm pre}\log q)$ stored bits when raw source is retained. The verifier separately performs $H$ trusted transitions and $H$ public-coordinate checks per query. Deployment uses $H$ table lookups and $O(H)$ generated symbols; no external shield is inserted.
\end{proof}

\begin{lemma}[Packing separated endpoint functions]\label[lemma]{lem:seq-packing}
For $m\ge512$ and $H\ge5$, restrict to mechanisms with
$\theta_{0,v}\ne\theta_{0,0}$ for a fixed $v\ne0$. There exists a packing of
$K_{\rm pack}=2^{m^2/128}$ mechanisms with pairwise endpoint disagreement at least $7/16$, even with a fixed known binding. \end{lemma}
\begin{proof}
We give the elementary packing calculation to specify the stronger information permissions of the lower bound.

For a random pair of such tables, its difference matrix has independent fair signs except for one tied entry. For the signed matrices $A,A_c$ above, couple each to an iid sign matrix; changing the tied entry costs at most $2/m$ in operator norm. Two $1/4$-nets of size $9^m$ and scalar Hoeffding bounds yield
\begin{equation}
 \Pr\!\left[\max(\norm A,\norm{A_c})>\tfrac12\right]
 \le4\exp\!\left[2m\ln9-\frac{(m-4)^2}{32}\right]
 \le e^{-m^2/64}.
\end{equation}
A union bound over fewer than $K_{\rm pack}^2$ pairs proves existence. The fixed initial binding contributes the diagonal sign matrix $D_\eta(z,z)=(-1)^{f(u_\eta,z)}$ and permutation $P_u(z,z')=\mathbf1\{z'=z+u\}$, both orthogonal. The signed test agreement is
\begin{equation}
 m^{-1}\mathbf1^\top D_\eta P_{u_\eta}
       (A^{H-1}-A_c^{H-1})\mathbf1,
\end{equation}
whose magnitude is at most $2^{2-H}$. Thus disagreement is at least
$(1-2^{2-H})/2\ge7/16$.

\end{proof}

\begin{lemma}[Mixed source-and-feedback information lower bound]\label[lemma]{prop:seq-mixed-lower}\label[lemma]{app:packing}
For $m\ge512$ and $H\ge5$, every learner with $n_{\rm rec}$ source records and at most $Q$ binary verification replies satisfies \eqref{eq:packed-risk}, even with known binding and adaptively chosen training prompts. In particular, worst-world expected endpoint risk at most $1/16$ requires \eqref{eq:main-lower}.
\end{lemma}
\begin{proof}
Use the packing of \cref{lem:seq-packing}. Let $J$ be a uniformly chosen packing index. Reveal the learner's independent random seed at the start of the transcript and condition on it; subsequent randomized choices are then deterministic functions of that seed and preceding observations. Raw controls and resets convey no mechanism information. Conditional public-coordinate transitions have a mechanism-independent law; each next state bit conveys at most one bit. Hence $n_{\rm rec}$ source records convey at most $n_{\rm rec}H$ bits. Any adaptive binary verification replies add at most $Q$ bits, even when the learner may choose its training prompts. Conditional on this seed and the preceding transcript, the chosen queries supply no additional mechanism information. Therefore
$I(J;\text{transcript})\le n_{\rm rec}H+Q$ bits.

Decode the closest packing endpoint function from the learned predictor. A mistaken index forces endpoint error at least $7/32$ by the triangle inequality. For a randomized predictor use its conditional output probabilities as a randomized endpoint function; malformed outputs only add error. The entropy encoding bound
$H(J\mid\text{transcript})\le1+p_{\rm err}\log_2K_{\rm pack}$ gives
\begin{equation}\label{eq:packed-risk}
 \sup_{\theta,\eta}\E[\text{endpoint risk}]
 \ge\frac7{32}\left[1-\frac{128(n_{\rm rec}H+Q+1)}{m^2}\right]_+.
\end{equation}
For worst-case risk at most $1/16$, this implies
$Q\ge5m^2/896-n_{\rm rec}H-1$.
\end{proof}

\begin{lemma}[Matched query order at logarithmic source budget]\label[lemma]{cor:seq-matched-query}
For $\delta=1/16$, fixed $c\in(0,1/64]$, $a=1-c/H$, and $L=\lceil2^{16}\ln(2m)\rceil$, the sequential learner has $N_{\rm pre}=O(\log m)$ and $Q=O(m^2)$, while every learner at this source budget with worst-world expected endpoint risk at most $1/16$ has $Q=\Omega(m^2)$ as $m\to\infty$.
\end{lemma}
\begin{proof}
At fixed confidence $\delta=1/16$, fixed $c\in(0,1/64]$,
$a=1-c/H$ and $L=\lceil2^{16}\ln(2m)\rceil$, the upper has
$N_{\rm pre}=O(\log m)$ and $Q=O(m^2)$. Its perfect-recovery event implies the required expected-risk upper. Equation~\eqref{eq:packed-risk} gives $Q=\Omega(m^2)$ at that source budget. This matches the query coordinate in this regime, while retaining the whole-record cost in the source coordinate.
\end{proof}

\begin{proof}[Proof of \cref{thm:resource}]
\Cref{prop:seq-low-source-resources} supplies the stated success event and all operational costs. By \eqref{eq:epsilon-star},
$\varepsilon_*^{-1}=\max\{24m^2/L,52m,800\}$, and the horizon condition implies $L\ge1$. Substitution into \eqref{eq:rank-budget}, together with $Q_{\rm id}=m\lceil\log_2(3/\delta)\rceil$, yields
$Q=O(m^2+(m+m^2/L)\ln(m/\delta)+m\ln(1/\delta))$; the factor $\ln B=O(\ln m)$ is absorbed here. Exact canonical execution gives endpoint success, target first-state retention and task utility one on the same event. \Cref{prop:seq-mixed-lower} supplies \eqref{eq:main-lower}, and \cref{cor:seq-matched-query} supplies the fixed-confidence specialization and query optimality.
\end{proof}

\subsection{A finite-budget boundary for a known pair of worlds}\label{app:rare-leakage}
We isolate the gauge ambiguity from the cost of learning an unknown transition table.
The following experiment supplies the learner with two candidate worlds and asks which one generated its observations. It does not supply the true member. This restriction permits an exact boundary; it is not an upper bound for the full unknown-table class.

\begin{definition}[Known-pair sequential observation experiment]\label[definition]{def:rare-experiment}
Fix $H\ge3$ and a sequential world $(\theta^0,\eta^0)$.
Let $\theta^1_{u,z}=\theta^0_{u,z}+\ell(u)$,
$u_{\eta^1}=u_{\eta^0}$ and
$b_{\eta^1}=b_{\eta^0}+\ell(u_{\eta^0})$, with all additions in $\mathbb F_2$.
Both tables and both bindings are known candidates. The unknown index is $W\in\{0,1\}$.
Let $D_0,D_1$ be the common prompt laws conditional on even and odd suffix tag parity, respectively.
An initial observation $O$ has law $P_w$ in world $w$ and includes all source records, inherited checkpoint weights and other side information jointly. Take its observation space to be standard Borel.
After $O$, at most $Q\in\mathbb N_0$ verifier queries receive fresh exogenous iid prompts from
$D_\varepsilon=(1-\varepsilon)D_0+\varepsilon D_1$, $0\le\varepsilon\le1$.
The prompt sequence is independent of $(W,O)$; every presented prompt costs a query, including a rejection. At a prompt, the learner chooses a candidate trajectory adaptively and receives the deterministic endpoint verifier's reply. There is no uncharged prompt selection.
The final test prompt has law $D_1$ independently of training and $W$.
The loss is endpoint failure, including malformed outputs. Procedures may randomize and have no computational restriction.
Write $\mu=P_0+P_1$, $p_w=dP_w/d\mu$, and
$\alpha=\int\min(p_0,p_1)\,d\mu=1-\operatorname{TV}(P_0,P_1)$.

\end{definition}

\begin{proposition}[Exact pair risk under rare informative prompts]\label[proposition]{prop:rare-pair}
In \cref{def:rare-experiment}, for a prior $\Pr(W=0)=\pi$, the optimal Bayes risk is
\begin{equation}\label{eq:rare-bayes}
 R_\pi^*(Q)=(1-\varepsilon)^Q
 \int\min\{\pi p_0,(1-\pi)p_1\}\,d\mu.
\end{equation}
Here $(1-\varepsilon)^0=1$, including at $\varepsilon=1$.
If $r_{\rm mm}(P_0,P_1)$ is the infimum, over randomized tests based on $O$, of their maximum error in the two worlds, then
\begin{equation}\label{eq:rare-minimax}
 R_{\rm mm}^*(Q)=(1-\varepsilon)^Q r_{\rm mm}(P_0,P_1).
\end{equation}
For the equal prior, $R_{1/2}^*(Q)=\tfrac12(1-\varepsilon)^Q\alpha$.
This also equals the minimax risk if a measurable involution of $O$ exchanges $P_0$ and $P_1$.
\end{proposition}

\begin{lemma}[Parity equivalence and an identifying query]\label[lemma]{lem:rare-separating}
In \cref{def:rare-experiment}, all verifier replies coincide in the two worlds on an even prompt. On an odd prompt the canonical trajectory of candidate $0$ receives reply $1-W$, identifying $W$ exactly.
\end{lemma}
\begin{proof}
Public state coordinates are the same in the two worlds.
Their private endpoint difference at a fixed prompt is
\[
 (b_{\eta^1}+b_{\eta^0})+
 (\theta^1_{u_\eta,z_0}+\theta^0_{u_\eta,z_0})+
 \sum_{i=2}^H(\theta^1_{u_i,z_{i-1}}+\theta^0_{u_i,z_{i-1}})
 =\sum_{i=2}^H\ell(u_i).
\]
The first two terms are both $\ell(u_\eta)$ and cancel.
On an even prompt, the endpoint targets and all verifier replies coincide: malformed strings fail in both worlds; admissibility checks the common public path, and the final bit is the same. Intermediate private bits are not verified.
On an odd prompt, submit the canonical trajectory of candidate world $0$. Its public path is legal in either world, while its endpoint is correct precisely when $W=0$. The reply is therefore $1-W$. One odd prompt identifies the candidate world exactly.
\end{proof}

\begin{lemma}[No information gain on the all-even event]\label[lemma]{lem:rare-lower}
In \cref{def:rare-experiment}, every adaptive procedure has Bayes risk at least the right-hand side of \eqref{eq:rare-bayes} and maximum-world risk at least the right-hand side of \eqref{eq:rare-minimax}.
\end{lemma}
\begin{proof}
If $(1-\varepsilon)^Q=0$, both proposed lower bounds are zero and follow from nonnegativity. Otherwise the following conditioning is well defined.
Presample $Q$ prompts even if a procedure stops early. Let $E$ be the event that all are even; $\Pr(E\mid W,O)=(1-\varepsilon)^Q$.
Conditional on $E$, the prompt law is world independent. At query $j$, a common history and the procedure's same random seed induce the same submitted string; \cref{lem:rare-separating} gives the same reply. Induction over $j=1,\ldots,Q$ shows that the full training transcript conditional on $E$ is a common Markov kernel applied to $O$.
The independent test prompt supplies only additional world-independent randomness.
From any output, define a test choosing world $0$ exactly when the output passes its endpoint verifier, and choosing $1$ otherwise. Since the two odd-prompt targets are disjoint, this test's error is at most the output's endpoint failure in each world. Thus, conditional on $E$, no procedure improves on a randomized test based on $O$.
For prior $\pi$, pointwise choice between the two weighted densities gives the error integral in \eqref{eq:rare-bayes}. Dropping nonnegative errors on $E^c$ proves its lower bound.
For the minimax claim, the same induced test has two conditional errors $e_0,e_1$, with $\max_w e_w\ge r_{\rm mm}$. The unconditional maximum error is at least $(1-\varepsilon)^Q\max_w e_w$, proving the lower bound in \eqref{eq:rare-minimax}.
\end{proof}

\begin{lemma}[Attaining pair risk and symmetrizing tests]\label[lemma]{lem:rare-attainment}
In \cref{def:rare-experiment}, the Bayes lower bound in \eqref{eq:rare-bayes} is attained, and the minimax lower bound in \eqref{eq:rare-minimax} is an attainable infimum. If a measurable involution exchanges $P_0,P_1$, their minimax testing risk equals their equal-prior Bayes testing risk.
\end{lemma}
\begin{proof}
Use the separating candidate at the first odd prompt. If there is one, execute the identified world at test time. If none occurs, use a Bayes-optimal test of $O$, or an arbitrarily near-optimal minimax test, and execute the chosen world. At an odd test prompt the output is correct exactly when this choice is correct. These procedures attain the Bayes bound and approach the minimax infimum.
For an involution $T$ exchanging the two laws, transform a test $f$ into $1-f\circ T$; its two errors are those of $f$ in reverse order. A fair mixture has equal errors, each the equal-prior average error of $f$. Applying this to a Bayes-optimal test proves that minimax equals equal-prior Bayes risk under symmetry. Without this assumption, the two quantities need not agree.
\end{proof}

\begin{proof}[Proof of \cref{prop:rare-pair}]
If $\varepsilon=1$ and $Q\ge1$, every prompt is odd; \cref{lem:rare-separating} attains zero risk. In all other cases \cref{lem:rare-lower,lem:rare-attainment} give the matching lower and upper values. For equal prior, the weighted-density integral is $\alpha/2$ by \cref{def:rare-experiment}. The symmetry assertion follows from \cref{lem:rare-attainment}. The convention at $Q=0$ corresponds to no prompt being observed.
\end{proof}

\begin{definition}[Independent-record specialization]\label[definition]{def:rare-source}
Suppose $O$ consists of $n$ independent raw source records from the usual even-suffix source law, together with an independent, world-independent initializer.
The initial state is uniform, $U_1$ is uniform, and the suffix is uniform conditional on even tag parity.
Each state transition uses $aP_U+(1-a)\Pi$ with $0\le a\le1$ and fresh transition noise. The endpoint values $a=0,1$ below concern this information experiment; the finite-training results keep their own prescribed $a=1-c/H$.
Put
\begin{align}
 G_H(z)&=\frac{(1+z)\bigl[(1+z)^{H-1}+(1-z)^{H-1}\bigr]}{2^H},\label{eq:rare-tag-pgf}\\
 h&=a+\frac{2(1-a)}q,\qquad
 r=\frac{a+(1-a)/q}{h},\qquad c_k=[z^k]G_H(z)^n.\label{eq:rare-channel}
\end{align}
For $j\ge0$ define the majority error with a fair tie break by
\begin{equation}\label{eq:rare-majority}
 b_j(r)=\sum_{t=0}^{\lfloor(j-1)/2\rfloor}\binom jt r^t(1-r)^{j-t}
 +\frac{\mathbf1_{\{j\text{ even}\}}}{2}\binom j{j/2}[r(1-r)]^{j/2}.
\end{equation}
For odd $j$, the tie term is absent; $b_0(r)=1/2$ and an empty sum is zero.
\end{definition}
\begin{corollary}[Source records and reward queries for the pair]\label[corollary]{cor:rare-source}
Under \cref{def:rare-source}, the equal-prior Bayes and minimax risks both equal
\begin{equation}\label{eq:rare-source-risk}
 (1-\varepsilon)^Q
 \sum_{k=0}^{nH}c_k\sum_{j=0}^k\binom kj h^j(1-h)^{k-j}b_j(r).
\end{equation}
In the noiseless case $a=1$, this reduces to
\begin{equation}\label{eq:rare-noiseless}
 R^*(n,Q)=\tfrac12\,2^{-n(H-1)}(1-\varepsilon)^Q.
\end{equation}
Consequently, in the noiseless case $a=1$, for $0<\delta<1/2$ and $0\le\varepsilon<1$, error at most $\delta$ is possible exactly when
\begin{equation}\label{eq:rare-budget}
 n(H-1)\log2+Q[-\log(1-\varepsilon)]\ge\log\frac1{2\delta}.
\end{equation}
\end{corollary}

\begin{lemma}[Symmetry of complete source records]\label[lemma]{lem:rare-source-symmetry}
Under \cref{def:rare-source}, there is a measurable involution of all observed records and the independent initializer exchanging $P_0$ and $P_1$. Hence equal-prior Bayes and minimax risks agree.
\end{lemma}
\begin{proof}
For a record with states $(s_j,z_j)$ and controls $(b_j,u_j)$, leave controls and public states fixed and replace each private state by
$s_j+\sum_{i=1}^j\ell(u_i)$, leaving $s_0$, PAD and EOS fixed.
Applying this map twice is the identity. A transition following $\theta^0$ becomes a transition following $\theta^1$; a uniform reset remains uniform. The control law is unchanged, so the map exchanges the two complete-record laws. Apply it separately to all records and leave the independent initializer fixed. The resulting involution exchanges $P_0,P_1$; \cref{prop:rare-pair} therefore equates Bayes and minimax risk.
\end{proof}

\begin{lemma}[The source innovation channel]\label[lemma]{lem:rare-innovations}
Under \cref{def:rare-source}, conditional on the controls, the sequential innovation transformation displayed below is invertible and has a product likelihood. A tag-zero position is uninformative. Each tag-one position independently gives an erasure with probability $1-h$ or a binary observation equal to $W$ with probability $r$, conditional on not being erased, with $h,r$ from \eqref{eq:rare-channel}.
\end{lemma}
\begin{proof}
Given the controls and initial state, define innovations relative to candidate $0$:
\[
 e_j^z=z_j-z_{j-1}-u_j,\qquad
 e_j^s=s_j-s_{j-1}-b_j-\theta^0_{u_j,z_{j-1}}.
\]
This is an invertible sequential transformation of the observed states.
When $\ell(u_j)=0$, the two conditional innovation laws coincide.
When $\ell(u_j)=1$, the innovation $(e_j^z,e_j^s)=(0,w)$ has probability $a+(1-a)/q$ in world $w$; each of the other $q-1$ innovations has probability $(1-a)/q$.
These probabilities do not depend on previous states. Iterating the conditional factorization shows that, given the controls, the innovation likelihood factors over positions.
An innovation with $e_j^z\ne0$ has world-independent likelihood.
The event $e_j^z=0$ has probability $h$ in either world; conditional on it, the private innovation equals $W$ with probability $r$. Thus each tag-one position gives either an erasure or an independent binary observation with correctness $r\ge1/2$. Tag-zero positions give no information. Because the erasure and control laws do not depend on $W$, they supply no additional likelihood factor.
\end{proof}

\begin{lemma}[Tag counts and majority error]\label[lemma]{lem:rare-majority}
Under \cref{def:rare-source}, the total number $K$ of tag-one positions has law $\Pr(K=k)=c_k$. Given $K=k$, the number $J$ of nonerased observations is $\operatorname{Bin}(k,h)$. For $J=j$, the optimal equal-prior classification error is $b_j(r)$ in \eqref{eq:rare-majority}, including $j=0$ and $r=1/2,1$.
\end{lemma}
\begin{proof}
The balanced tag $\ell(u)=t+x\cdot y$ is a fair bit under uniform $u$.
For one record, its first tag is an independent fair bit, while its $H-1$ suffix tags are uniform over even-parity bit strings. The generating function of the latter count is
$[(1+z)^{H-1}+(1-z)^{H-1}]/2^{H-1}$.
Multiplication by $(1+z)/2$ gives \eqref{eq:rare-tag-pgf}.
Independent records give $\Pr(K=k)=c_k$ for the total tag-one count.
Given $K=k$, the number $J$ of nonerased observations is $\operatorname{Bin}(k,h)$.
Given $J=j$, the number of correct binary observations is $\operatorname{Bin}(j,r)$.
For $1/2<r<1$, the likelihood ratio increases with the number voting for world $1$, so majority with a fair tie break is Bayes-optimal. At $r=1/2$ all choices have average error $1/2$, and at $r=1$ any nonempty sample identifies $W$; the same formula includes both endpoints.
Its error is exactly \eqref{eq:rare-majority}. Averaging first over $J$, then $K$, and multiplying by the all-even reward probability from \cref{prop:rare-pair} proves \eqref{eq:rare-source-risk}.
\end{proof}

\begin{proof}[Proof of \cref{cor:rare-source}]
\Cref{lem:rare-source-symmetry} gives symmetry, \cref{lem:rare-innovations} gives the channel likelihood, and \cref{lem:rare-majority} gives its count distribution and optimal classification error. Averaging $b_j(r)$ first against $\operatorname{Bin}(k,h)$ and then against $c_k$, and applying \cref{prop:rare-pair}, proves \eqref{eq:rare-source-risk} for both Bayes and minimax risk.

At $a=1$, $h=r=1$. Any tag-one transition identifies the world, so only $K=0$ contributes error $1/2$.
Since $G_H(0)=2^{-(H-1)}$, its probability is $2^{-n(H-1)}$.
This proves \eqref{eq:rare-noiseless}. Taking logarithms of its comparison to $\delta$ gives \eqref{eq:rare-budget}, including $n=0$ or $Q=0$.
\end{proof}

\begin{corollary}[Limiting observation regimes]\label[corollary]{cor:rare-limits}
Under \cref{def:rare-source}, for $n=0$ or $a=0$, the risk is $(1-\varepsilon)^Q/2$.
For $\varepsilon=0$, reward cannot improve the source-only risk; for $\varepsilon=1$ and $Q\ge1$ it is zero.
A checkpoint that already distinguishes the worlds has disjoint initial laws and zero risk even with $Q=0$.
\end{corollary}
\begin{proof}
For $n=0$ the initial laws coincide. At $a=0$ all transitions are uniform resets and the two source laws again coincide. In both cases $\alpha=1$ and symmetry gives the displayed risk by \cref{prop:rare-pair}. The values at $\varepsilon=0$ and $\varepsilon=1,Q\ge1$ follow from its factor $(1-\varepsilon)^Q$. Disjoint initial laws give $\alpha=0$ and identify the world from $O$ alone.
\end{proof}

\Cref{prop:rare-pair} accounts for checkpoint information through its joint law with source, including their dependence. This is an observation-based identification result; the optimization results use their specified random initializations.
The factor $(1-\varepsilon)^Q$ relies on charged, exogenous prompts: freely choosing an odd prompt reduces the identification cost to one query.
The source channel and likelihood test are available because both candidate tables are known. Learning an arbitrary unknown table still incurs the separate identification costs in \cref{thm:resource}; the pair formula does not remove them.

\subsection{Information supplied by contextual source observations}
Use the contextual observation model of \cref{ctx:sec:experiment}, with $\nctx\ge1$, odd cue dimension $k=p-2$, independent uniform contexts and exact numerical source observations. A world fixes the slot orientations $\xi$, contextual signs $s$ and cue vector $\tau$. The error below averages the four held-out amplitude pairs, as well as contexts, queries and cues.

\begin{theorem}[Information supplied by contextual source observations]\label{thm:context-information-formal}\label{ctx:prop:information-formal}
Two worlds are reward-equivalent on every training input if and only if
\begin{equation}\label{eq:context-equivalence-formal}
 \tau'=\varepsilon\tau,\qquad s'_\ell \xi'_\ell=\varepsilon s_\ell \xi_\ell
 \quad(\ell=1,\ldots,{\nctx}),\qquad \varepsilon\in\{\pm1\}.
\end{equation}
The ${\nctx}$ orientations may vary independently; the global sign leaves all predictions unchanged. Flipping an orientation reverses exactly the equal-sign test-amplitude labels in that context.

For the exact numerical source, let ${\Nobs}$ count contextual source records, and let $\mathcal R_{\Nobs}$ be the infimum, over all world-independent learners and finite fixed iid feedback-query budgets, of worst-world expected test error after ${\Nobs}$ iid source records. Then
\begin{equation}\label{eq:context-risk-formal}
 \mathcal R_{\Nobs}=\frac14(1-1/{\nctx})^{\Nobs}.
\end{equation}
For ${\nctx}=1$, the factor is one at ${\Nobs}=0$ and zero at ${\Nobs}\ge1$. For approximation tolerance $0<\epsilon<1$ and $K={\nctx}2^k$ context--cue types, $\lceil K\log(K/\epsilon)\rceil$ iid reward queries attain error at most $\mathcal R_{\Nobs}+\epsilon/2$ for every fixed world. With source observations alone, the minimax error is $1/2$ for every ${\Nobs}$.
\end{theorem}

The proof is given in \cref{app:context-information}.

\subsection{The information supplied by source observations}\label{app:context-information}
We prove \cref{thm:context-information-formal} for the contextual task in \cref{ctx:sec:experiment}. The proof also applies to any fixed $0<\rho_{\rm mem}<1$ and odd $k\ge1$. The count ${\Nobs}$ in this appendix is the number of observed source records.

\begin{definition}[Exact contextual source observation]\label[definition]{def:context-source-observation}
A source record reveals its context $\ell$, the two columns $\kappa_j=Re_j$ of an independent Haar frame in $O(2)$, and $x=U_\ell\kappa_J$ for independent uniform $J\in\{1,2\}$. In the full input these keys and query are embedded in the observed context block. The numerical fields $G_1,G_2$ are independent $N(0,I_{k+2})$ vectors, independent of the world, frame, context and index. The predicted bit is $\sgn(G_{J,1}G_{J,2})$. In particular, the joint source law in one context contains no information about another context's orientation or about $s,\tau$.

\end{definition}

\begin{lemma}[First Fourier coefficients of the cue rule]\label[lemma]{lem:context-fourier}
Let $k\ge1$ be odd, $\tau\in\{\pm1\}^k$, and $Z$ uniform on $\{\pm1\}^k$.
Set $a_\ell=s_\ell \xi_\ell$ and $M_\tau(z)=\sgn(\tau^\top z)$. For a uniform sign vector $Z$,
\begin{equation}\label{eq:cue-fourier}
 \E[M_\tau(Z)Z_i]=\mu_k\tau_i,\qquad
 \mu_k=2^{-(k-1)}\binom{k-1}{(k-1)/2}>0.
\end{equation}
\end{lemma}
\begin{proof}
To derive this identity, absorb $\tau_j$ into each $Z_j$ and condition on the other $k-1$ signs. The conditional average of
$\sgn(Z_i+\sum_{j\ne i}Z_j)Z_i$ is one when the even sum is zero and zero otherwise.

\end{proof}

\begin{lemma}[Contextual reward equivalence and test separation]\label[lemma]{lem:context-equivalence}
Under the contextual training and test distributions of \cref{ctx:sec:experiment}, two worlds have identical correctness rewards for every training input and action if and only if \eqref{eq:context-equivalence-formal} holds. Within this equivalence class, reversing a context orientation reverses precisely that context's equal-sign test-amplitude labels. The common sign transformation preserves every prediction.
\end{lemma}
\begin{proof}
Equality of rewards for every action implies equality of the training labels. Taking their first Fourier coefficients at $q=1$ using \cref{lem:context-fourier} gives
$a_\ell\tau_i=a'_\ell\tau'_i$ for every $\ell,i$. Since all entries are signs, fixing one coordinate gives a common $\varepsilon$ with
$\tau'=\varepsilon\tau$ and $a'=\varepsilon a$. These relations also suffice, proving \eqref{eq:context-equivalence}.

For equal-sign test amplitudes with common sign $d$, the label is
$s_\ell dM_\tau(z)$. Changing $\xi_\ell$ while preserving the training equivalence changes $s_\ell$ by $-\varepsilon$, reversing this label. For opposite-sign amplitudes with first sign $d$, the label is
$a_\ell v_q dM_\tau(z)$ and is invariant. The global transformation $(s,\tau)\mapsto(\varepsilon s,\varepsilon\tau)$ preserves both training and test predictions. Any changed orientation has equal-sign test witnesses, so this exhausts the prediction-inert transformations after test labels are included.
\end{proof}

\begin{definition}[Finite-query contextual minimax risk]\label[definition]{def:context-risk}
Precisely, define
\[
 \mathcal R_{\Nobs}=\inf_{Q\in\mathbb N_0}\ \inf_{\mathcal A:\,\text{at most }Q\text{ iid feedback queries}}
 \ \sup_{\xi,s,\tau}\E[\operatorname{err}(\mathcal A)].
\]
The algorithm and initialization are chosen independently of the world; observations may determine all subsequent computation. Source count ${\Nobs}$ and the feedback cap are fixed. Risk averages test inputs, training observations and algorithmic randomness.

\end{definition}

\begin{lemma}[Error contributed by an unobserved orientation]\label[lemma]{lem:context-missing}
For $\Nobs\in\mathbb N_0$ exact source records and every finite feedback-query budget, the worst-world expected test error is at least $\frac14(1-1/\nctx)^{\Nobs}$. This bound continues to hold for adaptive actions and chosen training prompts. At $\nctx=1,\Nobs=0$, the missing-context probability is interpreted as one.
\end{lemma}
\begin{proof}
For the lower bound, fix $a,\tau$, choose independent uniform orientations $\xi_\ell$, and set $s_\ell=a_\ell \xi_\ell$. All training labels are independent of these orientations. If a context has no source record, its orientation remains uniform conditional on the entire transcript: every observed source record belongs to another context, and every training reward has the same value under its flip. This also holds for adaptive actions or chosen training prompts, since each conditional reply distribution is unchanged. Equal-sign tests have probability $1/2$ and a fair hidden label, giving conditional error at least $1/4$ on a missing context. Its missing probability is $(1-1/{\nctx})^{\Nobs}$. Averaging over the uniform test context, then comparing the worst world to this prior, proves
\[
 \mathcal R_{\Nobs}\ge\tfrac14(1-1/{\nctx})^{\Nobs}.
\]
\end{proof}

\begin{lemma}[Orientation recovery from exact observations]\label[lemma]{lem:context-orientation}
Under \cref{def:context-source-observation}, one record identifies its context orientation almost surely. For any finite number of records, all visited orientations are therefore recovered simultaneously almost surely.
\end{lemma}
\begin{proof}
A visited orientation is identifiable almost surely from exact numerical observations. Its query belongs to $\{\kappa_1,\kappa_2\}$ for the identity and to $\{P\kappa_1,P\kappa_2\}$ for the swap. These sets intersect only at finitely many angles in each component of $O(2)$: a column must be fixed by $P$ or mapped to the other column. These angles have Haar probability zero. Finite ${\Nobs}$ therefore permits exact recovery of every visited orientation, even though $J$ is hidden.
\end{proof}

\begin{lemma}[Finite feedback-table attainment]\label[lemma]{prop:context-table}
Let $K=\nctx 2^k$, $0<\epsilon<1$, and $F_{\Nobs}=\frac14(1-1/\nctx)^{\Nobs}$. Under \cref{def:context-source-observation}, a learner with $\lceil K\log(K/\epsilon)\rceil$ fresh iid binary correctness queries has expected test error at most $F_{\Nobs}+\epsilon/2$ for every fixed world. It uses one binary action and one verification per query, zero optimizer updates, and the finite storage specified below.
\end{lemma}
\begin{proof}
For the upper bound, reduce a feedback prompt $(\ell,q,z)$ to $(\ell,v_qz)$. These reduced cells are uniform over $K={\nctx}2^k$ possibilities, and the label is $a_\ell M_\tau(v_qz)$. Choosing action $+1$ turns its binary correctness reply into the exact label. After $Q$ iid queries,
\[
 \Pr[\text{some cell missing}]\le K e^{-Q/K}.
\]
For $0<\epsilon<1$, take $Q=\lceil K\log(K/\epsilon)\rceil$. Recover visited orientations by \cref{lem:context-orientation}. On full coverage, the learner knows $g_\ell(z)=a_\ell M_\tau(z)$. It predicts $v_qd g_\ell(z)$ for an opposite-sign test pair with first sign $d$. For an equal-sign pair it predicts $\xi_\ell d g_\ell(z)$ when the source context was observed, and an independent fair sign otherwise. Its conditional expected error is exactly $F_{\Nobs}=\frac14(1-1/{\nctx})^{\Nobs}$. Full feedback coverage is independent of source coverage. If feedback coverage fails, fair prediction gives error $1/2$, so the unconditional error is
\[
 F_{\Nobs}+(1/2-F_{\Nobs})\Pr[\text{some cell missing}]
 \le F_{\Nobs}+\epsilon/2.
\]
This bound holds for every fixed world. Taking $\epsilon\downarrow0$ proves the risk infimum; no fixed finite budget is asserted to attain it exactly.

Each query generates one binary action and makes one verification. There are no optimizer updates, test queries or selection queries. The construction stores $K$ labels, $K$ occupancy bits, $O({\nctx})$ orientation and context flags, and $O(\log(Q+1))$ counting bits, in addition to its numerical input and working memory. This finite table establishes information attainability; its exponential dependence on $k$ is separate from the neural learning budgets.
\end{proof}

\begin{lemma}[Source-only error]\label[lemma]{lem:context-source-only}
In the contextual observation model, for every source count $\Nobs$, the source-only minimax expected test error is $1/2$.
\end{lemma}
\begin{proof}
Finally, fix $\xi,\tau$ and randomize the independent context signs $s_\ell$. Source observations have the same distribution for all signs, while every test label changes with its context's sign. The prior-average source-only error is $1/2$, and fair prediction attains that value. This proves the last assertion of \cref{thm:context-information}.
\end{proof}

\begin{proof}[Proof of \cref{thm:context-information-formal}]
\Cref{lem:context-equivalence} proves the reward-equivalence and test-separation assertions, using the cue identity of \cref{lem:context-fourier}. For the risk in \cref{def:context-risk}, \cref{lem:context-missing} gives the lower bound for every finite query cap. \Cref{prop:context-table} gives a world-uniform upper bound $F_{\Nobs}+\epsilon/2$ for each $0<\epsilon<1$ with a finite fixed cap. Taking the infimum over those caps and then $\epsilon\downarrow0$ gives equality. A single context is missing with probability one when no source record is observed and probability zero after any positive source count, giving the stated convention at $\nctx=1$. \Cref{lem:context-source-only} proves the final assertion.
\end{proof}

\section{Learning routes and resource comparisons}\label{app:learning-routes}
\subsection{What additional source data can replace}
The sequential family admits two resource endpoints. With little source, feedback learns the observable transition class and source resolves its remaining ambiguity. With abundant source, the mechanism can be learned first, leaving binding search. The formal guarantees and their exact budget substitutions are given below. At failure probability $\delta_{\rm tab}$ and source correlation $a>0$, a classical table learner uses $O(q^2a^{-2}\log(q/\delta_{\rm tab})+H)$ raw source tokens and at most $q-1$ fresh-prompt verifications, storing $q^2$ successor indices. The additive $H$ pays for the final whole source record.

\begin{table}[t]
\centering\small\setlength{\tabcolsep}{3pt}\renewcommand{\arraystretch}{1.12}
\caption{What source learning leaves for reward queries. The first two routes compare information resources at fixed confidence and the stated noise regimes; the last gives sufficient observation and query budgets for a finite Adam learning path. Here $q$ is the number of states and controls and $H$ is the execution horizon.}
\label{tab:routes}
\begin{tabular}{@{}p{.43\linewidth}p{.29\linewidth}p{.24\linewidth}@{}}
\toprule
Route and remaining reward problem & Observed source tokens & Training verifier queries\\
\midrule
\multicolumn{3}{@{}l}{\textit{Information resources: identify a mechanism and its task binding}}\\[2pt]
Few source observations; reward queries recover the mechanism class & $O(\log q)$ & $\Theta(q^2)$\\
\multicolumn{3}{@{}l}{\quad Logarithmic $H$; $a=1-c/H$ with fixed $c$; \cref{thm:resource}.}\\[3pt]
Learn a transition table first; rewards identify only the initial operation & $O(q^2\log q+H)$ & At most $q-1$\\
\midrule
\multicolumn{3}{@{}l}{\textit{Neural training: acquire execution, then adapt the same parameters}}\\[2pt]
Source Adam and two sampled reward updates & At most $N B_sT_s$ & $2B_R$\\
\bottomrule
\end{tabular}
\par\smallskip\begin{minipage}{\linewidth}\small
For the Adam route, $N$ is tokens per source record, $B_s$ records per source batch, $T_s$ the maximum source batches inspected, and $B_R$ completions per reward batch. There are at most $T_s-1$ source updates. Each of the two reward updates uses $B_R$ trajectories, totaling $2B_R$ verifier queries. Its horizon and rates are those of \cref{thm:neural}. Each verifier query additionally processes $H$ prompt tokens and $H+2$ generated decisions; verifier computation is separate.
\end{minipage}
\end{table}

The query optimum in \cref{thm:resource} applies to its fixed-confidence, logarithmic-source and logarithmic-horizon regime. The neural theorem provides another finite training route with its own horizon and block rates. In the contextual task, \cref{thm:context-information} instead gives an exact source-dependent error floor, approached by a finite label-table learner. Together these results distinguish the cost of identifying the task from the cost of learning a neural computation.

\begin{corollary}[The resource routes in the comparison table]\label[corollary]{cor:learning-resource-routes}
Under the sequential observation model, the low-source row of \cref{tab:routes} holds for the fixed-confidence regime of \cref{thm:resource}. For $H\ge5$, under the independent-triple table learner of \cref{seqadam:app:fb-table} with $a>0$ and $0<\delta_{\rm tab}<1$, the table-first row uses at most $6M_0+2H+2$ raw source tokens, where $M_0=\lceil8q^2a^{-2}\log(2q^3/\delta_{\rm tab})\rceil$, stores $q^2$ successor indices, and uses at most $q-1$ fresh-prompt verifications. Its exact-execution event has probability at least $1-\delta_{\rm tab}$. The Adam row has the budgets and success guarantee of \cref{thm:neural}.
\end{corollary}
\begin{proof}
The low-source row takes $\delta=1/16$, $m=q/2\ge512$, fixed $c\in(0,1/64]$, $a=1-c/H$ and $H-1=\lceil2^{16}\log(2m)\rceil$; its upper bound is fresh-iid and its matching lower bound even permits chosen prompts. It solves binary equations for the observable class and then estimates two source sign statistics. The table-first route uses the retained independent triples in \cref{seqadam:app:fb-table}: $M_0=\lceil8q^2a^{-2}\log(2q^3/\delta_{\rm tab})\rceil$ and at most $6M_0+2H+2$ raw tokens. It stores a transition table and enumerates bindings. Its guarantee is exact execution on the table-recovery event, whose probability is at least $1-\delta_{\rm tab}$. The Adam row instead gives the stated high-probability sampled-success bound under its prescribed horizon, model widths and real-arithmetic updates.
\end{proof}

\subsection{What remains for feedback and its alternatives}
\begin{proposition}[Task parameters absent from source observations]\label[proposition]{prop:source-only-binding}
In the sequential family, a decoder with binding-independent initialization using only task-independent source data and source replay has endpoint success at most $1/q$ averaged over an independent uniform binding. In the contextual family its source-only minimax error is $1/2$.
\end{proposition}
\begin{proof}
Source-only prediction and source replay contain no task-binding information in the sequential family. Under a uniform independent binding, any such decoder averages endpoint success at most $1/q$. In the contextual family, the corresponding source-only minimax error is $1/2$. More source prediction cannot identify those task parameters. For the sequential claim, condition on the mechanism, source observations, decoder randomness and a test prompt. The binding remains uniform on $q$ operations. Applied to the initial state, these $q$ operations give $q$ distinct states: the public component distinguishes different $u$, and the private bit distinguishes the two values of $b$. Every subsequent true transition is a permutation, so the $q$ possible endpoints are distinct. A decoder output can match at most one endpoint, giving success at most $1/q$; malformed outputs cannot increase it. Averaging the conditioned variables preserves this bound. The contextual assertion is \cref{lem:context-source-only}.
\end{proof}

Task labels or expert supervised fine-tuning (SFT) can supply the missing information. Binary correctness in the contextual model already identifies a binary label once the action is known, making supervised adaptation a legitimate comparison.

\begin{definition}[Reward-weighted sampled likelihood]\label[definition]{def:weighted-likelihood}
For batch item $b$, let $R_b$ be its terminal reward, $Y_{b,j}$ its $j$th sampled token, and $\pi_w(Y_{b,j}\mid\mathrm{prefix}_{b,j})$ the next-token probability given the preceding tokens. On the exact baseline-zero REINFORCE batch, the reward-weighted likelihood loss
\[
 -\frac1{B_R}\sum_{b=1}^{B_R}R_b
       \sum_{j=1}^{n}\log\pi_w(Y_{b,j}\mid\mathrm{prefix}_{b,j})
\]
defines the reward-weighted likelihood objective, with sampled tokens and rewards held fixed.
\end{definition}
\begin{proposition}[Equality of single-batch updates]\label[proposition]{prop:weighted-reinforce}
The objective in \cref{def:weighted-likelihood} gives the same update as baseline-zero REINFORCE on that exact batch when optimizer settings and state agree.
\end{proposition}
\begin{proof}
Differentiation with the sampled tokens and rewards fixed gives the negative batch average of $R_b\sum_{j=1}^n\nabla_w\log\pi_w(Y_{b,j}\mid\mathrm{prefix}_{b,j})$, which is precisely the negative baseline-zero REINFORCE score estimate. Thus both routes supply the same gradient to the same optimizer state and settings, yielding the same parameter and optimizer-state update.
\end{proof}

Additional off-policy passes or normalization by accepted samples define different updates. Verification-based selection at deployment introduces test-verifier access and its generation cost.

\begin{corollary}[Frozen source-informed execution by binding search]\label[corollary]{cor:source-controller}
On the source event in \cref{seqadam:app:fb-controller}, a frozen greedy controller identifies the sequential binding in at most $q-1$ fresh-prompt verifications and then executes every test prompt exactly. In addition to its source budget it uses at most $n(q-1)$ generated symbols and $(H+1)(q-1)$ forward evaluations, with zero further gradient updates. Deployment uses $H+1$ forwards and $n$ symbols without verification.
\end{corollary}
\begin{proof}
There is also a concrete source-informed controller in the sequential task. On the source event, local state and EOS argmaxes are correct. Freeze this model, choose a candidate first control, and greedily generate the rest. Enumerating at most $q-1$ candidates on fresh training prompts identifies the binding and gives exact deployment execution. In addition to the source budget, this costs at most $q-1$ verifications, $n(q-1)$ generated symbols and $(H+1)(q-1)$ forward evaluations, with zero additional gradient updates. Deployment uses $H+1$ forwards and $n$ symbols without verification; retained context and workspace are charged. \Cref{seqadam:app:fb-controller} proves the margins and identification property.
\end{proof}

The controller explicitly searches bindings; the Adam construction adapts the source model's own parameters using sampled trajectories. Both use the same acquired transition information, with different query and computation costs.

\section{Shared Gaussian network foundations}\label{app:shared-foundations}\label{app:foundations}\label{seqadam:app:foundations}
The following initialization and derivative statements are used by both sequential optimizers. Their distinct objectives, rates, stopping rules and probability allocations are verified in the separate applications.
\subsection{Finite Gaussian initialization}
\label{seqadam:app:init-main}

Let \(N\) be the full padded context length, \(V\) the vocabulary size,
\(R\) the head count, \(d\) the residual width, \(k_{\mathrm{head}}\)
the head width, and \(M_{\mathrm{ff}}\) the feed-forward width.
The local head width $k_{\mathrm{head}}$ equals $d_h=d/R$ in the source budget. The public-state cardinality remains \(m=q/2\); write
\(k_\mu=\norm{\mu}^2\) for the kernel-mean quantity in the feedback
reference, to distinguish it from the head width.

\begin{definition}[Prescribed Gaussian initialization and lag weights]\label[definition]{def:shared-gaussian}
For the complete block in \eqref{seqadam:app:couple-architecture}, initialize
all the following entries independently:
\begin{align}
 E_{ij}&\sim\mathcal N(0,d^{-1}),&
 (Q_h)_{ij},(K_h)_{ij},(V_h)_{ij}
   &\sim\mathcal N(0,k_{\mathrm{head}}^{-1}),\notag\\
 (O_h)_{ij}&\sim\mathcal N(0,d^{-1}),&
 b_h(\ell)&\sim\mathcal N(0,1),\notag\\
 B_{ij},C_{ij},b_j,c_j&\sim\mathcal N(0,1),&
 U_{ij}&\sim\mathcal N(0,(M_{\mathrm{ff}}d)^{-1}),\notag\\
 (b_{\mathrm{out}})_i&\sim\mathcal N(0,d^{-1}),&
 (W_{\mathrm{out}})_{yi}&\sim\mathcal N(0,d^{-1}).
 \label{seqadam:app:init-laws}
\end{align}
The attention-head sum has scale \(R^{-1/2}\).
Its fixed temperature is chosen from these dimensions and the confidence
level, without using the unknown mechanism. Every displayed parameter trains from these prescribed width-dependent Gaussian scales.

Let \(\ell_h\) be the lag with maximum relative bias in head \(h\).
The winners depend only on the independent relative biases.
Conditional on coverage
\(n_\ell:=\#\{h:\ell_h=\ell\}\geq R/(2N)\), put
\begin{equation}
 \omega_\ell=\frac{n_\ell}{R},\qquad
 w_\ell=\omega_\ell+\mathbf1_{\{\ell=0\}},\qquad
 p=NV,\qquad D_w=\operatorname{diag}_\ell(w_\ell I_V).
 \label{seqadam:app:init-lag-weights}
\end{equation}

\end{definition}

\begin{lemma}[Hard-lag encoder frame]
\label[lemma]{seqadam:app:init-encoder}
Under \cref{def:shared-gaussian}, conditional on its coverage event, the hard-lag reference has
\begin{equation}
 z_{\mathrm{hard}}(x)=\sum_{\ell=0}^{N-1}A_\ell e_{x_\ell},
 \qquad
 A_\ell=\mathbf1_{\{\ell=0\}}E+
 R^{-1/2}\sum_{h:\ell_h=\ell}O_hV_hE.
 \label{seqadam:app:init-hard-encoder}
\end{equation}
Here \(e_{x_\ell}\) is a categorical basis vector, not a supplied semantic
state representation. Set
\(\mathcal A=[A_0\ \cdots\ A_{N-1}]\) and \(T=\mathcal A D_w^{-1/2}\).
For \(0<\xi\leq1\), the sufficient widths
\begin{align}
 d&\geq2048\xi^{-2}
       \left[p\log9+\log\frac8{\delta_{\mathrm{enc}}}\right],\notag\\
 k_{\mathrm{head}}&\geq\max\left\{1,
 \left\lceil\frac{4096N}{R\xi^2}
 \left[V\log9+\log\frac{8N}{\delta_{\mathrm{enc}}}\right]
 \right\rceil\right\}
 \label{seqadam:app:init-encoder-widths}
\end{align}
give \(\norm{T^\top T-I_p}_{\mathrm{op}}\leq\xi\)
with conditional probability at least \(1-\delta_{\mathrm{enc}}\),
uniformly over winner allocations satisfying coverage.
\end{lemma}

\begin{proof}
Write \(\mathcal A D_w^{-1/2}=C_{\mathrm{res}}+OZ\), where \(O\)
concatenates the \(O_h\) and is independent of \(E\) and all \(V_h\).
Explicitly, $C_{\rm res}=[E/\sqrt{w_0},0,\ldots,0]$ and the head/lag block of $Z$ is
$Z_{h,\ell}=\mathbf1_{\{\ell_h=\ell\}}V_hE/\sqrt{Rw_\ell}$.

With \(\varepsilon=\xi/8\), a Gaussian \(1/4\)-net bound gives
\(\norm{E^\top E-I_V}_{\mathrm{op}}\leq\varepsilon\).
Conditional on \(E\), the stacked value maps in lag group \(\ell\)
have \(n_\ell k_{\mathrm{head}}\) independent Gaussian rows.
Set $H_E=E^\top E$ and stack the lag-group maps $V_hE$ as $Y_\ell$.
On the encoder event $H_E$ is positive definite. Conditional on $E$,
$G_\ell=\sqrt{k_{\rm head}}Y_\ell H_E^{-1/2}$ has $n_\ell k_{\rm head}$ independent standard Gaussian rows. Their normalized Gram matrices are Wishart; the net bound and a union over all $N$ groups give
\[
 \left\|\frac{G_\ell^\top G_\ell}{n_\ell k_{\rm head}}-I\right\|\le\varepsilon,
 \qquad
 \left\|\frac{Y_\ell^\top Y_\ell}{n_\ell}-H_E\right\|
 \le\|H_E\|\varepsilon\le(1+\varepsilon)\varepsilon.
\]
The second inequality follows by multiplying the first error on both sides by $H_E^{1/2}$. Thus
\begin{align}
 \norm{C_{\mathrm{res}}^\top C_{\mathrm{res}}+Z^\top Z-I}_{\mathrm{op}}
 &\leq\varepsilon+(1+\varepsilon)\varepsilon,\notag\\
 \norm{C_{\mathrm{res}}}_{\mathrm{op}}^2&\leq1+\varepsilon,
 &\norm{Z}_{\mathrm{op}}^2&\leq(1+\varepsilon)^2.
 \label{seqadam:app:init-encoder-intermediate}
\end{align}
Conditional on \(E\) and the value maps, Gaussian subspace isometry for
\(O\) on \(\operatorname{range}(Z)\) bounds its Gram error by
\(\varepsilon\norm{Z}_{\mathrm{op}}^2\).
The residual cross term is retained. With orthonormal bases \(U_C,U_Z\)
of the two fixed column spaces, \(U_C^\top O U_Z\) has independent
Gaussian entries of variance \(1/d\).
Two \(1/4\)-nets under \eqref{seqadam:app:init-encoder-widths} bound its norm
by \(\varepsilon\), giving
\(\norm{C_{\mathrm{res}}^\top OZ}_{\mathrm{op}}
\leq\varepsilon(1+\varepsilon)^{3/2}\).
The four failure allocations are \(\delta_{\mathrm{enc}}/4\) each.
The total error is at most
\begin{equation}
 \varepsilon+(1+\varepsilon)\varepsilon
 +\varepsilon(1+\varepsilon)^2
 +2\varepsilon(1+\varepsilon)^{3/2}\leq6\varepsilon<\xi.
 \label{seqadam:app:init-encoder-total}
\end{equation}
The elementary Gaussian net estimate used here states that an
\(M\times s\) standard Gaussian matrix \(G\) satisfies
\(\norm{G^\top G/M-I}_{\mathrm{op}}\leq\varepsilon\), except with
probability \(\eta\), whenever
\[
 M\geq32\varepsilon^{-2}
       \left[s\log9+\log\frac2\eta\right].
\]
Fixed-direction chi-square tails followed by a \(1/4\)-net prove it.
Gaussian scalar tails and two nets give the residual cross bound.
Content-score events are combined afterwards by a union bound;
conditioning on them first would invalidate the Gaussian independence.
\end{proof}

\begin{definition}[Augmented tensor features]\label[definition]{def:shared-tensor}
Define
\begin{align}
 r_0(x)&=D_w^{1/2}\operatorname{concat}_\ell(e_{x_\ell}),&
 \norm{r_0(x)}^2&=\sum_\ell w_\ell=2,\notag\\
 k_{\mathrm{aug}}&=p+1,&
 s_{\mathrm{sym}}&=\frac{k_{\mathrm{aug}}(k_{\mathrm{aug}}+1)}2,
 \notag\\
 D_{\mathrm{feat}}&=p+s_{\mathrm{sym}}+1,&
 \Phi(x)&=
 \begin{bmatrix}
 r_0(x)\\
 \operatorname{symvec}\!\left(
 \begin{bmatrix}r_0(x)\\1\end{bmatrix}
 \begin{bmatrix}r_0(x)\\1\end{bmatrix}^{\!\top}\right)\\
 1
 \end{bmatrix}.
 \label{seqadam:app:init-tensor}
\end{align}
Symmetric vectorization is orthonormal for the Frobenius inner product,
so off-diagonal coordinates carry a factor \(\sqrt2\).
\end{definition}

\begin{lemma}[Exact coefficient representation]\label[lemma]{lem:shared-coefficient-identity}
Under \cref{def:shared-gaussian,def:shared-tensor}, there exists a realized coefficient matrix $C_{\mathrm{all}}$ for which \(z_{\mathrm{hard}}=Tr_0\), and
\begin{equation}
 \norm{\Phi(x)}^2=2+9+1=12,\qquad
 h_{\mathrm{hard}}(x)=C_{\mathrm{all}}\Phi(x).
 \label{seqadam:app:init-coefficient-identity}
\end{equation}
\end{lemma}
\begin{proof}
The latter identity is exact, obtained by expanding the actual bilinear
feed-forward block and its biases; it assumes no output-value matching. Explicitly, substitute $z_{\rm hard}=Tr_0$ in \eqref{seqadam:app:couple-architecture}. Each factor $B_jTr_0+b_j$ or $C_jTr_0+c_j$ is a linear functional of $(r_0,1)$. Their product is the Frobenius inner product of $(r_0,1)(r_0,1)^\top$ with the symmetric outer product of the two coefficient vectors. Orthonormal symmetric vectorization gives its coefficient row; multiplication by $U$, addition of $Tr_0$, and addition of $b_{\rm out}$ assemble $C_{\rm all}$. Finally $\|r_0\|^2=2$, the squared Frobenius norm of $(r_0,1)(r_0,1)^\top$ is $3^2=9$, and the last coordinate contributes one, proving the norm identity.
\end{proof}

\begin{lemma}[Realized coefficient-operator concentration]
\label[lemma]{seqadam:app:init-bilinear}
Under \cref{def:shared-gaussian,def:shared-tensor}, suppose \(0<\gamma\leq1\) and
\(\norm{T^\top T-I}_{\mathrm{op}}\leq\gamma/16\).
The additional conditions
\begin{align}
 d&\geq4096\gamma^{-2}
 \left[D_{\mathrm{feat}}\log9+\log\frac8{\delta_{\mathrm{ffn}}}\right],
 \label{seqadam:app:init-output-width}\\
 J&=s_{\mathrm{sym}}\log9+\log\frac8{\delta_{\mathrm{ffn}}},
 &A_0&=1+\gamma^{-2}J,\notag\\
 L_0&=\log\frac{\mathrm e A_0 k_{\mathrm{aug}}}
                         {\delta_{\mathrm{ffn}}\gamma},
 &M_{\mathrm{ff}}&=
 \left\lceil2^{50}A_0^2(k_{\mathrm{aug}}+L_0)^8\right\rceil
 \label{seqadam:app:init-ff-width}
\end{align}
give
\(\norm{C_{\mathrm{all}}^\top C_{\mathrm{all}}
 -I_{D_{\mathrm{feat}}}}_{\mathrm{op}}\leq\gamma\)
with conditional probability at least \(1-\delta_{\mathrm{ffn}}\).
In particular \(d\geq D_{\mathrm{feat}}\).
\end{lemma}

\begin{proof}
The residual Gram error is a principal block of the full Gram error,
and must therefore meet the desired total accuracy.
Let \(H_{\mathrm{aug}}=\operatorname{diag}(T^\top T,1)\).
A neuron's two augmented coefficient vectors are independent
\(\mathcal N(0,H_{\mathrm{aug}})\).
Whitening gives independent \(a,b\sim\mathcal N(0,I_{k_{\mathrm{aug}}})\)
and
\[
 v=\operatorname{symvec}\!\left(\frac{ab^\top+ba^\top}{2}\right).
\]
For each symmetric \(S\),
\begin{equation}
 \E(a^\top S b)^2=\norm{S}_F^2,\qquad
 \E vv^\top=I_{\mathrm{sym}}.
 \label{seqadam:app:init-product-covariance}
\end{equation}
Furthermore,
\begin{align}
 \mathbb P\{\norm{v}^2>(2k_{\mathrm{aug}}+3t)^2\}
 &\leq2\exp(-t),\notag\\
 \E\norm{v}^4&\leq[k_{\mathrm{aug}}(k_{\mathrm{aug}}+2)]^2.
 \label{seqadam:app:init-product-tail}
\end{align}
Truncation at squared norm \(B_0=(2k_{\mathrm{aug}}+3t)^2\)
has covariance bias at most \(5k_{\mathrm{aug}}^2\exp(-t/2)\).
Set \(\varepsilon=\gamma/16\). It suffices that
\begin{align}
 t&\geq\max\left\{1,\log\frac{8M_{\mathrm{ff}}}{\delta_{\mathrm{ffn}}},
           2\log\frac{10k_{\mathrm{aug}}^2}{\varepsilon}\right\},
 \notag\\
 M_{\mathrm{ff}}&\geq8B_0^2\varepsilon^{-2}
 \left[s_{\mathrm{sym}}\log9+\log\frac8{\delta_{\mathrm{ffn}}}\right].
 \label{seqadam:app:init-truncation-conditions}
\end{align}
Scalar Hoeffding on a \(1/4\)-net, together with the truncation bias and
the event of any realized truncation, bounds the actual whitened
sample-covariance error by \(\varepsilon\), with failure probability
\(\delta_{\mathrm{ffn}}/2\).
The explicit choice \eqref{seqadam:app:init-ff-width} satisfies
\eqref{seqadam:app:init-truncation-conditions} with
\(t=128(k_{\mathrm{aug}}+L_0)\); no implicit \(\log M_{\mathrm{ff}}\)
condition remains.
Transforming back through \(H_{\mathrm{aug}}^{1/2}\), the actual
product-feature matrix \(F\) satisfies
\begin{equation}
 \norm{F^\top F/M_{\mathrm{ff}}-I_{\mathrm{sym}}}_{\mathrm{op}}
 \leq(1+\gamma/16)^2\varepsilon+2\gamma/16+(\gamma/16)^2
 \leq\gamma/4.
 \label{seqadam:app:init-product-gram}
\end{equation}
Conditional on \(T,F\), the realized \(Z=[UF,b_{\mathrm{out}}]\)
has distribution \(XQ^{1/2}/\sqrt d\), where
\[
 Q=\operatorname{diag}(F^\top F/M_{\mathrm{ff}},1)
\]
and \(X\) is standard Gaussian. This representation is independent of
the now-fixed \(T\).
Gaussian nets and \eqref{seqadam:app:init-output-width} imply, simultaneously,
\begin{equation}
 \norm{X^\top X/d-I}_{\mathrm{op}}\leq\gamma/8,\qquad
 \norm{T^\top X/\sqrt d}_{\mathrm{op}}\leq\gamma/4
 \label{seqadam:app:init-output-cross}
\end{equation}
except with probability \(\delta_{\mathrm{ffn}}/2\).
Hence
\[
 \norm{Z^\top Z-I}_{\mathrm{op}}\leq13\gamma/32,\qquad
 \norm{T^\top Z}_{\mathrm{op}}\leq\sqrt5\,\gamma/8.
\]
Since \(C_{\mathrm{all}}=[T,Z]\), the total block error is at most
\[
 \max\{\gamma/16,13\gamma/32\}+\sqrt5\,\gamma/8<\gamma.
\]
This proves concentration of the realized coefficient operator.
\end{proof}

\begin{lemma}[Tensor kernel and source coefficients]\label[lemma]{lem:shared-tensor-kernel}
For the features in \cref{def:shared-tensor}, the ideal kernel is
\begin{equation}
 \ip{\Phi(x)}{\Phi(x')}=
 k_{\mathrm{base}}(x,x')^2+3k_{\mathrm{base}}(x,x')+2,\qquad
 k_{\mathrm{base}}(x,x')=
 \sum_\ell w_\ell\mathbf1_{\{x_\ell=x'_\ell\}}.
 \label{seqadam:app:init-kernel}
\end{equation}
Thus \(K_{\max}=12\), and on coverage
\begin{equation}
 3w_\ell+w_\ell^2\geq\frac3{2N},\qquad
 k_{0H}=2w_0w_H\geq\frac1N.
 \label{seqadam:app:init-kernel-coefficients}
\end{equation}
These singleton weights and pair coefficients are those used in the
source reference.
\end{lemma}
\begin{proof}
Write $b=\langle r_0(x),r_0(x')\rangle=k_{\rm base}(x,x')$. The tensor-feature inner product is $(b+1)^2$, because symmetric vectorization preserves the Frobenius inner product. Adding the first and last feature blocks gives $b+(b+1)^2+1=b^2+3b+2$. Since $0\le b\le2$, its maximum is $12$. Expanding $b^2$ gives the singleton coefficient $w_\ell^2$ and the pair coefficient $2w_0w_H$. On coverage $w_\ell\ge1/(2N)$ and $w_0\ge1$, so the two displayed lower bounds follow.
\end{proof}

\paragraph{Finite soft attention and nonzero initial readout.}
\begin{lemma}[Coverage and separated bias winners]\label[lemma]{lem:shared-bias-event}
For $N\ge2$ and $0<\ell<N-1$, the independent bias event has coverage and every pairwise gap at least
\(g\), with probability at least \(1-\delta_{\mathrm{bias}}\), when
\begin{equation}
 R\geq8N\log\frac{2N}{\delta_{\mathrm{bias}}},\qquad
 g=\frac{\delta_{\mathrm{bias}}}{RN^2}.
 \label{seqadam:app:init-bias-event}
\end{equation}
\end{lemma}
\begin{proof}
For coverage, uniformly distributed winning lags give failure at most $N\exp(-R/(8N))\le\delta_{\rm bias}/2$. A pairwise difference of independent unit Gaussians has density at most $1/(2\sqrt\pi)$; the union over at most $RN^2/2$ pairs gives gap failure at most $RN^2g/(2\sqrt\pi)<\delta_{\rm bias}/2$.
\end{proof}

\begin{lemma}[Finite-temperature approximation of hard lag selection]\label[lemma]{lem:shared-soft-hard}
On the bias event of \cref{lem:shared-bias-event} and a block-norm event with bound $M_{\rm init}$, choose $0<\ell<N-1$ and a fixed positive temperature satisfying $t_{\rm att}\le g/(4M_{\rm init}^4)$ and \eqref{seqadam:app:init-temperature}. Then each head has off-winner attention mass at most $\ell$, and the soft and hard initial features obey \eqref{seqadam:app:init-soft-hard} uniformly over all input strings.
\end{lemma}
\begin{proof}
A separate Gaussian norm event supplies a deterministic
\(M_{\mathrm{init}}\) bounding every initialized block.
Content scores then have magnitude at most \(M_{\mathrm{init}}^4\).
Taking \(t_{\mathrm{att}}\leq g/(4M_{\mathrm{init}}^4)\)
keeps the winner gap, in unscaled score units, at least \(g/2\).
If also
\begin{equation}
 t_{\mathrm{att}}\leq\frac{g}{2\log((N-1)/\ell)},
 \label{seqadam:app:init-temperature}
\end{equation}
each head's off-winner mass is at most \(\ell\).
With \(Z_0=M_{\mathrm{init}}+\sqrt R\,M_{\mathrm{init}}^3\),
\begin{equation}
 \sup_x\norm{h_{\mathrm{soft},0}(x)-h_{\mathrm{hard},0}(x)}
 \leq2\sqrt R\,M_{\mathrm{init}}^3\ell
       [1+2M_{\mathrm{init}}^3(Z_0+1)].
 \label{seqadam:app:init-soft-hard}
\end{equation}
To obtain the feature estimate, the difference of an attention-weighted average from its winning value has norm at most $2\ell M_{\rm init}^2$ before multiplication by $O_h$. Summing the scaled heads gives a raw residual difference at most $2\sqrt R M_{\rm init}^3\ell$. Both residual vectors have norm at most $Z_0$. Along their connecting segment, the derivative of the bilinear feed-forward map with respect to its residual input has norm at most $1+2M_{\rm init}^3(Z_0+1)$. Multiplying these bounds proves \eqref{seqadam:app:init-soft-hard}.
This holds uniformly over all strings, including repeated padding
and wrong generated tokens. The derivatives are bounded in \cref{app:shared-derivatives}; subsequent motion is controlled separately in \cref{seqadam:app:adam-joint,app:couple-main}. No infinite-width or zero-temperature limit is interchanged.
\end{proof}

\begin{lemma}[A simultaneous initialization norm event]\label[lemma]{lem:shared-global-norm}
For the Gaussian law in \cref{def:shared-gaussian}, let $P$ be its number of initialized scalar parameters and let $0<\delta_{\rm norm}<1$. With probability at least $1-\delta_{\rm norm}$, every initialized block Frobenius, spectral or vector norm is at most the deterministic $M_{\rm init}$ in \eqref{seqadam:app:init-global-norm}.
\end{lemma}
\begin{proof}
For an explicit coarse norm event, let \(P\) be the number of initialized
scalar parameters, all with variance at most one in \eqref{seqadam:app:init-laws}.
With failure probability at most \(\delta_{\mathrm{norm}}\), every
block Frobenius, spectral or vector norm is bounded by
\begin{equation}
 M_{\mathrm{init}}=
 \max\left\{1,\sqrt{2P\log\frac{2P}{\delta_{\mathrm{norm}}}}\right\}.
 \label{seqadam:app:init-global-norm}
\end{equation}
For each scalar Gaussian of variance at most one, the probability that its absolute value exceeds $\sqrt{2\log(2P/\delta_{\rm norm})}$ is at most $\delta_{\rm norm}/P$. A union bound over the $P$ scalars and the inequality between spectral and Frobenius norms give \eqref{seqadam:app:init-global-norm}. Bound every scalar Gaussian and then each block's Frobenius norm.
This conservative bound is polynomial in the prescribed dimensions.
It permits the temperature and all positive rates to be selected before
drawing initialization or source data.
\end{proof}

\subsection{Uniform derivatives of the complete block}\label{seqadam:app:couple-main}\label{app:shared-derivatives}
\begin{definition}[Complete raw attention and bilinear block]\label[definition]{def:shared-architecture}
Use \(R\) heads, \(N\) retained positions, relative biases \(b_h(\ell)\)
and fixed temperature \(t_{\mathrm{att}}>0\). Define
\begin{align}
 e_\ell&=E(x_\ell),\notag\\
 \alpha_h(\ell)&=\softmax_\ell\!\left[
 \frac{b_h(\ell)}{t_{\mathrm{att}}}
 +\frac{\ip{Q_he_0}{K_he_\ell}}{\sqrt{k_{\mathrm{head}}}}
 \right],\notag\\
 z&=e_0+R^{-1/2}\sum_{h=1}^R O_h
       \sum_{\ell=0}^{N-1}\alpha_h(\ell)V_he_\ell,\notag\\
 h_\psi&=z+U[(Bz+b)\odot(Cz+c)]+b_{\mathrm{out}}.
 \label{seqadam:app:couple-architecture}
\end{align}
Every displayed matrix, embedding and bias is included in \(\psi\)
with positive rates specified separately in the Adam and SGD applications.
Causal attention sees every real token in the padded context; the displayed equations specify the complete raw architecture.

\end{definition}

\begin{lemma}[Uniform raw feature and derivative bounds]\label[lemma]{lem:shared-raw-derivative}
For \cref{def:shared-architecture}, if every block matrix spectral norm, bias-vector norm and
embedding-column norm is at most \(M\geq1\), put
\begin{align}
 Z&=M+\sqrt R\,M^3,\notag\\
 A_{\mathrm{raw}}&=Z+M^3(Z+1)^2+M,\notag\\
 D_z&=1+\sqrt R[3M^2+M^3N(4M^3+t_{\mathrm{att}}^{-1})],\notag\\
 D&=D_z[1+2M^3(Z+1)]+3M^2(Z+1)^2+1.
 \label{seqadam:app:couple-raw-envelope}
\end{align}
Then \(\norm{h_\psi}\leq A_{\mathrm{raw}}\) and
\(\norm{D_\psi h_\psi}_{\mathrm{op}}\leq D\) in the concatenated
Euclidean/Frobenius metric.
\end{lemma}
\begin{proof}
For a global unit parameter direction, each individual matrix or bias
direction has norm at most one. The attention-score derivative is at
most \(4M^3+t_{\mathrm{att}}^{-1}\), while
\[
 \norm{\dot\alpha}_1\leq N\norm{\dot{\mathrm{score}}}_\infty.
\]
Differentiating the \(O,V,e\) paths gives \(D_z\).
The feature bound follows from $\|z\|\le Z$ and the block norm bounds in the displayed architecture. Both feed-forward factors have norm at most \(M(Z+1)\), yielding the
displayed direct-factor and through-\(z\) terms. No block is omitted.
\end{proof}

\subsection{The nonlinear normalized member}
\label{seqadam:app:norm-main}

\begin{definition}[Normalized block and trainable gain]\label[definition]{def:shared-normalized}
Retain every raw attention and bilinear block, replacing only the final
output feature by
\begin{equation}
 h_{\mathrm{norm}}=g\odot\mathcal R(h_{\mathrm{raw}}),\qquad
 \mathcal R(x)=\frac{\sqrt2\,x}{\sqrt{1+\norm{x}^2/K}},
 \qquad K=12.
 \label{seqadam:app:norm-definition}
\end{equation}
The gain \(g\in\R^d\) is trainable and starts at the task-independent
vector of ones. Every original random parameter retains its prescribed
Gaussian law. Each original backbone block and the gain update at the
same positive backbone rate; the readout uses the specified faster
rate. The gain initializer is thus deterministic.

This nonlinear final RMS-type normalization has a fixed positive
denominator offset and trainable coordinate gain. It changes the feature
and its derivatives away from the constant-norm ideal feature manifold.

\end{definition}

\begin{lemma}[Normalized initial-feature approximation]\label[lemma]{lem:shared-normalized-initial}
Use \cref{def:shared-normalized}.
The normalization has global derivative norm at most \(\sqrt2\)
and global output norm at most \(\sqrt{2K}\).
On the actual hard-feature event,
\(\norm{C_{\mathrm{all}}^\top C_{\mathrm{all}}-I}_{\mathrm{op}}
\leq\gamma\) and \(\norm{\Phi}^2=K\). For \(\gamma\leq1/4\),
\begin{equation}
 \norm{\mathcal R(C_{\mathrm{all}}\Phi)-C_{\mathrm{all}}\Phi}
 \leq4\gamma.
 \label{seqadam:app:norm-hard-error}
\end{equation}
If the raw soft/hard features differ by at most \(\zeta_{\mathrm{raw}}\),
then
\begin{equation}
 \norm{h_{\mathrm{norm},0}-C_{\mathrm{all}}\Phi}
 \leq\sqrt2\,\zeta_{\mathrm{raw}}+4\gamma.
 \label{seqadam:app:norm-initial-error}
\end{equation}
\end{lemma}
\begin{proof}
The derivative of $\mathcal R$ has tangential eigenvalue $\sqrt2(1+\|x\|^2/K)^{-1/2}$ and radial eigenvalue $\sqrt2(1+\|x\|^2/K)^{-3/2}$, both at most $\sqrt2$; its output norm is at most $\sqrt{2K}$. To see this, write \(\norm{C_{\mathrm{all}}\Phi}^2/K=1+t\),
where \(|t|\leq\gamma\). The factor \(\sqrt2/\sqrt{2+t}\)
differs from one by at most \(\gamma\), and
\(\norm{C_{\mathrm{all}}\Phi}<4\).
Thus the estimate concerns the realized random operator.
The Lipschitz bound and triangle inequality give the soft/hard estimate in the statement, because $g_0$ is the vector of ones.
\end{proof}

\begin{lemma}[Normalized feature and derivative envelopes]\label[lemma]{lem:shared-normalized-derivative}
Use \cref{def:shared-normalized} and the norm event in \cref{lem:shared-global-norm}.
On a radius-one ball around initialization in all backbone parameters,
\(\norm{g}_\infty\leq2\). Therefore
\begin{equation}
 \norm{h_{\mathrm{norm}}}\leq2\sqrt{2K}<12=A.
 \label{seqadam:app:norm-feature-envelope}
\end{equation}
Let \(D_{\mathrm{raw}}\) be the raw derivative bound
\eqref{seqadam:app:couple-raw-envelope}, evaluated at \(M_{\mathrm{init}}+1\).
The concatenated derivative norm is bounded by
\begin{equation}
 D_{\mathrm{norm}}=2\sqrt2\,D_{\mathrm{raw}}+\sqrt{2K}.
 \label{seqadam:app:norm-derivative}
\end{equation}
\end{lemma}
\begin{proof}
A radius-one displacement from $g_0=\mathbf1$ gives $\|g\|_\infty\le2$. Apply the global output and derivative bounds of \cref{lem:shared-normalized-initial} and the raw derivative bound of \cref{lem:shared-raw-derivative}. The first term differentiates every original attention, embedding and
feed-forward parameter through \(\mathcal R\).
The second differentiates the coordinate gain, using
\(\norm{\mathcal R(h_{\mathrm{raw}})}_\infty\leq\sqrt{2K}\).
Both the gain and all original blocks are included in the motion bound.

\end{proof}

\section{Sequential acquisition and sampled Adam adaptation}\label{app:sequential-adam}
\subsection{Finite source learning and sampled Adam improvement}
Use the two architectures and source-to-feedback update rule in \cref{sec:results}: all parameters train, source cross-entropy has the coupled positive ridge penalty, and bias-corrected Adam uses $(\beta_1,\beta_2)=(0.9,0.999)$. At the source stopping state, retain every parameter, remove the penalty and reset both moments. Feedback uses fresh on-policy trajectories, baseline zero, and retains every trajectory and every output score. The complete choices of horizon, width, Gaussian scales, positive block rates, stopping threshold and batch sizes are those in \cref{seqadam:app:adam-joint}. Write $\varepsilon_{\rm logit}:=h$ for the tolerance in \eqref{seqadam:eq:adam-h}. The metrics $E,M,S,V$ are defined in \cref{sec:setup}.

\begin{theorem}[Finite source learning and sampled Adam improvement]\label[theorem]{thm:neural-formal}
Fix $q=2^{2r+2}$ with $r\ge1$, $c=1/64$ and $\delta\in(0,1)$. For either network member in \eqref{eq:network}--\eqref{eq:network-output}, with the explicit horizon, dimensions and training choices of \cref{seqadam:app:adam-joint}, the source algorithm uses at most $T_s$ batches of $B_s$ records and makes at most $T_s-1$ updates; it is followed by exactly two feedback Adam updates, each on $B_R$ sampled trajectories. For every fixed world, with probability at least $1-\delta$, source stopping succeeds and, simultaneously for every held-out prompt $X$,
\begin{align}
 \min\{E(w_f,X),M(w_f,X),S(w_f,X)\}&\ge7/8,\\
 \max\{E(w_s,X),M(w_s,X),S(w_s,X)\}&\le1/q+2n\varepsilon_{\rm logit},\\
 \min_{J\in\{E,M,S\}}\big[J(w_f,X)-J(w_s,X)\big]&\ge3/4,\\
 V(w_f,X)&\le9c/8+1/128,
\end{align}
where $w_s$ is the source stopping state and $w_f$ the second feedback iterate. With $N=2H+2$ source tokens per record and $n=H+2$ decisions per trajectory,
\begin{align}
 N_{\rm pre}&\le NB_sT_s,& Q&=2B_R,\\
 N_{\rm generated}&=2nB_R,&
 N_{\rm updates,source}&\le T_s-1,
 \qquad N_{\rm updates,feedback}=2.
\end{align}
\end{theorem}

The \hyperref[seqadam:proof:neural-formal]{complete proof} closes \cref{seqadam:app:adam-joint}, following its parameter construction.

\subsection{Applying the shared bounds to Adam}\label{app:adam-foundations-application}
\begin{proposition}[Adam substitution into the shared initialization bounds]\label[proposition]{prop:adam-shared-application}
Use the dimensions, temperature and probability allocations prescribed in \cref{seqadam:app:adam-joint}, with $\gamma=d^{-1/8}$. These satisfy the hypotheses of the shared encoder, coefficient, bias, norm, soft-attention and derivative lemmas. Their four initialization events have total failure at most $\delta/4$. For the normalized member, the initial feature discrepancy is at most $3\varepsilon_{\rm src}/8$. The raw or normalized derivative envelopes apply throughout every radius-one backbone ball controlled by the Adam displacement bounds.
\end{proposition}
\begin{proof}
The substitutions into \cref{seqadam:app:init-encoder,seqadam:app:init-bilinear} are
$N=2H+2$, $V=2q+2$, $R=R_{\rm head}$, $k_{\rm head}=d/R_{\rm head}$,
$D_{\rm feat}=k$, and $\xi_{\rm enc}=\gamma/16$ from
\cref{seqadam:app:adam-joint}. In particular,
$2048\xi_{\rm enc}^{-2}=2^{19}\gamma^{-2}$ and
$4096\xi_{\rm enc}^{-2}=2^{20}\gamma^{-2}$.
Thus the encoder and head premises are exactly the first and third terms
in $D_\Gamma\gamma^{-2}$; the second term supplies the output Gram premise.
Since $\gamma=d^{-1/8}$, $d\ge D_\Gamma^{4/3}$ implies all three.
Rounding $d$ to a multiple of $R_{\rm head}$ preserves them and makes the head width integral.
The FFN width is the shared lemma's width with
$(J,A_0,L_0)=(J_{\rm ff},A_{\rm ff},L_{\rm ff})$.
The bias, encoder, FFN and global-norm events each use $\delta/16$;
the encoder and FFN conditional bounds are combined by the tower property,
then with the other events by a union bound, without conditioning the Gaussian laws on content scores.
The temperature satisfies both inequalities in \cref{seqadam:app:init-temperature}
and the preceding gap condition; its choice of $\ell$ makes the raw feature error at most $z_{\rm raw}$.
For the normalized member, $\gamma\le\varepsilon_{\rm src}/16$ and
$z_{\rm raw}\le\varepsilon_{\rm src}/(8\sqrt2)$ give
$\sqrt2z_{\rm raw}+4\gamma\le3\varepsilon_{\rm src}/8$.
The separate row, readout, geometry, batch and reward events keep their allocations
in \eqref{seqadam:eq:adam-failure-allocation}; none is replaced by the SGD allocation.

Evaluate the displayed derivative bound at $M=M_{\rm init}+1$ for the whole radius-one parameter ball. The source and feedback displacement bounds in \cref{seqadam:app:adam-joint} keep every supplied backbone parameter in that ball. The actual initialization has bounded features, and its smaller source tube gives the feature envelope $A_{\rm src}=12$. The amplitude-free coordinate Adam bound pays motion even when coupled regularizer gradients are large.
The gain is included in the backbone parameter count and in every source and feedback update. The source feature tolerance and subsequent motion are selected in \cref{seqadam:app:adam-joint}; their errors include the initial $4\gamma$ normalization discrepancy. The proof uses the same raw coefficient operator as an analysis reference and makes no rotation-equivariance claim for the trained normalized network.
\end{proof}

\subsection{Finite source acquisition by coupled-regularized Adam}\label{seqadam:app:source-budgets}
The source stage learns a transition function and output format before any terminal reward is observed. We first identify an invariant population ridge optimum, transfer it to the realized hard-feature head, and then show that the actual sampled Adam algorithm reaches a source-observable stopping condition. The finite initializer and full-model derivative bounds are in \cref{seqadam:app:foundations}; \cref{seqadam:app:adam-joint} chooses the joint horizon, widths, source precision, and failure allocations. Throughout this section, $N=2H+2$, $H\ge5$, $V=2q+2$, and $K=12$. We write $T_s,B_s,\eta_s$ for the source inspection limit, complete-record batch size, and effective head step. The physical head step is $\eta_s/\sqrt d$.

Let $F_0(\Theta)$ and $F_+(\Theta)$ denote the unconditioned and tag-plus population source cross-entropies, averaged over all $N$ raw next-token targets. In the unconditioned law, $S_0$ and all controls are independent uniform. Conditional on the controls, the state transitions use $aP_U+(1-a)\Pi$. The canonical feature map has kernel
\begin{equation}\label{seqadam:adam-src-kernel}
 \ip{\Phi(x)}{\Phi(x')}=k_{\rm base}(x,x')^2+3k_{\rm base}(x,x')+2,
 \qquad k_{\rm base}(x,x')=\sum_{l=0}^{N-1}w_l\mathbf1\{x_l=x'_l\}.
\end{equation}
\begin{lemma}[Canonical feature geometry]\label[lemma]{seqadam:lem:canonical-geometry}\label[lemma]{seqadam:adam-src-canonical}
For the canonical kernel \eqref{seqadam:adam-src-kernel} on the encoder event of \cref{seqadam:app:foundations}, the squared feature norm is $K=12$, the singleton weights are at least $3/(2N)$, and the lag-pair weight $k_{0H}$ is at least $1/N$. Its categorical and tensor representations are isometric on their spans.
\end{lemma}
\begin{proof}
Its squared feature norm is $K$, its singleton coefficients are at least $k_{\min}=3/(2N)$, and its lag-pair coefficient satisfies $k_{0H}\ge1/N$. The categorical singleton/pair representation and the tensor representation in \cref{seqadam:app:foundations} have the same kernel. Mapping finite linear combinations of one representation to the other preserves their Gram quadratic form, and hence gives an isometry of their spans. Projecting any comparator onto that span cannot increase its norm.
\end{proof}

\begin{definition}[Semantic source coordinates]\label[definition]{seqadam:def:semantic-source}
For a state-output row $y$, define
\begin{align}\label{seqadam:adam-src-semantic}
 h_{1,y}(x)&=\mathbf1\{x_0\in\mathcal U,x_H\in\mathcal S\}
       \left(\mathbf1\{y=\phi_{x_0}(x_H)\}-q^{-1}\right),\\
 h_{2,y}(x)&=\mathbf1\{x_0\in\mathcal S,x_H\in\mathcal U\}
       \left(\mathbf1\{y=\phi_{x_H}(x_0)\}-q^{-1}\right),\nonumber
\end{align}
and set both functions to zero on nonstate rows. Their pair supports are disjoint. The active lag pair is $(U_1,S_0)$ at the first-state prediction, $(S_{i-1},U_i)$ at subsequent state predictions, and $(S_H,U_1)$ at EOS. Both functions vanish at the initial control-choice prediction because lag $H$ is PAD.

\end{definition}

\begin{lemma}[Invariant population source dynamics]\label[lemma]{seqadam:lem:source-closure}
Under the unconditioned source law above, Euclidean gradient descent on $F_0$ preserves the mask/class space plus the semantic span in \cref{seqadam:def:semantic-source}. Its two semantic coordinates satisfy \eqref{seqadam:adam-src-closure}, with $b_j$ defined by \eqref{seqadam:adam-src-transition-parameter}. These identities hold on every legal class/PAD pattern, including generated state histories.
\end{lemma}
\begin{proof}
The gradient of $F_0$ preserves the sum of mask/class logits and $\alpha_1h_1+\alpha_2h_2$. Here $\alpha_j$ are semantic logit amplitudes, distinct from Adam's moment parameters. Put
\begin{equation}\label{seqadam:adam-src-transition-parameter}
 b_j=\frac{\exp(\alpha_j)-1}{\exp(\alpha_j)+q-1}.
\end{equation}
If $v_i$ is the correct state-class mass at state target $i$ and $v_E$ is the incorrect state-class mass at EOS, the semantic coordinates of an auxiliary Euclidean descent step are
\begin{align}\label{seqadam:adam-src-closure}
 \alpha_1^+-\alpha_1&=\frac{\eta k_{0H}}{Nq^2}(a-v_1b_1),\\
 \alpha_2^+-\alpha_2&=\frac{\eta k_{0H}}{Nq^2}
       \left[\sum_{i=2}^H(a-v_ib_2)-v_Eb_2\right].\nonumber
\end{align}
The EOS contribution is essential because $h_2$ is active on the incorrect state class there.

To verify closure, center in both interaction arguments. At the first-state query, independence leaves only the pair $(U_1,S_0)$. At a repeated-state query, averaging the current control eliminates coordinates omitting it; a surviving coordinate pairs that control with the immediately preceding state. At EOS, integrating uniform $S_0$ makes $(U_2,\ldots,U_H,S_1,\ldots,S_H)$ independent of $U_1$. Coordinates omitting both $U_1$ and $S_0$ vanish by averaging $U_1$. Among coordinates retaining $U_1$, an earlier state leaves a later control to average, and a control leaves uniform $S_0$; only $(U_1,S_H)$ survives. Among coordinates retaining $S_0$ but not $U_1$, an earlier state leaves a later control, $S_H$ leaves a suffix average followed by $U_1$, and a suffix control leaves another suffix control when $H\ge3$. The surviving semantic coefficient has factor $q^{-2}$. Distinct individual record variables are pairwise independent uniform, which also preserves the mask/class space. These categorical identities hold on every prefix with the legal class/PAD pattern, including incorrect generated states.
\end{proof}

\begin{lemma}[Finite source comparator]\label[lemma]{seqadam:lem:source-comparator}
Under the unconditioned source law with $0<a<1$, its target-averaged entropy is $H_0$ in \eqref{seqadam:adam-src-entropy}. For every $0<\varepsilon\le1$, the invariant feature class contains a finite comparator with excess cross-entropy at most $\varepsilon$ and squared norm at most $D^*(\varepsilon)$ in \eqref{seqadam:adam-src-comparator}.
\end{lemma}
\begin{proof}
Let $h_q(a)$ be the entropy of a point mass mixed with the uniform $q$-state law at weights $a,1-a$. The unconditioned entropy per optimized target is
\begin{equation}\label{seqadam:adam-src-entropy}
 H_0=\frac{(H+1)\log q+Hh_q(a)}{N}.
\end{equation}
For $\varepsilon>0$, set
\begin{equation}\label{seqadam:adam-src-comparator-parameters}
 \alpha^*=\log\frac{1+(q-1)a}{1-a},\qquad
 M=\alpha^*+\log(V/\varepsilon),\qquad P_l=\mathbf1\{x_l=\mathrm{PAD}\}.
\end{equation}
Use the class logits
\begin{align}\label{seqadam:adam-src-format}
 c_S&=M(P_0-P_H+P_{2H}),&c_U&=M(P_H-P_0),\\
 c_E&=M(1-P_{2H}),&c_{\rm PAD}&=0,\nonumber
\end{align}
and add $\alpha^*(h_1+h_2)$. They select the correct class at every raw target; the extra $\alpha^*$ in $M$ pays for the active incorrect state partition at EOS. The constant term of $c_E$ can be represented by all lag-zero indicators. This comparator has excess CE at most $\varepsilon$ and squared norm bounded by
\begin{equation}\label{seqadam:adam-src-comparator}
 D^*(\varepsilon)=\frac{(5q+V+1)M^2}{k_{\min}}
       +\frac{2q(q-1)(\alpha^*)^2}{k_{0H}}.
\end{equation}
For deterministic algorithmic choices, we use the upper envelope obtained by substituting $k_{\min}=3/(2N)$ and $k_{0H}=1/N$ in this expression.
\end{proof}

\begin{lemma}[Excess loss controls execution]\label[lemma]{seqadam:lem:source-excess}
For an invariant reference from \cref{seqadam:lem:source-closure} with excess $E=F_0(\Theta)-H_0$, the decomposition \eqref{seqadam:adam-src-excess-decomposition} and bounds \eqref{seqadam:adam-src-transition-error} hold. In particular, at $a=1-c/H$, $E\le c^2(q-1)^2/(32q^2H^2N)$ implies $|b_j-a|\le c/(8H)$ for $j=1,2$.
\end{lemma}
\begin{proof}
Let $\operatorname{KL}$ denote Kullback--Leibler divergence. For any invariant reference, let $D_j=\operatorname{KL}(T_a\Vert T_{b_j})$, where $T_b$ denotes the relevant true transition permutation mixed with uniform at weights $b,1-b$. The exact excess decomposition is
\begin{equation}\label{seqadam:adam-src-excess-decomposition}
 D_1+(H-1)D_2+\sum_{\text{all stages }s}[-\log v_s]
       =N\bigl[F_0(\Theta)-H_0\bigr].
\end{equation}
Consequently, if the excess is $E$, then
\begin{equation}\label{seqadam:adam-src-transition-error}
 |b_1-a|\le\frac q{q-1}\sqrt{NE/2},\qquad
 |b_2-a|\le\frac q{q-1}\sqrt{\frac{NE}{2(H-1)}}.
\end{equation}
For $a=1-c/H$, the condition $E\le c^2(q-1)^2/(32q^2H^2N)$ gives $|b_j-a|\le c/(8H)$ for both transitions. The first transition is observed at one position per record, which determines the stronger precision requirement.
\end{proof}

\begin{lemma}[Tag contraction and source-gradient transfer]\label[lemma]{seqadam:lem:tag-transfer}\label[lemma]{seqadam:adam-src-ridge}
For the Sequential source law, let $C_{\rm tag}=\prod_{i=2}^H(-1)^{\ell(U_i)}$ and $\rho_c=2^{-r}$. A bounded observable retaining $r_S$ states and $r_{\rm tag}$ suffix controls satisfies \eqref{seqadam:adam-src-signed-coordinate}. At each invariant reference of \cref{seqadam:lem:source-closure}, the gradient discrepancy is at most $\mu_F$ in \eqref{seqadam:adam-src-mixing}, uniformly over the unknown mechanism.
\end{lemma}
\begin{proof}
Write $C_{\rm tag}=\prod_{i=2}^H(-1)^{\ell(U_i)}$ and $\rho_c=2^{-r}$.
The signed contraction holds on the full state space for every unknown
mechanism. Indeed, in the Koopman convention $P_Af=f\circ\phi_A$, decompose
$f(s,z)=f_+(z)+(-1)^sf_-(z)$. In
$M_c=q^{-1}\sum_{b,u}(-1)^{\ell(u)}P_{(b,u)}$, averaging the control bit
$b$ annihilates the odd component. The even component reduces to normalized
convolution by $(-1)^{t+x\cdot y}$ on the public $Z$ coordinates,
independently of $\theta$. Its Walsh eigenvalues are zero at dual
$t$-frequency zero and $2^{-r}(-1)^{\alpha\cdot\beta}$ at dual
$t$-frequency one. Thus $\|M_c\|_{\rm op}=2^{-r}$. For the noisy kernels
$T_A=aP_A+(1-a)\Pi$, balanced tags annihilate the $\Pi$ term, giving signed
operator norm $a\rho_c$.
For a bounded observable $G$ retaining $r_S$ states and $r_{\rm tag}$ suffix controls, the signed-coordinate estimate is
\begin{equation}\label{seqadam:adam-src-signed-coordinate}
 \left|\E_0[C_{\rm tag}G]\right|
 \le q^{r_S}(a\rho_c)^{H-1-r_{\rm tag}}\norm G_\infty.
\end{equation}
Insert diagonal projectors for retained states into the Markov product. Each unretained tagged control contributes an operator of norm $a\rho_c$, while a retained control has norm at most one. Summing retained state values costs at most $q^{r_S}$. The control $U_1$ supplies no signed contraction. At an invariant reference, the probability and label terms of each gradient coordinate retain at most three tagged controls and three states. The sum of squared categorical feature weights is at most $V^2K$. Summing over $V$ output rows and taking a square root therefore gives
\begin{equation}\label{seqadam:adam-src-mixing}
 \norm{\nabla F_+(\Theta)-\nabla F_0(\Theta)}_F
 \le\mu_F,
 \qquad \mu_F=2q^3(a\rho_c)^{H-4}V^{3/2}\sqrt K.
\end{equation}
This bound will be used only at the invariant ridge optimum.
\end{proof}

\begin{lemma}[Invariant ridge optimum]\label[lemma]{seqadam:lem:ridge-optimum}
For $\lambda>0$, the unique minimizer $\Theta_\lambda$ of $J_0=F_0+\lambda\|\Theta\|_F^2/2$ belongs to the invariant space of \cref{seqadam:lem:source-closure} and satisfies \eqref{seqadam:adam-src-ridge-risk}. Under \eqref{seqadam:adam-src-ridge-choice}, its excess is at most $e$, its norm at most $R_\lambda$, and every correct output-class mass on a legal-pattern prefix is at least $\exp(-Ne)$.
\end{lemma}
\begin{proof}
Define $J_0(\Theta)=F_0(\Theta)+\lambda\norm\Theta_F^2/2$. Its unique minimizer $\Theta_\lambda$ belongs to the invariant space above: auxiliary zero-start Euclidean GD with step at most $1/(K+\lambda)$ stays in that space by \eqref{seqadam:adam-src-closure}, ridge scales its coefficients, and strong convexity gives convergence to the unique optimum. This auxiliary sequence only characterizes the optimum; the source algorithm below is Adam.

The comparator yields
\begin{align}\label{seqadam:adam-src-ridge-risk}
 F_0(\Theta_\lambda)-H_0&\le\varepsilon+\lambda D^*(\varepsilon)/2,\\
 \norm{\Theta_\lambda}_F^2&\le D^*(\varepsilon)+2\varepsilon/\lambda.\nonumber
\end{align}
Choose
\begin{equation}\label{seqadam:adam-src-ridge-choice}
 \varepsilon=e/2,\qquad \overline D=\max\{D^*(e/2),e\},\qquad
 \lambda=e/\overline D\le1,\qquad R_\lambda=\sqrt{2\overline D}.
\end{equation}
Then the excess is at most $e$ and $\norm{\Theta_\lambda}_F\le R_\lambda$. Class masses depend only on the legal class/PAD pattern. Equation~\eqref{seqadam:adam-src-excess-decomposition} therefore gives $v_s\ge\exp(-Ne)$ on every legal-pattern prefix, including unseen words and incorrect state histories.
\end{proof}

\begin{lemma}[Transfer to the realized physical head]\label[lemma]{seqadam:lem:physical-transfer}
Let $\Theta_\lambda$ be as in \cref{seqadam:lem:ridge-optimum}, and assume $\|C^\top C-I\|_{\rm op}\le\gamma<1$. The minimizer $W^*$ of \eqref{seqadam:adam-src-physical-objective} satisfies \eqref{seqadam:adam-src-physical-transfer} and the uniform logit bound \eqref{seqadam:adam-src-logit-transfer}. A total logit error at most $h$ relative to $\Theta_\lambda\Phi$ gives correct class mass at least $\exp(-Ne-2h)$ and leakage at most $Ne+2h$.
\end{lemma}
\begin{proof}
Let $C$ be the realized hard-feature operator, $G_C=C^\top C$, and $\norm{G_C-I}_\op\le\gamma<1$. The fixed-feature physical head objective
\begin{equation}\label{seqadam:adam-src-physical-objective}
 J_C(W)=F_+(WC)+\lambda\norm W_F^2/2
\end{equation}
is $\lambda$-strongly convex in the complete physical head, including its nullspace components. At $\widetilde W=\Theta_\lambda C^\top$,
\begin{align}\label{seqadam:adam-src-physical-transfer}
 \nabla J_C(\widetilde W)
   &=[\nabla F_+(\Theta_\lambda G_C)+\lambda\Theta_\lambda]C^\top,\\
 r:=\norm{\nabla J_C(\widetilde W)}_F
   &\le\norm C_\op(\mu_F+K\gamma R_\lambda),\qquad
 \norm{W^*-\widetilde W}_F\le r/\lambda,\nonumber
\end{align}
where $W^*$ minimizes $J_C$. Thus, uniformly over all input strings,
\begin{equation}\label{seqadam:adam-src-logit-transfer}
 \norm{(W^*C-\Theta_\lambda)\Phi(x)}_\infty
 \le\sqrt K\left[\gamma R_\lambda+
          \frac{(1+\gamma)(\mu_F+K\gamma R_\lambda)}\lambda\right].
\end{equation}
The tag-conditioned source perturbation is paid once through strong convexity. A total uniform logit error at most $h$ relative to $\Theta_\lambda\Phi$ gives class mass at least $\exp(-Ne-2h)$ and leakage at most $Ne+2h$.
\end{proof}

\begin{lemma}[Descent with a current-sample adaptive denominator]\label[lemma]{seqadam:lem:adaptive-descent}\label[lemma]{seqadam:adam-src-descent}
We use an inexact first-moment estimate that allows the preconditioner to depend on all current samples. Let $F$ be globally $L$-smooth and bounded below by $F^*$. Write $g_t=\nabla F(x_t)$, $z_t=g_t+\xi_t$, $\norm{\xi_t}\le b$, and let $\widehat m_t(z)$ be the bias-corrected first moment with parameter $\beta\in[0,1)$ and zero initial moment. Suppose
\begin{equation}\label{seqadam:adam-src-inexact-update}
 x_{t+1}=x_t-\eta P_t\widehat m_t(z)+r_t,\qquad
 \norm{r_t}\le\eta b_r,\qquad
 \lambda_pI\preceq P_t\preceq\Lambda_pI.
\end{equation}
Set $e_{\rm err}=\Lambda_pb+b_r$ and $u=1-\beta$. If
\begin{equation}\label{seqadam:adam-src-descent-step}
 \eta\le\frac{\lambda_pu^{3/2}}{8L\Lambda_p^2},
\end{equation}
then every finite prefix satisfies
\begin{equation}\label{seqadam:adam-src-descent-inequality}
 F(x_{T+1})+\frac{\eta\lambda_p}{2}\sum_{t=1}^T\norm{g_t}^2
 \le F(x_1)+\frac{3\eta T e_{\rm err}^2}{2\lambda_p}.
\end{equation}
In particular, with $\Delta=F(x_1)-F^*$,
$\sum_t\norm{g_t}^2\le2\Delta/(\eta\lambda_p)+3Te_{\rm err}^2/\lambda_p^2$.

\end{lemma}
\begin{proof}
For the proof, write $h_t=\widehat m_t(g)$ and absorb the corrected noise into an update error of norm at most $e_{\rm err}$. Let $S_g=\sum_t\norm{g_t}^2$ and $S_h=\sum_t\norm{h_t}^2$. Corrected-weight Jensen gives $S_h\le S_g/u$, and the first-moment lag identity gives
\begin{equation}\label{seqadam:adam-src-momentum-lag}
 \sqrt{\sum_t\norm{h_t-g_t}^2}
 \le\frac{L\eta\Lambda_p\beta}{u^{3/2}}\sqrt{S_g}
       +\frac{L\eta e_{\rm err}\beta}{u}\sqrt T.
\end{equation}
Smoothness and the squared-step bound imply
\begin{align}\label{seqadam:adam-src-descent-proof}
 F(x_{T+1})&\le F(x_1)-\frac{3\eta\lambda_p}{4}S_g
       +\eta D_m e_{\rm err}\sqrt{TS_g}+L\eta^2Te_{\rm err}^2,\\
 D_m&=1+L\eta\Lambda_p\beta/u\le9/8.\nonumber
\end{align}
The lag and squared-step losses cost at most
$\lambda_p(\beta+\sqrt u)/8\le\lambda_p/4$ in the descent coefficient.
Young's inequality absorbs the mixed term using
$D_m^2/\lambda_p+L\eta\le89/(64\lambda_p)<3/(2\lambda_p)$, proving \eqref{seqadam:adam-src-descent-inequality}.

\end{proof}

\begin{corollary}[Finite stopping and a closed parameter tube]\label[corollary]{seqadam:cor:descent-stop}
Under \cref{seqadam:lem:adaptive-descent}, the choices \eqref{seqadam:adam-src-descent-consequence} give an inspected gradient of norm at most $r$. If in addition $F\ge\mu\|x\|^2/2$ and $F(x_1)\le F_{\rm init}$, the prefix-radius bound below holds. With the stated headroom $B_0$, radius and step cap, \eqref{seqadam:adam-src-tube} closes the stopped tube by first exit.
\end{corollary}
\begin{proof}
For a target $r>0$, the conditions
\begin{equation}\label{seqadam:adam-src-descent-consequence}
 e_{\rm err}\le\lambda_pr/4,\qquad
 T\ge\max\left\{1,\left\lceil\frac{4\Delta}{\eta\lambda_pr^2}\right\rceil\right\}
\end{equation}
ensure $\norm{g_t}\le r$ at some inspected iterate. If $F\ge\mu\norm x^2/2$ and $F(x_1)\le F_{\rm init}$, the squared radius of every prefix is at most
$2F_{\rm init}/\mu+3\eta T e_{\rm err}^2/(\mu\lambda_p)$.
To close a stopped tube without a circular radius, choose a headroom $B_0>0$, radius squared $2(2F_{\rm init}+B_0)/\mu$, impose also $\eta\le B_0/(\lambda_pr^2)$, and use $T=\max\{1,\lceil4F_{\rm init}/(\eta\lambda_pr^2)\rceil\}$. Then
\begin{equation}\label{seqadam:adam-src-tube}
 \eta T\lambda_pr^2\le4F_{\rm init}+B_0,\qquad
 F(x_t)\le F_{\rm init}+3F_{\rm init}/8+3B_0/32<2F_{\rm init}+B_0.
\end{equation}
Bounds established inside this predetermined tube therefore close by first exit.
\end{proof}

\begin{lemma}[Coordinate confinement under coupled ridge Adam]\label[lemma]{seqadam:lem:coordinate-confinement}\label[lemma]{seqadam:adam-src-coordinates}
Assume $0\le\beta_1<1$, $0<\beta_2<1$, $\beta_1^2<\beta_2$, $\lambda>0$, and $\epsilon_0>0$. Suppose each head-gradient coordinate is $g_{t,j}=c_{t,j}+\lambda w_{t,j}$ with $|c_{t,j}|\le C_h/\sqrt d$. Use physical step $\eta_s/\sqrt d$, $\eta_s\le1$, and physical denominator offset $\epsilon_0/\sqrt d$. Let the moments start at zero and put
\begin{equation}\label{seqadam:adam-src-moment-constant}
 K_{\rm mom}=\big[(1-\beta_2)(1-\beta_1^2/\beta_2)\big]^{-1/2},\qquad
 M_c=\max\left\{M_0,\frac{C_h}\lambda+
                       \frac{\eta_sK_{\rm mom}}{1-\beta_1}\right\}.
\end{equation}
If $|w_{1,j}|\le M_0/\sqrt d$, then $|w_{t,j}|\le M_c/\sqrt d$ at every finite step.

\end{lemma}
\begin{proof}
Indeed, weighted Cauchy--Schwarz for the complete corrected histories gives
$|\widehat m_{t,j}|\le K_{\rm mom}\sqrt{\widehat v_{t,j}}$, so each coordinate step is at most $\delta_w=\eta_sK_{\rm mom}/\sqrt d$. The corrected first-moment weights have mean age
\begin{equation}\label{seqadam:adam-src-age}
 \sum_s a_{ts}(t-s)=\frac{\beta_1}{1-\beta_1}
       -\frac{t\beta_1^t}{1-\beta_1^t}\le\frac{\beta_1}{1-\beta_1}.
\end{equation}
Consequently,
\begin{equation}\label{seqadam:adam-src-coordinate-lag}
 |\widehat m_{t,j}-\lambda w_{t,j}|
 \le C_h/\sqrt d+\lambda\delta_w\beta_1/(1-\beta_1).
\end{equation}
Outside the interval of radius
$[C_h/\lambda+\eta_sK_{\rm mom}\beta_1/(1-\beta_1)]/\sqrt d$, momentum has the coordinate's sign, so the step points inward. Inside it, one step stays within $M_c/\sqrt d$; an overshoot across zero is at most the step bound. This proves confinement by induction.

Define
\begin{equation}\label{seqadam:adam-src-preconditioner}
 G_c=C_h+\lambda\max\left\{M_0,C_h/\lambda+K_{\rm mom}/(1-\beta_1)\right\}.
\end{equation}
Then $|g_{t,j}|\le G_c/\sqrt d$, and the actual update is exactly
\begin{equation}\label{seqadam:adam-src-exact-adam}
 W_{t+1}=W_t-\eta_sP_t\widehat m_t,
 \quad P_t=\diag_j\frac1{\sqrt{d\widehat v_{t,j}}+\epsilon_0},
 \quad\frac I{G_c+\epsilon_0}\preceq P_t\preceq\frac I{\epsilon_0}.
\end{equation}
The ridge coefficient is constant throughout the moment history. These are spectral bounds on the realized denominator for coupled regularization.

\end{proof}

\begin{proposition}[Finite sampled source stopping]\label[proposition]{seqadam:prop:source-stopping}\label[proposition]{seqadam:adam-src-algorithm}
Use the source law and architecture of \cref{seqadam:app:source-budgets,seqadam:app:foundations}. Choose the precision, Gram tolerance, envelopes, positive rates and batches by \eqref{seqadam:adam-src-precision}--\eqref{seqadam:adam-src-batch}, with \eqref{seqadam:adam-src-S1} and \eqref{seqadam:adam-src-S2} satisfied. Conditional on the stated initialization, row and Gram events, the all-parameter source Adam procedure below stops within $T_s$ inspections with probability at least $1-\delta_{\rm batch}$. Its uniform legal-pattern source-logit error relative to $\Theta_\lambda\Phi$ is less than $h/2$, including unseen words and generated state histories. It uses at most $B_sT_s$ fresh records, $NB_sT_s$ targets and $T_s-1$ updates, with no verifier queries.
\end{proposition}
\begin{proof}
We now use $\beta_1=0.9$, $\beta_2=0.999$, $u=1-\beta_1$, and $\epsilon_0=1$. Every supplied parameter, including the backbone and any normalization gain, is optimized with the complete source loss plus $\lambda/2$ times the squared norm of all supplied parameters. In the fixed-head comparison \eqref{seqadam:adam-src-physical-objective}, the constant backbone penalty is omitted; it remains in the actual backbone gradients. All moments start at zero and remain carried throughout source training.

Choose $h\in(0,1)$ and
\begin{equation}\label{seqadam:adam-src-precision}
 0<e\le\min\left\{h/N,\frac{c^2(q-1)^2}{32q^2H^2N}\right\},
 \qquad\gamma=d^{-1/8}\le1/8,
\end{equation}
and use \eqref{seqadam:adam-src-ridge-choice}. Require the realized hard Gram event and
\begin{equation}\label{seqadam:adam-src-S1}
 \norm{C^\top C-I}_\op\le\gamma,\qquad
 \mu_F\le\frac{\lambda h}{64\sqrt K},\qquad
 \gamma\le\frac{\lambda h}{64\sqrt K K(R_\lambda+1)}.
\end{equation}
Then \eqref{seqadam:adam-src-logit-transfer} is less than $h/8$.

Here the source feature envelope is $A_s=12$, distinct from the canonical norm $\sqrt K$. For an initialization failure allocation $\delta_0$, put
\begin{align}\label{seqadam:adam-src-initial-envelopes}
 M_0&=\sqrt{2\log(2Vd/\delta_0)},&W_{\max}&=\sqrt V M_0,\\
 C_{\rm row}&=2\sqrt{2k+4\log(2d/\delta_0)},&
 C_h&=2\sqrt K C_{\rm row}+3,\nonumber
\end{align}
where $k$ is the canonical feature dimension. The row and readout events are
$\max_i\sqrt d\norm{C_i}\le C_{\rm row}$ and
$\max_{y,i}|W_{0,yi}|\le M_0/\sqrt d$.
Use \eqref{seqadam:adam-src-preconditioner} and define
\begin{align}\label{seqadam:adam-src-budgets}
 \lambda_p&=(G_c+1)^{-1},&\Lambda_p&=1,&L&=2K+1,\\
 F_{\rm init}&=\log V+2A_sW_{\max}+\lambda W_{\max}^2/2,
 &R&=\sqrt{2(2F_{\rm init}+1)/\lambda},\nonumber\\
 \rho&=\lambda h/(64\sqrt K),&r_{\rm goal}&=\rho/4,\nonumber\\
 \eta_s&=\min\{1,\lambda_pu^{3/2}/(8L),1/(\lambda_pr_{\rm goal}^2)\},
 &b&=\lambda_p\rho/32,\nonumber\\
 T_s&=\max\left\{1,\left\lceil\frac{4F_{\rm init}}
                   {\eta_s\lambda_pr_{\rm goal}^2}\right\rceil\right\},\nonumber\\
 \varepsilon_{\rm feat}&=\min\{1,b/[2(A_sR+\sqrt2)],h/(8R)\}.
 \nonumber
\end{align}
In addition require
\begin{equation}\label{seqadam:adam-src-S2}
 \gamma\le\varepsilon_{\rm feat}/16.
\end{equation}
Choose the initial raw soft/hard error using the finite temperature construction in \cref{seqadam:app:foundations} to be at most
\begin{equation}\label{seqadam:adam-src-raw-tolerance}
 z_{\rm raw}=\min\{\varepsilon_{\rm feat}/(8\sqrt2),1/(4\sqrt d)\}.
\end{equation}
The normalized member then has initial discrepancy at most
$\sqrt2z_{\rm raw}+4\gamma\le3\varepsilon_{\rm feat}/8$; the base member has a smaller bound. Both initial coordinate envelopes are at most $(C_h-1)/\sqrt d$.

Let $D$ be the complete feature-derivative bound on the radius-one backbone ball and let $P_\psi$ count all supplied backbone coordinates. Set
\begin{equation}\label{seqadam:adam-src-backbone-rate}
 r_\psi=\min\{1,1/D,\varepsilon_{\rm feat}/(2D),1/(D\sqrt d)\},
 \qquad
 \rho_{\rm back}=\min\left\{1,
       \frac{r_\psi\sqrt d}{2\eta_sT_sK_{\rm mom}\sqrt{P_\psi}}\right\}>0.
\end{equation}
Every backbone coordinate uses physical step $\eta_s\rho_{\rm back}/\sqrt d$. The Adam coordinate-step bound gives total displacement at most $r_\psi/2$, independently of the batches and including the coupled regularizer gradients. Thus all feature coordinates remain bounded by $C_h/\sqrt d$ and the full-feature discrepancy stays at most $\varepsilon_{\rm feat}$. This also includes the normalization gain when present.

With $P_{\rm head}=Vd$, choose
\begin{equation}\label{seqadam:adam-src-batch}
 B_s=\left\lceil8P_{\rm head}A_s^2b^{-2}
                 \log\frac{2P_{\rm head}T_s}{\delta_{\rm batch}}\right\rceil.
\end{equation}
At each current source iterate, draw $B_s$ fresh complete records and compute the averaged all-target gradient, including the coupled regularizer. If the head gradient's Frobenius norm is at most $\rho/2$, stop before any update. Otherwise update all blocks using actual bias-corrected Adam, unless this was the $T_s$th check; a failed final check returns failure without another update. A stopping batch is not inserted into a later moment. If the final check fails, mark source acquisition as failed and pass the final inspected parameter state to the two prescribed feedback batches; take no further source update. This algorithm uses at most $T_s$ batches and $T_s-1$ updates on every outcome. Its selection statistic uses only source observations.

Explicitly, for each performed update and every supplied coordinate, its actual minibatch gradient $g_t$ enters
\begin{equation}\label{seqadam:adam-src-moment-recursion}
 m_t=\beta_1m_{t-1}+(1-\beta_1)g_t,\quad
 v_t=\beta_2v_{t-1}+(1-\beta_2)g_t^2,\quad
 \widehat m_t=\frac{m_t}{1-\beta_1^t},\quad
 \widehat v_t=\frac{v_t}{1-\beta_2^t}.
\end{equation}
The coordinate subtracts its prescribed physical step times
$\widehat m_t/(\sqrt{\widehat v_t}+1/\sqrt d)$. All blocks update simultaneously, with their strictly positive head or backbone rates.

Within $\norm W_F\le R$, the difference between actual-feature and hard-reference population head gradients is at most
$(A_sR+\sqrt2)\varepsilon_{\rm feat}\le b/2$.
Each coordinate of a complete-record CE gradient lies in $[-A_s,A_s]$. Conditional scalar Hoeffding and a union bound over the $P_{\rm head}$ coordinates and $T_s$ predictable inspections give a batch/mean Frobenius error at most $b/2$, with failure at most $\delta_{\rm batch}$. Thus every inspected actual gradient differs from $\nabla J_C(W)$ by at most $b$ inside the stopped tube.

Coordinate confinement makes the effective preconditioner in \eqref{seqadam:adam-src-exact-adam} lie between $\lambda_pI$ and $I$. Apply \eqref{seqadam:adam-src-descent-inequality} with $\eta=\eta_s$, noise bound $b$, no additive update error, and target $r_{\rm goal}$. The tube construction uses $\mu=\lambda$ and headroom one. If the final check fails, extend its already computed gradient by one counterfactual update only in this argument. The bound then gives some inspected gradient norm at most $r_{\rm goal}$, while every prefix loss remains below $2F_{\rm init}+1$. First exit closes $\norm W_F<R$. Since $b\le\rho/32$, such a reference gradient necessarily triggers the observable threshold $\rho/2$. Hence failure cannot occur on the good event. Conversely, a triggering gradient gives $\norm{\nabla J_C(W)}_F\le\rho/2+b<\rho$.

Strong convexity now yields
\begin{equation}\label{seqadam:adam-src-stopped-accuracy}
 \norm{W-W^*}_F\le\rho/\lambda,\qquad
 \sqrt K\norm C_\op\rho/\lambda<h/32,\qquad
 R\varepsilon_{\rm feat}\le h/8.
\end{equation}
Together with \eqref{seqadam:adam-src-S1}, these bounds give uniform source-logit discrepancy less than $h/2$ from $\Theta_\lambda\Phi$, including unseen legal-pattern words and incorrect generated state histories. The canonical class masses are at least $\exp(-Ne)$ and $|b_j-a|\le c/(8H)$. The actual source policy inherits the corresponding explicit leakage and transition perturbation bounds, and its initial control choice is close to uniform because the canonical control rows agree. Its worst-case consumption is $B_sT_s$ fresh records, $NB_sT_s$ raw next-token targets, at most $T_s-1$ updates, and $T_s$ gradient-norm checks, with zero verifier calls.
\end{proof}

\begin{lemma}[Width requirements without self-reference]\label[lemma]{seqadam:lem:source-width}\label[lemma]{seqadam:adam-src-width}
For the source parameters in \cref{seqadam:prop:source-stopping}, the envelopes \eqref{seqadam:adam-src-dimension-envelopes} are independent of $d$. The explicit bounds \eqref{seqadam:adam-src-width-requirements} ensure the feature condition \eqref{seqadam:adam-src-S2} and the Gram-accuracy condition in \eqref{seqadam:adam-src-S1}. With $\gamma=d^{-1/8}$, an initialization requirement $d\ge D_\Gamma\gamma^{-2}$ is satisfied by $d\ge D_\Gamma^{4/3}$.
\end{lemma}
\begin{proof}
Write $t_d=1+\log d$ and define quantities independent of $d$:
\begin{align}\label{seqadam:adam-src-dimension-envelopes}
 \overline m&=\sqrt{2\max\{1,\log(2V/\delta_0)\}},\\
 \overline C&=2\sqrt{2k+4\log(2/\delta_0)+4},\qquad
 \overline c_h=2\sqrt K\,\overline C+3,\nonumber\\
 \overline g&=1+\overline c_h+
       \max\{\lambda\overline m,\overline c_h+\lambda K_{\rm mom}/u\},\nonumber\\
 \overline f&=\log V+2A_s\sqrt V\,\overline m
                     +(\lambda V/2)\overline m^2,\qquad
 \overline r=\sqrt{2(2\overline f+1)/\lambda}.\nonumber
\end{align}
Then $M_0\le\overline m\sqrt{t_d}$, $C_h\le\overline c_h\sqrt{t_d}$,
$G_c+1\le\overline g\sqrt{t_d}$, $F_{\rm init}\le\overline f t_d$, and $R\le\overline r\sqrt{t_d}$. Put
\begin{equation}\label{seqadam:adam-src-Bdim}
 B_{\rm dim}=\max\{16,1024\overline g(A_s\overline r+\sqrt2)/\rho,
                                      128\overline r/h\}.
\end{equation}
The prescribed feature tolerance satisfies
$\varepsilon_{\rm feat}\ge16/[B_{\rm dim}(1+\log d)]$.
Since $\log d\le16d^{1/16}/\exp(1)$, the explicit requirements
\begin{equation}\label{seqadam:adam-src-width-requirements}
 d\ge[B_{\rm dim}(1+16/\exp(1))]^{16},\qquad
 d\ge\left[\frac{64\sqrt K K(R_\lambda+1)}{\lambda h}\right]^8
\end{equation}
imply \eqref{seqadam:adam-src-S2} and the Gram-accuracy inequality in \eqref{seqadam:adam-src-S1}. An initializer requirement $d\ge D_\Gamma\gamma^{-2}$ becomes $d\ge D_\Gamma^{4/3}$ under $\gamma=d^{-1/8}$. The feature and head widths can then be instantiated without self-reference. The separate signed-source inequality in \eqref{seqadam:adam-src-S1}, and any feedback-driven tightening of $\varepsilon_{\rm feat}$, are handled in \cref{seqadam:app:adam-joint}.
\end{proof}

\begin{lemma}[Realized row envelope]\label[lemma]{seqadam:lem:source-row}
Condition on coverage and the actual Gram event $\|C^\top C-I\|_{\rm op}\le\gamma\le1/8$. If \eqref{seqadam:adam-src-row-width} holds, then $\max_i\sqrt d\|C_i\|\le C_{\rm row}$, with $C_{\rm row}$ in \eqref{seqadam:adam-src-initial-envelopes}, except with conditional probability at most $\delta_0$.
\end{lemma}
\begin{proof}
For completeness, the row event used above follows from left invariance conditional on coverage and the actual Gram event. In distribution,
\begin{equation}\label{seqadam:adam-src-polar-row}
 C=\frac Z{\sqrt d}H_Z^{-1/2}R_C,\qquad
 H_Z=Z^\top Z/d,\qquad R_C=(C^\top C)^{1/2},
\end{equation}
where $Z$ has independent standard Gaussian entries and is independent of $R_C$. If
\begin{equation}\label{seqadam:adam-src-row-width}
 d\ge128[k\log9+\log(4/\delta_0)],
\end{equation}
then $\norm{H_Z-I}_\op\le1/2$ except with probability $\delta_0/2$. On this event,
$\norm{H_Z^{-1/2}R_C}_\op\le3/2<2$ because $\gamma\le1/8$.
The Gaussian moment-generating function at $1/4$ gives
\begin{equation}\label{seqadam:adam-src-row-tail}
 \Pr\{\norm{z_i}^2>2k+4\log(2d/\delta_0)\}\le\delta_0/(2d).
\end{equation}
A row union bound therefore supplies $\max_i\sqrt d\norm{C_i}\le C_{\rm row}$ with conditional failure at most $\delta_0$. This does not assert independence of the realized feature rows. The readout-coordinate event has a separate failure allocation $\delta_0$; coverage, full-Gram, parameter-envelope, and source-batch failures are added in \cref{seqadam:app:adam-joint}.
\end{proof}

\subsection{Two sampled Adam updates after source training}
\label{seqadam:app:fb-main}

The source model constructed in Appendix~\ref{seqadam:app:source-budgets} supplies a
uniform approximation to a canonical executor. We show that two complete,
on-policy feedback batches identify the downstream binding while every supplied
parameter continues to train. The argument first controls the actual Adam
histories, then compares the correlated finite feature operator with a Gaussian
operator uniformly over adaptive histories, and finally couples complete
generated trajectories. Appendix~\ref{seqadam:app:adam-joint} selects the dimensions and
precisions simultaneously and combines the probability events.

Throughout this appendix, $k$ denotes the canonical feature dimension,
$K=12$ is its squared feature norm, and $A=\sqrt K$.
The hard feature map from Appendix~\ref{seqadam:app:foundations} is
$h_{\rm hard}(x)=C\Phi(x)$, with $C\in\R^{d\times k}$ and
$\norm{\Phi(x)}^2=K$. We write $\norm{\cdot}_F$ for the Frobenius norm.
The source and feedback stages use $\beta_1=0.9$ and $\beta_2=0.999$.
At feedback entry the parameters are retained, the source regularizer is removed,
and all first and second moments are reset. Feedback ascent uses the complete
terminal-reward score, baseline zero, and the original batch denominator.

We record a bound that remains valid when the gradients and denominators are
dependent. For a stage-local history $g_1,\ldots,g_t$, initialized with zero
moments, define
\begin{align}
 a_{t,s}&=\frac{(1-\beta_1)\beta_1^{t-s}}{1-\beta_1^t},&
 b_{t,s}&=\frac{(1-\beta_2)\beta_2^{t-s}}{1-\beta_2^t},\label{seqadam:adam-fb-weights}\\
 \widehat m_t(g)&=\sum_{s=1}^t a_{t,s}g_s,&
 \widehat v_t(g)&=\sum_{s=1}^t b_{t,s}g_s^{\odot2}.
\end{align}
Both weight sequences sum to one. For $0\le\beta_1<1$,
$0<\beta_2<1$, and $\beta_1^2<\beta_2$, put
\begin{equation}
 K_{\rm mom}=\left[(1-\beta_2)
                 (1-\beta_1^2/\beta_2)\right]^{-1/2}.
 \label{seqadam:adam-fb-kmom}
\end{equation}

\begin{lemma}[Magnitude and history perturbation]
\label[lemma]{seqadam:adam-fb-history-lemma}\label[lemma]{seqadam:adam-fb-histories}
For every coordinate and every finite history,
$|\widehat m_t(g)|\le K_{\rm mom}\sqrt{\widehat v_t(g)}$.
For $\epsilon>0$, let
$\mathcal D_t(g)=\widehat m_t(g)/(\sqrt{\widehat v_t(g)}+\epsilon)$,
coordinatewise. Two histories with the same zero initialization satisfy
\begin{equation}
 \norm{\mathcal D_t(g)-\mathcal D_t(h)}_\infty
 \le\frac{1+K_{\rm mom}}{\epsilon}
       \max_{s\le t}\norm{g_s-h_s}_\infty.
 \label{seqadam:adam-fb-history-bound}
\end{equation}
In particular, every Adam coordinate direction has magnitude at most
$K_{\rm mom}$, without an upper gradient bound or a positive lower second moment.
\end{lemma}

\begin{proof}
For one coordinate, weighted Cauchy--Schwarz gives
\begin{align}
 |\widehat m_t(g)|^2
 &\le K_t^2\widehat v_t(g),\\
 K_t^2=\sum_{s=1}^t\frac{a_{t,s}^2}{b_{t,s}}
 &=\frac{(1-\beta_1)^2}{(1-\beta_1^t)^2}
   \frac{1-\beta_2^t}{1-\beta_2}
   \sum_{j=0}^{t-1}(\beta_1^2/\beta_2)^j
 \le K_{\rm mom}^2.\label{seqadam:adam-fb-weight-cauchy}
\end{align}
Here $1-\beta_1^t\ge1-\beta_1$; when $\beta_1=0$, the first-moment
weight is concentrated at the newest gradient and the zeroth power is one.
Write $r_g=(\sum_s b_{t,s}g_s^2)^{1/2}$ and define $r_h$ similarly.
If $d_t=\max_{s\le t}|g_s-h_s|$, then
\[
 |\widehat m_t(g)-\widehat m_t(h)|\le d_t,
 \qquad |r_g-r_h|\le d_t
\]
by the reverse triangle inequality in the weighted Euclidean norm.
The identity
\[
 \frac{m_g}{r_g+\epsilon}-\frac{m_h}{r_h+\epsilon}
 =\frac{m_g-m_h}{r_g+\epsilon}
  +\frac{m_h}{r_h+\epsilon}\frac{r_h-r_g}{r_g+\epsilon}
\]
now proves~\eqref{seqadam:adam-fb-history-bound}, including coordinates with zero
second moment.
\end{proof}

The bound uses the complete history since the common zero initialization.
Source moments are retained during source training and reset once at the stage
boundary; both feedback histories enter the second update.

The finite coefficient rows are correlated. Their relevant symmetry follows
from the complete initialization in Appendix~\ref{seqadam:app:foundations}.

\begin{lemma}[Initial left invariance]
\label[lemma]{seqadam:adam-fb-left-invariance}
Conditional on any relative-bias realization satisfying lag coverage, the
base member's actual hard coefficient operator obeys
$C\stackrel{\rm law}=QC$ for every deterministic $Q\in O(d)$.
Its canonical feature coordinates and lag weights are unchanged by this
transformation.
\end{lemma}

\begin{proof}
Distinguish the FFN matrices $B_{\rm ff},C_{\rm ff},U_{\rm ff}$ from
the coefficient operator. Apply the deterministic transformation
\begin{align*}
 E&\mapsto QE,& O_h&\mapsto QO_h,\\
 Q_h&\mapsto Q_hQ^\top,&K_h&\mapsto K_hQ^\top,
 &V_h&\mapsto V_hQ^\top,\\
 B_{\rm ff}&\mapsto B_{\rm ff}Q^\top,
 &C_{\rm ff}&\mapsto C_{\rm ff}Q^\top,
 &U_{\rm ff}&\mapsto QU_{\rm ff},\\
 b_{\rm out}&\mapsto Qb_{\rm out},
 &W_{\rm out}&\mapsto W_{\rm out}Q^\top.
\end{align*}
Leave the relative biases and FFN input biases unchanged. Each Gaussian block
has its original law under its indicated left or right orthogonal transform;
the transformations also preserve joint independence. This remains true
conditional on the unchanged relative biases.

The query, key, and value vectors are unchanged, so all soft attention scores
and weights are unchanged. The residual feature transforms as $z\mapsto Qz$;
the FFN preactivations are unchanged and its output transforms by $Q$.
For the hard reference, the lag blocks satisfy $A_\ell' =QA_\ell$ and
$T'=QT$, since the lag-weight matrix is unchanged and invertible on coverage.
The augmented FFN coefficient rows $[B_{\rm ff}T,b_{\rm ff}]$ and
$[C_{\rm ff}T,c_{\rm ff}]$ are unchanged, hence so is their symmetric-product
coefficient matrix $F$. The actual expansion therefore gives
\[
 C'=[QT,QU_{\rm ff}F,Qb_{\rm out}]=QC,
 \qquad (C')^\top C'=C^\top C.
\]
This proves the identity for the coefficient columns themselves, including
their constant and linear terms, rather than only on sampled inputs.
\end{proof}

This lemma concerns the raw initial operator. For the normalized member the
same operator is used as a reference, and the normalization and trainable gain
are paid through the feature approximation. No trajectory equivariance of
coordinatewise Adam is required.

For rows $c_i^\top$ of $C$, define the finite and Gaussian operators
\begin{align}
 \mathcal A_C(S)
 &=\sum_{i=1}^d\frac{c_ic_i^\top}
                  {\sqrt{d\,c_i^\top Sc_i}+\epsilon},&
 \mathcal A(S)
 &=\E_{z\sim N(0,I_k)}
       \frac{zz^\top}{\sqrt{z^\top Sz}+\epsilon}.
 \label{seqadam:adam-fb-operators}
\end{align}
The next estimate is uniform over the second-moment matrices, which permits
their dependence on the realized features and training batches.

\begin{lemma}[Uniform correlated adaptive geometry]
\label[lemma]{seqadam:adam-fb-uniform-geometry}\label[lemma]{seqadam:app:adam-geometry}
Let $k\ge1$, $d\ge k$, $G\ge0$, $\epsilon>0$,
$0<\tau\le1$, and $0<\delta_{\rm geo}<1$. Define
\begin{align}
 K_3&=\frac{8[k(k+2)]^{3/4}}{\delta_{\rm geo}},&
 L_{\rm pert}&=\frac{6k}{\epsilon}
                    +\frac{GK_3}{\epsilon^2},\\
 \gamma_{\rm geo}&=\min\left\{\frac18,
                  \frac{\tau}{16L_{\rm pert}}\right\},&
 a_{\rm geo}&=\frac\tau2,\\
 T_{\rm tr}^2&=\max\left\{1,
            \frac{64k(k+2)}{\epsilon a_{\rm geo}\delta_{\rm geo}}\right\},&
 h_{\rm net}&=\left(\frac{a_{\rm geo}\epsilon^2}
                              {16T_{\rm tr}^3}\right)^2,\\
 D_{\rm sym}&=\frac{k(k+1)}2,&
 N_{\rm net}&=\left(1+\frac{2\sqrt{k}G^2}{h_{\rm net}}\right)^{D_{\rm sym}}.
 \label{seqadam:adam-fb-geometry-constants}
\end{align}
Suppose $C$ has a left-orthogonally invariant law and the event
$\mathcal E_C=\{\norm{C^\top C-I}_{\op}\le\gamma_{\rm geo}\}$
has positive probability. Set
\begin{equation}
 d_{\rm geo}=\left\lceil\max\left\{
 \begin{aligned}
 &32\gamma_{\rm geo}^{-2}
       \left[k\log9+\log\frac{16}{\delta_{\rm geo}}\right],\\
 &\frac{32k^2T_{\rm tr}^4}{\epsilon^2a_{\rm geo}^2}
       \log\frac{16k^2N_{\rm net}}{\delta_{\rm geo}}
 \end{aligned}
 \right\}\right\rceil.
 \label{seqadam:adam-fb-geometry-width}
\end{equation}
If $d\ge d_{\rm geo}$, then, conditional on $\mathcal E_C$, with probability
at least $1-\delta_{\rm geo}$,
\begin{equation}
 \sup_{S\succeq0,\,\norm{S}_{\op}\le G^2}
       \norm{\mathcal A_C(S)-\mathcal A(S)}_{\op}\le\tau.
 \label{seqadam:adam-fb-uniform-event}
\end{equation}
The unconditional failure probability is at most
$\Pr(\mathcal E_C^c)+\delta_{\rm geo}$.
\end{lemma}

\begin{proof}
On $\mathcal E_C$, let $R=(C^\top C)^{1/2}$, which is positive definite and
satisfies $\norm{R-I}_{\op}\le\gamma_{\rm geo}$. Conditional on $R$, left
invariance makes $CR^{-1}$ a uniform Stiefel matrix. Averaging an independent
uniform left rotation identifies this conditional law and its independence
from $R$. Thus, conditional on $\mathcal E_C$, we may represent
\begin{equation}
 C=d^{-1/2}ZM,\qquad M=H_Z^{-1/2}R,\qquad
 H_Z=Z^\top Z/d,
 \label{seqadam:adam-fb-polar}
\end{equation}
where $Z$ has independent standard Gaussian entries and is independent of $R$.
This representation does not assert independence of the rows of $C$.

For a fixed unit vector $v$, the chi-square moment generating function gives
\[
 \Pr\bigl(|v^\top(H_Z-I)v|>b\bigr)
 \le2\exp(-db^2/8),\qquad 0<b<1.
\]
A $1/4$-net of size at most $9^k$ and the symmetric-matrix net inequality give
\[
 \Pr\bigl(\norm{H_Z-I}_{\op}>\gamma_{\rm geo}\bigr)
 \le2\exp\{k\log9-d\gamma_{\rm geo}^2/32\}
 \le\delta_{\rm geo}/8.
\]
On the complementary event,
\begin{equation}
 \begin{aligned}
 \norm{H_Z^{-1/2}-I}_{\op}&\le2\gamma_{\rm geo},&
 \norm{R}_{\op}&\le1+\gamma_{\rm geo},\\
 \norm{M-I}_{\op}&\le4\gamma_{\rm geo}\le\tfrac12,&
 d^{-1}\sum_i\norm{z_i}^2&\le2k.
 \end{aligned}
 \label{seqadam:adam-fb-polar-bounds}
\end{equation}
Since $\E\norm{z}^3\le[k(k+2)]^{3/4}$, Markov's inequality also gives
$d^{-1}\sum_i\norm{z_i}^3\le K_3$ except with probability
$\delta_{\rm geo}/8$.

Write
$\mathcal A_Z(S)=d^{-1}\sum_i z_iz_i^\top/
(\sqrt{z_i^\top Sz_i}+\epsilon)$.
Superscript ${\rm tr}$ denotes truncation to $\norm{z}\le T_{\rm tr}$.
Uniformly over $S$,
\begin{equation}
 \norm{\mathcal A(S)-\mathcal A^{\rm tr}(S)}_{\op}
 \le\frac{\E[\norm z^2\mathbf1_{\{\norm z>T_{\rm tr}\}}]}\epsilon
 \le\frac{k(k+2)}{\epsilon T_{\rm tr}^2}
 \le\frac{a_{\rm geo}\delta_{\rm geo}}{64}.
 \label{seqadam:adam-fb-truncation}
\end{equation}
The corresponding empirical tail is at most $a_{\rm geo}/8$ except with
probability $\delta_{\rm geo}/8$, by Markov's inequality.

The positive semidefinite (PSD) matrices of operator norm at most $G^2$ lie in a Frobenius ball of
radius $\sqrt kG^2$ in dimension $D_{\rm sym}$. A packing argument supplies
an $h_{\rm net}$-net, with centers in that set, of size at most $N_{\rm net}$.
Each truncated matrix entry has magnitude at most $T_{\rm tr}^2/\epsilon$.
Scalar Hoeffding at entry tolerance $a_{\rm geo}/(4k)$, unioned over the net
and $k^2$ entries, has failure probability at most
\begin{equation}
 2k^2N_{\rm net}
 \exp\left\{-\frac{da_{\rm geo}^2\epsilon^2}{32k^2T_{\rm tr}^4}\right\}
 \le\delta_{\rm geo}/8.
 \label{seqadam:adam-fb-net-tail}
\end{equation}
At every net point the operator error is at most $a_{\rm geo}/4$, since an
entrywise bound $b$ gives operator norm at most $kb$.
For two PSD matrices,
\[
 |\sqrt{z^\top Sz}-\sqrt{z^\top S'z}|
 \le\norm z\sqrt{\norm{S-S'}_{\op}}.
\]
Consequently each truncated kernel changes by at most
$T_{\rm tr}^3\sqrt{h_{\rm net}}/\epsilon^2$ within a net ball.
The empirical and population extension costs together are at most
$2T_{\rm tr}^3\sqrt{h_{\rm net}}/\epsilon^2=a_{\rm geo}/8$.
Combining both tails, the net bound, and its extension gives
\begin{equation}
 \sup_S\norm{\mathcal A_Z(S)-\mathcal A(S)}_{\op}
 \le\frac{a_{\rm geo}}8+\frac{a_{\rm geo}}{64}
       +\frac{a_{\rm geo}}4+\frac{a_{\rm geo}}8
 =\frac{33a_{\rm geo}}{64}<a_{\rm geo}.
 \label{seqadam:adam-fb-gaussian-uniform}
\end{equation}

It remains to transfer through $M$. For
$\mathcal K_S(x)=xx^\top/(\sqrt{x^\top Sx}+\epsilon)$ and
$\kappa=\norm{M-I}_{\op}\le1/2$, direct subtraction gives
\begin{equation}
 \norm{\mathcal K_S(M^\top y)-\mathcal K_S(y)}_{\op}
 \le\kappa\left[
      \frac{(2+\kappa)\norm y^2}{\epsilon}
          +\frac{G\norm y^3}{\epsilon^2}\right].
 \label{seqadam:adam-fb-kernel-perturbation}
\end{equation}
Here the numerator difference is bounded by
$(\norm{M^\top y}+\norm y)\norm{(M^\top-I)y}$; both denominators
are at least $\epsilon$, and their difference is at most
$G\norm{(M^\top-I)y}$.
Averaging~\eqref{seqadam:adam-fb-kernel-perturbation} and using the second- and
third-moment bounds gives
\[
 \sup_S\norm{\mathcal A_C(S)-\mathcal A_Z(S)}_{\op}
 \le4\gamma_{\rm geo}L_{\rm pert}\le\tau/4.
\]
Together with~\eqref{seqadam:adam-fb-gaussian-uniform}, this is at most
$3\tau/4\le\tau$. The Gaussian auxiliary failures total at most
$\delta_{\rm geo}/2$, uniformly over the admissible $R$; integrating its
conditional law proves the lemma.
\end{proof}

\begin{corollary}[Adaptive histories and exact coefficient scaling]\label[corollary]{seqadam:cor:coefficient-scaling}
On the event of \cref{seqadam:adam-fb-uniform-geometry}, every coefficient history with $\|u_s\|\le G$ satisfies the moment, subspace-leakage and Gaussian lower bounds below. With physical gradients $Cu_s$, denominator $\epsilon/\sqrt d$ and rate $\eta/\sqrt d$, the coefficient increment is exactly \eqref{seqadam:adam-fb-exact-scaling}, including the current adaptive denominator.
\end{corollary}
\begin{proof}
For complete coefficient histories $u_s$, set
\begin{equation}
 m_t=\sum_s a_{t,s}u_s,\qquad
 S_t=\sum_s b_{t,s}u_su_s^\top.
 \label{seqadam:adam-fb-coefficient-moments}
\end{equation}
If $\norm{u_s}\le G$, then $\norm{m_t}\le G$ and
$\norm{S_t}_{\op}\le G^2$. Equation~\eqref{seqadam:adam-fb-uniform-event}
therefore applies to histories chosen using $C$ and any training information.
For histories in a subspace $H_0$, Gaussian reflection gives
$P_{H_0^\perp}\mathcal A(S_t)m_t=0$ and hence
$\norm{P_{H_0^\perp}\mathcal A_C(S_t)m_t}\le\tau G$.
The same Gaussian operator satisfies
\[
 \mathcal A(S)\succeq
 \frac{I}{2(G\sqrt{2(k+2)}+\epsilon)}.
\]
Indeed, retain $\norm z\le\sqrt{2(k+2)}$ and use
$\E[(v^\top z)^2\norm z^2]=k+2$ for each unit $v$.
This lower bound concerns action on the momentum vector; current gradients
need not align with previous momentum.

For fixed features, a physical gradient $Cu_s$ and coefficient coordinate
$\vartheta=C^\top w$ yield the exact ascent identity
\begin{equation}
 \epsilon_{\rm physical}=\epsilon/\sqrt d,\qquad
 \eta_{\rm physical}=\eta/\sqrt d
 \quad\Longrightarrow\quad
 \Delta\vartheta=\eta\mathcal A_C(S_t)m_t.
 \label{seqadam:adam-fb-exact-scaling}
\end{equation}
The current denominator remains inside $\mathcal A_C$.
\end{proof}

\begin{lemma}[Rewards of the learned executor]\label[lemma]{seqadam:lem:executor-rewards}\label[lemma]{seqadam:adam-fb-reference}
Let $q\ge16$, $c=1/64$, $H\ge5$, and $|b_j-(1-c/H)|\le c/(8H)$ for $j=1,2$. Conditional on correct output classes, the executor rewards satisfy \eqref{seqadam:adam-fb-rewards} and \eqref{seqadam:adam-fb-reward-bounds}, including trajectories with repaired hidden-bit errors. In particular, $r_*-\bar r\ge1/2$ and $\bar r<r_*/q$.
\end{lemma}
\begin{proof}
Let $q\ge16$, $c=1/64$, and let the source conditional transition parameters
satisfy $|b_j-(1-c/H)|\le c/(8H)$ for $j=1,2$, as proved in
Appendix~\ref{seqadam:app:source-budgets}. Define
\begin{equation}
 B_{\rm exec}=b_1b_2^{H-1},\qquad
 d_j=b_j+2(1-b_j)/q,\qquad
 D_{\rm exec}=d_1d_2^{H-1}.
 \label{seqadam:adam-fb-executor-products}
\end{equation}
At a valid transition of type $j$, the correct and hidden-bit-flipped states
have probabilities $\alpha_j=b_j+(1-b_j)/q$ and
$\gamma_j=(1-b_j)/q$. Their sum is $d_j$ and difference is $b_j$.
Summing all even and odd hidden-error histories gives, conditional on the
choice and correct output classes,
\begin{equation}
 r_*=\frac{D_{\rm exec}+B_{\rm exec}}2,\qquad
 \bar r=\frac{D_{\rm exec}-B_{\rm exec}}2,\qquad
 r_a=0\quad\text{for the other public arms}.
 \label{seqadam:adam-fb-rewards}
\end{equation}
Here $r_*$ is the reward of the bound operation and $\bar r$ that of its
hidden-bit opposite. Other public operations cannot reach the required public
endpoint while satisfying the grammar and public-coordinate checks. Repaired hidden-bit histories are
included in~\eqref{seqadam:adam-fb-rewards}. Conditional on the correct output classes, the public-coordinate checks pass with probability $D_{\rm exec}$, independently of the selected control, and
\begin{equation}
 r_*\ge B_{\rm exec}\ge1-9c/8,\qquad
 0\le\bar r\le\sum_j\gamma_j\le\frac{9c}{8q},\qquad
 1-D_{\rm exec}\le9c/8.
 \label{seqadam:adam-fb-reward-bounds}
\end{equation}
The sum over $j$ counts one first transition and $H-1$ repeated transitions.
Thus $\Delta=r_*-\bar r\ge1/2$ and $\bar r<r_*/q$.
\end{proof}

\begin{lemma}[Prompt-independent canonical choice geometry]\label[lemma]{seqadam:lem:choice-mean}
For the canonical source feature map on a legal choice prompt, with $H\ge5$, let $\mu$ be its mean under the training prompt law. Then \eqref{seqadam:adam-fb-choice-mean} holds on every training and held-out choice prompt.
\end{lemma}
\begin{proof}
Let $\mu$ be the mean canonical feature of a training choice prompt. A
degree-two coordinate retains at most two controls, so $H\ge5$ leaves an
unused balanced suffix tag. Its expectation is therefore the same in either
tag class. Against each fixed legal choice prompt, averaging the kernel's
matching indicators also gives the same constant. Consequently
\begin{equation}
 \mu^\top\Phi(x)=\norm\mu^2=:\kappa\in[2,K]
 \quad\text{for every training and held-out choice prompt}.
 \label{seqadam:adam-fb-choice-mean}
\end{equation}
This identity keeps the reference choice probabilities prompt-independent.
\end{proof}

\begin{definition}[Two-step Gaussian reference]\label[definition]{seqadam:def:gaussian-reference}
The Gaussian reference starts with zero coefficient increments $Y_0=0$ and
uniform choice probabilities. Its row gradient is
\begin{equation}
 u_{t,a}=c_{t,a}\mu,\qquad
 c_{t,a}=p_{t-1,a}(r_a-J_{t-1}),\qquad
 J_{t-1}=\sum_a p_{t-1,a}r_a.
 \label{seqadam:adam-fb-bandit-gradient}
\end{equation}
Each row uses its own corrected moments and update
$\eta_t\mathcal A(S_{t,a})m_{t,a}$.
Set
\begin{align}
 \epsilon_f&=\frac1{64q},&
 \rho_0&=\frac{2e^{-2}}{\sqrt{2\pi}},&
 C_0&=\frac{2(1-\beta_1)\rho_0}{64\sqrt2+1},\\
 \eta_1&=\frac1{100K_{\rm mom}\sqrt K},&
 \eta_2&=\frac{\log(15(q-1))}{C_0},&
 S_{\rm step}&=\eta_1+\eta_2.
 \label{seqadam:adam-fb-two-step-rates}
\end{align}

\end{definition}

\begin{lemma}[Two Gaussian reference updates]
\label[lemma]{seqadam:adam-fb-two-step-reference}
With~\eqref{seqadam:adam-fb-two-step-rates}, the reference chooses the bound operation
with probability at least $15/16$ after the second update, on every training
and held-out choice prompt. Its total Frobenius path length is at most
\begin{equation}
 R_{\rm ref}=S_{\rm step}C_{\rm move},\qquad
 C_{\rm move}=K_{\rm mom}+32\sqrt K.
 \label{seqadam:adam-fb-reference-radius}
\end{equation}
\end{lemma}

\begin{proof}
For a row history, write $m_{t,a}=\bar c_{t,a}\mu$ and
$S_{t,a}=v_{t,a}\mu\mu^\top$. Gaussian symmetry gives
\begin{equation}
 \mathcal A(v\mu\mu^\top)\mu=\mu\mathcal B(v),\qquad
 \mathcal B(v)=\E_{Z\sim N(0,1)}
       \frac{Z^2}{\sqrt{\kappa v}|Z|+\epsilon_f}>0.
 \label{seqadam:adam-fb-rank-one-field}
\end{equation}
Every prompt logit increment is
$\eta_t\kappa\bar c_{t,a}\mathcal B(v_{t,a})$.
Whenever $p_{t-1,*}\ge1/q$, $J_{t-1}\ge r_*/q>\bar r$: the correct row has
positive gradient and momentum, and every wrong row has negative gradient
and momentum. Thus the correct probability is nondecreasing, and prompt
independence follows from~\eqref{seqadam:adam-fb-choice-mean}.

Before $p_{t-1,*}$ reaches $15/16$,
\begin{align}
 c_{t,*}&\ge\Delta p_{t-1,*}(1-p_{t-1,*})\ge\Delta/(16q),\\
 c_{s,*}&\le r_*p_{s-1,*}\le r_*p_{t-1,*}
       \le(16r_*/\Delta)c_{t,*}\le32c_{t,*}
       \qquad(s\le t).
 \label{seqadam:adam-fb-correct-gradient}
\end{align}
The newest first-moment weight is at least $1-\beta_1$, so
$\bar c_{t,*}\ge(1-\beta_1)c_{t,*}$ and
$\sqrt{v_{t,*}}\le32c_{t,*}$.
The event $1\le|Z|\le2$ has probability at least $\rho_0$. Therefore
\[
 \mathcal B(v_{t,*})\ge
       \frac{\rho_0}{2\sqrt{\kappa v_{t,*}}+\epsilon_f}.
\]
Writing $\chi=16q\epsilon_f/\Delta\le1$, the correct logit increment is
at least
\begin{equation}
 \eta_t(1-\beta_1)\rho_0
       \frac{\kappa}{64\sqrt\kappa+\chi}
 \ge\eta_t C_0.
 \label{seqadam:adam-fb-positive-logit}
\end{equation}
The fraction increases for $\kappa\ge2$. Wrong logits decrease. If the target
has not already been reached, the second step alone therefore raises every
correct-versus-wrong logit gap by at least $\log(15(q-1))$. Since the gaps
started at zero and increased on the first step, $p_{2,*}\ge15/16$.

For the path bound, Lemma~\ref{seqadam:adam-fb-history-lemma} and Gaussian dual
Cauchy--Schwarz bound each row field by $K_{\rm mom}$. The first step thus
changes each logit by at most $1/100$. At both gradient evaluations,
\[
 p_{t-1,a}\le e^{0.02}/q\le50/(49q),\qquad t=1,2.
\]
Using $J_{t-1}\le(3/2)(50/49)r_*/q$ and negativity of wrong coefficients gives
$|c_{t,a}|\le2r_*/q^2$ for wrong rows. Their corrected first moments have
norm at most $2r_*\sqrt K/q^2$. Since
$\norm{\mathcal A(S)}_{\op}\le1/\epsilon_f$, each wrong row field has
norm at most $128r_*\sqrt K/q$. Their aggregate norm is at most
$128\sqrt K/\sqrt q\le32\sqrt K$, while the correct row contributes at
most $K_{\rm mom}$. Summing the two step lengths proves
\eqref{seqadam:adam-fb-reference-radius}.
\end{proof}

\begin{definition}[Sampled feedback and coefficient increments]\label[definition]{seqadam:def:sampled-feedback}\label[definition]{seqadam:adam-fb-sampled-comparison}
Generate exactly $n=H+2$ full-vocabulary decisions in the designated choice,
state, and EOS slots. The verifier checks grammar, public-coordinate consistency, and the
terminal endpoint. Every sampled trajectory, including a zero-reward one,
contributes through the original full batch denominator. Define
\[
 X_t=(W_{{\rm fast},t}-W_{{\rm fast},s})C,\qquad X_0=0,
\]
where $W_{{\rm fast},s}$ is the source control-row readout. The fixed-executor
choice-only population gradient, with the actual training-prompt distribution,
is denoted $U(X)$. \end{definition}

\begin{lemma}[Choice gradients and two-step history fields]\label[lemma]{seqadam:lem:feedback-lipschitz}
Under \cref{seqadam:def:sampled-feedback}, the fixed-executor choice gradient is globally $6K$-Lipschitz as in \eqref{seqadam:adam-fb-choice-lipschitz}. For histories of length at most two, the Gaussian row fields are Lipschitz in the maximum Frobenius history error, with constant $D_A$ in \eqref{seqadam:adam-fb-field-lipschitz}.
\end{lemma}
\begin{proof}
Bounded rewards and $\norm\Phi\le A$ give the global bound
\begin{equation}
 \norm{U(X)-U(Y)}_F\le L_c\norm{X-Y}_F,
 \qquad L_c=6K.
 \label{seqadam:adam-fb-choice-lipschitz}
\end{equation}
At the reference iterates this gradient is~\eqref{seqadam:adam-fb-bandit-gradient}.

For two histories, the Gaussian row fields together are Lipschitz in the
maximum Frobenius history error, with constant
\begin{equation}
 D_A=\frac{\sqrt2(1+K_{\rm mom})}{\epsilon_f}.
 \label{seqadam:adam-fb-field-lipschitz}
\end{equation}
To see this, apply~\eqref{seqadam:adam-fb-history-bound} to each scalar history
$z^\top u_s$. The squared scalar maximum is bounded by the sum of its two
squared errors. Summing over output rows and using dual Cauchy--Schwarz against
a unit Frobenius test matrix gives~\eqref{seqadam:adam-fb-field-lipschitz}; Gaussian
second moments replace each scalar square by its coefficient norm squared.
There is no factor depending on the number of canonical coordinates or rows.
The physical-coordinate version uses epsilon $\epsilon_f/\sqrt d$.
\end{proof}

\begin{lemma}[Complete-trajectory population comparison]\label[lemma]{seqadam:lem:population-coupling}
Under \cref{seqadam:def:sampled-feedback}, set $G=n\sqrt2 A$ and impose the uniform geometry event. While the coefficient discrepancy from \cref{seqadam:def:gaussian-reference} is at most $E_*$ in \eqref{seqadam:adam-fb-stopped-radius}, assume source, slow-head and feature-motion logit error at most $2h$, canonical class leakage at most $h\le1$, and feature error at most $e_f$. Then the actual reward-weighted coefficient score satisfies \eqref{seqadam:adam-fb-F1}, with $C_{\rm pop}$ as displayed there.
\end{lemma}
\begin{proof}
Set $G=n\sqrt2 A$. Apply Lemma~\ref{seqadam:adam-fb-uniform-geometry} with this $G$,
$\epsilon=\epsilon_f$, and tolerance $\tau$. A complete coefficient score has
norm at most $G$, so the same event covers all actual row histories.

Compare the iterates until their coefficient error exceeds
\begin{equation}
 E_*=(1024A)^{-1},\qquad B_{\exp}=A(R_{\rm ref}+1).
 \label{seqadam:adam-fb-stopped-radius}
\end{equation}
Within this region $\norm{X_t}_F\le R_{\rm ref}+1$.
Let $h\le1$ be the source logit tolerance. The source entry and slow-rate
choices in Appendices~\ref{seqadam:app:source-budgets} and~\ref{seqadam:app:adam-joint}
supply three uniform bounds: the source hard-logit error plus slow-head and
feature-motion errors is at most $2h$; the actual feature error is at most
$e_f$; and the canonical source class leakage is at most $h$, using $Ne\le h$.

On each matching legal-class history, the actual incorrect-class mass is at
most $\ell=h\exp(2B_{\exp}+4)$. The factor pays the change in class odds
caused by arbitrary fast-row coefficients in the stopped region. Once an
incorrect-class token is generated the coupling has already failed, so no
class-mass bound is required on its malformed continuation. Conditional state
laws only change through the $2h$ perturbation of state rows; fast control rows
do not change the within-state normalization. Conditional choice and state
laws differ from the fixed-executor laws by at most $8h$ in total variation.

Let $g_{{\rm coeff},t}$ denote the complete reward-weighted score expressed
using $\Phi$, with the probabilities and trajectories of the actual policy.
Maximal conditional couplings on matching histories give
\begin{equation}
 \norm{\E[g_{{\rm coeff},t}\mid\mathcal F_{t-1}]-U(X_{t-1})}_F
 \le C_{\rm pop}h,\qquad
 C_{\rm pop}=256n^2A\exp(2B_{\exp}+4).
 \label{seqadam:adam-fb-F1}
\end{equation}
Here the update indexed $t$ is evaluated at iterate $t-1$.
Indeed the path total variation is at most $n(8h+\ell)$, and both complete
coefficient score vectors have norm at most $n\sqrt{2K}$. On a matched valid
path, each non-choice fast-control score has norm at most $\sqrt{2K}\ell$,
whereas the reference fast score is zero there. The choice fast-score
difference is at most $8\sqrt{2K}(h+\ell)$. The same deterministic verifier
returns the same reward on matched paths, including histories whose hidden
errors repair. Hence the expectation error is at most
\[
 2n\sqrt{2K}\,n(8h+\ell)+8n\sqrt{2K}(h+\ell)
 \le C_{\rm pop}h.
\]
This accounts for non-choice prefixes, EOS, distribution change, and malformed
trajectories without a teacher-prefix substitution.
\end{proof}

\begin{lemma}[Concentration for the two fresh feedback batches]\label[lemma]{seqadam:lem:feedback-sampling}
For the two predictable iterates in \cref{seqadam:def:sampled-feedback}, let each fresh batch have size \eqref{seqadam:adam-fb-F2}. With probability at least $1-\delta_R$, both sampled coefficient gradients differ from their conditional means in Frobenius norm by at most $\varepsilon_g$. Independence is required within each fresh batch, conditional on its past.
\end{lemma}
\begin{proof}
For a desired sampling error $\varepsilon_g$, take
\begin{equation}
 B_R=\left\lceil
 8qk n^2A^2\varepsilon_g^{-2}
       \log\frac{4qk}{\delta_R}\right\rceil.
 \label{seqadam:adam-fb-F2}
\end{equation}
Each complete-trajectory coefficient entry lies in $[-nA,nA]$.
Conditional Hoeffding and a union over $qk$ coordinates and the two predictable
iterates give both Frobenius errors at most $\varepsilon_g$ with probability
at least $1-\delta_R$; the smaller threshold $\varepsilon_g/2$ already meets
this allocation. Only trajectories within each fresh batch are independent.
\end{proof}

\begin{lemma}[History-aware moving-feature comparison]\label[lemma]{seqadam:lem:feedback-recursion}
On the events and within the stopped region of \cref{seqadam:lem:population-coupling,seqadam:lem:feedback-sampling}, suppose the actual feature error is at most $e_f$ and $\|C\|_{\rm op}\le2$. The complete physical Adam updates, including both moment histories, satisfy the coefficient-error recurrence \eqref{seqadam:adam-fb-F3} for $t=1,2$.
\end{lemma}
\begin{proof}
The actual physical fast gradient differs from $g_{{\rm coeff},t}C^\top$ by
at most $n\sqrt2 e_f$. Apply the physical history bound before projection:
for two histories the squared sum of coordinatewise maximal errors is at most
the sum of the two squared Frobenius errors. Multiplying the physical rate
$\eta_t/\sqrt d$ by the reciprocal physical epsilon $\sqrt d/\epsilon_f$
cancels $d$. Postmultiplication by $C$, with $\norm C_{\op}\le2$, therefore
costs at most $2\eta_tD_A n\sqrt2e_f$.
The fixed-feature update is exactly~\eqref{seqadam:adam-fb-exact-scaling}; its uniform
operator error costs at most $\eta_t\tau G$ across all rows.
Writing $E_t=\max_{0\le s\le t}\norm{X_s-Y_s}_F$, Gaussian history comparison
and~\eqref{seqadam:adam-fb-choice-lipschitz} give, for $t=1,2$,
\begin{equation}
 E_t\le(1+\eta_tD_A L_c)E_{t-1}
 +\eta_t\left[
     D_A(C_{\rm pop}h+\varepsilon_g)+\tau G
                       +2D_A n\sqrt2e_f\right].
 \label{seqadam:adam-fb-F3}
\end{equation}
Both the moment numerator and the current random denominator are retained.
The equality $X_0=Y_0=0$ concerns increments from the learned source;
its nonzero source function error enters~\eqref{seqadam:adam-fb-F1}.
\end{proof}

\begin{proposition}[Two sampled updates preserve the reference margin]\label[proposition]{seqadam:prop:feedback-transfer}
Use \cref{seqadam:def:sampled-feedback}, the source guarantee of \cref{seqadam:prop:source-stopping}, and the reference of \cref{seqadam:def:gaussian-reference}. On the geometry and sampling events, choose tolerances by \eqref{seqadam:adam-fb-error-allocation} and \eqref{seqadam:adam-fb-F4}, and impose the source-error and slow-motion bounds below. Then $\max_{t\le2}\|X_t-Y_t\|_F\le E_*/2$. The final choice-logit discrepancy is at most $AE_*+2h$, and the remaining complete-path discrepancy is at most $n(8h+h\exp(2B_{\exp}+4))$. The procedure uses exactly two updates and $2B_R$ verifier queries.
\end{proposition}
\begin{proof}
Define
\begin{equation}
 Q_{\rm amp}=(1+\eta_1D_A L_c)(1+\eta_2D_A L_c),\qquad
 \xi=\frac{E_*}{8S_{\rm step}Q_{\rm amp}}.
 \label{seqadam:adam-fb-error-allocation}
\end{equation}
Choose
\begin{align}
 h&\le\min\left\{\frac1{2^{20}n\exp(2B_{\exp}+4)},
                      \frac{\xi}{D_A C_{\rm pop}}\right\},&
 \varepsilon_g&\le\xi/D_A,\\
 \tau&\le\xi/G,&
 e_f&\le\min\left\{\frac{\xi}{2D_A n\sqrt2},
                            \frac{h}{32R_W}\right\}.
 \label{seqadam:adam-fb-F4}
\end{align}
Here a deterministic bound for the complete physical readout is
$R_W=R+S_{\rm step}K_{\rm mom}\sqrt q+1$, where $R$ is the source head bound.
The forcing in~\eqref{seqadam:adam-fb-F3} is at most $4\xi$, so
$E_2\le4S_{\rm step}Q_{\rm amp}\xi=E_*/2$ and the stopped region cannot be
exited. This argument controls the large second update from its gradient
evaluation inside the region.

The source feature error is included in the total $e_f$ allowance, rather than
reset at feedback entry. It suffices to cap it at $e_f/2$, to require slow
readout movement at most $h/(32A)$, and to bound additional backbone movement
by
\begin{equation}
 \min\left\{\frac12,\frac{e_f}{2D},\frac{h}{32DR_W}\right\}
 \label{seqadam:adam-fb-slow-motion}
\end{equation}
inside the remaining derivative ball. Appendix~\ref{seqadam:app:adam-joint} uses the
stronger source allowance $e_f/4$, reserves the radius remaining after source
training, and specifies a positive common slow multiplier. The bound on each
Adam coordinate from Lemma~\ref{seqadam:adam-fb-history-lemma} makes these movement
bounds deterministic. They include every backbone parameter and the normalized
member's gain. The full $R_W$ also includes physical readout nullspace
components and pays the head/backbone cross term through the feature error.

At the final iterate the choice logit error from $X_2-Y_2$ is at most
$AE_*$, with an additional base error at most $2h$. The remaining complete-path
error is bounded by $n(8h+h\exp(2B_{\exp}+4))$.
The reference task utility and first-state retention are at least
$(15/16)(1-9c/8)$, and its grammar-or-public-coordinate violation probability is at most $9c/8$.
Appendix~\ref{seqadam:app:adam-joint} combines these margins with the source baseline,
the finite-width events, and~\eqref{seqadam:adam-fb-F4} to obtain the stated capability
and gain bounds.

The two feedback batches use exactly $2B_R$ fresh training prompts, $2B_R$
binary verifier calls, $2nB_R$ unrestricted generated decisions, and two
optimizer updates. The $2HB_R$ state, $2B_R$ control, and $2B_R$ EOS counts
refer to designated positions; the actual emitted token classes may differ.
There is no feedback checkpoint selection or test-time verifier use.
\end{proof}

\subsection{An explicit joint source and feedback budget}
\label{seqadam:app:adam-joint}

We give an explicit order for choosing parameters for the Gaussian architecture of
Appendix~\ref{seqadam:app:foundations}, the sampled ridge-Adam source procedure of
Appendix~\ref{seqadam:app:source-budgets}, and the two sampled feedback updates of
Appendix~\ref{seqadam:app:fb-main}. All logarithms are natural unless their base is
displayed. Fix
\[
 q=2^{2r+2}\ge16,\qquad V=2q+2,\qquad c=\frac1{64},\qquad K=12,
 \qquad \beta_1=0.9,\quad\beta_2=0.999.
\]
The source estimates use $A_{\mathrm{src}}=12$, whereas feedback uses
$A_{\mathrm{fb}}=\sqrt K$. These two constants have different roles. Here $S$ is $S_{\rm step}$ in Appendix~\ref{seqadam:app:fb-main}, and $\varepsilon_f$ below is its feature tolerance $e_f$.
For a requested failure probability $0<\delta<1$, allocate
\begin{align}
 \delta_{\mathrm{bias}}=\delta_{\mathrm{enc}}
 =\delta_{\mathrm{ffn}}=\delta_{\mathrm{norm}}&=\delta/16,
 &\delta_0&=\delta/16,\nonumber\\
 \delta_{\mathrm{geo}}=\delta_{\mathrm{batch}}
 =\delta_R&=\delta/8.
 \label{seqadam:eq:adam-failure-allocation}
\end{align}
The allocation $\delta_0$ is used separately for the row envelope and the
readout-coordinate event.

\begin{lemma}[An admissible horizon and source precision]\label[lemma]{seqadam:lem:joint-horizon}
Fix the task and optimizer parameters above. The ordered choices \eqref{seqadam:eq:neural-horizon} and \eqref{seqadam:eq:adam-h} satisfy the signed-source requirement $\mu_F\le\lambda h/(64\sqrt K)$ of \cref{seqadam:prop:source-stopping} and both feedback bounds on $h$ in \eqref{seqadam:adam-fb-F4}. Every choice depends only on the declared task parameters; the minimum allowed horizon is determined before $H$ is chosen.
\end{lemma}
\begin{proof}
First define constants that depend only on $q$ and the fixed optimizer
parameters:
\begin{align*}
 u&=1-\beta_1,
 &K_{\mathrm{mom}}&=
 \bigl[(1-\beta_2)(1-\beta_1^2/\beta_2)\bigr]^{-1/2},\\
 \epsilon_f&=\frac1{64q},
 &\rho_0&=\frac{2\exp(-2)}{\sqrt{2\pi}},\\
 C_0&=\frac{2(1-\beta_1)\rho_0}{64\sqrt2+1},
 &\eta_1&=\frac1{100K_{\mathrm{mom}}\sqrt K},\\
 \eta_2&=\frac{\log(15(q-1))}{C_0},
 &S&=\eta_1+\eta_2,\\
 C_{\mathrm{move}}&=K_{\mathrm{mom}}+32\sqrt K,
 &R_{\mathrm{ref}}&=S C_{\mathrm{move}},\\
 B_{\mathrm{exp}}&=A_{\mathrm{fb}}(R_{\mathrm{ref}}+1),
 &E_{\mathrm{exp}}&=\exp(2B_{\mathrm{exp}}+4),\\
 D_A&=\frac{\sqrt2(1+K_{\mathrm{mom}})}{\epsilon_f},
 &L_c&=6K,\\
 E_*&=\frac1{1024A_{\mathrm{fb}}},
 &Q_{\mathrm{amp}}&=(1+\eta_1D_A L_c)(1+\eta_2D_A L_c),\\
 \xi&=\frac{E_*}{8S Q_{\mathrm{amp}}}.
\end{align*}
Set
\begin{align*}
 h_0&=\min\left\{\frac1{2^{20}E_{\mathrm{exp}}},
       \frac{\xi}{256A_{\mathrm{fb}}D_AE_{\mathrm{exp}}}\right\},\\
 C_q&=\frac23(5q+V+1)+2q(q-1),\\
 L_0&=\max\left\{1,\log\frac{24qV}{ch_0}\right\},\\
 B_H&=\frac{7372800K C_q q^3 V^{3/2}L_0^2}{h_0^2}.
\end{align*}
Choose any integer horizon satisfying
\begin{equation}
 H\ge\left\lceil\max\{128,\,2\log_2 B_H\}\right\rceil,
 \qquad n=H+2,\qquad N=2H+2.
 \label{seqadam:eq:neural-horizon}
\end{equation}
For the minimum prescribed horizon, the right side determines $H$ directly
from $q$; larger prescribed horizons are also allowed. Set the tolerances exactly as follows:
\begin{equation}
 \begin{gathered}
 h=\frac{h_0}{(H+2)^2},\qquad
 e=\min\left\{\frac hN,
          \frac{c^2(q-1)^2}{32q^2H^2N}\right\},\\
 \overline D=\max\{D_*(e/2),e\},\qquad
 \lambda=\frac e{\overline D},\qquad
 R_\lambda=\sqrt{2\overline D}.
 \end{gathered}
 \label{seqadam:eq:adam-h}
\end{equation}
Here $D_*$ is the deterministic comparator envelope from
Appendix~\ref{seqadam:app:source-budgets}: its kernel weights are replaced by
$k_{\min}=3/(2N)$ and $k_{0H}=1/N$. Explicitly, with
\[
 a=1-c/H,\qquad
 \beta_* =\log\frac{1+(q-1)a}{1-a},\qquad
 M=\beta_*+\log\frac{2V}{e},
\]
the value used in \eqref{seqadam:eq:adam-h} is
\[
 D_*(e/2)=\frac{2N}{3}(5q+V+1)M^2
                  +2Nq(q-1)\beta_*^2.
\]
Thus these choices use neither realized lag frequencies nor the hidden
mechanism nor any test score.

We check the horizon substitution explicitly. For $H\ge5$, the above
choices give
\begin{align*}
 h&\ge\frac{h_0}{4H^2},
 &e&\ge\frac{h_0}{12H^3},
 &\beta_*&\le\log(qH/c),\\
 \overline D&\le3C_qH(L_0+4\log H)^2
                 \le75C_qL_0^2H^3,
 &\lambda h&\ge\frac{h_0^2}{3600C_qL_0^2H^8}.
\end{align*}
In the source-tag estimate, $\rho_c\le1/2$ and
\[
 \mu_F=2q^3(a\rho_c)^{H-4}V^{3/2}\sqrt K,
 \qquad
 64\sqrt K\,\mu_F\le2048Kq^3V^{3/2}2^{-H}.
\]
For $H\ge128$, $H^8\le2^{H/2}$; moreover,
\eqref{seqadam:eq:neural-horizon} gives $B_H\le2^{H/2}$. Hence
$2^H\ge B_HH^8$, which proves
\[
 \mu_F\le\frac{\lambda h}{64\sqrt K}.
\]
The constant $B_H$ contains no $H$. In particular, this step does not require
a source tolerance exponentially small in the horizon. The equality choice
of $h$ also gives both feedback requirements
\[
 h\le\frac1{2^{20}nE_{\mathrm{exp}}},\qquad
 h\le\frac{\xi}{D_A C_{\mathrm{pop}}},\qquad
 C_{\mathrm{pop}}=256n^2 A_{\mathrm{fb}}E_{\mathrm{exp}}.
\]
\end{proof}

\begin{lemma}[Joint finite widths]\label[lemma]{seqadam:lem:joint-width}
Use the horizon and tolerances of \cref{seqadam:lem:joint-horizon} and the failure allocations \eqref{seqadam:eq:adam-failure-allocation}. The dimensions selected below, culminating in \eqref{seqadam:eq:adam-joint-width}, meet the encoder, coefficient-Gram, row-envelope and uniform-geometry requirements of \cref{seqadam:app:foundations,seqadam:lem:source-row,seqadam:adam-fb-uniform-geometry}. They also satisfy $\gamma\le\varepsilon_{\rm src}/16$, $\gamma\le\gamma_{\rm geo}$ and the Gram part of \eqref{seqadam:adam-src-S1}. All dimension choices are made without self-reference.
\end{lemma}
\begin{proof}
Define the feature dimensions and feedback precisions before selecting the
residual width:
\begin{align*}
 p&=NV,& k_{\mathrm{aug}}&=p+1,
 &s_{\mathrm{sym}}&=k_{\mathrm{aug}}(k_{\mathrm{aug}}+1)/2,
 &k&=p+s_{\mathrm{sym}}+1,\\
 G&=n\sqrt2 A_{\mathrm{fb}},
 &\varepsilon_g&=\xi/D_A,
 &\tau&=\xi/G.
\end{align*}
For completeness, instantiate \cref{seqadam:app:adam-geometry} with
$k,G,\epsilon_f,\tau$, and $\delta_{\mathrm{geo}}$. Its explicit constants are
\begin{align*}
 K_3&=\frac{8[k(k+2)]^{3/4}}{\delta_{\mathrm{geo}}},
 &L_{\mathrm{pert}}&=\frac{6k}{\epsilon_f}
                         +\frac{GK_3}{\epsilon_f^2},\\
 \gamma_{\mathrm{geo}}&=\min\{1/8,\tau/(16L_{\mathrm{pert}})\},
 &a_{\mathrm{geo}}&=\tau/2,\\
 T_{\mathrm{geo}}^2&=\max\left\{1,
 \frac{64k(k+2)}{\epsilon_f a_{\mathrm{geo}}\delta_{\mathrm{geo}}}\right\},
 &h_{\mathrm{geo}}&=\left(
 \frac{a_{\mathrm{geo}}\epsilon_f^2}{16T_{\mathrm{geo}}^3}\right)^2,\\
 D_{\mathrm{geo}}&=k(k+1)/2,
 &\mathcal N_{\mathrm{geo}}&=
 \left(1+\frac{2\sqrt k G^2}{h_{\mathrm{geo}}}\right)^{D_{\mathrm{geo}}}.
\end{align*}
The sufficient width is
\begin{equation}
 d_{\mathrm{geo}}=\left\lceil\max\left\{
 32\gamma_{\mathrm{geo}}^{-2}
       \left[k\log9+\log\frac{16}{\delta_{\mathrm{geo}}}\right],\,
 \frac{32k^2T_{\mathrm{geo}}^4}{\epsilon_f^2 a_{\mathrm{geo}}^2}
       \log\frac{16k^2\mathcal N_{\mathrm{geo}}}{\delta_{\mathrm{geo}}}
 \right\}\right\rceil.
 \label{seqadam:eq:adam-geometry-width-joint}
\end{equation}
All these quantities are fixed before $d$.

We next recall the dimension-independent source envelopes, retaining
$A_{\mathrm{src}}=12$:
\begin{align*}
 \overline m&=\sqrt{2\max\{1,\log(2V/\delta_0)\}},\\
 \overline C&=2\sqrt{2k+4\log(2/\delta_0)+4},
 &\overline c_h&=2\sqrt K\,\overline C+3,\\
 \overline g&=1+\overline c_h+
       \max\{\lambda\overline m,
                        \overline c_h+\lambda K_{\mathrm{mom}}/u\},\\
 \overline f&=\log V+2A_{\mathrm{src}}\sqrt V\,\overline m
                         +\frac{\lambda V}{2}\overline m^2,
 &\overline r&=\sqrt{\frac{2(2\overline f+1)}{\lambda}},\\
 \rho&=\frac{\lambda h}{64\sqrt K},
 &B_{\mathrm{dim}}&=\max\left\{16,
 \frac{1024\overline g(A_{\mathrm{src}}\overline r+\sqrt2)}{\rho},
 \frac{128\overline r}{h}\right\}.
\end{align*}
The source formulas below satisfy, for $t=1+\log d$,
\[
 R(d)\le\overline r\sqrt t,\qquad
 \varepsilon_{\mathrm{feat,src}}(d)
       \ge\frac{16}{B_{\mathrm{dim}}t}.
\]
Put
\begin{align*}
 R_W(d)&=R(d)+S K_{\mathrm{mom}}\sqrt q+1,
 &\overline r_W&=\overline r+S K_{\mathrm{mom}}\sqrt q+1,\\
 \varepsilon_{f,0}&=\frac{\xi}{2D_A n\sqrt2},
 &\overline\varepsilon_f&=
       \min\{\varepsilon_{f,0},h/(32\overline r_W)\},\\
 \varepsilon_f(d)&=\min\{\varepsilon_{f,0},h/(32R_W(d))\},
 &\varepsilon_{\mathrm{src}}(d)&=
       \min\{\varepsilon_{\mathrm{feat,src}}(d),\varepsilon_f(d)/4\},\\
 B_{\mathrm{joint}}&=\max\{B_{\mathrm{dim}},64/\overline\varepsilon_f\}.
\end{align*}
Here $\epsilon_f$ denotes Adam's feedback denominator constant, whereas
$\varepsilon_f$ denotes feature approximation error.
Since $R_W(d)\le\overline r_W\sqrt t$, we have
\[
 \varepsilon_f(d)\ge\overline\varepsilon_f/\sqrt t,
 \qquad
 \varepsilon_{\mathrm{src}}(d)\ge\frac{16}{B_{\mathrm{joint}}t}.
\]
Consequently, using $\log d\le16d^{1/16}/\mathrm e$, the bound
$d\ge[B_{\mathrm{joint}}(1+16/\mathrm e)]^{16}$ ensures
$d^{-1/8}\le\varepsilon_{\mathrm{src}}(d)/16$.

Choose the head count and concentration constants by
\begin{align*}
 R_{\mathrm{head}}&=\left\lceil
             8N\log\frac{2N}{\delta_{\mathrm{bias}}}\right\rceil,\\
 D_{\mathrm{head}}&=2^{20}N
                \left[V\log9+\log\frac{8N}{\delta_{\mathrm{enc}}}\right],\\
 D_\Gamma&=\max\left\{
 2^{19}\left[p\log9+\log\frac8{\delta_{\mathrm{enc}}}\right],\,
 4096\left[k\log9+\log\frac8{\delta_{\mathrm{ffn}}}\right],\,
 D_{\mathrm{head}}\right\}.
\end{align*}
Define
\begin{equation}
 \begin{split}
 d_{\mathrm{req}}=\max\biggl\{&
 [B_{\mathrm{joint}}(1+16/\mathrm e)]^{16},\,
 \left[\frac{64\sqrt K\,K(R_\lambda+1)}{\lambda h}\right]^8,\,
 \gamma_{\mathrm{geo}}^{-8},\,d_{\mathrm{geo}},\\
 &128\left[k\log9+\log\frac4{\delta_0}\right],\,
 D_\Gamma^{4/3}\biggr\},\qquad
 d=R_{\mathrm{head}}\left\lceil\frac{d_{\mathrm{req}}}{R_{\mathrm{head}}}\right\rceil,
 \qquad\gamma=d^{-1/8}.
 \end{split}
 \label{seqadam:eq:adam-joint-width}
\end{equation}
Every quantity on the right of $d_{\mathrm{req}}$ is independent of $d$.
The condition $d\ge D_\Gamma^{4/3}$ is precisely the substitution of
$\gamma=d^{-1/8}$ into $d\ge D_\Gamma\gamma^{-2}$. It supplies the
actual encoder and full-feature Gram bounds with encoder tolerance
$\gamma/16$. In particular the $D_{\mathrm{head}}$ term ensures the
head-width bound while preserving the architecture's integer relation
\[
 k_{\mathrm{head}}=d/R_{\mathrm{head}}.
\]
After fixing $d$, choose the FFN width by
\begin{align*}
 J_{\mathrm{ff}}&=s_{\mathrm{sym}}\log9+\log(8/\delta_{\mathrm{ffn}}),
 &A_{\mathrm{ff}}&=1+\gamma^{-2}J_{\mathrm{ff}},\\
 L_{\mathrm{ff}}&=\log\frac{\mathrm e A_{\mathrm{ff}}k_{\mathrm{aug}}}
                              {\delta_{\mathrm{ffn}}\gamma},
 &m_{\mathrm{ffn}}&=\left\lceil
       2^{50}A_{\mathrm{ff}}^2(k_{\mathrm{aug}}+L_{\mathrm{ff}})^8
                         \right\rceil.
\end{align*}
\end{proof}

\begin{lemma}[Positive rates and sampled implementation]\label[lemma]{seqadam:lem:joint-rates}
At the dimensions of \cref{seqadam:lem:joint-width}, the finite temperature, batch sizes and strictly positive block rates below meet the hypotheses of \cref{seqadam:prop:source-stopping,seqadam:prop:feedback-transfer}. On their good events, source stopping succeeds, the total backbone displacement across both stages is at most one, and $\max_{t\le2}\|X_t-Y_t\|_F\le E_*/2$. Source and feedback retain the exact bias-corrected moment and stopping rules specified below.
\end{lemma}
\begin{proof}
At the now chosen width, the source constants of
Appendix~\ref{seqadam:app:source-budgets} are explicitly
\begin{align*}
 M_0&=\sqrt{2\log(2Vd/\delta_0)},
 &W_{\max}&=\sqrt V M_0,\\
 C_{\mathrm{row}}&=2\sqrt{2k+4\log(2d/\delta_0)},
 &C_h&=2\sqrt K C_{\mathrm{row}}+3,\\
 G_c&=C_h+\lambda\max\{M_0,C_h/\lambda+K_{\mathrm{mom}}/u\},
 &\lambda_p&=(G_c+1)^{-1},\\
 L_s&=2K+1,
 &F_{\mathrm{init}}&=\log V+2A_{\mathrm{src}}W_{\max}
                         +\lambda W_{\max}^2/2,\\
 R&=\sqrt{2(2F_{\mathrm{init}}+1)/\lambda},
 &r_{\mathrm{goal}}&=\rho/4,\\
 \eta_s&=\min\left\{1,\frac{\lambda_p u^{3/2}}{8L_s},
                      \frac1{\lambda_p r_{\mathrm{goal}}^2}\right\},
 &b&=\lambda_p\rho/32,\\
 T_s&=\max\left\{1,\left\lceil
       \frac{4F_{\mathrm{init}}}{\eta_s\lambda_p r_{\mathrm{goal}}^2}\right\rceil\right\},\\
 \varepsilon_{\mathrm{feat,src}}&=\min\left\{1,
       \frac b{2(A_{\mathrm{src}}R+\sqrt2)},\frac h{8R}\right\},\\
 B_s&=\left\lceil\frac{8Vd A_{\mathrm{src}}^2}{b^2}
                    \log\frac{2VdT_s}{\delta_{\mathrm{batch}}}\right\rceil.
\end{align*}
These formulas also define the width-dependent functions used above; their
dimension-independent envelopes were fixed first. Source Adam starts with
fresh zero moments, uses physical denominator offset $1/\sqrt d$, and uses
head rate $\eta_s/\sqrt d$. At every inspected iterate it computes the
complete-record minibatch head gradient, including the coupled term $\lambda W$.
It stops before an update when that gradient's Frobenius norm is at most
$\rho/2$. A failed $T_s$th check declares failure without another update.
On the joint good event, the source result proves that this failure cannot
occur and that the output has the required uniform source logit accuracy.
The norm criterion uses only source observations.

All dimensions are fixed before drawing initialization. Let $P_{\mathrm{total}}$
be the number of initialized scalars and $P_\psi$ the number of supplied
backbone scalars. Define
\begin{align*}
 M_{\mathrm{init}}&=\max\left\{1,
       \sqrt{2P_{\mathrm{total}}\log(2P_{\mathrm{total}}/\delta_{\mathrm{norm}})}
                         \right\},\\
 Z_0&=M_{\mathrm{init}}+\sqrt{R_{\mathrm{head}}}M_{\mathrm{init}}^3,\\
 z_{\mathrm{raw}}&=\min\left\{
       \frac{\varepsilon_{\mathrm{src}}}{8\sqrt2},\frac1{4\sqrt d}\right\},\\
 \ell&=\min\left\{\frac1{2N},\,
 \frac{z_{\mathrm{raw}}}
 {2\sqrt{R_{\mathrm{head}}}M_{\mathrm{init}}^3
       [1+2M_{\mathrm{init}}^3(Z_0+1)]}\right\},\\
 g_{\mathrm{bias}}&=\frac{\delta_{\mathrm{bias}}}{R_{\mathrm{head}}N^2},\\
 t_{\mathrm{att}}&=\min\left\{
       \frac{g_{\mathrm{bias}}}{4M_{\mathrm{init}}^4},
       \frac{g_{\mathrm{bias}}}{2\log((N-1)/\ell)}\right\}>0.
\end{align*}
Thus the implemented attention has a finite positive temperature. For the
radius-one derivative bound, write $M_1=M_{\mathrm{init}}+1$ and set
\begin{align*}
 Z_1&=M_1+\sqrt{R_{\mathrm{head}}}M_1^3,\\
 D_z&=1+\sqrt{R_{\mathrm{head}}}
       [3M_1^2+M_1^3N(4M_1^3+t_{\mathrm{att}}^{-1})],\\
 D_{\mathrm{raw}}&=D_z[1+2M_1^3(Z_1+1)]
                             +3M_1^2(Z_1+1)^2+1,\\
 D&=\begin{cases}
 D_{\mathrm{raw}},&\text{base member},\\
 2\sqrt2 D_{\mathrm{raw}}+\sqrt{2K},&\text{normalized gain member}.
 \end{cases}
\end{align*}
The Gaussian parameter envelope and gain initialization are included in this
radius-one bound. In the source precision choices use the capped
$\varepsilon_{\mathrm{src}}$, and put
\begin{align*}
 r_\psi&=\min\left\{1,\frac1D,
                 \frac{\varepsilon_{\mathrm{src}}}{2D},\frac1{D\sqrt d}\right\},\\
 \rho_{\mathrm{back}}&=\min\left\{1,
 \frac{r_\psi\sqrt d}{2\eta_s T_sK_{\mathrm{mom}}\sqrt{P_\psi}}\right\}>0.
\end{align*}
Every source backbone coordinate uses the physical rate
$\eta_s\rho_{\mathrm{back}}/\sqrt d$. Its total source displacement is at
most $r_\psi/2\le1/2$, and the source feature error is at most
$\varepsilon_{\mathrm{src}}\le\varepsilon_f/4$.

At feedback, remove the source regularizer and reset all first and second
moments. Choose the positive common slow multiplier
\begin{equation}
 \rho_R=\min\left\{1,\,
 \frac{h}{32A_{\mathrm{fb}}S K_{\mathrm{mom}}\sqrt V},\,
 \frac{\sqrt{d/P_\psi}}{S K_{\mathrm{mom}}}
       \min\left\{\frac12,\frac{\varepsilon_f}{2D},
                         \frac h{32DR_W}\right\}\right\}.
 \label{seqadam:eq:adam-positive-feedback-rates}
\end{equation}
Use physical rate $\eta_t/\sqrt d$ for fast control-output rows and
$\eta_t\rho_R/\sqrt d$ for every slow supplied parameter, at $t=1,2$.
Every feedback block uses physical denominator offset $\epsilon_f/\sqrt d$.
Indeed, \eqref{seqadam:eq:adam-positive-feedback-rates} gives
\begin{align*}
 \rho_R S K_{\mathrm{mom}}\sqrt V&\le h/(32A_{\mathrm{fb}}),\\
 \rho_R S K_{\mathrm{mom}}\sqrt{P_\psi/d}
 &\le\min\{1/2,\varepsilon_f/(2D),h/(32DR_W)\}.
\end{align*}
Source and feedback backbone displacements total at most one, so the same
$D$ remains valid during both phases. The bound $R_W$ includes physical
readout components in the nullspace of the feature operator. These choices
supply total feature error at most $\varepsilon_f$ and base, slow-row, and
feature-motion logit discrepancy at most $2h$; fast coefficients are
controlled by the feedback comparison.

Each of the two feedback updates averages complete on-policy trajectories
using batch size
\begin{equation}
 B_R=\left\lceil
       \frac{8qk n^2 A_{\mathrm{fb}}^2}{\varepsilon_g^2}
                    \log\frac{4qk}{\delta_R}\right\rceil.
 \label{seqadam:eq:adam-feedback-batch-joint}
\end{equation}
Zero-reward trajectories and all token-score terms are retained. The
baseline is zero. The history-sensitive comparison in
Appendix~\ref{seqadam:app:fb-main} has per-step forcing at most $4\xi$ and therefore
\[
 \max_{t\le2}\|X_t-Y_t\|_F\le4S Q_{\mathrm{amp}}\xi=E_*/2.
\]
Both nonzero Adam histories participate in the second update. The width and
precision choices control the current random denominator before coefficient
projection, through the uniform geometry of
\cref{seqadam:app:adam-geometry}.
\end{proof}

\begin{lemma}[Joint probability and held-out capability]\label[lemma]{seqadam:lem:joint-margins}
For every fixed unknown world, the construction of \cref{seqadam:lem:joint-horizon,seqadam:lem:joint-width,seqadam:lem:joint-rates} succeeds with probability at least $1-\delta$. On this event, simultaneously for all held-out prompts, each source metric $E,M,S$ is at most $1/q+2nh$, each final metric exceeds $7/8$, and each gain exceeds $3/4$. The final grammar-or-public-coordinate violation probability is at most $9c/8+1/128$.
\end{lemma}
\begin{proof}
The bias, encoder, FFN/full-Gram, and parameter-envelope failures total
$4\delta/16$. The row and readout-coordinate failures total $2\delta/16$.
Uniform geometry, source batches, and the two feedback batches contribute
$3\delta/8$. Thus the union bound is
\[
 4\delta/16+2\delta/16+3\delta/8=3\delta/4<\delta.
\]
The encoder's four internal allocations total $\delta_{\mathrm{enc}}$ and
the FFN's two internal allocations total $\delta_{\mathrm{ffn}}$.
The row envelope is conditional on coverage and the actual Gram event;
the geometry event is uniform over histories. These facts suffice for the
union bound without assuming independent realized feature rows. Gaussian
initialization is not conditioned on later content-score events when its
independence lemmas are applied. For every fixed unknown world, fresh batches
obey the stopped conditional concentration estimates. This is not a
simultaneous guarantee for one dataset across all unknown worlds.

The canonical source selects a uniform operation. Sharp transitivity makes
its first state uniform, and subsequent doubly stochastic noisy-permutation
kernels keep its endpoint uniform. Its valid-class endpoint success and
first-state retention are consequently $1/q$, and its task utility is at
most $1/q$. Full-vocabulary and actual-logit coupling bounds each of the
three actual source metrics by
\[
 1/q+2nh.
\]
Writing $r_{\mathrm{opp}}$ for the opposite arm's reward, the separate exact
identity is $r_*+r_{\mathrm{opp}}=D_{\mathrm{exec}}\le1$;
the source uniformity conclusion uses the task structure, rather than the
individual reward upper bounds.

After the two reference updates, the correct choice has probability at
least $15/16$. Reference task utility and first-state retention are at least
\[
 \frac{15}{16}\left(1-\frac{9c}{8}\right)=\frac{7545}{8192}.
\]
The coefficient comparison contributes choice-logit error at most
$A_{\mathrm{fb}}E_*=1/1024$. The base, format, and state contribution is
\[
 n(8h+hE_{\mathrm{exp}})\le9/2^{20},
\]
and the choice-distribution contribution is at most $1/256$. Their total
deployment-law discrepancy is strictly less than $1/128$. Hence all three
actual metrics $E,S,M$ exceed $7/8$, and each exceeds its corresponding
source value by more than $3/4$. The probability of a grammar or public-coordinate violation is at most
\[
 \frac{9c}{8}+\frac1{128}.
\]
The bound uses the original verifier, which checks grammar, public-coordinate consistency and the endpoint. In the reference, only the correct arm and its hidden-bit opposite can receive reward, including histories whose private-state errors repair.
\end{proof}

\begin{proposition}[Polynomial implementation and exact query accounting]\label[proposition]{seqadam:prop:joint-resources}\label[proposition]{seqadam:app:fb-resources}
For the ordered construction of \cref{seqadam:lem:joint-horizon,seqadam:lem:joint-width,seqadam:lem:joint-rates}, the dimensions, batch sizes, iteration bound, inverse temperature and reciprocals of all positive physical rates have polynomial real-arithmetic bounds in expanded $q,H,\delta^{-1}$, up to logarithms. Source consumes at most $B_sT_s$ records and $NB_sT_s$ targets, with at most $T_s-1$ updates and no verifier queries. Feedback consumes exactly $2B_R$ fresh prompts and terminal queries, generates $2nB_R$ decisions, and makes two updates.
\end{proposition}
\begin{proof}
We make the resource dependence explicit, including the late choices of
temperature and positive rates. With fixed optimizer parameters,
$\eta_1$ is constant, $\eta_2=O(\log q)$, and
$R_{\mathrm{ref}}=O(\log q)$. In fact
\[
 E_{\mathrm{exp}}
 =\exp\!\bigl(2A_{\mathrm{fb}}(C_{\mathrm{move}}\eta_1+1)+4\bigr)
   [15(q-1)]^{2A_{\mathrm{fb}}C_{\mathrm{move}}/C_0}.
\]
This is a fixed power of $q$. Moreover $D_A=O(q)$,
$Q_{\mathrm{amp}}=O(q^2\log q)$, and
$\xi^{-1}=O(q^2(\log q)^2)$, so $h_0^{-1}$ is polynomial in $q$ up to
logarithms. Thus $B_H$ is explicit and the minimum horizon in
\eqref{seqadam:eq:neural-horizon} is finite. For any admitted $H$,
\[
 h^{-1}=\frac{(H+2)^2}{h_0},\qquad
 e^{-1}\le\frac{12H^3}{h_0},\qquad
 \lambda^{-1}\le\frac{900C_qL_0^2H^6}{h_0}.
\]
These bounds make the dimension-independent source envelopes polynomial in
the expanded variables $q,H,\delta^{-1}$, up to logarithms.

The geometry net does not introduce an exponential width: it enters only
through
\[
 \log\mathcal N_{\mathrm{geo}}
 =D_{\mathrm{geo}}\log\left(1+\frac{2\sqrt k G^2}{h_{\mathrm{geo}}}\right).
\]
All other geometry constants are powers, products, or positive reciprocals
of already chosen quantities. Thus $d_{\mathrm{geo}}$,
$B_{\mathrm{joint}}$, and $d_{\mathrm{req}}$ are polynomial in those expanded
variables. Rounding adds less than $R_{\mathrm{head}}$ to $d_{\mathrm{req}}$.
The number of heads is polynomial, and $k_{\mathrm{head}}=d/R_{\mathrm{head}}$
is integral. Since $\gamma^{-2}=d^{1/4}$,
\[
 A_{\mathrm{ff}}=1+d^{1/4}J_{\mathrm{ff}},\qquad
 m_{\mathrm{ffn}}=
 O\!\left((1+d^{1/4}J_{\mathrm{ff}})^2
 [k_{\mathrm{aug}}+\log d+\log(\delta_{\mathrm{ffn}}^{-1})]^8\right),
\]
with the explicit multiplier $2^{50}$ given above. The complete architecture
of Appendix~\ref{seqadam:app:foundations} has polynomially many scalars in these
dimensions, hence $P_{\mathrm{total}}$, $P_\psi$, and $M_{\mathrm{init}}$ are
polynomial up to logarithms.

The source cap gives
$\varepsilon_{\mathrm{src}}^{-1}\le
B_{\mathrm{joint}}(1+\log d)/16$. Therefore
$z_{\mathrm{raw}}^{-1}$ and $\ell^{-1}$ are polynomial as well. In particular
\[
 t_{\mathrm{att}}^{-1}=
 \max\left\{\frac{4M_{\mathrm{init}}^4}{g_{\mathrm{bias}}},
       \frac{2\log((N-1)/\ell)}{g_{\mathrm{bias}}}\right\}
\]
is polynomial. The displayed derivative formula is a fixed-degree
polynomial in $M_1$, $N$, $\sqrt{R_{\mathrm{head}}}$, and
$t_{\mathrm{att}}^{-1}$; consequently $D$ is polynomial for both members.

Finally, the displayed formulas make $\eta_s^{-1}$, $T_s$, $B_s$, and $B_R$
polynomial. For the late backbone rate,
\[
 r_\psi^{-1}=\max\{1,D,2D/\varepsilon_{\mathrm{src}},D\sqrt d\},\qquad
 \rho_{\mathrm{back}}^{-1}
 =\max\left\{1,\frac{2\eta_sT_sK_{\mathrm{mom}}\sqrt{P_\psi/d}}{r_\psi}\right\},
\]
and \eqref{seqadam:eq:adam-positive-feedback-rates} similarly gives a polynomial
bound for $\rho_R^{-1}$. Multiplying by $\sqrt d$ and the reciprocals of
$\eta_s,\eta_1,\eta_2$ proves the same statement for every positive physical
learning-rate reciprocal. Neither $D$ nor these late rates occurs on the
right side of the residual-width choice. This ordered substitution gives a polynomial real-arithmetic bound in expanded $q,H,\delta^{-1}$, up to logarithms, with the absolute exponents specified above.

The source phase uses at most $B_sT_s$ fresh complete raw records and
$NB_sT_s$ predicted next-token observations. On every outcome it performs
at most $T_s-1$ optimizer updates and $T_s$ gradient-norm checks, and makes
zero task-verifier calls. The feedback phase uses exactly $2B_R$ fresh
prompts, containing $2HB_R$ prompt tokens, and exactly $2B_R$ terminal
verifier calls. It generates $2nB_R$ unrestricted tokens and performs
exactly two optimizer updates. There are $2HB_R$ designated state positions,
$2B_R$ designated control positions, and $2B_R$ designated EOS positions.
These are position counts: the full-vocabulary sampler can emit an
incorrect type at any position, so each emitted-type count is bounded by
$2nB_R$ unless measured from the actual sequence. There is no test-time
verifier, reward filtering, source-success labeling, or feedback checkpoint
selection. \Cref{app:learning-routes} compares the corresponding learning routes and resources.
\end{proof}

\begin{proposition}[Frozen-source binding identification]\label[proposition]{seqadam:prop:frozen-controller}\label[proposition]{seqadam:app:fb-controller}
On the source event established in \cref{seqadam:prop:source-stopping}, with the joint tolerances \eqref{seqadam:eq:adam-h}, externally fixing the first control and greedily decoding the frozen source produces its exact trajectory and EOS. Binding enumeration therefore uses at most $q-1$ verifier queries and no additional gradient updates. Its generation, forward-evaluation, padding and storage costs are as detailed below.
\end{proposition}
\begin{proof}
On the same source event, the canonical correct-state probability at every
legal prefix is at least $(1-h)(1-9c/(8H))$. The actual source logit error
is below $h/2$, so its probability is at least
$(1-h)(1-9c/(8H))-h>1/2$ under \eqref{seqadam:eq:adam-h}.
EOS also has probability above $1/2$. Thus externally forcing any candidate
first control and greedily decoding the frozen source model yields its exact
state trajectory and EOS, including on unseen words. By sharp transitivity,
a training query accepts exactly when that candidate equals the unknown
binding. Enumerate at most $q-1$ candidates on fresh training prompts and
infer the last if all reject. This uses at most $q-1$ verifier calls,
$n(q-1)$ generated symbols, $H(q-1)$ prompt tokens, and $(H+1)(q-1)$
neural forward evaluations, with zero further gradient updates. Store the
binding index and an enumeration counter in addition to the model and its
retained context. Deployment requires $H+1$ forwards and $n$ generated
symbols with no verifier. This comparator retains the original source cost.

Padding is separate from observed information. Each source record starts
with $N$ PAD tokens; separate full-prefix evaluation of its $N$ targets
uses $N^2$ context-token occurrences, including $N(N+1)/2$ PAD occurrences.
Each feedback trajectory uses $nN$ context-token occurrences and $H$ prompt
symbols. Each verifier performs $H$ private transitions and $H$ public-coordinate checks. Source stopping uses an already-computed head gradient
norm, without extra data or verifier access. The model stores
$P_{\mathrm{total}}$ real parameters and the corresponding optimizer moments.
These costs use real arithmetic and real-valued parameter storage.
\end{proof}

\begin{proposition}[Opaque-table recovery and binding selection]\label[proposition]{seqadam:prop:table-controller}\label[proposition]{seqadam:app:fb-table}
Under the raw Sequential source law with $H\ge5$ and $0<a<1$ and $0<\delta_{\rm tab}<1$, the estimator using \eqref{seqadam:app:fb-table-budget} recovers all $q^2$ opaque transition-table rows with probability at least $1-\delta_{\rm tab}$. Its charged source-token count is at most $6M_0+2H+2$. On recovery, at most $q-1$ fresh-prompt verifier queries identify the binding and yield perfect deployment utility and public-coordinate compliance, with no gradient updates or test-time oracle.
\end{proposition}
\begin{proof}
Under the raw source law, execution controls $U_1,\ldots,U_{H-1}$ are
jointly iid uniform: $U_1$ is independent, and $U_2,\ldots,U_{H-1}$ form
a strict subset of the tag-conditioned suffix. Retain odd execution
transitions through $H-1$ and average the intervening even controls.
Their inputs become independent uniform states because
$q^{-1}\sum_{U\in\mathcal U} T_U=\Pi$. Thus a complete record supplies
$\lfloor H/2\rfloor$ iid triples in this marginal retained experiment.
Take
\begin{equation}
 M_0=\left\lceil8q^2a^{-2}\log\frac{2q^3}{\delta_{\rm tab}}\right\rceil,
 \qquad
 n_{\rm rec}=\left\lceil\frac{M_0}{\lfloor H/2\rfloor}\right\rceil.
 \label{seqadam:app:fb-table-budget}
\end{equation}
The charged raw-token count satisfies
\[
 Nn_{\rm rec}=(2H+2)n_{\rm rec}\le6M_0+2H+2.
\]
The empirical mode for every opaque state/control row recovers all $q^2$
successors with probability at least $1-\delta_{\rm tab}$. This estimator
requires neither a coordinate chart nor downstream goal labels.

For completeness, let $M=n_{\rm rec}\lfloor H/2\rfloor$ be the number
of retained iid triples and put
$t=\log(2q^3/\delta_{\rm tab})$. Each row count has mean
$\mu=M/q^2\ge8a^{-2}t$. Binomial lower-tail concentration bounds the
probability of fewer than $\mu/2$ observations by $e^{-\mu/8}$.
Conditional on $n$ observations of a row, true-minus-competitor indicator
differences lie in $[-1,1]$ and have mean $a$. Their centered log moment
generating function is at most $\lambda^2/2$, so an incorrect empirical
mode, including a tie, has probability at most $e^{-na^2/2}$. Union
bounding the $q^2$ row counts and $q^2(q-1)$ comparisons gives failure
probability at most $\delta_{\rm tab}$. This proves recovery with full-
record rounding. It does not assume independence of all $H$ transitions
within a record.

On recovery, enumerate candidate bindings on separately drawn fresh
uniform training prompts. Generate each candidate's exact path using the
recovered table. The binary terminal reward equals one exactly when that
candidate is $\eta$: suffix invertibility and sharp transitivity of the
first transition prove the equivalence. Test at most $q-1$ candidates,
stopping at success or inferring the last binding if all reject. No
canonical-prompt query selection is needed.

The additional costs are at most $q-1$ verifier calls,
$(H+2)(q-1)$ generated symbols, and $H$ table lookups per candidate or
deployment. There are zero gradient updates and no test-time oracle.
Final storage is $q^2$ successor indices and one binding index; training
may use $q^3$ counts. Deployment has perfect utility and public-coordinate
compliance on the recovery event. This comparator uses the same observations with a table-based computation, separating information requirements from the neural training route. The $2H+2$ granularity term and every
unused raw token remain charged.
\end{proof}

\phantomsection\label{seqadam:proof:neural-formal}
\begin{proof}[Proof of \cref{thm:neural-formal}]
Fix $q=2^{2r+2}$, $r\ge1$, $c=1/64$ and $\delta\in(0,1)$ as in the theorem. Apply \cref{seqadam:lem:joint-horizon} to select $H$ and $(h,e,\lambda)$, then \cref{seqadam:lem:joint-width} to choose the integer head, residual and FFN dimensions. These choices satisfy the source precision, signed-gradient, Gram and geometry inequalities before initialization is drawn. For either architecture member, \cref{seqadam:lem:joint-rates} supplies the appropriate derivative bound, finite temperature and positive physical rates. The coupled ridge term and source moments are retained throughout the source stage; the penalty is removed and both moments are reset only at the feedback boundary.

By \cref{seqadam:prop:source-stopping}, on the source events the observable stopping test succeeds within $T_s$ inspections and at most $T_s-1$ actual updates. Its uniform logit bound applies to the legal-pattern prefixes needed for freely generated trajectories. The same parameter choices satisfy the sampling, motion and history assumptions of \cref{seqadam:prop:feedback-transfer}, so the two actual sampled updates stay within $E_*/2$ of the reference in \cref{seqadam:adam-fb-two-step-reference}. This comparison uses full reward-weighted trajectories and the current adaptive denominators.

The union bound and metric conversion are exactly those in \cref{seqadam:lem:joint-margins}: the allocated failures sum to $3\delta/4<\delta$, and, on the resulting event, the source upper bound, final $7/8$ lower bounds, $3/4$ gains and violation bound hold simultaneously over all held-out prompts. Substituting $\varepsilon_{\rm logit}=h$ gives the four displayed assertions of the theorem. Finally, \cref{seqadam:prop:joint-resources} gives $N_{\rm pre}\le NB_sT_s$, $Q=2B_R$, $N_{\rm generated}=2nB_R$, at most $T_s-1$ source updates and exactly two feedback updates, with $N=2H+2$ and $n=H+2$. This establishes every stated resource count for the same training construction.
\end{proof}

\subsection{Output format and regularized source learning}\label{seqadam:app:format}
The source theorem uses a regularizer to select a structured optimum. The following obstruction explains why accurate conditional execution and a small unregularized source gradient would not suffice. It concerns population risk in the canonical coefficient class, rather than the trajectory selected by neural Adam.

\begin{proposition}[A source near-minimizer with incorrect held-out format]\label{seqadam:prop:format-obstruction}
Use the raw source law with $H\ge5$ and $0<a<1$. Let $f_\Theta(x)=\Theta\Phi(x)$ contain all positively weighted categorical singleton and pair features, with $\norm{\Phi(x)}=\sqrt K$, $K=12$, and let $F_+$ be population source CE. Fix a forbidden suffix $w=(w_2,\ldots,w_H)$. For every $\varepsilon>0$ and $\delta_f\in(0,1)$, there is a finite $\Theta$ such that
\[
 F_+(\Theta)-\inf_{\Theta'}F_+(\Theta')\le\varepsilon,
 \qquad \norm{\nabla F_+(\Theta)}_F\le\varepsilon,
\]
all within-state conditional predictions equal the noisy true transitions, but EOS has probability at least $1-\delta_f$ at the first-state prediction on $w$, uniformly over the initial state and preceding chosen control.
\end{proposition}
\begin{proof}
Let $P_l=\mathbf1\{x_l=\mathrm{PAD}\}$. The polynomial
$D_1=P_{H+1}-P_H$ is one exactly when the raw prefix length is $H+1$, the first-state target, and zero at every other source target. At that stage $U_j$ lies at lag $H+1-j$. Define
\begin{equation}\label{seqadam:eq:format-spike}
 d_w(x)=\sum_{j=2}^H\bigl(1-\mathbf1\{x_{H+1-j}=w_j\}\bigr),
 \qquad \Delta f_{\rm EOS}(x)=M\bigl(\tfrac12-d_w(x)\bigr)D_1(x).
\end{equation}
This is a degree-two categorical function. Every training suffix differs from $w$, so $d_w\ge1$ and the change suppresses the incorrect EOS logit by at least $M/2$ wherever it acts. On $w$ it raises that logit by $M/2$.

Choose finite base logits with exact noisy state kernels, uniform initial-state and control conditionals, and a format separator whose leakage is at most $\zeta$. The separator in \eqref{seqadam:adam-src-format}, with sufficiently large finite amplitude, provides this property even at the actual EOS, where the state interaction remains active.

The resulting within-class population objective has zero gradient in every coefficient. State predictions equal their true conditional laws; the initial state and early controls are uniform. At the last suffix control $U_H$, the total tag constrains one bit, but any degree-two prefix feature retains at most two of the $H-2\ge3$ preceding suffix controls. An unused independent balanced tag annihilates the target-feature correlation. Uniform control logits therefore have zero population gradient at this target as well. Convexity implies the within-class minimum
\[
 H_0=\frac{(H+1)\log q+Hh_q(a)}N.
\]
Full CE is that objective plus nonnegative class loss, and finite separators approach zero class loss. Hence $\inf F_+=H_0$. The unrestricted conditional entropy is lower by $\log2/N$; the coefficient-class infimum is not being identified with it.

The perturbation changes no correct-class conditional law and only reduces source class leakage. Consequently
\[
 F_+(\Theta)\le H_0-\log(1-\zeta),\qquad
 \norm{\nabla F_+(\Theta)}_F\le2\sqrt K\,\zeta,
\]
independently of $M$. The second bound follows because the full probability vector differs from its within-class counterpart by at most $2\zeta$ in $\ell_1$; average the residual outer products over all targets. Taking
$\zeta=\min\{1/2,1-e^{-\varepsilon},\varepsilon/(2\sqrt K)\}$ supplies the source bounds.

At a forbidden first-state prefix, the EOS probability is
\[
 p_{\rm EOS}=\frac1{1+R(x)e^{-M/2}},\qquad
 R(x)=\sum_{y\ne\mathrm{EOS}}e^{f_y(x)-f_{\rm EOS}(x)}.
\]
The base logits and the set of initial-state/control combinations are finite, so $R_{\max}<\infty$. Choose
$M\ge\max\{0,2\log[R_{\max}(1-\delta_f)/\delta_f]\}$.
This proves the uniform EOS bound. Under the fixed-length sampler an EOS token in a state slot is a grammar failure, regardless of subsequent tokens.
\end{proof}

The perturbation increases the coefficient norm with $M$ and targets one forbidden suffix. It isolates the role of regularization in extending source prediction to the required output format.

\section{Contextual acquisition and sampled Adam adaptation}\label{app:context-adam}
\subsection{Acquisition, contextual transfer and intervention}
Use the task and attention--bilinear architecture in \cref{ctx:sec:experiment}, with $\nctx\ge1$ and odd $p\ge17$. The fixed normalizer is the finite positive mixture in \eqref{ctx:eq:class}, with weights summing to one, powers in $[0,1]$ and offsets in $[0,Z_{\max}]$. Initialize every factor coordinate independently from a standard Gaussian. Use $0\le\beta_1<1$, $0<\beta_2<1$, $\beta_1^2<\beta_2$, and the physical step sizes and denominator offsets \ifbridgeReview specified below\else in \cref{ctx:sec:result}\fi, which implement a common normalized update. Retain all parameters between stages and reset only the two Adam moments.

\ifbridgeReview
The physical query/key rows are $Q_i=q_i/\sqrt{\rctx}$ and $K_i=k_i/\sqrt{\rctx}$; physical head rows are $a_i,b_i,o_i$. A normalized step $\eta_t$ and positive denominator offset $\epsilon$ use query/key step $\eta_t/\sqrt{\rctx}$ and offset $\epsilon/\sqrt{\rctx}$, and head step $\eta_t$ and offset $\epsilon/\rctx$. Every normalized coordinate therefore has the same nonzero rate and offset. At an exceptional zero attention pool, define $g=0$ and its selected derivative as zero; the proof events avoid this point.
\fi

\begin{theorem}[Acquisition, contextual transfer and intervention]\label[theorem]{ctx:thm:main-formal}
Fix the task dimensions, an architecture \eqref{ctx:eq:class}, Adam parameters $(\beta_1,\beta_2)$ as above, a world $(U,s,\tau)$ and $\delta\in(0,1)$. Choose width, batches, steps, offsets and input precision as in \cref{ctx:sec:budgets,ctx:sec:alphabet}; they are explicit and finite. With probability at least $1-\delta$ over initialization and all training samples, jointly:
\begin{enumerate}
\item \emph{Acquisition.} Initial source matching is at most $0.51$; after source training it is at least $0.9$ uniformly over contexts, frames and queries, and greedy source-bit accuracy exceeds $0.9$.
\item \emph{Transfer.} At the final feedback iterate, sampled success is at least $19/20$ on every held-out context, query, cue and amplitude pair, an improvement of at least $2/5$ over the source checkpoint.
\item \emph{Intervention.} The same schedules and budgets with the initial query/key parameters frozen in both stages, and every head factor trained, end with sampled test success at most $1/2+1/64$; the paired gap exceeds $2/5$.
\end{enumerate}
The conclusions hold for the finite coordinate alphabet of \cref{ctx:sec:alphabet} with supplied numerical decoding and real-valued optimizer states. At fixed ${\nctx},p,Z_{\max},m_{\rm norm}$ the budget formulas may be shared across mixture members, with probability evaluated for each fixed member.
\end{theorem}

The complete proof is in \cref{ctx:sec:alphabet}, using the budgets of \cref{ctx:sec:budgets}.

\subsection{Source learning in the radial class}\label{ctx:sec:sourceproof}

We prove \cref{ctx:thm:main} with explicit constants. All core norms below are Frobenius norms, combined by the Euclidean direct sum. The row coordinates are the normalized parameters of \cref{ctx:sec:experiment}. Time derivatives in this and the next section describe a deterministic reference; \cref{ctx:sec:adamproof} constructs that reference and couples the actual finite Adam runs to this analytical reference.

\begin{lemma}[Source symmetry and Gaussian reduction]\label[lemma]{ctx:lem:source-gaussian}
Under the source experiment of \cref{ctx:sec:experiment}, consider the unique self-consistent reference from independent standard-Gaussian factor rows, with coordinatewise update $F_\epsilon$ as defined below. Its attention core has form $M=\beta U$, and its head has only $C_{12}=C_{21}=t/2$ in the source-active content block. For $\beta\ge0$, the pooled content has law $h=\sqrt S G$ and conditional label mean $m_\rho(G_1)m_\rho(G_2)$, with $S,\rho,m_\rho$ defined in the proof. The population head gradient has no entries outside the same two positions. Existence and uniqueness of this reference are supplied by \cref{ctx:lem:reference-existence}.
\end{lemma}
\begin{proof}
At a source core $M=\beta U$, $\beta\ge0$, write
\[
 a=\sigma(\beta),\quad S=a^2+(1-a)^2,\quad
 \varrho=a^2/S,\quad \rho=\sqrt\varrho.
\]
Then $S_\beta/S=2(\varrho-a)$ and $\varrho_\beta=2\varrho(1-\varrho)$. Gaussian conditioning gives a pooled content vector $h=\sqrt S G$, $G\sim N(0,I_p)$, and
\[
 \E[y\mid G]=m_\rho(G_1)m_\rho(G_2),\qquad
 m_\rho(z)=2\Phi\!\left(\frac{\rho z}{\sqrt{1-\rho^2}}\right)-1.
\]
Here $\Phi$ and $\phi$ denote the standard Gaussian cumulative distribution function (CDF) and density.

Source sign and permutation symmetries force the reference head core to have only $C_{12}=C_{21}=t/2$ in its $p$-dimensional content block. The indices in this subsection refer to that block. Flipping an inactive value coordinate and its two factor coordinates preserves the source law and kills entries with exactly one such index. Negating $o$ together with both first-channel factors preserves source alignment and kills diagonal and inactive--inactive mean entries. Exchanging $a,b$ equates the remaining two entries. Context-factor sign flips apply because source context inputs vanish. These are signed parameter permutations, so uniqueness of the coordinatewise reference preserves the row law. At the resulting core, the population head gradient itself has only its $12,21$ entries: the same Gaussian sign symmetries kill the other derivatives individually. This stronger fact will preserve the inactive row factors.
\end{proof}

\begin{lemma}[Strict source drift throughout the normalization class]\label[lemma]{ctx:lem:source-strict-drift}
Use the source law and symmetric core of \cref{ctx:lem:source-gaussian}, with $p\ge17$ and the finite positive mixture \eqref{ctx:eq:class}. For $t>0$ and finite $\beta\ge0$, the source loss satisfies $L_\beta<0$. For each component with $\alpha>0$, the uncertainty term $D$ has the integral representation \eqref{ctx:eq:Dintegral} and the logarithmic derivative \eqref{ctx:eq:drift}; for $\alpha=0$, $D=(1-\varrho)/\pi$ and that logarithmic derivative equals $2a$.
\end{lemma}
\begin{proof}
The source logit is $f=t h_1h_2\psi(\norm h^2)$, and its loss has the exact decomposition
\begin{equation}\label{ctx:eq:sourceCE}
 L=\E\log(1+e^{-|f|})+\frac12\E\left[|f|\{1-y\sgn(G_1G_2)\}\right].
\end{equation}
For one component $(\alpha,\zeta)$ with $\alpha>0$, put $v=\zeta/S$ and define
\[
 D=\frac12\E\left[\frac{|G_1G_2|}{(\norm G^2+v)^\alpha}
 \{1-m_\rho(G_1)m_\rho(G_2)\sgn(G_1G_2)\}\right].
\]
The Gamma integral and elementary Gaussian integration yield
\begin{equation}\label{ctx:eq:Dintegral}
 D=\frac{1-\varrho}{\pi\Gamma(\alpha)}
 \int_0^\infty\frac{u^{\alpha-1}e^{-vu}(1+2u)^{-p/2}}
 {1+2u(1-\varrho)}\,du.
\end{equation}
Indeed,
\[
 \E[Ze^{-uZ^2}m_\rho(Z)]
 =\frac{\sqrt{2/\pi}\rho}{(1+2u)\sqrt{1+2u(1-\rho^2)}}.
\]
Multiply this identity for the two active Gaussian coordinates and by $(1+2u)^{-(p-2)/2}$ for the rest, then subtract from the perfect-sign expression. The Gamma representation follows by substitution in its defining integral; nonnegative sign-adjusted integrands justify Tonelli.

Let $\E_w$ integrate under the normalized positive integrand in \eqref{ctx:eq:Dintegral}, and write $B_u=1+2u(1-\varrho)$. Integration of the derivative of $u$ times that integrand gives
\[
 \E_w B_u^{-1}=1-\alpha+p\E_w\frac{u}{1+2u}+v\E_wu.
\]
The origin boundary vanishes for $\alpha>0$, and the infinity boundary and derivatives are integrable for $p\ge17$, including $v=0$. Differentiating with $v_\beta=-2(\varrho-a)v$ gives
\begin{equation}\label{ctx:eq:drift}
 -\frac{\partial_\beta(S^{1-\alpha}D)}{S^{1-\alpha}D}
 =2\left[(1-\alpha)a+\varrho p\E_w\frac{u}{1+2u}+av\E_wu\right]>0.
\end{equation}
For $\alpha=0$ the direct formula is $D=(1-\varrho)/\pi$, and the ratio is $2a$; no $\Gamma(0)$ is used.

For every component, the derivative with respect to $S$ of its margin factor is
\[
 \partial_S\frac{S}{(S\norm G^2+\zeta)^\alpha}
 =\frac{\zeta+(1-\alpha)S\norm G^2}
 {(S\norm G^2+\zeta)^{\alpha+1}}\ge0.
\]
The first term of \eqref{ctx:eq:sourceCE} therefore cannot increase with $\beta$. Its second term is $t$ times the positive weighted sum of $S^{1-\alpha}D$. Equation \eqref{ctx:eq:drift} proves $L_\beta<0$ whenever $t>0$ and $\beta$ is finite. The argument averages the noise terms componentwise; it does not replace the logistic loss of a mixture by a mixture of logistic losses.
\end{proof}

\begin{definition}[Uniform source constants]\label[definition]{ctx:def:source-constants}
Set $a_\star=19/20$, $\varrho_\star=a_\star^2/[a_\star^2+(1-a_\star)^2]$, $\rho_\star=\sqrt{\varrho_\star}$, $s_\star=\rho_\star/\sqrt{1-\rho_\star^2}$, and
\begin{align}
 P_E&=\phi(2)^2\Pr(|Z|\le1)^{p-2},& m_0&=2\phi(1),\nonumber\\
 m_\star&=2\Phi(2s_\star)-1,&
 A_{\min}&=\frac{P_E m_0^2}{2(p+6+2Z_{\max})},\nonumber\\
 D_{\min}&=\frac{P_E(1-m_\star^2)}{2(p+6+2Z_{\max})},&&\nonumber\\
 \gamma_s&=\min\left\{\frac{D_{\min}}2,
 \frac{p(1-\varrho_\star)e^{-4Z_{\max}}5^{-p/2-2}}{3\pi}\right\}.
 \label{ctx:eq:sourceconstants}
\end{align}
\end{definition}

\begin{lemma}[Uniform drift and head warmup bounds]\label[lemma]{ctx:lem:source-uniform-drift}
Under the assumptions of \cref{ctx:lem:source-strict-drift}, the constants in \cref{ctx:def:source-constants} are positive. For $a\le a_\star$ and $t\ge0$, $-L_\beta\ge t\gamma_s/2$. Put $t_{\min}=A_{\min}/8$ and $d_s=3A_{\min}/8$. For all $\beta\ge0$, $L_{tt}<1$ and $-L_t\ge d_s$ whenever $0\le t\le t_{\min}$.
\end{lemma}
\begin{proof}
These are explicit positive real numbers. On $G_1,G_2\in[1,2]$ with all other coordinates in $[-1,1]$, the probability is at least $P_E$ and $\norm G^2+\zeta/S\le p+6+2Z_{\max}$. Its correlation and remaining uncertainty give $A_{\min},D_{\min}$ uniformly until $a_\star$. For $\alpha\le1/2$, the first term of \eqref{ctx:eq:drift} lower bounds the unscaled decrease by $D_{\min}/2$. For $\alpha\ge1/2$, restrict the integral to $u\in[1,2]$, and use $\Gamma(\alpha)\le3$, $v\le2Z_{\max}$ and $1+2u\le5$ to obtain the second term in $\gamma_s$. Since $S^{1-\alpha}\ge1/2$,
\begin{equation}\label{ctx:eq:sourcebound}
 -L_\beta\ge t\gamma_s/2\qquad(a\le a_\star).
\end{equation}
The coefficient of $t$ in $f$ is bounded by $|G_1G_2|+1/2$, whence $L_{tt}<1$. At $t=0$, $-L_t\ge A_{\min}/2$. Thus, with
\[
 t_{\min}=A_{\min}/8,\qquad d_s=3A_{\min}/8,
\]
we have $-L_t\ge d_s$ on $0\le t\le t_{\min}$. The correlation-event bound also holds beyond $a_\star$ whenever $\beta\ge0$, which will make $t_{\min}$ an invariant lower boundary.
\end{proof}

\begin{proposition}[Finite-time source acquisition by the reference]\label[proposition]{ctx:prop:source-reference-acquisition}\label[proposition]{ctx:sec:sourcelaw}
Start the source reference from the Gaussian row law in \cref{ctx:lem:source-gaussian}. With the constants of \cref{ctx:def:source-constants} and the offset and horizon in \eqref{ctx:eq:sourcehorizon}, its parity coefficient reaches and stays above $t_{\min}$ after time $S_{\rm warm}$. Its correct-slot matching reaches $a_\star=19/20$ by $S_s$ and remains at least $a_\star$ thereafter. Its attention gradient is \eqref{ctx:eq:sourceM}, and every normalized coordinate has speed at most one.
\end{proposition}
\begin{proof}
Let $F_\epsilon(z)=z/(|z|+\epsilon)$ act coordinatewise on each row's scaled population gradient. Align $k'=Uk$. The signed-coordinate source symmetries give $M=\beta U$ and the head form above. Haar averaging in the active block and uniform contexts give the full gradient
\begin{equation}\label{ctx:eq:sourceM}
 \nabla_M L=\frac{L_\beta}{2{\nctx}}U.
\end{equation}
The head gradient has no extra $1/{\nctx}$: each context has the same head loss. Each aligned query/key pair obeys
\[
 \dot q=F_\epsilon(\chi k'),\quad \dot k'=F_\epsilon(\chi q),
 \qquad \chi=-L_\beta/(2{\nctx}).
\]
Its product derivative is nonnegative for $\chi\ge0$. Initially $q,k'\in[1,2]$ has probability at least $\phi(2)^2$; there both remain at least one, and their product derivative is at least one when $\chi\ge\epsilon$. The negative event gives the same contribution. Consequently
\[
 \dot\beta\ge p_0:=e^{-4}/\pi=2\phi(2)^2\qquad(\chi\ge\epsilon).
\]

The active head coordinates $(a_1,a_2,b_1,b_2,o)$ have alignment $t=\E[o(a_1b_2+a_2b_1)]$. With $d=-L_t$, their equations are
\begin{align*}
 \dot a_1&=F_\epsilon(do b_2),&\dot a_2&=F_\epsilon(do b_1),\\
 \dot b_1&=F_\epsilon(do a_2),&\dot b_2&=F_\epsilon(do a_1),\\
 \dot o&=F_\epsilon\bigl(d(a_1b_2+a_2b_1)\bigr).
\end{align*}
Every alignment contribution has the form $uF_\epsilon(du)$. On the event that all five initial entries are in $[1,2]$, with probability at least $p_5=\phi(2)^5$, all remain there or above during warmup, and
\[
 \dot t\ge6p_5\frac{d}{d+\epsilon}.
\]
At $t=0$ the direction is positive; at $t=t_{\min}$ it is strictly inward. Thus $t$ reaches and then stays above $t_{\min}$, even if its derivative later changes sign. Choose
\begin{equation}\label{ctx:eq:sourcehorizon}
 \epsilon_s=\frac{t_{\min}\gamma_s}{4{\nctx}},\quad
 S_{\rm warm}=\frac{t_{\min}(d_s+\epsilon_s)}{6p_5d_s},\quad
 S_s=S_{\rm warm}+\frac{\log19}{p_0}+1.
\end{equation}
After warmup, \eqref{ctx:eq:sourcebound} implies $\chi\ge\epsilon_s$ until $a_\star$, so matching reaches $a_\star$ by $S_s$. Thereafter $\beta$ cannot decrease because $L_\beta<0$ and $t>0$. Every block remains trainable, and every normalized coordinate has speed at most one.
\end{proof}

\begin{lemma}[Inactive factors and the output sign at the source endpoint]\label[lemma]{ctx:lem:source-retained-law}
For the reference of \cref{ctx:prop:source-reference-acquisition}, cross-context attention moments vanish. At the source switch, the $\nctx$ context head coordinates and $k$ inactive cue coordinates retain their independent standard-Gaussian law, independently of the active five-coordinate evolution. The output factor $o$ has a symmetric absolutely continuous law and $\Pr(o\ge0)=1/2$.
\end{lemma}
\begin{proof}
Sparse context support supplies the block decomposition. Simultaneous query/key sign flips in each block kill cross-context reference moments already at initialization. Their finite deviations remain in the full-core norm of \cref{ctx:sec:adamproof}; acquired matching is within the supplied blocks.

At the switch, the ${\nctx}$ context head coordinates and $k$ inactive cue coordinates retain their independent standard-Gaussian reference law, independent of the active five-coordinate evolution. Their population source gradients vanish; source context sample gradients vanish identically. The transformation $(o,a_1,b_1)\mapsto(-o,-a_1,-b_1)$ preserves source alignment and the coordinatewise flow. Hence $o$ has a symmetric continuous law. Finite-time locally Lipschitz flow preserves absolute continuity, so $\Pr(o\ge0)=1/2$. This entire source-produced law is retained in the feedback proof.
\end{proof}

\subsection{Contextual feedback from the acquired state}\label{ctx:sec:feedbackproof}

Let $\beta_s$ be the source contrast at $S_s$. The two source-active content coordinates are zero on all target inputs. Thus the reference source checkpoint has zero target logit without replacing any parameter. The starting law of this stage is exactly the law constructed in \cref{ctx:sec:sourcelaw}.

\begin{lemma}[Invariant feedback cores and population gradients]\label[lemma]{ctx:lem:feedback-core}
Run the feedback reference from the source endpoint in \cref{ctx:prop:source-reference-acquisition,ctx:lem:source-retained-law}, under the contextual task in \cref{ctx:sec:experiment}. The target-active head has form \eqref{ctx:eq:targetcore}; the attention blocks have a common contrast parameter $\lambda$ and fixed sum coefficient $\beta_s$. With $c=\tanh(\lambda/2)$, its population success and scalar coefficients are \eqref{ctx:eq:targetscalar}, and its full core gradients are \eqref{ctx:eq:targetgradient}.
\end{lemma}
\begin{proof}
The target-active head core has the invariant form
\begin{equation}\label{ctx:eq:targetcore}
 C_{\ell i}+C_{i\ell}=B s_\ell\tau_i,\qquad
 C_{\rm context,context}=C_{\rm cue,cue}=0,
\end{equation}
where $i$ ranges over cue coordinates. To see this for arbitrary $s$, conjugate context factors by the diagonal signs $s$ and replace the uniform cue in context $\ell$ by $s_\ell z$. The quadratic logit is invariant under an overall input sign, reducing to all-positive $s$. The inactive source factors retain their Gaussian law under these transformations. Context permutations and signed cue permutations equate the aligned cross coefficients. Finally, negating $o$, both first source-channel factors and every cue factor preserves the source law; on target inputs it maps $f(z)$ to $-f(-z)$. Labels also change sign with the cue, so this symmetry kills even target blocks and preserves the cross block. All these transformations are signed permutations, under which $F_\epsilon$ is equivariant.

Let $P_-$ project onto $(1,-1)$ and $P_+=I-P_-$. The reference attention blocks are
\[
 M_{\ell\ell}=(\beta_sP_++\lambda P_-)U_\ell,\qquad
 c=\tanh(\lambda/2),
\]
with common $\lambda$ and zero cross-context blocks. A training pool has cue $cvz$, where $v\in\{\pm1\}$ is the selected slot sign. Define
\begin{align}
 T&=\tau^\top z,&w(c)&=c\psi(1+kc^2),\nonumber\\
 J(B,c)&=\E\sigma(Bw(c)|T|),&
 E_T&=\E[|T|\sigma'(Bw(c)|T|)],\nonumber\\
 d_f&=w(c)E_T/k,&g_f&=(1-c^2)Bw'(c)E_T/2.
 \label{ctx:eq:targetscalar}
\end{align}
Direct differentiation gives the full, unsymmetrized core gradients
\begin{equation}\label{ctx:eq:targetgradient}
 (\nabla_CJ)_{\ell i}=(\nabla_CJ)_{i\ell}
 =\frac{d_f}{{\nctx}}s_\ell\tau_i,\qquad
 (\nabla_MJ)_{\ell\ell}=\frac{g_f}{{\nctx}}U_\ell P_-.
\end{equation}
Cue symmetry makes its aligned gradient proportional to $\tau$; the scalar product with $\tau$ is $w(c)E_T$, giving $1/k$. Uniform contexts give $1/{\nctx}$. The attention factor follows from two-slot softmax, query averaging and $dc/d\lambda=(1-c^2)/2$. There is no additional source factor $1/2$ in $g_f$.
\end{proof}

\begin{lemma}[Coordinate dynamics and positive task-rule drift]\label[lemma]{ctx:lem:feedback-head-drift}
Under \cref{ctx:lem:feedback-core}, the signed head coordinates satisfy \eqref{ctx:eq:targetrows}. As long as $d_f>0$, all contributions to $\dot B$ are nonnegative and the positive-probability event inherited from \cref{ctx:lem:source-retained-law} yields \eqref{ctx:eq:Bdrift}, with $p_g=\phi(2)^4/2$.
\end{lemma}
\begin{proof}
For a head row define
\[
 A_c=\frac{s^\top a_{\rm context}}{\sqrt {\nctx}},\quad
 D_c=\frac{s^\top b_{\rm context}}{\sqrt {\nctx}},\quad
 X_c=\frac{\tau^\top a_{\rm cue}}{\sqrt k},\quad
 Y_c=\frac{\tau^\top b_{\rm cue}}{\sqrt k}.
\]
For the output factor, contracting the core gradient with its row feature gives the normalized row gradient
\begin{align*}
 g_o&:=\langle\nabla_C J,ab^\top\rangle_F\\
 &=\frac{d_f}{\nctx}\bigl[(s^\top a_{\rm context})(\tau^\top b_{\rm cue})
 +(\tau^\top a_{\rm cue})(s^\top b_{\rm context})\bigr]\\
 &=d_f\sqrt{\frac{k}{\nctx}}(A_cY_c+X_cD_c).
\end{align*}
Summing the coordinatewise equations with their signs therefore yields
\begingroup\allowdisplaybreaks[0]
\begin{align}
 \dot A_c&=\sqrt {\nctx} F_\epsilon\!\left(\frac{d_f\sqrt k}{{\nctx}}oY_c\right),&
 \dot D_c&=\sqrt {\nctx} F_\epsilon\!\left(\frac{d_f\sqrt k}{{\nctx}}oX_c\right),\nonumber\\
 \dot X_c&=\sqrt k F_\epsilon\!\left(\frac{d_f}{\sqrt {\nctx}}oD_c\right),&
 \dot Y_c&=\sqrt k F_\epsilon\!\left(\frac{d_f}{\sqrt {\nctx}}oA_c\right),\nonumber\\
 \dot o&=F_\epsilon\!\left(d_f\sqrt{\frac{k}{\nctx}}(A_cY_c+X_cD_c)\right),&
 B&=\frac{\E[o(A_cY_c+X_cD_c)]}{\sqrt{{\nctx}k}}.
 \label{ctx:eq:targetrows}
\end{align}
\endgroup
These equations follow by summing the original coordinatewise updates. All contributions to $\dot B$ have sign $d_f$. At the switch, $A_c,D_c,X_c,Y_c$ are independent standard Gaussians, independent of the symmetric sign of $o$. The event that these four variables are at least one and $o\ge0$ has probability at least $p_g=\phi(2)^4/2$. These inequalities persist while $d_f>0$. The $o$ contribution alone implies
\begin{equation}\label{ctx:eq:Bdrift}
 \dot B\ge\frac{2p_g}{\sqrt{{\nctx}k}}
 \frac{2d_f\sqrt{k/{\nctx}}}{2d_f\sqrt{k/{\nctx}}+\epsilon_f}.
\end{equation}
The acquired contrast makes the right side positive; with $c=0$ the training cue disappears.
\end{proof}

\begin{lemma}[Feedback attention dynamics]\label[lemma]{ctx:lem:feedback-attention}
Under \cref{ctx:lem:feedback-core}, the aligned query/key contrasts obey \eqref{ctx:eq:contrast}. The sign of $\dot\lambda$ is the sign of $g_f$, the sum coordinates remain fixed, and the attention-core form in \cref{ctx:lem:feedback-core} is preserved.
\end{lemma}
\begin{proof}
For attention align $k'=Uk$ and set $u=q_1-q_2$, $v=k'_1-k'_2$ within a block. Then
\begin{align}
 \dot u&=2F_\epsilon(g_fv/(2{\nctx})),&
 \dot v&=2F_\epsilon(g_fu/(2{\nctx})),\nonumber\\
 \lambda&=\E[uv]/2,&
 \dot\lambda&=\E\left[vF_\epsilon(g_fv/(2{\nctx}))+uF_\epsilon(g_fu/(2{\nctx}))\right].
 \label{ctx:eq:contrast}
\end{align}
Thus $\dot\lambda$ has the sign of $g_f$. Sum coordinates are pointwise constant. Simultaneous coordinate exchange preserves vanishing mixed sum/contrast moments of the source law; their independence is unnecessary. Block sign symmetries and context permutations retain the complete attention-core form.
\end{proof}

\begin{lemma}[An invariant contrast region]\label[lemma]{ctx:lem:feedback-barrier}
For the positive mixture \eqref{ctx:eq:class}, each component satisfies \eqref{ctx:eq:wprime} and has nonnegative derivative at $\cmin=1/\sqrt k$. Along the feedback reference started at the acquired source endpoint, the region $B\ge0$, $c\ge\cmin$ is invariant.
\end{lemma}
\begin{proof}
For every component,
\begin{equation}\label{ctx:eq:wprime}
 \frac{d}{dc}\frac{c}{(1+kc^2+\zeta)^\alpha}
 =\frac{1+\zeta+(1-2\alpha)kc^2}{(1+kc^2+\zeta)^{\alpha+1}}.
\end{equation}
It is nonnegative at $\cmin=1/\sqrt k$. Equations \eqref{ctx:eq:Bdrift} and \eqref{ctx:eq:contrast}, the source bound $c\ge0.9$, and uniqueness give joint invariance of $B\ge0$ and $c\ge\cmin$. Some members decrease contrast above that boundary; actual Adam deviations from the reference barrier will be paid separately.
\end{proof}

\begin{lemma}[Finite time to the target coefficient]\label[lemma]{ctx:lem:feedback-hitting}
In the invariant region of \cref{ctx:lem:feedback-barrier}, let $w_{\min},\Btar,\mu_k,d_{\min,f},\epsilon_f,S_f$ be as in \eqref{ctx:eq:feedbackhorizon} and its preceding sentence. Then $w_{\min}\le w(c)\le1$, $\E|T|/k=\mu_k$, and $d_f\ge d_{\min,f}$ until $B=\Btar$. With offset $\epsilon_f$, the reference reaches $B\ge\Btar$ by $S_f$ and remains there.
\end{lemma}
\begin{proof}
In this region $w_{\min}\le w(c)\le1$, where $w_{\min}=\cmin/(1+k+Z_{\max})$. Put
\begin{align}
 L_\star&=\log19+1,&
 \Btar&=\frac{2(1+k+Z_{\max})L_\star}{\cmin},\nonumber\\
 \mu_k&=\frac{\binom{k-1}{(k-1)/2}}{2^{k-1}},&
 d_{\min,f}&=w_{\min}\mu_ke^{-\Btar k}/4,\nonumber\\
 \epsilon_f&=d_{\min,f}\sqrt{k/{\nctx}},&
 S_f&=\frac{3\sqrt{{\nctx}k}\Btar}{4p_g}+1.
 \label{ctx:eq:feedbackhorizon}
\end{align}
Conditioning one sign in the odd Boolean sum, or cancelling paired binomial terms, gives $\E|T|/k=\mu_k$. Since $|T|\le k$ and $\sigma'(z)\ge e^{-z}/4$ for $z\ge0$, we have $d_f\ge d_{\min,f}$ until $B$ reaches $\Btar$. Then \eqref{ctx:eq:Bdrift} gives $\dot B\ge4p_g/(3\sqrt{{\nctx}k})$. Hence $B\ge\Btar$ by $S_f$ and remains so.
\end{proof}

\begin{lemma}[Held-out margin and pooled-cue sensitivity]\label[lemma]{ctx:lem:feedback-margin}
At the endpoint supplied by \cref{ctx:lem:feedback-hitting}, every context, query, Boolean cue and pair of amplitudes from $\{\pm1,\pm\rho_{\rm mem}\}$ has signed reference logit at least $L_\star=\log19+1$. For $|z|\le1$, the positive normalizer mixture satisfies the derivative bound \eqref{ctx:eq:zlip}.
\end{lemma}
\begin{proof}
For any two amplitudes from $\{\pm1,\pm\rho_{\rm mem}\}$, let the correct weight be $a=(1+c)/2$ and the pooled cue multiplier be $z_0=a u_{\rm selected}+(1-a)u_{\rm other}$. Then
\begin{align*}
 \sgn(u_{\rm selected})z_0
 &\ge a\rho_{\rm mem}-(1-a)
 =c-\frac{1+c}{2}2^{-m_{\rm mem}}\ge\cmin/2,\\
 |z_0|&\le1.
\end{align*}
The signed logit is at least $B\cmin/[2(1+k+Z_{\max})]$, since $|T|\ge1$. The reference therefore has margin at least $L_\star$ on all sixteen such pairs, including the four disjoint test pairs. The bound
\begin{equation}\label{ctx:eq:zlip}
 \left|\frac{d}{dz}\{z\psi(1+kz^2)\}\right|\le1
\end{equation}
holds componentwise and hence for the positive mixture, and will control finite core error.
\end{proof}

\subsection{The frozen intervention and source-information bound}\label{ctx:sec:comparisonproof}

\begin{proposition}[Stationary target prediction under frozen reference attention]\label[proposition]{ctx:prop:frozen-reference}
Use the same Gaussian initialization law, source and feedback schedules and moment reset as in \cref{ctx:thm:main-formal}, but set query/key updates to zero in both stages. Its population reference has $M=0$; at the source switch its target-active head coefficients vanish. During feedback every reference head row is stationary, and its test logit remains zero on every amplitude pair.
\end{proposition}
\begin{proof}
The frozen branch shares initialization and schedules with the main run but sets query/key updates to zero in both stages. Its reference $M=0$ is an expectation identity; the actual finite initialization is retained in the coupling. Source head symmetry still leaves only the parity-channel coefficients at uniform retrieval. No source-acquisition conclusion is needed for this branch.

On target training records its reference pool is exactly $e_\ell$. Conditional labels are balanced, so
\[
 \E_y\sigma(y C_{\ell\ell})=1/2,\qquad\nabla_CJ=0.
\]
All target-active reference coefficients vanish at the switch; the target logit is zero for every attention core there, and $\nabla_MJ=0$ as well. After the moment reset, every reference head row is stationary during feedback. Its logit stays zero on every test pair. \Cref{ctx:sec:budgets} bounds the finite branch with its random initial attention and trained head.
\end{proof}

\begin{proposition}[The error contributed by an unobserved context]\label[proposition]{ctx:prop:missing-context-lower}
In the exact numerical contextual experiment of \cref{thm:context-information}, for $N_{\rm pre}$ iid source observations (or a worst-case cap of that many observations), every world-independent learner has worst-world expected test error at least $\frac14(1-1/\nctx)^{N_{\rm pre}}$, regardless of its number of permitted adaptive feedback queries. A missing context contributes conditional error at least $1/4$ under the independent uniform prior on orientation flips.
\end{proposition}
\begin{proof}
For the lower-bound direction of \cref{thm:context-information}, pair worlds by independently flipping any subset of
\[
 (U_\ell,s_\ell)\longmapsto(U_\ell P_{\rm swap},-s_\ell).
\]
Every permitted antipodal training prompt has the same reward for each action in paired worlds. This pointwise equality survives adaptive prompts, randomized policies, parameter updates, arbitrary computation and any number of permitted feedback queries. Source records from other contexts are also unaffected.

If context $\ell$ is absent from $N_{\rm pre}$ iid source records, its flip remains uniform under an independent uniform prior on the ${\nctx}$ flips, conditional on the full transcript and supplied $\tau$. Equal-sign test amplitude pairs have opposite paired labels, whereas opposite-sign pairs agree. Equal-sign mass is $1/2$, so a missing context has conditional test error at least $1/4$. It is missing with probability $(1-1/{\nctx})^{N_{\rm pre}}$. Averaging over uniform test contexts and then bounding the supremum by the prior mean proves the lower bound.

Receiving fewer than a worst-case cap of iid observations cannot improve on receiving the full cap. An expected stopping budget is different: a coupon-collector rule can wait until every context appears. Chosen-context source queries also change the experiment. The information result concerns unobserved orientations; the frozen-attention comparison concerns the specified neural training path (\cref{ctx:sec:comparisons}).
\end{proof}

\subsection{A sufficient contrast region and its signal scale}\label{app:contrast-scale}
The contextual reference path stays in $c\ge1/\sqrt{k}$, and the sampled-path error budget controls its deviation. The reciprocal normalization below separates this sufficient invariant region from the presence of a learning signal.
\begin{proposition}[Nonmonotone signal within the reciprocal member]\label[proposition]{prop:contrast-scale}
Let $k>1$, $0\le c\le1$, and $\psi(u)=u^{-1}$, corresponding to a single mixture component with weight one, exponent one and shift zero. Then
$w(c)=c\psi(1+kc^2)=c/(1+kc^2)$ increases strictly on $[0,1/\sqrt{k}]$, decreases strictly on $[1/\sqrt{k},1]$, and has maximum $1/(2\sqrt{k})$. It is positive at every $c>0$.
\end{proposition}
\begin{proof}
The denominator is positive throughout the interval. The quotient rule gives
\[
 w'(c)=\frac{(1+kc^2)-2kc^2}{(1+kc^2)^2}
       =\frac{1-kc^2}{(1+kc^2)^2}.
\]
Its sign is positive for $c<1/\sqrt{k}$, zero at that point and negative above it. Integrating its strict sign between distinct points on either side proves the two monotonicity statements. Substitution gives
$w(1/\sqrt{k})=(1/\sqrt{k})/(1+1)=1/(2\sqrt{k})$.
For $c>0$, numerator and denominator are positive. Thus this member has a positive signal below $1/\sqrt{k}$, while the convergence argument uses the invariant region above it.
\end{proof}

\subsection{Actual sampled Adam and the cubic moment map}\label{ctx:sec:adamproof}

We now justify the reference and control both finite runs.

\begin{lemma}[Physical Adam and normalized coordinates]\label[lemma]{ctx:lem:adam-normalization}
Use the physical query/key and head step sizes and offsets of \cref{ctx:thm:main-formal}, and reset the moments at each stage. The actual bias-corrected Adam update in every normalized coordinate is exactly $\mp\eta_tD_t$, with minus sign for source and plus sign for feedback, and $D_t$ given by the moment recurrence below.
\end{lemma}
\begin{proof}
Within each stage, scalar gradients $g_t$ generate
\begingroup\allowdisplaybreaks[0]
\begin{align*}
 m_t&=\beta_1m_{t-1}+(1-\beta_1)g_t,&
 v_t&=\beta_2v_{t-1}+(1-\beta_2)g_t^2,\\
 \widehat m_t&=m_t/(1-\beta_1^t),&
 \widehat v_t&=v_t/(1-\beta_2^t),\qquad
 D_t=\frac{\widehat m_t}{\sqrt{\widehat v_t}+\epsilon}.
\end{align*}
\endgroup
Source subtracts $\eta_tD_t$; feedback adds it. If $H_i=D\varphi_i^\top$ times the batch core gradient, physical query/key and head gradients are $H_i/\sqrt {\rctx}$ and $H_i/{\rctx}$, respectively. Homogeneity of both moments with the prescribed physical offsets yields exactly $\eta_tD_t$ in each normalized coordinate. This uses the physical first and second moments; it does not replace them by a population gradient.
\end{proof}

\begin{definition}[History constants]\label[definition]{ctx:def:adam-history-constants}
Define
\begin{equation}\label{ctx:eq:Adamconstants}
 K_A=\frac{1-\beta_1}{\sqrt{(1-\beta_2)(1-\beta_1^2/\beta_2)}},\qquad
 \Lambda=\frac{\beta_1}{1-\beta_1}
 +\frac{\sqrt{\beta_2(1+\beta_2)}}{1-\beta_2}.
\end{equation}
\end{definition}

\begin{lemma}[A uniform bound for every Adam history]\label[lemma]{ctx:lem:adam-history-bound}
For $0\le\beta_1<1$, $0<\beta_2<1$, $\beta_1^2<\beta_2$ and $\epsilon>0$, every scalar gradient history satisfies $|D_t|\le K_A$, where $K_A$ is defined in \cref{ctx:def:adam-history-constants}; moreover $K_A\ge1$.
\end{lemma}
\begin{proof}
Weighted Cauchy--Schwarz gives $|D_t|\le K_A$ for every coordinate history. Before adding $\epsilon$, the squared ratio is bounded by
\[
 \frac{(1-\beta_1)^2}{1-\beta_2}
 \frac{1-\beta_2^t}{(1-\beta_1^t)^2}
 \frac{1-(\beta_1^2/\beta_2)^t}{1-\beta_1^2/\beta_2}.
\]
The inequality $(1-x)(1-y)\le(1-\sqrt{xy})^2$ bounds this by $K_A^2$, and also shows $K_A\ge1$. A positive offset can only decrease the ratio. Thus all-zero rewards and a zero current gradient with nonzero old momentum are covered.
\end{proof}

\begin{lemma}[Gaussian row envelopes and core moments]\label[lemma]{ctx:lem:row-envelopes}
Use the stage horizons from \cref{ctx:prop:source-reference-acquisition,ctx:lem:feedback-hitting} and the history bound of \cref{ctx:lem:adam-history-bound}. With the row dimensions, radii and moments defined below, every actual and reference row remains in its radius-$\rho_i$ ball up to $H$. The quadratic and cubic moment maps satisfy \eqref{ctx:eq:momentderivs}. For $0<\delta_c<1$, event \eqref{ctx:eq:B3event} has failure probability at most $\delta_c/4$ and bounds the actual, empirical-reference and population cores by $B_3$, and their speeds by $V=3v_0B_3$. The same bounds hold for the frozen branch.
\end{lemma}
\begin{proof}
Put
\begin{align*}
 d_1&=4{\nctx},&d_2&=2({\nctx}+p)+1,&H&=S_s+S_f+2,\\
 v_0&=K_A\sqrt{\max(d_1,d_2)},&a_0&=1+v_0H,&
 \rho_i&=a_0+\norm{Z_i},
\end{align*}
where $Z_i$ is a row's initial Gaussian vector. Every actual or reference row lies in its radius-$\rho_i$ ball through time $H$. For dimension $d$ define
\begin{align*}
 \mu_3(d)&=4\left[a_0^3+2^{3/2}\frac{\Gamma((d+3)/2)}{\Gamma(d/2)}\right],\\
 \mu_6(d)&=32[a_0^6+d(d+2)(d+4)].
\end{align*}
The two row moment maps, embedded in the direct-sum core, are
\[
 \varphi_1(q,k)=qk^\top,\qquad\varphi_2(a,b,o)=oab^\top.
\]
Since $\rho_i\ge1$, both satisfy
\begin{equation}\label{ctx:eq:momentderivs}
 \norm{\varphi_i}\le\rho_i^3,\qquad
 \norm{D\varphi_i}\le3\rho_i^2,\qquad
 \norm{D^2\varphi_i}\le6\rho_i.
\end{equation}
For an allocation $\delta_c$ set
\[
 B_3=\frac4{\delta_c}\{\mu_3(d_1)+\mu_3(d_2)\}.
\]
Markov's inequality gives, with failure at most $\delta_c/4$,
\begin{equation}\label{ctx:eq:B3event}
 \sum_{\rm groups}\frac1{\rctx}\sum_i\rho_i^3\le B_3.
\end{equation}
It bounds both actual cores, their coupled empirical reference cores and their speeds by $V=3v_0B_3$. Deterministic population cores satisfy the same envelope. Freezing query/key velocities only reduces it.
\end{proof}

\begin{lemma}[Normalizer and sample-logit derivatives]\label[lemma]{ctx:lem:core-derivatives}
Let $\psi$ satisfy \eqref{ctx:eq:class}, let the source pool be nonzero, and restrict the head core to $\|C\|\le B_3$. With $Y=\sqrt p W$, $X=Y/\|h\|$ and $W$ the largest of the $2p$ Gaussian magnitudes, the normalizer, pooled feature and sample logit satisfy the derivative bounds below; in particular \eqref{ctx:eq:corederivs} holds with $C_1=64(B_3+1)$ and $C_2=4096(B_3+1)$.
\end{lemma}
\begin{proof}
For $u>0$, the mixture obeys
\[
 \psi(u)\le\max(1,u^{-1}),\qquad
 |u\psi'(u)|\le\psi(u),\qquad u^2\psi''(u)\le2\psi(u).
\]
For $A({\rctx}_0)=\sqrt{\psi({\rctx}_0^2)}$, differentiation gives $|A'|\le A/{\rctx}_0$ and $|A''|\le6A/{\rctx}_0^2$. Hence $\norm{D_hg}\le2A$ and $\norm{D_h^2g}\le9A/{\rctx}_0$; the latter follows by bounding ${\rctx}_0|A''|+3|A'|$ in the derivative of $A({\rctx}_0)h$.

On source records let $W$ be the maximum of the $2p$ Gaussian magnitudes, $Y=\sqrt p W$, and $X=Y/\norm h$. Unit keys/query and softmax give
\[
 \norm h\le Y,\quad\norm{D_Mh}\le2Y,\quad\norm{D_M^2h}\le8Y.
\]
Therefore
\begin{align*}
 \norm g&\le1+Y,&\norm{D_Mg}&\le4(Y+X),\\
 \norm{D_M^2g}&\le16(Y+X)+36(YX+X^2).
\end{align*}
On $\norm C\le B_3$, conservative joint-core bounds are
\begin{equation}\label{ctx:eq:corederivs}
 \norm{Df}\le C_1(1+Y^2+X^2),\quad
 \norm{D^2f}\le C_2(1+Y^3+X^3),
\end{equation}
where $C_1=64(B_3+1)$ and $C_2=4096(B_3+1)$.
\end{proof}

\begin{lemma}[Uniform source inverse-moment bounds]\label[lemma]{ctx:lem:inverse-moments}
For $p\ge17$ and any deterministic attention core, the source variables $Y,X$ of \cref{ctx:lem:core-derivatives} satisfy $\E Y^j\le Y_j$ and $\E X^j\le X_j$ for $1\le j\le8$, with the explicit constants in \eqref{ctx:eq:inversemoments}. These estimates require no independence between $W$ and $h$.
\end{lemma}
\begin{proof}
For $1\le j\le16$ define $m_j=2^{j/2}\Gamma((j+1)/2)/\sqrt\pi$. For $1\le j\le8$ put
\begin{equation}\label{ctx:eq:inversemoments}
 Y_j=p^{j/2}(2p)m_j,\qquad
 X_j=p^{j/2}\sqrt{\frac{(2p)m_{2j}2^j}{\prod_{\ell=1}^j(p-2\ell)}}.
\end{equation}
These bound $\E Y^j$ and $\E X^j$ uniformly over attention cores. Conditional on the keys/query, $h$ is isotropic Gaussian with variance $S_\pi=\sum_j\pi_j^2\ge1/2$. Cauchy--Schwarz bounds the product $W^j\norm h^{-j}$ using its $2j$ moments; inverse chi-square integration gives the denominator. Independence of $W$ and $h$ is not used. The strongest denominator remains positive at $p=17$.
\end{proof}

\begin{lemma}[Population source regularity at zero shifts]\label[lemma]{ctx:lem:source-regularity}
On the core region bounded by $B_3$, the source population gradient norm, its Lipschitz constant and the sample-gradient second moment are bounded respectively by $G_s,L_s,\sigma_s^2$ in \eqref{ctx:eq:sourcereg}. The expected source loss is $C^2$ on compact core sets, including mixture members with zero architectural shifts.
\end{lemma}
\begin{proof}
Consequently, the population source gradient bound, its Lipschitz constant, and the sample-gradient second moment are bounded respectively by
\begin{align}
 G_s&=C_1(1+Y_2+X_2),&
 \sigma_s^2&=3C_1^2(1+Y_4+X_4),\nonumber\\
 L_s&=\sigma_s^2/4+C_2(1+Y_3+X_3).
 \label{ctx:eq:sourcereg}
\end{align}
To justify derivatives at zero architectural shifts, first add a common auxiliary positive squared shift. The envelopes above are uniform in that shift. Along deterministic convergent core/shift sequences the limiting pool is nonzero almost surely, so sample derivatives converge. The eighth moments bound squared Hessians and give uniform integrability. Vitali convergence establishes continuity of expected derivatives and the $C^2$ limit on compact core sets. No sample-adaptive supremum of inverse radii is asserted; the algorithm uses the prescribed shifts.
\end{proof}

\begin{lemma}[Actual reward-gradient regularity]\label[lemma]{ctx:lem:target-regularity}
On the same core region, every target pool has norm in $[1,\sqrt{k+1}]$. The actual terminal-reward sample gradient has norm at most $A_f$ and conditional expectation $y\sigma'(f)Df$. Its second moment and the population feedback gradient norm and Lipschitz constant are bounded by $\sigma_f^2,G_f,L_f$ in \eqref{ctx:eq:targetreg}.
\end{lemma}
\begin{proof}
Target pools have norm between one and $\sqrt{k+1}$ for arbitrary $M$. Define
\begin{align}
 A_f&=64(B_3+1)(k+1)^2,&H_f&=4096(B_3+1)(k+1)^3,\nonumber\\
 G_f&=A_f/4,&L_f&=(A_f^2+H_f)/4,&\sigma_f^2&=A_f^2.
 \label{ctx:eq:targetreg}
\end{align}
The actual sample gradient is $R_{\rm rew}A\sigma(-Af)Df$. Its norm is at most $A_f$; its conditional action expectation is $y\sigma'(f)Df$, exactly the derivative of sampled success. Thus the stated variance bound applies to actual terminal rewards.
\end{proof}

\begin{definition}[Joint regularity constants]\label[definition]{ctx:def:joint-regularity}
Set
\[
 G=\max(G_s,G_f),\quad L=\max(L_s,L_f),\quad
 \sigma^2=\max(\sigma_s^2,\sigma_f^2),\quad
 \epsilon_{\min}=\min(\epsilon_s,\epsilon_f).
\]
\end{definition}

\begin{lemma}[Existence and uniqueness of the self-consistent reference]\label[lemma]{ctx:lem:reference-existence}
For the source and feedback objectives of \cref{ctx:sec:experiment}, use the Gaussian row initialization, offsets and horizons specified above. The population row equations admit a unique self-consistent reference on both stages, retaining row states at the switch. A frozen branch has the same property after projecting query/key velocities to zero. For prescribed bounded core paths the row gradients satisfy \eqref{ctx:eq:rowsensitivity}, and their averaged finite-time sensitivity is controlled by the finite Gaussian moment $\E[\rho^4e^{6GH\rho/\epsilon_{\min}}]$.
\end{lemma}
\begin{proof}
Write $\Psi=(M,C)$, and let $F=L$ in the source stage and $F=J$ in feedback. For a prescribed continuous core path, each reference row solves the locally Lipschitz ordinary differential equation (ODE) with velocity $\pm F_\epsilon(D\varphi_i^\top\nabla F)$; the sign depends on the stage. A frozen row projects its query/key velocities to zero. Bounded coordinate speed prevents finite-time divergence. For two row/core pairs, \eqref{ctx:eq:momentderivs} gives
\begin{equation}\label{ctx:eq:rowsensitivity}
 \norm{H_i(X)-H_i(Y)}\le
 6G\rho_i\norm{X_i-Y_i}+3L\rho_i^2\norm{\Psi_X-\Psi_Y}.
\end{equation}
Solving this inequality for each row before averaging gives a finite Volterra Lipschitz constant proportional to $\E[\rho^4e^{6GH\rho/\epsilon_{\min}}]$. The map from a prescribed core path to the expectation of the evolved row moment is therefore a contraction on sufficiently short time intervals and maps the bounded core region into itself. Iterating these intervals constructs the unique self-consistent reference on both stages. The same inequality proves uniqueness; the switch concatenates the paths with their row states retained. This justifies the symmetry arguments above.
\end{proof}

\begin{lemma}[An averaged exponential-radius event]\label[lemma]{ctx:lem:exponential-radius}
With $\lambda_0=6GH/\epsilon_{\min}$ and $E_4,B_4$ from \eqref{ctx:eq:B4def}, $\E[\rho^4e^{\lambda_0\rho}]\le E_4(d)$ in dimension $d$. Event \eqref{ctx:eq:B4event} has failure probability at most $\delta_c/4$.
\end{lemma}
\begin{proof}
Put $\lambda_0=6GH/\epsilon_{\min}$ and
\begin{align}
 E_4(d)&=8\,2^{d/2}[a_0^4+4d(d+2)]e^{\lambda_0a_0+\lambda_0^2},\nonumber\\
 B_4&=\frac4{\delta_c}\{E_4(d_1)+E_4(d_2)\}.
 \label{ctx:eq:B4def}
\end{align}
The inequality $e^{\lambda_0\norm Z}\le e^{\lambda_0^2+\norm Z^2/4}$, a Gaussian tilt and $(a+{\rctx}_0)^4\le8(a^4+{\rctx}_0^4)$ give this moment bound. A second unconditional Markov event, with failure at most $\delta_c/4$, gives
\begin{equation}\label{ctx:eq:B4event}
 \sum_{\rm groups}\frac1{\rctx}\sum_i\rho_i^4 e^{\lambda_0\rho_i}\le B_4.
\end{equation}
The averaged radius moment controls the cubic row maps without a
width-dependent maximum-radius bound.
\end{proof}

\begin{lemma}[Comparing a history with a current gradient]\label[lemma]{ctx:lem:adam-current}
For any vector $h_0$ and bias-corrected Adam history with positive offset $\epsilon$, inequality \eqref{ctx:eq:Adamcurrent} holds. The normalized first-moment mean lag is at most $\beta_1/(1-\beta_1)$ and the normalized second-moment squared lag is at most $\beta_2(1+\beta_2)/(1-\beta_2)^2$.
\end{lemma}
\begin{proof}
For any current vector $h_0$ and bias-corrected history, let $b_{ts}$ be the normalized nonnegative second-moment weights. Subtracting fractions and using weighted reverse triangle inequalities gives
\begin{equation}\label{ctx:eq:Adamcurrent}
 \norm{D_t-F_\epsilon(h_0)}
 \le\frac{\norm{\widehat m_t-h_0}
 +\sqrt{\sum_s b_{ts}\norm{g_s-h_0}^2}}{\epsilon}.
\end{equation}
The normalized first-moment lag mean is at most $\beta_1/(1-\beta_1)$, and the second-moment lag second moment is at most $\beta_2(1+\beta_2)/(1-\beta_2)^2$, producing $\Lambda$ in \eqref{ctx:eq:Adamconstants}.
\end{proof}

\begin{lemma}[Sample, lag and Euler errors within a stage]\label[lemma]{ctx:lem:adam-local-error}
On event \eqref{ctx:eq:B3event}, suppose all deterministic steps are at most $\eta$ and each fresh core batch gradient differs from its conditional population gradient by at most $\nu$. With $A_i=6Gv_0\rho_i+3LV\rho_i^2$, the actual direction error is bounded by \eqref{ctx:eq:Adamsampleerror}, and the reference Euler local error is at most $\eta_t^2A_i/(2\epsilon_{\min})$. These estimates apply within each stage after its moment reset, with accumulated parameter error retained.
\end{lemma}
\begin{proof}
Let $\eta$ bound all deterministic steps. On \eqref{ctx:eq:B3event}, population row gradients at two indices in the same stage differ by at most $\eta$ times their index distance times
\[
 A_i=6Gv_0\rho_i+3LV\rho_i^2.
\]
If the fresh core batch error is at most $\nu$, its row error is at most $3\rho_i^2\nu$. Equation \eqref{ctx:eq:Adamcurrent} then bounds the actual direction error relative to the normalized current population gradient by
\begin{equation}\label{ctx:eq:Adamsampleerror}
 \frac{\eta\Lambda A_i+6\rho_i^2\nu}{\epsilon_{\min}}.
\end{equation}
The reference Euler local error is at most $\eta_t^2 A_i/(2\epsilon_{\min})$. Both moments reset at the boundary, so history estimates stay within their objective; accumulated parameter error remains. Shortening the final step improves these lag and Euler bounds.
\end{proof}

\begin{proposition}[Joint finite-Adam core coupling]\label[proposition]{ctx:prop:adam-coupling}
Couple each actual row to the reference of \cref{ctx:lem:reference-existence} using its initial Gaussian row. Assume \eqref{ctx:eq:B3event}, \eqref{ctx:eq:B4event}, the batch-error condition of \cref{ctx:lem:adam-local-error}, and empirical-reference core error at most $w$ at each deterministic grid time. Then the actual-to-population core errors satisfy \eqref{ctx:eq:discretecore} and \eqref{ctx:eq:coupling}, including the stage switch. The same bounds apply to the frozen branch using its own reference and the shared radius events.
\end{proposition}
\begin{proof}
Couple actual and reference rows using the same initial Gaussian row. Let $e_{i,\ell}$ be row discrepancy and $\Delta_\ell$ the actual core error relative to its population reference. Then
\begin{align*}
 e_{i,\ell+1}\le{}&\left(1+\eta_\ell\frac{6G\rho_i}{\epsilon_{\min}}\right)e_{i,\ell}
 +\eta_\ell\frac{3L\rho_i^2}{\epsilon_{\min}}\Delta_\ell\\
 &+\frac{\eta_\ell}{\epsilon_{\min}}
 \left\{\eta(\Lambda+1/2)A_i+6\rho_i^2\nu\right\}.
\end{align*}
Solve rowwise, multiply by $3\rho_i^2$, and average using \eqref{ctx:eq:B4event}. If the reference empirical-core discrepancy at all deterministic grid times is at most $w$, the result is
\begin{align}
 \Delta_\ell&\le w+\sum_{j<\ell}\eta_j(K_H\Delta_j+R_H),\nonumber\\
 K_H&=9LB_4/\epsilon_{\min},\nonumber\\
 R_H&=\frac{B_4}{\epsilon_{\min}}
 \left\{\eta(\Lambda+1/2)(18Gv_0+9LV)+18\nu\right\}.
 \label{ctx:eq:discretecore}
\end{align}
Discrete Gronwall gives, including at the stage boundary,
\begin{equation}\label{ctx:eq:coupling}
 \Delta_\ell\le(w+HR_H)e^{K_HH}.
\end{equation}
The same argument controls the frozen branch with its own reference. Shared Gaussian radius events suffice. Its finite initial attention core is included in $w$; actual trained rows are not asserted to remain independent.
\end{proof}

\subsection{Explicit budgets and functional outcomes}\label{ctx:sec:budgets}

\begin{definition}[Explicit continuous-input training budgets]\label[definition]{ctx:def:finite-adam-budgets}
Fix the task, architecture and Adam coefficients first. Equations \eqref{ctx:eq:sourcehorizon} and \eqref{ctx:eq:feedbackhorizon} determine $\epsilon_s,S_s,\epsilon_f,S_f$, hence $H,\epsilon_{\min}$. Set $\delta_c=\delta/4$, and form all constants in \cref{ctx:sec:adamproof}. They do not depend on width, step count, or a sampled maximum row norm.

Define $c_p=\Gamma(p/2)/[\sqrt\pi\Gamma((p-1)/2)]$ and
\begin{align}
 \kappa&=\min\left\{\frac1{256\pi},\frac{t_{\min}}{160000c_p^2},
 \frac{\cmin}8,\frac1{32(k+1+\sqrt{2k}B_3)}\right\},\nonumber\\
 \tau_{\rm core}&=\kappa/4,\quad E_0=e^{K_HH},\quad w=\tau_{\rm core}/(4E_0),\nonumber\\
 \nu&=\frac{\epsilon_{\min}\tau_{\rm core}}{144HB_4E_0},\nonumber\\
 \eta&=\min\left\{1,\frac{\epsilon_{\min}\tau_{\rm core}}
 {8HB_4E_0(\Lambda+1/2)(18Gv_0+9LV)}\right\}.
 \label{ctx:eq:steps}
\end{align}
Put $T_s=\lceil S_s/\eta\rceil$, $T_f=\lceil S_f/\eta\rceil$, and $T=T_s+T_f$. Use step $\eta$, shortening the final step of each stage to make its total time exactly $S_s$ or $S_f$. Every step is positive. Choose integers, rounded up and at least one,
\begin{equation}\label{ctx:eq:widthbatch}
 {\rctx}\ge\frac{8(T+1)\{\mu_6(d_1)+\mu_6(d_2)\}}{\delta_cw^2},\qquad
 B_{\rm mb}\ge\frac{8T\sigma^2}{\delta_c\nu^2}.
\end{equation}
These choices make the width and sample budget explicit for every fixed
member.
\end{definition}

\begin{lemma}[A joint high-probability core event]\label[lemma]{ctx:lem:joint-concentration}
For the budgets in \cref{ctx:def:finite-adam-budgets}, with probability at least $1-\delta_c$, both trained and frozen branches have core discrepancy at most $\tau_{\rm core}/2$, hence at most $\kappa/4$, at every deterministic grid time. The event jointly includes the two Gaussian moment events, empirical-reference concentration and fresh-batch concentration; it does not require independence between branches.
\end{lemma}
\begin{proof}
At a deterministic reference time the core is an average of iid row moments with total variance at most $\{\mu_6(d_1)+\mu_6(d_2)\}/{\rctx}$. Chebyshev and the union over $T+1$ times give error at most $w$ with failure at most $\delta_c/8$ per branch. We apply this concentration bound before intersecting with the radius
events, preserving row independence. Given each actual past, a fresh batch has variance at most $\sigma^2/B_{\rm mb}$ while \eqref{ctx:eq:B3event} holds. Chebyshev and the update union bound give batch error at most $\nu$, failing with probability at most $\delta_c/8$ per branch. The branches may share samples; no interbranch independence is needed.

The two moment events cost $\delta_c/2$, the two reference-grid events $\delta_c/4$, and the two batch events $\delta_c/4$. On their intersection, \eqref{ctx:eq:coupling} and \eqref{ctx:eq:steps} give $\Delta\le\tau_{\rm core}/2$, in particular $\Delta\le\kappa/4$, for both branches throughout the grid. Source parameter error is carried into feedback.
\end{proof}

\begin{lemma}[Continuous-input source acquisition]\label[lemma]{ctx:lem:continuous-source-outcome}
On the joint event of \cref{ctx:lem:joint-concentration}, initial source matching is below $0.51$, and source-endpoint matching is at least $0.9$ uniformly over source contexts, frames and queries. The endpoint greedy source-bit error is less than $0.1$.
\end{lemma}
\begin{proof}
Initially $\norm M\le w$, so correct source attention is at most $e^{2w}/2<0.51$. At the source endpoint the reference odds are at least $19$; core error at most $\kappa$ gives actual odds at least $19e^{-2\kappa}>9$. Matching is therefore at least $0.9$ uniformly over unit source frames and queries.

Conditional on a fixed frame/query and trained parameters, each relevant pooled Gaussian channel has correlation at least $9/\sqrt{82}$ with its selected channel. Gaussian planar angles bound the parity-sign error by $2\arctan(1/9)/\pi$. Coefficient error changes this sign only if
\[
 |h_1h_2|/\norm h^2\le\kappa/t_{\min}.
\]
The direction $h/\norm h$ is uniform on the $(p-1)$-sphere; each of its first two coordinate densities is at most $c_p$. A union bound gives probability at most $4c_p\sqrt{\kappa/t_{\min}}\le0.01$. The radial multiplier is positive. The total source error is below $0.1$.
\end{proof}

\begin{lemma}[Continuous-input transfer and entry comparison]\label[lemma]{ctx:lem:continuous-transfer}
On the joint event of \cref{ctx:lem:joint-concentration}, each source-checkpoint target logit has magnitude at most $1/32$. At the feedback endpoint, every held-out signed logit exceeds $\log19$, so sampled success exceeds $19/20$. Source-checkpoint success is at most $1/2+1/128$, and the gain exceeds $2/5$.
\end{lemma}
\begin{proof}
At the source checkpoint the reference target-active coefficients vanish, giving actual target logit magnitude at most $(k+1)\kappa$. At the final endpoint the reference signed logit is at least $\log19+1$. A core attention error $\Delta_M$ changes a softmax weight by at most $\Delta_M/2$ and the pooled scalar by at most $\Delta_M$. Equation \eqref{ctx:eq:zlip} bounds the resulting logit error by $kB_{\rm final}\Delta_M$. Moreover,
\[
 \norm{C_{\rm ref}}_F^2\ge {\nctx}kB_{\rm final}^2/2,\qquad
 kB_{\rm final}\le\sqrt{2k/{\nctx}}B_3\le\sqrt{2k}B_3.
\]
Coefficient error contributes at most $(k+1)\Delta_C$. The choice of $\kappa$ makes the total final error and source-entry magnitude at most $1/32$. Hence final success exceeds $19/20$, entry success is at most $1/2+1/128$, and their difference exceeds $2/5$.
\end{proof}

\begin{corollary}[Continuous-input frozen-attention gap]\label[corollary]{ctx:cor:continuous-frozen}
On the same event, the frozen branch has sampled test success at most $1/2+1/64$ on every held-out input, and its gap below the trained branch exceeds $2/5$.
\end{corollary}
\begin{proof}
For the frozen branch, the reference head uses absent source channels and has zero target logit for every $M$. Its actual test logit, even on unequal-amplitude memories, has magnitude at most $(k+1)\Delta_C\le1/16$. The $1/4$-Lipschitz sigmoid gives sampled success at most $1/2+1/64$, proving the paired gap. This completes the continuous-input part of \cref{ctx:thm:main}.
\end{proof}

\subsection{Finite coordinate alphabets}\label{ctx:sec:alphabet}

We fix all continuous budgets first, then specify a deterministic observation encoding. Write $N_s=B_{\rm mb}T_s$ and $N_f=B_{\rm mb}T_f$. Both branches can share source records, so a single source-data event suffices. Optimizer states remain real-valued.

\begin{definition}[Source pool separation constants]\label[definition]{ctx:def:pool-event-constants}
Set $\delta_v=\delta/8$ and
\begin{equation}\label{ctx:eq:QRconstants}
 s_0=\sqrt{\frac{p-1}{e}}\left(\frac{\delta_v}{4N_s}\right)^{1/(p-1)},\quad
 L_v=2\sqrt{p+\log(2N_s/\delta_v)},\quad m_v=s_0^2/L_v.
\end{equation}
\end{definition}

\begin{lemma}[A uniform lower bound for all source attention pools]\label[lemma]{ctx:lem:uniform-pool}
For the $N_s$ iid two-column Gaussian source value matrices and the constants in \cref{ctx:def:pool-event-constants}, event \eqref{ctx:eq:QRevent} holds with probability at least $1-\delta_v$. On that event every simplex pool has norm at least $m_v/\sqrt2$, and an operator-norm value perturbation at most $m_v/2$ leaves every interpolation pool at norm at least $m_v/(2\sqrt2)$.
\end{lemma}
\begin{proof}
For each $p$-by-two Gaussian value matrix, a chi-square Chernoff bound allocates probability at most $\delta_v/(4N_s)$ to each of two events: its first-column norm is below $s_0$, or the distance of its second column from the first-column span is below $s_0$. The first dimension is $p\ge p-1$, so the latter bound suffices for both. Their union costs $\delta_v/(2N_s)$. The Frobenius upper-tail event above $L_v$ also costs at most $\delta_v/(2N_s)$. The QR determinant identity therefore gives $\sigma_{\min}\ge m_v$. Union over the $N_s$ records yields, with failure at most $\delta_v$,
\begin{equation}\label{ctx:eq:QRevent}
 \norm V_F\le L_v,\qquad\sigma_{\min}(V)\ge m_v,\qquad
 \norm{V\pi}\ge m_v/\sqrt2\quad\text{for every simplex weight }\pi.
\end{equation}
In the first expression, $\norm V_F$ denotes the Frobenius norm of the two-column value matrix. The event controls adaptive weights without selecting or rejecting any record. A value perturbation of operator norm at most $m_v/2$ leaves every interpolation pool at norm at least $m_v/(2\sqrt2)$.
\end{proof}

\begin{definition}[Sign-preserving observation encoding]\label[definition]{ctx:def:observation-quantizer}
Use an odd, sign-preserving saturating quantizer with spacing $\gamma=2^{-q}$. It rounds each nonzero magnitude upward to the next grid point and leaves zero unchanged, with clipping level $\lceil L_{\rm clip}/\gamma\rceil\gamma$. Error within the clipping range is at most $\gamma$. It encodes source queries, keys and values, commutes with $U$, and preserves the parity bit. Embedded zeros remain exact. Rounded frames need not be orthonormal; we couple their paths to the continuous ones.
\end{definition}

\begin{definition}[Finite computation-graph constants]\label[definition]{ctx:def:graph-constants}
Let $P={\rctx}(6{\nctx}+2p+1)$, $d_h={\nctx}+p$, $\epsilon_{\rm dec}=1/100$, and define
\begin{align}
 R_0&=\sqrt{2\log(16P/\delta)},&&\nonumber\\
 R_{\rm bd}&=1+R_0+K_AH,&&\nonumber\\
 L_{\rm clip}&=\max\{L_v,2,\sqrt{2\log(8p/\epsilon_{\rm dec})}\},&&\nonumber\\
 h_0&=\min\{1,m_v/(2\sqrt2)\},&&\nonumber\\
 V_{\rm bd}&=L_{\rm clip}+2,&&\nonumber\\
 F_{\max}&=d_h^2R_{\rm bd}^3V_{\rm bd}^2(1+h_0^{-1})^2,&&\nonumber\\
 B_{\rm prim}&=\exp\!\left(100[1+{\rctx}F_{\max}+{\rctx}R_{\rm bd}^2+d_hV_{\rm bd}^2
                  +h_0^{-2}+Z_{\max}+m_{\rm norm}]\right),&&\nonumber\\
 N_{\rm op}&=100{\rctx}(3{\nctx}+p+m_{\rm norm}+1)^2,&&\nonumber\\
 D&=P+8{\nctx}+2p+20,&&\nonumber\\
 K_{\rm graph}&=(4B_{\rm prim})^{4N_{\rm op}},&&\nonumber\\
 L_{\rm graph}&={\rctx}DK_{\rm graph},&&\nonumber\\
 L_{\rm state}&=(7{\nctx}+3p)(R_{\rm bd}+1)^2.&&
 \label{ctx:eq:graphconstants}
\end{align}
\end{definition}

\begin{lemma}[Finite-graph bounds for encoded inputs]\label[lemma]{ctx:lem:graph-regularity}
Use the budgets in \cref{ctx:def:finite-adam-budgets}, the constants in \cref{ctx:def:graph-constants} and the quantizer of \cref{ctx:def:observation-quantizer}. The Gaussian coordinate event has failure probability at most $\delta/8$ and keeps both continuous and encoded paths within $R_{\rm bd}$. On that event and \eqref{ctx:eq:QRevent}, restrict interpolation pools to norms at least $h_0$. The scalar computation graph has derivatives through order two bounded by $K_{\rm graph}$; $L_{\rm graph}$ bounds normalized row-gradient state/input Lipschitz constants and logit differences in maximum norm, and $L_{\rm state}$ bounds core differences caused by maximum-norm parameter perturbations.
\end{lemma}
\begin{proof}
A Gaussian maximum event, failing with probability at most $\delta/8$, bounds all initialized coordinates by $R_0$. The Adam coordinate bound then keeps both continuous and encoded paths within $R_{\rm bd}$. Scores have magnitude at most $16R_{\rm bd}^2$, since an embedded query/key has only two nonzero coordinates.

The graph may compute the full $2{\nctx}$-by-$2{\nctx}$ core, softmax, pool and head. The operation count covers scalar additions, products, reciprocals, exponentials and logarithms. Unaveraged core/head sums cost at most ${\rctx}R_{\rm bd}^2$ or ${\rctx}F_{\max}$. Evaluate a radial component as $\exp[-\alpha_j\log(u+\zeta_j)]$ and its mixture square root as the exponential of half its logarithm. On the interpolation domain,
\[
 h_0^2\le u\le d_hV_{\rm bd}^2,\qquad
 \frac1{1+d_hV_{\rm bd}^2+Z_{\max}}\le\psi(u)\le\max(1,h_0^{-2}).
\]
Thus no inverse small mixture weight is needed. The bound $B_{\rm prim}$ covers every primitive's value and first two derivatives, including softmax/logistic denominators and the normalizer logarithms. Induction over unary and binary nodes bounds scalar graph derivatives through order two by $K_{\rm graph}$.

Normalized row gradients equal ${\rctx}$ times literal normalized-coordinate derivatives. Hence $L_{\rm graph}$ bounds their state/input Lipschitz constants in maximum norm and also logit differences. Batch averaging preserves this bound. Subtracting quadratic and cubic moment entries and taking Frobenius norms gives the core map's maximum-norm state bound $L_{\rm state}$.
\end{proof}

\begin{lemma}[Lipschitz comparison of actual Adam histories]\label[lemma]{ctx:lem:adam-history-lipschitz}
For two same-length actual Adam gradient histories with the same decay coefficients and offset $\epsilon>0$, their directions satisfy \eqref{ctx:eq:AdamLipschitz}, where $K_A$ is the history constant of \cref{ctx:def:adam-history-constants}.
\end{lemma}
\begin{proof}
For any two actual Adam histories, subtracting moment fractions and applying weighted reverse triangle inequalities yields
\begin{equation}\label{ctx:eq:AdamLipschitz}
 \norm{D_t(g)-D_t(g')}_\infty
 \le\frac{1+K_A}{\epsilon}\max_{s\le t}\norm{g_s-g'_s}_\infty.
\end{equation}
\end{proof}

\begin{lemma}[Parameter coupling before the first action mismatch]\label[lemma]{ctx:lem:encoded-parameter-path}
Under the graph conditions of \cref{ctx:lem:graph-regularity}, couple continuous and encoded runs with the same initial parameters and samples. While their sampled actions agree, the running maximum-norm parameter discrepancy obeys \eqref{ctx:eq:encodedpath}, with $A_0=(1+K_A)/\epsilon_{\min}$. The bound retains accumulated parameter error across the moment reset.
\end{lemma}
\begin{proof}
Let $E_t$ be the running parameter discrepancy while all sampled actions in a paired continuous/encoded history agree. With $A_0=(1+K_A)/\epsilon_{\min}$,
\begin{align}
 E_{t+1}&\le(1+\eta_tA_0L_{\rm graph})E_t
             +\eta_tA_0L_{\rm graph}\gamma,\nonumber\\
 E_{\rm final}&\le\gamma\{e^{A_0L_{\rm graph}H}-1\}.
 \label{ctx:eq:encodedpath}
\end{align}
Both paths reset moments at the switch while carrying accumulated parameter error.
\end{proof}

\begin{definition}[Precision and permitted parameter error]\label[definition]{ctx:def:observation-precision}
Define
\[
 E_\star=\min\left\{1,\frac\kappa{8L_{\rm state}},
                          \frac\delta{32N_fL_{\rm graph}}\right\}.
\]
Choose an integer $q\ge m_{\rm mem}$ such that $\gamma=2^{-q}$ is no larger than each of
\begin{equation}\label{ctx:eq:precision}
 \frac14,\quad\frac{m_v}{2\sqrt{2p}},\quad
 \frac\kappa{100(R_{\rm bd}+1)^2},\quad
 \frac{\pi\epsilon_{\rm dec}^2}{96p^{3/2}L_{\rm clip}},\quad
 \frac{E_\star}{2}e^{-A_0L_{\rm graph}H}.
\end{equation}
\end{definition}

\begin{proposition}[On-policy encoding coupling and endpoint margins]\label[proposition]{ctx:prop:encoded-coupling}
Choose precision as in \cref{ctx:def:observation-precision}. On the Gaussian coordinate and source-pool events, the grid represents all target inputs exactly, source perturbations satisfy the pool interpolation bound, and parameter discrepancy before an action mismatch is at most $E_\star/2$. Across both branches the probability of any action mismatch, conditional on the preceding bounds up to that mismatch, is at most $\delta/128$. Together with the continuous core event, all required events fail with probability at most $65\delta/128<\delta$. On their intersection the combined core discrepancy is at most $5\kappa/16$; the matching and target endpoint conclusions in \cref{ctx:lem:continuous-source-outcome,ctx:lem:continuous-transfer,ctx:cor:continuous-frozen} retain their stated margins. Fresh encoded source-bit accuracy is established separately in \cref{ctx:lem:encoded-source-bit}.
\end{proposition}
\begin{proof}
This is a finite, explicit choice after width and sample counts. Target amplitudes $0,\pm1,\pm\rho_{\rm mem}$ are exact on the grid, source perturbations satisfy the interpolation condition, and parameter discrepancy is at most $E_\star/2$.

Couple actual policy actions by shared uniform random variables and stop at the first differing action. Until that event, target inputs coincide; the sigmoid Lipschitz bound makes its conditional mismatch probability at most $L_{\rm graph}E_\star/8$. Across both branches' $2N_f$ samples, the union costs at most $\delta/128$. On the complementary event all sampled labels/rewards coincide and \eqref{ctx:eq:encodedpath} applies to the actual algorithms. Deterministic path Lipschitzness alone would not justify this on-policy step.

The continuous two-branch proof costs $\delta/4$, the maximum event $\delta/8$, the QR event $\delta/8$, and mismatches at most $\delta/128$. Their sum is $65\delta/128<\delta$. Combined core discrepancy is at most $\kappa/4+\kappa/16=5\kappa/16$. Rounded queries and keys alter a source score by at most $9R_{\rm bd}^2\gamma$, within the remaining matching slack. Target inputs are exact, so the endpoint margins in \cref{ctx:sec:budgets} still apply.
\end{proof}

\begin{lemma}[Quadratic anti-concentration and rounding error]\label[lemma]{ctx:lem:quadratic-encoding}
Condition on trained parameters and fresh rounded source keys/query. Let $C$ be the nonzero symmetric part of the $p\times p$ restriction of the head to the source content coordinates, rescaled to Frobenius norm one, and let $h$ be its $p$-dimensional Gaussian content pool. Pooled values with variance $S_\pi\ge1/2$ satisfy
\[
 \Pr(|h^\top Ch-z_0|\le u)\le \frac2{\sqrt\pi}p^{1/4}\sqrt{u/S_\pi}
\]
for every fixed $z_0$ and $u\ge0$. Rounding with spacing $\gamma$ and clipping threshold at least $M_{\rm clip}$ changes the quadratic sign with probability at most
\[
 4p e^{-M_{\rm clip}^2/2}+2\sqrt{2/\pi}\,p^{3/4}\sqrt{\gamma(2M_{\rm clip}+\gamma)}.
\]
\end{lemma}
\begin{proof}
Source-bit accuracy on a fresh encoded record requires an additional sign argument. Condition on the trained head and fresh rounded keys/query; its attention weights are independent of fresh Gaussian values. Normalize a nonzero symmetric quadratic coefficient to Frobenius norm one. Some eigenvalue has magnitude at least $1/\sqrt p$. Conditioning on the other Gaussian coordinates and using the squared-Gaussian density gives, for every fixed center $z_0$,
\[
 \Pr(|h^\top Ch-z_0|\le u)
 \le\frac2{\sqrt\pi}p^{1/4}\sqrt{u/S_\pi},\qquad S_\pi\ge1/2.
\]
An interval of fixed length has maximal squared-Gaussian mass at zero. On all $2p$ magnitudes being at most $M_{\rm clip}$, rounding changes a quadratic score by at most $p\gamma(2M_{\rm clip}+\gamma)$. Thus its sign-change probability is at most
\[
 4p e^{-M_{\rm clip}^2/2}
 +2\sqrt{2/\pi}\,p^{3/4}\sqrt{\gamma(2M_{\rm clip}+\gamma)}.
\]
\end{proof}

\begin{lemma}[Greedy accuracy on a fresh encoded source record]\label[lemma]{ctx:lem:encoded-source-bit}
Under \cref{ctx:prop:encoded-coupling} and precision \cref{ctx:def:observation-precision}, rounding a fresh source record adds at most $\epsilon_{\rm dec}=1/100$ to its Gaussian-proxy greedy source-bit error. The resulting error remains less than $0.1$.
\end{lemma}
\begin{proof}
The reference source-content block has norm $t/\sqrt2\ge t_{\min}/\sqrt2$. The core bounds in \eqref{ctx:eq:steps} and \cref{ctx:prop:encoded-coupling} keep the trained restriction within $\kappa<t_{\min}/\sqrt2$, and restriction and symmetrization cannot increase Frobenius error, so its symmetric part is nonzero and \cref{ctx:lem:quadratic-encoding} applies. Take $M_{\rm clip}=\sqrt{2\log(8p/\epsilon_{\rm dec})}$. Equation \eqref{ctx:eq:precision} makes this at most $\epsilon_{\rm dec}$. A positive radial multiplier preserves the greedy sign, including the zero-pool convention. Adding this error to the Gaussian-proxy source error still gives less than $0.1$. This uniform argument is specific to quadratic heads; unrestricted finite-grid decoders are not covered.
\end{proof}

\begin{proposition}[Observation and computation costs]\label[proposition]{ctx:prop:encoded-resources}
For the quantizer and precision of \cref{ctx:def:observation-quantizer,ctx:def:observation-precision}, a coordinate needs $b_{\rm coord}$ bits as specified below. A literal source record has $8\nctx+2p$ coordinate fields and one predicted bit; sparse serialization has $2p+6$ coordinate fields, a context index and that bit. One run uses $(N_s,N_f,N_f,T_s+T_f)$ source records, generated actions, verifier queries and updates, with no validation or selection queries, $P$ parameters and $m_{\rm norm}$ normalizer components per pool. The encoding preserves the missing-context lower-bound direction of \cref{ctx:prop:missing-context-lower}.
\end{proposition}
\begin{proof}
Each coordinate has at most $2\lceil L_{\rm clip}/\gamma\rceil+1$ symbols, requiring
\[
 b_{\rm coord}=\left\lceil\log_2[2\lceil L_{\rm clip}/\gamma\rceil+1]\right\rceil
\]
bits in a fixed-length encoding. A literal source record contains $8{\nctx}+2p$ coordinate fields and one predicted bit. Public sparse serialization uses $2p+6$ coordinate fields, a context index and the next bit; it does not reveal the selected-slot index. Numerical decoding is part of the interface and is not learned by a token embedding. The trained and frozen neural runs use this same alphabet. Deterministic encoding preserves the lower-bound direction of \cref{thm:context-information}, since it cannot reveal an unvisited orientation. The equality and its attainability proof in \cref{app:context-information} use exact numerical source observations.

For one algorithm run, the resources are $N_s$ source records and predicted bits, $N_f$ generated binary actions, $N_f$ verifier queries, and $T_s+T_f$ updates, with zero validation or selection queries. Coordinate bits are additionally charged by the chosen serialization. The normalizer evaluates $m_{\rm norm}$ components per pool and the network has $P$ parameters. Optimization uses real-valued states; the resource counts additionally specify observation precision and normalizer evaluation. All conclusions of \cref{ctx:thm:main} now follow.
\end{proof}

\begin{proof}[Proof of \cref{ctx:thm:main-formal}]
Fix the task, world, normalizer member, Adam coefficients and $\delta$ as in the theorem. The normalized implementation is \cref{ctx:lem:adam-normalization}. The reference exists uniquely by \cref{ctx:lem:reference-existence}; \cref{ctx:prop:source-reference-acquisition,ctx:lem:source-retained-law} supply its acquired matching and switch law. The target-core and coordinate results in \cref{ctx:lem:feedback-core,ctx:lem:feedback-head-drift,ctx:lem:feedback-attention} imply the invariant region and finite hitting time in \cref{ctx:lem:feedback-barrier,ctx:lem:feedback-hitting}; \cref{ctx:lem:feedback-margin} then supplies the held-out margin. The frozen reference is \cref{ctx:prop:frozen-reference}.

Set $\delta_c=\delta/4$ and choose the deterministic budgets of \cref{ctx:def:finite-adam-budgets}. The row, gradient and history bounds in \cref{ctx:lem:row-envelopes,ctx:lem:source-regularity,ctx:lem:target-regularity,ctx:lem:exponential-radius,ctx:lem:adam-current,ctx:lem:adam-local-error} verify the hypotheses of \cref{ctx:prop:adam-coupling}; \cref{ctx:lem:joint-concentration} realizes them jointly for both branches. Thus \cref{ctx:lem:continuous-source-outcome,ctx:lem:continuous-transfer,ctx:cor:continuous-frozen} give acquisition, transfer and intervention for continuous observations.

Finally choose the encoding and precision of \cref{ctx:def:observation-quantizer,ctx:def:observation-precision}. The uniform-pool and graph bounds in \cref{ctx:lem:uniform-pool,ctx:lem:graph-regularity}, together with the actual-history comparison \cref{ctx:lem:adam-history-lipschitz}, verify the parameter coupling \cref{ctx:lem:encoded-parameter-path}. Proposition~\ref{ctx:prop:encoded-coupling} carries the matching and target margins to the actual encoded, on-policy runs with total failure probability at most $65\delta/128<\delta$. Lemma~\ref{ctx:lem:encoded-source-bit} gives the remaining fresh source-bit accuracy. These jointly prove all three conclusions on an event of probability at least $1-\delta$. Proposition~\ref{ctx:prop:encoded-resources} supplies the observation and computation costs, with real-valued optimizer states. Every budget constant uses the fixed dimensions, Adam coefficients, $Z_{\max}$ and $m_{\rm norm}$ rather than a sampled row maximum; the mixture bounds used above are uniform over the stated members, so the same formulas apply at the fixed dimensions with probability evaluated separately for each fixed member.
\end{proof}

\section{The supplementary SGD path}\label{app:sgd-path}
\subsection{Supplementary SGD learning path}\label{app:sgd-result}
The sequential task and the two architecture members in \eqref{eq:network} also admit a finite stochastic gradient descent (SGD) learning path. This supplementary result uses an unregularized source objective, a predetermined source endpoint and its own horizon and rate choices. Its probability and resource statements apply to this schedule; the Adam schedule is specified separately in \cref{thm:neural}. The following subsections provide the complete source, optimizer-specific coupling and feedback proofs, using the common initialization in \cref{app:shared-foundations}.

Source SGD uses a readout rate and a smaller strictly positive rate for every other block. Feedback uses the actual complete-trajectory terminal REINFORCE score with baseline zero and fresh minibatches. Control-output rows have predictable rate $\eta_t=\eta_0/p_{\max,t}$, where $p_{\max,t}$ is the maximum conditional control probability on one fixed training prompt. It is inspected before the fresh batch and uses no reward or test information. Here $\eta_0$ is the base feedback step and $\rho_R>0$ the relative step multiplier for non-control blocks, which have rate $\eta_t\rho_R$. These rates are distinct from the task binding $\eta$. Batch denominators are unchanged even when all rewards are zero.

Write $w_s,w_f$ for the source-trained and final parameters, $B_{\rm src},B_R$ for source-record and feedback-trajectory batch sizes, and $T_{\rm src},T$ for their update counts. In the bounds below, $h$ is the prescribed per-decision approximation tolerance, $\nu$ the feedback gradient-noise tolerance, and $A$ the feature-norm envelope (6 for the base architecture and 12 for its normalized variant), distinct from the generated control $A$ in the task definition.

\begin{theorem}[Finite source acquisition and feedback improvement]\label[theorem]{thm:neural-sgd}
Fix $q=2^{2r+2}$, $r\ge1$, $0<c\le1/64$ and $0<\delta<1$. For either architecture member, choose the explicit nonempty horizon range \eqref{eq:neural-horizon}, widths, initialization and rates of \cref{app:source-budgets,app:foundations,app:fb-main}. After $T_{\rm src}$ raw next-token SGD updates and $T$ sampled REINFORCE updates of the same parameter vector, for every fixed world, with probability at least $1-\delta$, simultaneously for every test prompt $X$,
\begin{align}
 \min\{E(w_f,X),M(w_f,X),S(w_f,X)\}&\ge7/8,\\
 \max\{E(w_s,X),M(w_s,X),S(w_s,X)\}&\le1/q+nh,\\
 \min_{J\in\{E,M,S\}}\big[J(w_f,X)-J(w_s,X)\big]&\ge3/4,\\
 V(w_f,X)&\le9c/8+4nh.
\end{align}
Here $w_s$ is the predetermined source checkpoint, and $h=\nu/(128n^2A^2)$ with all constants given in the appendices. The budgets are
\begin{equation}
 N_{\rm pre}=NB_{\rm src}T_{\rm src},\quad
 Q=B_RT,\quad N_{\rm generated}=nQ,\quad
 N_{\rm updates}=T_{\rm src}+T.
\end{equation}
There are zero source verifier calls and zero checkpoint/test selection. Rate inspections add $T$ inference-only evaluations and $HT$ prompt-token appearances; padding and full-context costs are listed in \cref{app:fb-resources}. Under real arithmetic, the finite budgets are polynomial in expanded $q,H,c^{-1},\delta^{-1}$ up to logarithms; the explicit exponents are given in the appendices.
\end{theorem}

The \hyperref[sgd:proof:main]{complete proof} closes \cref{app:fb-main}, using the source event in \cref{sgd:prop:source-event} and the final-metric conversion in \cref{sgd:prop:final-metrics}.

\subsection{Finite neural source acquisition}\label{app:source}\label[section]{app:source-budgets}
We give the complete source construction and its finite budgets. The feedback constants in \cref{app:fb-population} and \eqref{app:fb-budgets} are chosen first and depend only on $q$: $K=12$, $\eta_0$, $T$, $s$, $\Lambda$, $\nu$, $B_0$ and $B_{\exp}$. The horizon and source accuracy are chosen next, followed by widths, temperature, derivative envelopes and strictly positive slow rates. This order prevents a width--accuracy circularity. In this and the neural appendices, $N=2H+2$ is both the record length and padded context length; $n=H+2$ is the number of generated decisions.

\begin{lemma}[Permutation family and signed contraction]\label[lemma]{sgd:lem:family}\label[lemma]{app:source-family}
For the sequential family of \cref{thm:neural-sgd}, every $\phi_{(b,u)}$ is a permutation and the uniform control average is $\Pi$. There are $2^{m^2}$ labeled tables, including $2^{m^2-1}$ with $\theta_{0,0}\ne\theta_{0,v}$ for a fixed $v\ne0$, which contain the noncommuting pair $\phi_{(0,0)},\phi_{(0,v)}$. For the Koopman signed operator $M_c$ defined below, $\|M_c\|=\rho_c=2^{-r}$; the noisy source operator has signed norm $a\rho_c$.
\end{lemma}
\begin{proof}
Each $\phi_{(b,u)}$ is a permutation: recover $z=z'+u$ and then $s=s'+b+\theta_{u,z}$. For any fixed initial and desired final state there is exactly one control attaining that final state. Consequently
\begin{equation}
 q^{-1}\sum_{b,u}P_{(b,u)}=\Pi,
\end{equation}
where $\Pi$ is uniform averaging on the state alphabet. The family has $2^{m^2}$ distinct labeled tables and needs no group-closure assumption. The subfamily $\theta_{0,0}\ne\theta_{0,v}$ for fixed $v\ne0$ has $2^{m^2-1}$ elements and contains noncommuting operations: compare $\phi_{(0,0)}$ and $\phi_{(0,v)}$ at public coordinate zero.

For the Koopman convention $P_Uf=f\circ\phi_U$ for a full control $U=(b,u)$, decompose
$f(s,z)=f_+(z)+(-1)^sf_-(z)$. Averaging over $b$ in
$M_c=q^{-1}\sum_{b,u}(-1)^{\ell(u)}P_{(b,u)}$ kills the odd state-bit subspace. On the even subspace it is precisely the signed XOR convolution in \cref{app:coarse}. Hence $\norm{M_c}=2^{-r}=\rho_c$ for every unknown table. With source noise $aP+(1-a)\Pi$, the signed norm is $a\rho_c$. Transposing to forward distributions preserves the norm.
\end{proof}

\begin{lemma}[Invariant all-target source dynamics]\label[lemma]{sgd:lem:source-invariant}\label[lemma]{app:source-reference}
Under the unconditioned source law, the diagonal-lag kernel in \cref{seqadam:app:init-kernel} has $K=12$, $k_{\min}\ge3/(2N)$ and $k_{0H}\ge1/N$. For $H\ge3$, zero-start population CE descent is invariant in the mask/class span plus $\beta_1h_1+\beta_2h_2$, with $h_1,h_2$ defined below. Its semantic coefficients satisfy \eqref{eq:source-beta} on every prefix with the prescribed class/PAD pattern, including incorrect state histories.
\end{lemma}
\begin{proof}
Consider first the unconditioned source distribution, with all controls independent uniform. The diagonal-lag quadratic kernel realized by the Gaussian feature construction in \cref{seqadam:app:init-kernel} is
\begin{equation}
 K(x,x')=k_{\rm base}(x,x')^2+3k_{\rm base}(x,x')+2,
 \quad k_{\rm base}=\sum_{l=0}^{N-1}w_l\mathbf1\{x_l=x'_l\}.
\end{equation}
It has diagonal $K=12$, singleton coefficients at least $3/(2N)$ and the lag-pair coefficient $k_{0H}\ge1/N$. Here $x_l$ denotes the token at lag $l$ from the prediction position, with leading PAD tokens as needed.
Define the two centered state-output functions
\begin{align}
 h_{1,y}(x)&=\mathbf1\{x_0\in\mathcal U,x_H\in\mathcal S\}
       \left(\mathbf1\{y=\phi_{x_0}(x_H)\}-\frac1q\right),\\
 h_{2,y}(x)&=\mathbf1\{x_0\in\mathcal S,x_H\in\mathcal U\}
       \left(\mathbf1\{y=\phi_{x_H}(x_0)\}-\frac1q\right),
\end{align}
and set them to zero on nonstate output rows. Their categorical supports are disjoint. Zero-initialized population CE descent stays in the sum of mask/class logits and $\beta_1h_1+\beta_2h_2$. Write
$b_j=(e^{\beta_j}-1)/(e^{\beta_j}+q-1)$, let $v_i$ be the correct state-class mass at state prediction $i$, and let $v_E$ be the incorrect state-class mass at EOS. The simultaneous updates are exactly
\begin{align}\label{eq:source-beta}
 \beta_1^+-\beta_1&=\frac{\eta k_{0H}}{Nq^2}(a-v_1b_1),\\
 \beta_2^+-\beta_2&=\frac{\eta k_{0H}}{Nq^2}
       \left[\sum_{i=2}^H(a-v_ib_2)-v_Eb_2\right].
\end{align}
In particular EOS contributes a negative term; there is no source grammar mask.

For completeness, the categorical-coordinate argument checks more than pairwise independence. At the first-state query, all input states and controls are jointly independent; centering both arguments retains only the lag pair $(0,H)$. At a repeated-state query, averaging its current control eliminates coordinates omitting that control; a surviving coordinate must pair it with the current state. At EOS, integrating the uniform $S_0$ makes $(U_2,\ldots,U_H,S_1,\ldots,S_H)$ independent of $U_1$. Coordinates omitting both $U_1,S_0$ vanish by averaging $U_1$. Among coordinates containing $U_1$, only its pair with $S_H$ survives: an earlier state leaves a later control to average, and a control leaves uniform $S_0$. Among coordinates containing $S_0$ but not $U_1$, an earlier state leaves a later control, $S_H$ leaves a suffix average followed by $U_1$, and a suffix control leaves another suffix control when $H\ge3$. Thus the only semantic EOS contribution is $(S_H,U_1)$, with factor $q^{-2}$. Mask/class residuals remain in their span because distinct individual record variables are pairwise independent uniform. The resulting identities are functions of categorical prefixes and therefore also hold on incorrect generated state histories with the same class/PAD pattern.
\end{proof}

\begin{lemma}[Source comparator and transition precision]\label[lemma]{sgd:lem:source-comparator}
For the invariant reference of \cref{sgd:lem:source-invariant}, let $0<a<1$ and $0<\epsilon\le1$, as in the prescribed source accuracy. The comparator below has squared norm bounded by \eqref{eq:source-comparator} and CE excess at most $\epsilon$. At $\eta\le1/K$ and $T_{\rm src}\ge D^*/(2\eta\epsilon)$, the final excess satisfies $\mathcal E\le2\epsilon$. Its first and repeated transition errors obey the Pinsker bounds below, with denominators $1$ and $H-1$, respectively.
\end{lemma}
\begin{proof}
Let $h_q(a)$ be the entropy of a point mass mixed with the uniform $q$-state law at weights $a,1-a$. The irreducible unconditioned entropy per optimized token is
$[(H+1)\log q+Hh_q(a)]/N$. It is positive. Define
\begin{equation}
 \beta^*=\log\frac{1+(q-1)a}{1-a},\quad
 M=\beta^*+\log(V/\epsilon),\quad P_l=\mathbf1\{x_l=\mathrm{PAD}\}.
\end{equation}
The class logits
\begin{equation}
 c_S=M(P_0-P_H+P_{2H}),\quad c_U=M(P_H-P_0),\quad
 c_E=M(1-P_{2H}),\quad c_{\rm PAD}=0
\end{equation}
select the correct class at every raw target. Add $\beta^*(h_1+h_2)$. The extra $\beta^*$ in $M$ pays for the active wrong-class state partition at EOS. A squared comparator-norm bound is
\begin{equation}\label{eq:source-comparator}
 D^*\le\frac{(5q+V+1)M^2}{k_{\min}}
       +\frac{2q(q-1)(\beta^*)^2}{k_{0H}},
 \qquad k_{\min}\ge\frac3{2N}.
\end{equation}
The constant part of $c_E$ uses all lag-zero indicators. The comparator's CE excess is at most $\epsilon$. Convex smooth descent at $\eta\le1/K$ and $T_{\rm src}\ge D^*/(2\eta\epsilon)$ gives last-iterate excess $\mathcal E\le2\epsilon$.
For any invariant reference, if $D_j=\mathrm{KL}(T_a\|T_{b_j})$, the exact decomposition is
\begin{equation}
 D_1+(H-1)D_2+\sum_{\rm stages}-\log(\text{correct-class mass})=N\mathcal E.
\end{equation}
Pinsker's inequality gives
$|b_1-a|\le\frac q{q-1}\sqrt{N\mathcal E/2}$ and
$|b_2-a|\le\frac q{q-1}\sqrt{N\mathcal E/[2(H-1)]}$.
The first-state transition has only one observation position per record and requires the stronger accuracy.
\end{proof}

\begin{lemma}[Conditioning error for source gradients]\label[lemma]{sgd:lem:source-conditioning}\label[lemma]{app:source-conditioning}
Assume $H\ge5$ and the source family of \cref{sgd:lem:family}. A bounded observable retaining $r_S$ states and $r_{\rm tag}$ tagged controls has signed expectation at most $q^{r_S}(a\rho_c)^{H-1-r_{\rm tag}}\|F\|_\infty$. For the reference CE gradient this gives coordinate error $\mu_{\rm src}=2q^3(a\rho_c)^{H-4}$, gradient error at most $\mu_{\rm src}V^{3/2}\sqrt K$, and uniform logit error at source time $s_{\rm src}$ at most $s_{\rm src}\mu_{\rm src}V^{3/2}K$.
\end{lemma}
\begin{proof}
For a signed low-coordinate observable $F$, retaining $r_S$ state values and $r_{\rm tag}$ of the tagged controls $U_2,\ldots,U_H$, Markov-product contraction gives
\begin{equation}
 |\E_0[C_{\rm suffix}F]|
 \le q^{r_S}(a\rho_c)^{H-1-r_{\rm tag}}\norm F_\infty.
\end{equation}
Indeed insert coordinate projectors for retained states. A fixed control has transition norm at most one, each averaged tagged control has norm $a\rho_c$, and $U_1$ supplies no signed contraction. Summing retained state values gives the factor $q^{r_S}$. Reference CE-gradient coordinates retain conservatively at most three tagged controls and three states. Their discrepancy is at most
\begin{equation}
 \mu_{\rm src}=2q^3(a\rho_c)^{H-4}.
\end{equation}
The squared sum of categorical feature weights is at most $V^2K$, so the gradient discrepancy is at most $\mu_{\rm src}V^{3/2}\sqrt K$. Nonexpansiveness of the conditional population CE map couples it to the unconditioned reference, giving uniform logit error at most $s_{\rm src}\mu_{\rm src}V^{3/2}K$. Take $H\ge5$ to retain an unused balanced tag also in operation-label terms. The reference risk is unchanged by conditioning, but the conditional Bayes entropy is lower by $\log2/N$. The comparison therefore retains the conditional reference approximation gap of $\log 2/N$.
\end{proof}

\begin{definition}[SGD source accuracy and horizon]\label[definition]{sgd:def:source-budget}
Let $V=2q+2$, $\alpha=-\log\rho_c$ and $0<c\le1/64$. Use $A=6$ for the base architecture and $A=12$ for the normalized member. With the feedback constants chosen first as stated above, put
\begin{align}
 C_\epsilon&=\min\{c^2/1024,\ \nu e^{-B_{\exp}}/(2^{14}A^2)\},&
 C_e&=\frac{\nu}{13824A^2(1+3As)},\\
 C_M&=\log\frac{qV}{cC_\epsilon}+4,& C_B&=\log(q/c)+1,\\
 C_D&=2(5q+V+1)C_M^2+6q(q-1)C_B^2,&
 C_s&=C_D/(2C_\epsilon)+1,\\
 K_{\rm cond}&=2C_sq^3\rho_c^{-4}V^{3/2}K/C_e.
\end{align}
Choose any integer
\begin{equation}\label{eq:neural-horizon}
 H\ge\left\lceil\max\left\{5,(18/\alpha)^2,
                    2\max(0,\log K_{\rm cond})/\alpha\right\}\right\rceil.
\end{equation}
Set $N=2H+2$, $n=H+2$, $a=1-c/H$, and
\begin{align}
 \epsilon_{\rm src}&=C_\epsilon/H^3,\\
 \beta^*&=\log(qH/c-q+1),\\
 M&=\beta^*+\log(V/\epsilon_{\rm src}),\\
 D^*&=(2N/3)(5q+V+1)M^2+2Nq(q-1)(\beta^*)^2,\\
 \eta_{\rm src}&=1/A^2,\\
 T_{\rm src}&=\left\lceil D^*/(2\eta_{\rm src}\epsilon_{\rm src})\right\rceil,\\
 s_{\rm src}&=\eta_{\rm src}T_{\rm src},\\
 h&=\frac{\nu}{128n^2A^2},\\
 e_0&=\frac{h}{48(1+\sqrt2nAs)}.
\end{align}
\end{definition}

\begin{lemma}[Nonempty source horizon and error budgets]\label[lemma]{sgd:lem:source-budget}
The choices in \cref{sgd:def:source-budget} give, for both transition types, $|b_j-a|<c/(8H)$, correct-class leakage at most $he^{-B_{\exp}}/2$, and conditioning logit error at most $e_0$. Moreover $s_{\rm src}\le C_sH^6$ and $e_0\ge C_eH^{-3}$. The source endpoint is chosen before initialization or sampling.
\end{lemma}
\begin{proof}
All checkpoints are fixed before data and initialization. The reference satisfies
$|b_j-a|\le2\sqrt{3C_\epsilon}/H<c/(8H)$ for both $j$.
Correct-class leakage is at most $N\mathcal E\le6C_\epsilon/H^2\le he^{-B_{\exp}}/2$.
Using $\beta^*\le C_BH$, $M\le C_MH$, $D^*\le C_DH^3$ gives $s_{\rm src}\le C_sH^6$; also $e_0\ge C_eH^{-3}$. Thus the conditioning discrepancy is at most $e_0$ provided $K_{\rm cond}H^9e^{-\alpha H}\le1$. Equation~\eqref{eq:neural-horizon} ensures this because $9\log H\le9\sqrt H\le\alpha H/2$ and $\log K_{\rm cond}\le\alpha H/2$.
\end{proof}

\begin{definition}[SGD widths, temperature, and positive block rates]\label[definition]{sgd:def:source-widths}
Let $\delta_j=\delta/20$ for each of seven source estimates, and choose
\begin{align}
 B_{\rm src}&=\left\lceil16s_{\rm src}^2K^2\log(2T_{\rm src}/\delta_j)/e_0^2\right\rceil,\\
 W_H&=\sqrt V+\sqrt{2\log(1/\delta_j)},&
 R_{\rm src}&=W_H+\sqrt2As_{\rm src},\\
 R_{\rm tot}&=R_{\rm src}+\sqrt2nAs,&
 \zeta_{\rm tot}&=\min\left\{1,\frac{e_0}{As_{\rm src}(AR_{\rm src}+\sqrt2)+R_{\rm src}}\right\}.
\end{align}
For the base member choose $\gamma=\min\{1/16,e_0/(25s_{\rm src}),h\}$ and $\zeta_{\rm raw}=\zeta_{\rm tot}$. For the normalized member include $\zeta_{\rm tot}/8$ in that minimum and set $\zeta_{\rm raw}=\zeta_{\rm tot}/(2\sqrt2)$. These choices precede every width.

Put $\xi_{\rm enc}=\gamma/16$, $R=\lceil8N\log(2N/\delta_j)\rceil$, $p=NV$, $k_{\rm aug}=p+1$, $s_{\rm sym}=k_{\rm aug}(k_{\rm aug}+1)/2$, and $D_{\rm feat}=p+s_{\rm sym}+1$. Choose the smallest $d$ divisible by $R$, with head width $d_h=d/R$, satisfying
\begin{align}
 d&\ge2048\xi_{\rm enc}^{-2}[p\log9+\log(8/\delta_j)],\\
 d_h&\ge\max\left\{1,\left\lceil\frac{4096N}{R\xi_{\rm enc}^2}
                       [V\log9+\log(8N/\delta_j)]\right\rceil\right\},\\
 d&\ge4096\gamma^{-2}[D_{\rm feat}\log9+\log(8/\delta_j)],\\
 d&\ge50[\sqrt{VD_{\rm feat}}+\sqrt{2\log(1/\delta_j)}]^2/e_0^2.
\end{align}
With $J=s_{\rm sym}\log9+\log(8/\delta_j)$, $A_0=1+\gamma^{-2}J$, and
$L_0=\log(eA_0k_{\rm aug}/(\delta_j\gamma))$, set
\begin{equation}
 M_{\rm ff}=\left\lceil2^{50}A_0^2(k_{\rm aug}+L_0)^8\right\rceil.
\end{equation}
The parameter count is
$P=2dV+4Rd_hd+3dM_{\rm ff}+2M_{\rm ff}+d+RN$,
plus $d$ for the normalized gain. Define
\begin{align}
 M_{\rm init}&=\max\{1,\sqrt{2P\log(2P/\delta_j)}\},&
 Z_0&=M_{\rm init}+\sqrt R M_{\rm init}^3,\\
 L_{\rm att}&=2\sqrt R M_{\rm init}^3[1+2M_{\rm init}^3(Z_0+1)],\\
 \ell_{\rm att}&=\min\{1/2,\zeta_{\rm raw}/L_{\rm att}\},&
 g_{\rm gap}&=\delta_j/(RN^2),\\
 t_{\rm att}&=\min\left\{\frac{g_{\rm gap}}{4M_{\rm init}^4},
 \frac{g_{\rm gap}}{2\log((N-1)/\ell_{\rm att})}\right\}.
\end{align}
At $M_*=M_{\rm init}+1$, put
\begin{align}
 Z_*&=M_*+\sqrt R M_*^3,\\
 D_z&=1+\sqrt R[3M_*^2+M_*^3N(4M_*^3+1/t_{\rm att})],\\
 D_{\rm raw}&=D_z[1+2M_*^3(Z_*+1)]+3M_*^2(Z_*+1)^2+1.
\end{align}
Use $D=D_{\rm raw}$ for the base member or $D=2\sqrt2D_{\rm raw}+\sqrt{24}$ for the normalized member, and $r_{\rm ball}=\min(1,1/D)$. The source readout rate is $\eta_{\rm src}$ and every other supplied parameter has rate $\eta_{\rm src}\rho_{\rm src}$, with
\begin{equation}\label{eq:source-slow-rate}
\rho_{\rm src}=\min\left\{
1,\frac{r_{\rm ball}}{2\sqrt2DR_{\rm src}s_{\rm src}},
\frac{e_0}{D^2\left[AR_{\rm src}(AR_{\rm src}+\sqrt2)s_{\rm src}^2/\sqrt2
                       +\sqrt2R_{\rm src}^2s_{\rm src}\right]}
\right\}.
\end{equation}
Every rate is strictly positive. There is no momentum state, reinitialization or source-score selection. The feedback rate schedule is separately prescribed in \cref{app:fb-budgets}.
\end{definition}

\paragraph{Checking the shared bounds for SGD.}
\begin{lemma}[Shared-foundation and moving-feature premises for SGD]\label[lemma]{sgd:lem:source-applicability}
For the choices in \cref{sgd:def:source-budget,sgd:def:source-widths}, the shared encoder and coefficient concentration premises hold. The initial coefficient-readout bound is $e_0/5$, the initial soft or normalized feature error is at most $\zeta_{\rm tot}$, and \cref{app:couple-proposition} applies with $W_{\max}=W_H$, $R_W=R_{\rm src}$, $s=s_{\rm src}$ and $r=r_{\rm ball}$. Its logit error is at most $e_0$ and its backbone displacement at most $r_{\rm ball}/2$.
\end{lemma}
\begin{proof}
Here $\xi_{\rm enc}=\gamma/16$, $D_{\rm feat}=p+s_{\rm sym}+1$ and
$\delta_{\rm bias}=\delta_{\rm enc}=\delta_{\rm ffn}=\delta_{\rm norm}=\delta_j$.
The first three width inequalities above are precisely the premises of
\cref{seqadam:app:init-encoder,seqadam:app:init-bilinear}; the displayed FFN width supplies its remaining premise.
The fourth width inequality and $1+\gamma\le2$ give initial coefficient-readout norm
at most $e_0/5$ in \eqref{app:init-small-readout}.
The temperature bounds preserve the bias winner gap and make the soft/hard error at most
$\zeta_{\rm raw}$. For the normalized member,
$\gamma\le\zeta_{\rm tot}/8$ and
$\zeta_{\rm raw}=\zeta_{\rm tot}/(2\sqrt2)$ give
$\sqrt2\zeta_{\rm raw}+4\gamma\le\zeta_{\rm tot}$;
for the base member the raw bound already equals $\zeta_{\rm tot}$.
Apply \cref{app:couple-proposition} with
$W_{\max}=W_H$, $R_W=R_{\rm src}$, $s=s_{\rm src}$,
$r=r_{\rm ball}$, $\eta=\eta_{\rm src}$ and the chosen $A$ ($6$ or $12$).
The second term defining $\rho_{\rm src}$ implies
$\sqrt2\rho_{\rm src}DR_{\rm src}s_{\rm src}\le r_{\rm ball}/2$;
the third bounds \eqref{app:couple-logit-error} by $e_0$.
Thus both the ball and error premises hold for the actual common-batch path.
Conditional encoder and coefficient bounds are integrated before taking the union;
the seven source failure allocations remain $\delta_j=\delta/20$ each.
\end{proof}

\begin{proposition}[Actual SGD source event]\label[proposition]{sgd:prop:source-event}\label[proposition]{app:source-event}
For either architecture, use \cref{sgd:def:source-budget,sgd:def:source-widths}. For each fixed world, with failure probability at most $7\delta/20$, the actual source endpoint has uniform logit error at most $6e_0\le h$ relative to the invariant reference, feature error at most $2e_0/R_{\rm src}\le h/[12(R_{\rm tot}+1)]$, readout norm at most $R_{\rm src}$ and backbone displacement at most $r_{\rm ball}/2$. These bounds hold on all legal-pattern prefixes, including incorrect generated histories. Together with a conditional feedback failure allowance $\delta/2$, the total failure is at most $17\delta/20$.
\end{proposition}
\begin{proof}
The seven estimates proved in \cref{app:shared-foundations,app:sgd-foundations,app:couple-source-details} concern bias coverage/gaps, encoder frame, FFN/output Gram matrix, whole-initialization norms, initial coefficient readout, the full readout norm and fresh complete-record sampling. Their union failure is at most $7\delta/20$. The population reference, conditioned population, sampled ideal coefficients, actual hard features, soft or normalized fixed features, and moving all-parameter network are compared on the same batches. The total uniform source logit error is at most $6e_0\le h$; feature error is at most $2e_0/R_{\rm src}$, readout norm at most $R_{\rm src}$, and backbone displacement at most $r_{\rm ball}/2$.
Since $R_{\rm src}\ge1$ and $R_{\rm tot}+1\le2R_{\rm src}(1+\sqrt2nAs)$,
\begin{equation}
 2e_0/R_{\rm src}\le h/[12(R_{\rm tot}+1)]<h/[4(R_{\rm tot}+1)].
\end{equation}
These estimates provide all entrance conditions used by the finite feedback proof. In particular the checkpoint was produced by the actual source algorithm, and the uniform prefix bound includes incorrect generated histories. Combined with the target failure allowance $\delta/2$, the joint failure is at most $17\delta/20<\delta$ for each fixed unknown world.
\end{proof}

\begin{corollary}[Source-stage metric bounds]\label[corollary]{sgd:cor:source-metrics}
On the event in \cref{sgd:prop:source-event}, the actual ancestral source sampler obeys $\max\{E(w_s,X),M(w_s,X),S(w_s,X)\}\le1/q+nh$ simultaneously for every test prompt $X$.
\end{corollary}
\begin{proof}
The source reference selects its first control uniformly. Averaging the first transition over all controls produces a uniform state, and subsequent noisy permutations preserve uniformity. Class masses depend only on the mask. Thus its endpoint, first-state and task-utility probabilities are at most $1/q$ on every fixed prompt. Sequential coupling gives at most $1/q+nh$ for the actual source sampler. Arbitrary decoders require the separate average-over-binding information bound in \cref{app:fb-information}.
\end{proof}

\begin{proposition}[Source resource accounting]\label[proposition]{sgd:prop:source-resources}
The source schedule in \cref{sgd:def:source-budget,sgd:def:source-widths} uses $B_{\rm src}T_{\rm src}$ records, $NB_{\rm src}T_{\rm src}$ raw next-token targets, $T_{\rm src}$ updates and zero verifier queries. All budgets are finite and polynomial in expanded $q,H,c^{-1},\delta^{-1}$ up to logarithms under real arithmetic, with the padding and context costs below.
\end{proposition}
\begin{proof}
The source uses $T_{\rm src}B_{\rm src}$ records, $NB_{\rm src}T_{\rm src}$ raw next-token targets, $NB_{\rm src}T_{\rm src}$ prepended PADs, zero verifier calls and $T_{\rm src}$ updates. Separate target evaluation processes $N^2$ context-token appearances per record, with $N(N+1)/2$ PAD appearances. Final feedback and rate-inspection accounting appears in \cref{app:fb-resources}. The explicit bounds are finite polynomials in the expanded variables $q,H,c^{-1},\delta^{-1}$ up to logarithms, with the exponents specified in the budget formulas. The task alphabet has finite-precision symbols; the neural optimizer is analyzed in real arithmetic.
\end{proof}

\subsection{SGD readout and moving-feature bounds}\label{app:sgd-foundations}
Use the shared initializer and network in \cref{app:shared-foundations}, including the same Gaussian laws, encoder frame, realized coefficient operator, soft/hard approximation and derivative bounds. The remaining estimates below concern the SGD schedule.
\paragraph{Initial readout events.}
\begin{lemma}[Gaussian readout events]\label[lemma]{sgd:lem:initial-readout}
Conditional on the shared coefficient-operator event and its independence from the Gaussian readout $W_0$, the coefficient-readout bound \eqref{app:init-small-readout} fails with probability at most $\delta_{\rm head}$. The separate full-readout norm bound below has its own failure probability at most $\delta_{\rm head}$.
\end{lemma}
\begin{proof}
The readout \(W_0\) is independent of \(C_{\mathrm{all}}\), with entry
variance \(1/d\). Conditional on the coefficient-operator event,
Gaussian norm concentration gives
\begin{equation}
 \norm{W_0C_{\mathrm{all}}}_F
 \leq\sqrt{\frac{1+\gamma}{d}}
 \left[\sqrt{V D_{\mathrm{feat}}}
             +\sqrt{2\log(1/\delta_{\mathrm{head}})}\right]
 \label{app:init-small-readout}
\end{equation}
except with probability \(\delta_{\mathrm{head}}\).
Increasing \(d\) thus gives a small but nonzero initial coefficient
error, without setting the initial function to zero.
The separate full-readout Gaussian tail gives
\[
 \norm{W_0}_F\leq\sqrt V+
 \sqrt{\frac{2\log(1/\delta_{\mathrm{head}})}d},
\]
with its own failure allocation.
\end{proof}

\begin{proposition}[Deterministic shared-batch coupling]
\label[proposition]{app:couple-proposition}\label[proposition]{app:couple-main}
Write the complete logits as \(Wh_\psi(x)\), where \(\psi\) contains
all trainable blocks except the readout \(W\).
On the Euclidean/Frobenius ball \(\norm{\psi-\psi_0}\leq r\), suppose,
uniformly over allowed prefixes,
\begin{equation}
 \norm{h_\psi(x)}\leq A,\qquad
 \norm{D_\psi h_\psi(x)}_{\mathrm{op}}\leq D.
 \label{app:couple-envelope}
\end{equation}
Let \(\norm{W_0}_F\leq W_{\max}\), \(\eta\leq A^{-2}\),
\(T<\infty\), \(s=\eta T>0\), and
\(R_W=W_{\max}+\sqrt2\,As\).
The actual minibatch cross-entropy algorithm uses rate \(\eta\)
on \(W\) and \(\eta\rho\) on every block in \(\psi\), with \(\rho>0\).
Each complete-record loss is the arithmetic mean over its $N$ prescribed next-token targets, and the minibatch loss averages complete-record losses.
The reference uses the same batch sequence and \(W_0\), holding features
at \(h_{\psi_0}\). Writing the batch loss as \(\mathcal L_t\), the
simultaneous updates are
\begin{align}
 W_{t+1}&=W_t-\eta\nabla_W\mathcal L_t(W_t,\psi_t),\notag\\
 \psi_{t+1}&=\psi_t-\eta\rho\nabla_\psi\mathcal L_t(W_t,\psi_t),
 \notag\\
 V_{t+1}&=V_t-\eta\nabla_W\mathcal L_t(V_t,\psi_0),
 \qquad V_0=W_0.
 \label{app:couple-updates}
\end{align}
The reference is an analysis process, not the actual algorithm.
If
\begin{equation}
 \sqrt2\,\rho D R_Ws\leq r/2,
 \label{app:couple-ball-condition}
\end{equation}
the actual path stays in the radius-\(r\) ball and satisfies
\begin{align}
 &\sup_x\norm{W_T h_{\psi_T}(x)-V_T h_{\psi_0}(x)}_2\notag\\
 &\qquad\leq\rho D^2\left[
 \frac{A R_W(A R_W+\sqrt2)s^2}{\sqrt2}+\sqrt2\,R_W^2s
 \right].
 \label{app:couple-logit-error}
\end{align}
This holds deterministically for every shared batch sequence, without
an additional source-sampling or test-word-distribution premise.
\end{proposition}

\begin{proof}
The softmax residual has norm at most \(\sqrt2\), so
\(\norm{\nabla_W\mathcal L_t}_F\leq\sqrt2\,A\), and both readouts
have norm at most \(R_W\).
The full non-readout gradient satisfies
\(\norm{\nabla_\psi\mathcal L_t}\leq\sqrt2\,D R_W\). Therefore
\begin{equation}
 \norm{\psi_t-\psi_0}\leq\sqrt2\,\rho D R_W\eta t.
 \label{app:couple-backbone-motion}
\end{equation}
Induction with \eqref{app:couple-ball-condition} prevents leaving the
ball. A fixed-feature batch objective is convex and \(A^2\)-smooth in
\(W\), so its step is nonexpansive at the stated rate.
At the same readout,
\begin{equation}
 \norm{\nabla_W\mathcal L_t(W,\psi)
          -\nabla_W\mathcal L_t(W,\psi_0)}_F
 \leq D\norm{\psi-\psi_0}(A R_W+\sqrt2).
 \label{app:couple-gradient-perturbation}
\end{equation}
Add and subtract the two residual--feature products and use softmax
Lipschitzness to obtain this inequality.
Summing the nonexpansive comparison gives
\begin{equation}
 \norm{W_T-V_T}_F\leq
 \frac{\rho D^2R_W(A R_W+\sqrt2)s^2}{\sqrt2}.
 \label{app:couple-head-error}
\end{equation}
The final logit difference is at most
\(A\norm{W_T-V_T}_F+R_WD\norm{\psi_T-\psi_0}\), proving
\eqref{app:couple-logit-error}.
Its feature-motion term contains \(D^2\): the parameter gradient and
the conversion of parameter motion to feature motion each supply one
factor of \(D\).
\end{proof}

\paragraph{Raw-network radius and rates.}
\begin{corollary}[A positive slow rate on the raw-network ball]\label[corollary]{sgd:cor:positive-rate}
Suppose the initial block norms are bounded by $M_{\rm init}$, the initial feature envelope is $A_{\rm init}$, and $D$ bounds the feature derivative on the radius-one block ball. The choices \eqref{app:couple-small-ball} and \eqref{app:couple-positive-rate} satisfy the premises of \cref{app:couple-proposition} and yield logit error at most $\varepsilon_{\rm net}>0$.
\end{corollary}
\begin{proof}
If initial block norms are at most \(M_{\mathrm{init}}\), evaluate
\(D\) at \(M_{\mathrm{init}}+1\).
If the realized initializer satisfies
\(\sup_x\norm{h_{\psi_0}(x)}\leq A_{\mathrm{init}}\), choose
\begin{equation}
 r\leq\min\{1,D^{-1}\},\qquad A=A_{\mathrm{init}}+1.
 \label{app:couple-small-ball}
\end{equation}
The mean-value bound gives \(\norm{h_\psi}\leq A\) throughout this
ball, without using the much looser \(A_{\mathrm{raw}}\) for the
readout rate. Any prescribed \(\varepsilon_{\mathrm{net}}>0\)
is attained by the explicit positive choice
\begin{equation}
 0<\rho\leq\min\left\{
 1,\ \frac{r}{2\sqrt2\,D R_Ws},\
 \frac{\varepsilon_{\mathrm{net}}}{
 D^2[A R_W(A R_W+\sqrt2)s^2/\sqrt2+\sqrt2\,R_W^2s]}
 \right\}.
 \label{app:couple-positive-rate}
\end{equation}
The initialization and source-budget construction specifies every quantity
in this rate choice. The backbone-to-readout rate ratio \(\rho\) may be any positive value
satisfying the displayed bound, while readout learning proceeds on the \(\eta\) time scale.
\end{proof}

\paragraph{Normalized readout and accuracy choices.}
\paragraph{A dimension-independent initial readout bound.}
\begin{lemma}[Width-independent full-readout envelope]\label[lemma]{sgd:lem:full-readout}
The independent Gaussian readout obeys \eqref{app:norm-head-envelope} with probability at least $1-\delta_{\rm headnorm}$. Consequently the deterministic source readout envelope $R_W=W_H+\sqrt2As_{\rm src}$ can be chosen before the widths.
\end{lemma}
\begin{proof}
Use a separate Gaussian norm event for the complete initial readout:
\begin{equation}
 \norm{W_0}_F\leq W_H:=
 \sqrt V+\sqrt{2\log(1/\delta_{\mathrm{headnorm}})}.
 \label{app:norm-head-envelope}
\end{equation}
The failure probability is at most \(\delta_{\mathrm{headnorm}}\),
since the sharper tail is
\(\sqrt V+\sqrt{2\log(1/\delta_{\mathrm{headnorm}})/d}\).
The bound is independent of \(d\).
Set \(R_W=W_H+\sqrt2\,A s_{\mathrm{src}}\).
The deterministic source/update couplings need this full-readout
bound, including all nullspace components, rather than the larger
all-parameter envelope \(M_{\mathrm{init}}\).

Using \(M_{\mathrm{init}}\) in \(R_W\) would introduce a circular
choice: normalization error depends on the coefficient-Gram accuracy,
while \(M_{\mathrm{init}}\) grows with the widths selected to improve
that accuracy. Equation~\eqref{app:norm-head-envelope} avoids this
problem. The raw derivative and attention-temperature bounds still use
\(M_{\mathrm{init}}\).
\end{proof}

\begin{proposition}[Normalized-member precision choices]\label[proposition]{sgd:prop:normalized-precision}
Use the normalized architecture, the accuracy and horizon in \cref{sgd:def:source-budget}, and the source reference of \cref{sgd:lem:source-comparator}. For $A=12$, the accuracy choices \eqref{app:norm-precision}, made with \cref{sgd:lem:full-readout}, bound the initial normalized feature error by $\zeta_{\rm total}$. With the normalized derivative envelope and radius specified below, the source schedule \eqref{app:norm-source-budget} uses the same comparator and satisfies $s_{\rm src}\le C_sH^6$.
\end{proposition}
\begin{proof}
Given source time \(s_{\mathrm{src}}\) and a per-error budget \(e_0\),
set \(A=12\) and
\begin{align}
 \zeta_{\mathrm{total}}
 &=\min\left\{1,\
 \frac{e_0}{A s_{\mathrm{src}}(A R_W+\sqrt2)+R_W}\right\},
 \notag\\
 \gamma&\leq\min\left\{\frac1{16},
 \frac{e_0}{25s_{\mathrm{src}}},\frac{\zeta_{\mathrm{total}}}8
 \right\},\notag\\
 \zeta_{\mathrm{raw}}&\leq\frac{\zeta_{\mathrm{total}}}{2\sqrt2}.
 \label{app:norm-precision}
\end{align}
Equation~\eqref{seqadam:app:norm-initial-error} is then at most
\(\zeta_{\mathrm{total}}\).
These quantities precede \(d,M_{\mathrm{ff}}\), since \(R_W\) uses
\eqref{app:norm-head-envelope}.
The same actual coefficient-operator concentration proof applies to
\(C_{\mathrm{all}}\) at the smaller \(\gamma\).
Choose the temperature from the raw attention bound using
\(\zeta_{\mathrm{raw}}\).
The fixed-feature coupling handles the actual normalized map as a
uniformly bounded perturbation, not as an exactly polynomial map.

Choose \(r_{\mathrm{ball}}=\min\{1,D_{\mathrm{norm}}^{-1}\}\)
and use \(D_{\mathrm{norm}}\) in every source/feedback rate and stopping
inequality.

Take
\begin{equation}
 \eta_{\mathrm{src}}=A^{-2}=\frac1{144},\qquad
 T_{\mathrm{src}}=
 \left\lceil\frac{D^*}{2\eta_{\mathrm{src}}\varepsilon}
 \right\rceil.
 \label{app:norm-source-budget}
\end{equation}
At the prescribed source precision, the explicit construction gives
\(s_{\mathrm{src}}\leq C_sH^6\).
The reference kernel and comparator remain at \(K=12\).
All other source choices use the same formulas with
\(A,D_{\mathrm{norm}},W_H,\zeta_{\mathrm{total}}\) substituted as
indicated.
No normalization second-derivative assertion is required: the finite
feedback proof uses a fixed-backbone readout Hessian and pays the entire
backbone update through a first-derivative trajectory bound.
\end{proof}

\paragraph{Semantic functions and reference geometry.}
\begin{definition}[Source semantic reference and entropy]\label[definition]{sgd:def:semantic-reference}
We use the same $h_1,h_2$, $D^*$, $\mu_{\rm src}$ and $s_{\rm src}$ as the source construction. Here the reference CE excess is denoted $\mathcal E$, while $E(w,X)$ remains the endpoint metric.
For a state-output index \(y\), define
\begin{align}
 h_{1,y}(x)
 &=\mathbf1_{\{x_0\in\mathcal U,\ x_H\in\mathcal S\}}
 \left(\mathbf1_{\{y=\phi_{x_0}(x_H)\}}-\frac1q\right),\notag\\
 h_{2,y}(x)
 &=\mathbf1_{\{x_0\in\mathcal S,\ x_H\in\mathcal U\}}
 \left(\mathbf1_{\{y=\phi_{x_H}(x_0)\}}-\frac1q\right).
 \label{app:couple-semantic-functions}
\end{align}
They are zero for nonstate outputs.
At the first-state prediction, lag zero is \(U_1\) and lag \(H\)
is \(S_0\). At prediction \(i\geq2\), these lags contain \(S_{i-1}\)
and \(U_i\); at the end-of-sequence prediction they contain \(S_H\)
and \(U_1\).
Both functions vanish at the initial choice prediction because lag \(H\)
is padding. The reference is its mask/class component plus
\(\beta_1h_1+\beta_2h_2\).
Its unconditioned Bayes entropy per optimized token is
\begin{equation}
 \frac{(H+1)\log q+Hh_q(a)}N,\qquad N=2H+2,
 \label{app:couple-source-entropy}
\end{equation}
where \(h_q(a)\) is the entropy of a point mass mixed with uniform at
weights \(a\) and \(1-a\).
Meaningful state histories, including inconsistent histories, share
the same class/padding layout. The source coefficient identity controls
their logits as functions, not merely on teacher prefixes.
\end{definition}

\begin{lemma}[Isometric source-reference representations]\label[lemma]{sgd:lem:feature-isometry}
The categorical singleton/pair feature map and tensor map $\Phi$ are isometric on their spans. Comparator norms and zero-start coefficient descent therefore transfer between these representations.
\end{lemma}
\begin{proof}
The categorical singleton/pair map and tensor map \(\Phi\) have the
same kernel. For any finite linear combination of feature vectors,
sending one representation to the other preserves its Gram quadratic
form. The map is consequently well-defined and isometric between their
spans. Project a comparator into the span if needed; this cannot
increase its norm.
This transfers the source zero-start coefficient descent and comparator
bounds without equating arbitrary ambient parameter norms.
\end{proof}

\begin{lemma}[Last-iterate source excess]\label[lemma]{sgd:lem:last-iterate}
In the unconditioned reference, use the comparator of \cref{sgd:lem:source-comparator}, rate $0<\eta\le1/K$, and the predetermined budget $T\ge D^*/(2\eta\varepsilon_{\rm src})$. Then the final CE excess satisfies \eqref{app:couple-reference-risk}.
\end{lemma}
\begin{proof}
The unconditioned reference readout cross-entropy is convex and
\(K\)-smooth, since logits are linear in \(\Phi\) and
\(\norm{\Phi}^2=K\).
For \(\eta\leq1/K\), descent is nonexpansive and decreases risk.
The squared-distance telescope to any comparator gives average excess
over that comparator at most \(D^*/(2\eta T)\).
Monotonicity gives the last-iterate version. The comparator's own
excess above irreducible entropy is at most \(\varepsilon_{\mathrm{src}}\),
so the prescribed source horizon yields
\begin{equation}
 \mathcal E\leq2\varepsilon_{\mathrm{src}}.
 \label{app:couple-reference-risk}
\end{equation}
\end{proof}

\paragraph{Conditioning and complete-record sampling.}
\begin{lemma}[Conditional-population trajectory comparison]\label[lemma]{sgd:lem:conditional-trajectory}
For $H\ge5$, the source choices in \cref{sgd:def:source-budget} give the gradient-coordinate bound \eqref{app:couple-conditioning-coordinate} and conditional versus unconditioned reference logit error \eqref{app:couple-conditioning-logits}.
\end{lemma}
\begin{proof}
For a signed low-coordinate observable, write its Markov transition
product and insert a diagonal coordinate projector for each retained
state. Fixed controls have norm at most one; each averaged tagged control
has norm \(a\rho_c\).
The independent \(U_1\) has norm at most one and supplies no signed
contraction. Summing retained state values contributes the conservative
factor \(q^{r_S}\).
After multiplying a singleton/pair feature, probability and label terms
in a reference categorical gradient coordinate touch at most three
retained controls and three states.
Every normalized coordinate discrepancy is therefore at most
\begin{equation}
 \mu_{\rm src}=2q^3(a\rho_c)^{H-4}.
 \label{app:couple-conditioning-coordinate}
\end{equation}
The squared sum of categorical feature weights is at most \(V^2K\);
there are \(V\) output coordinates.
Thus the Frobenius gradient discrepancy is at most
\(\mu_{\rm src}V^{3/2}\sqrt K\).
Apply nonexpansiveness of the conditional population cross-entropy map
to the unconditioned reference at every step to obtain uniform logit
discrepancy at most
\begin{equation}
 s_{\mathrm{src}}\mu_{\rm src}V^{3/2}K\leq e_0.
 \label{app:couple-conditioning-logits}
\end{equation}
This controls local observables of the reference trajectory, without
asserting a full-history density-ratio bound.
\end{proof}

\begin{lemma}[Fresh complete-record sampling]\label[lemma]{sgd:lem:source-sampling}
At predictable source iterates with independent complete-record minibatches of size $B_{\rm src}$ from \cref{sgd:def:source-widths}, all $T_{\rm src}$ coefficient-gradient errors satisfy \eqref{app:couple-source-noise} except with probability $\delta_j$. On this event, the sampled versus population uniform logit error is at most $e_0$ as in \eqref{app:couple-sampling-logits}.
\end{lemma}
\begin{proof}
A complete-record coefficient gradient has norm at most \(\sqrt{2K}\),
regardless of correlations among its \(N\) target positions.
For each fresh independent record batch, a Hilbert-space bounded-difference
estimate and a union bound over predictable \(T_{\mathrm{src}}\) iterates
give gradient noise at most
\begin{equation}
 4\sqrt{\frac{K\log(2T_{\mathrm{src}}/\delta_j)}
                   {B_{\mathrm{src}}}}.
 \label{app:couple-source-noise}
\end{equation}
The same conditional population map is nonexpansive, so the sampling
logit discrepancy is at most
\begin{equation}
 4s_{\mathrm{src}}K
 \sqrt{\frac{\log(2T_{\mathrm{src}}/\delta_j)}{B_{\mathrm{src}}}}
 \leq e_0.
 \label{app:couple-sampling-logits}
\end{equation}
Independent batches consist of complete records; individual
within-record tokens are not assumed independent.
\end{proof}

\paragraph{Transport to realized and moving features.}
\begin{lemma}[Realized hard-feature coefficient comparison]\label[lemma]{sgd:lem:hard-features}
Let $h_{\rm hard}=C_{\rm all}\Phi$ with $\|C_{\rm all}^{\top}C_{\rm all}-I\|\le\gamma\le1/16$. Couple the realized and ideal readout updates on the same batches, starting from $W_0$ and zero respectively. At $\eta\le[(1+\gamma)K]^{-1}$, the coefficient discrepancy satisfies \eqref{app:couple-hard-coefficient-error}. Under \cref{sgd:def:source-widths}, it is at most $2e_0$.
\end{lemma}
\begin{proof}
For \(h_{\mathrm{hard}}=C_{\mathrm{all}}\Phi\), put
\(G=C_{\mathrm{all}}^\top C_{\mathrm{all}}\).
The same-batch readout step induces
\begin{equation}
 Z=W C_{\mathrm{all}},\qquad
 Z^+=Z-\eta\nabla\mathcal L_{\mathrm{batch}}(Z)\,G.
 \label{app:couple-coefficient-update}
\end{equation}
Ideal coefficient descent uses \(G=I\) and starts at zero.
For \(\eta\leq[(1+\gamma)K]^{-1}\), the first map is nonexpansive in
\begin{equation}
 \norm{D}_{G^{-1}}^2=\operatorname{tr}(DG^{-1}D^\top),
 \label{app:couple-weighted-norm}
\end{equation}
by convex smoothness after the invertible coordinate change.
Each batch gradient has norm at most \(\sqrt{2K}\), so
\begin{align}
 &\sup_x\norm{(Z_T-Z_{\mathrm{ideal},T})\Phi(x)}\notag\\
 &\quad\leq\sqrt K\sqrt{\frac{1+\gamma}{1-\gamma}}
 \left[\norm{W_0C_{\mathrm{all}}}_F+
                 s_{\mathrm{src}}\sqrt{2K}\,\gamma\right]\notag\\
 &\quad\leq5\norm{W_0C_{\mathrm{all}}}_F+
                 25s_{\mathrm{src}}\gamma\leq2e_0.
 \label{app:couple-hard-coefficient-error}
\end{align}
\end{proof}

\begin{lemma}[Fixed and moving feature comparisons]\label[lemma]{sgd:lem:soft-moving-features}
Use the feature envelopes, source schedule and positive slow rate of \cref{sgd:def:source-widths}. The fixed soft or normalized feature comparison has logit error at most $e_0$ in \eqref{app:couple-fixed-feature-error}; the moving-feature comparison adds at most $e_0$. Final feature error is at most $2e_0/R_{\rm src}$ and backbone displacement at most $r_{\rm ball}/2$.
\end{lemma}
\begin{proof}
For two fixed feature maps within uniform distance
\(\zeta_{\mathrm{total}}\), starting from the same full \(W_0\),
the gradients at a fixed readout differ by at most
\(\zeta_{\mathrm{total}}(A R_{\mathrm{src}}+\sqrt2)\).
Their fixed-feature step is nonexpansive for \(\eta\leq A^{-2}\).
Hence the fixed soft/normalized versus hard logit discrepancy is at most
\begin{equation}
 \zeta_{\mathrm{total}}
 [A s_{\mathrm{src}}(A R_{\mathrm{src}}+\sqrt2)+R_{\mathrm{src}}]
 \leq e_0.
 \label{app:couple-fixed-feature-error}
\end{equation}
Proposition~\ref{app:couple-proposition} supplies the final
moving-feature error at most \(e_0\), and backbone displacement
at most \(r_{\mathrm{ball}}/2\).
The error denominator in \eqref{app:couple-positive-rate}
also ensures feature motion at most \(e_0/R_{\mathrm{src}}\).
Initial soft/normalized feature error is at most
\(\zeta_{\mathrm{total}}\leq e_0/R_{\mathrm{src}}\),
so the final feature error is at most \(2e_0/R_{\mathrm{src}}\).
The full \(W_0\) norm, including nullspace components, is retained.
\end{proof}

\paragraph{The joint probability event.}
\begin{lemma}[Joint source and feedback probability]\label[lemma]{sgd:lem:joint-event}\label[lemma]{app:couple-source-details}
Allocate $\delta_j=\delta/20$ to each of the seven source estimates and $\delta/2$ to feedback sampling conditionally on the source event. With the widths and bias gap in \cref{sgd:def:source-widths}, their joint failure probability is at most $17\delta/20$, without assuming independence among the resulting events.
\end{lemma}
\begin{proof}
Combine the seven source-event allocations by a union bound,
without requiring mutual independence after conditioning.
Maxima of iid continuous biases select each lag uniformly.
A binomial Chernoff bound gives coverage failure at most
\begin{equation}
 N\exp\!\left(-\frac R{8N}\right)\leq\delta_j/2.
 \label{app:init-coverage-probability}
\end{equation}
A difference of two \(\mathcal N(0,1)\) biases has density at most
\(1/(2\sqrt\pi)\).
The probability that any of at most \(RN^2/2\) pairwise gaps has
magnitude below \(g\) is therefore at most
\begin{equation}
 \frac{RN^2g}{2}\leq\delta_j/2.
 \label{app:init-gap-probability}
\end{equation}
The encoder/feed-forward concentration proofs condition only on the
independent previously drawn blocks specified above.
Each of the seven source events fails with probability at most
\(\delta_j=\delta/20\).
Conditional on all source information on their joint event, fresh target
sampling fails with probability at most \(\delta/2\).
Thus total failure is at most
\begin{equation}
 7\delta/20+\delta/2=17\delta/20.
 \label{app:couple-joint-probability}
\end{equation}
Target concentration conditions on pre-batch history, never on future
adaptive rates.
\end{proof}

\subsection{Finite neural feedback and information comparisons}
\label{app:fb-main}

This appendix continues the same parameter vector produced by the source
analysis. Throughout, $N=2H+2$ is the source-record and context length,
$n=H+2$ is the number of generated decisions, and $L=H-1$ retains its
meaning as the displayed suffix length. The fixed kernel bound is $K=12$.
We write $A$ for the feature-norm envelope ($A=6$ for the base member and
$A=12$ for the normalized member); an operation selected by the policy is
identified explicitly as a control token below.

\begin{lemma}[Exact reward of the fixed source executor]\label[lemma]{sgd:lem:executor-reward}\label[lemma]{app:fb-reward}
Let $0<c\le1/64$ and $|b_j-(1-c/H)|\le c/(8H)$ for $j=1,2$. For the fixed source reference, conditional on the selected control and correct output classes, the expected rewards are exactly \eqref{app:fb-repaired-reward} and satisfy \eqref{app:fb-reward-bounds}. Its public-coordinate checks pass with probability $D_{\rm exec}$ independently of the selected control.
\end{lemma}
\begin{proof}
Let $0<c\le 1/64$ and suppose the source reference has
\[
 \left|b_j-\left(1-\frac cH\right)\right|
 \le \frac{c}{8H},\qquad j\in\{1,2\}.
\]
Set
\[
 B_{\rm exec}=b_1b_2^{H-1},\qquad
 d_j=b_j+\frac{2(1-b_j)}q,\qquad
 D_{\rm exec}=d_1d_2^{H-1}.
\]
At a valid state step of type $j$, the probabilities of the correct local
successor and its state-bit flip are respectively
\[
 \alpha_j=b_j+\frac{1-b_j}{q},\qquad
 \gamma_j=\frac{1-b_j}{q}.
\]
Their sum is $d_j$ and their difference is $b_j$. Multiplying the two-state
even/odd transition matrices shows that all even and odd state-bit-error
histories have total masses $(D_{\rm exec}+B_{\rm exec})/2$ and
$(D_{\rm exec}-B_{\rm exec})/2$. Thus, conditional on the selected control
and correct output classes, the exact conditional expected rewards are
\begin{equation}
 r_\eta=\frac{D_{\rm exec}+B_{\rm exec}}2,\qquad
 r_{\bar\eta}=\frac{D_{\rm exec}-B_{\rm exec}}2,\qquad
 r_a=0\quad(a\notin\{\eta,\bar\eta\}),
 \label{app:fb-repaired-reward}
\end{equation}
where $\bar\eta$ flips the binary control coordinate of $\eta$ and preserves its
public coordinate. A different public control coordinate cannot reach the
target public endpoint while satisfying the public-coordinate consistency checks. The opposite
state-bit choice can be repaired; its reward is explicitly included in
\eqref{app:fb-repaired-reward}. Conditional on the correct output classes, the public-coordinate checks pass with probability $D_{\rm exec}$, independently of the selected control in this fixed reference.
Product and union bounds give
\begin{equation}
 r_\eta\ge B_{\rm exec}\ge 1-\frac{9c}{8},\qquad
 r_{\bar\eta}\le \gamma_1+(H-1)\gamma_2
       \le\frac{9c}{8q},\qquad
 1-D_{\rm exec}\le\frac{9c}{8}.
 \label{app:fb-reward-bounds}
\end{equation}
The transfer argument below separately pays class leakage, EOS, and
feature/parameter drift. These fixed-executor equalities are not asserted
to hold exactly for the moving network.
\end{proof}

\begin{lemma}[Prompt-uniform choice-feature mean]\label[lemma]{sgd:lem:choice-mean}
For the ideal choice feature of the SGD source reference, $q\ge16$ and $H\ge5$, the training and test means coincide: $\mathbb E_+\Phi=\mathbb E_-\Phi=\mu$, and $\langle\mu,\Phi(x)\rangle=\|\mu\|^2=k_\mu\in[2,K]$ on every valid choice prompt. Also $\|\Phi(x)\|^2=K=12$.
\end{lemma}
\begin{proof}
Write $*=\eta$ and $\bar\eta$ for the two possibly rewarded choices. For
$q\ge16$, the ideal choice feature obeys
\[
 \norm{\Phi(x)}^2=K,\qquad
 \E_+\Phi=\E_-\Phi=\mu,\qquad
 \ip{\mu}{\Phi(x)}=\norm{\mu}^2=k_\mu\in[2,K]
\]
at every valid choice prompt. To obtain these identities, average the
categorical singleton and pair features over the uniform initial state and
the $H-1$ suffix controls. Since $H\ge5$, every coordinate leaves an unused
balanced suffix tag. At degree at most two, all non-PAD coordinates are
independent and uniform within their prescribed classes, so the kernel
mean against any fixed valid choice prompt is constant. The constant
kernel term gives $k_\mu\ge2$. This mean identity concerns choice prompts,
not arbitrary later prefixes.
\end{proof}

\begin{definition}[Reference reward path and predictable rate]\label[definition]{sgd:def:reward-reference}
Starting from equal conditional control logits, define the reference
coefficient increments by
\begin{equation}
 \Delta_{t+1}=\Delta_t+\eta_t g(\Delta_t),\qquad
 g_a=p_a(r_a-J)\mu,\qquad
 J=\sum_a p_ar_a,\qquad \Delta_0=0.
 \label{app:fb-reference-update}
\end{equation}
Here $p=\softmax(\Delta_t\Phi(x))$ on the control class. Use exactly the
same predictable rates as the actual network; do not differentiate the
rate rule. Let
\[
 \eta_t=\frac{\eta_0}{p_{\max,\mathrm{actual},t}},\qquad
 \eta_0=\frac1{64K}.
\]
The maximum is over \emph{conditional} control probabilities on one fixed
canonical training prompt, inspected before the fresh batch at step $t$.
This inspection uses no reward, binding identity, or test information.
\end{definition}

\begin{lemma}[Actual and reference rate comparison]\label[lemma]{sgd:lem:rate-comparison}
For the predictable rate in \cref{sgd:def:reward-reference}, $\eta_t\le\eta_0q$ pathwise. If the actual and reference conditional control logits differ uniformly by at most $1/16$ and the correct reference control is maximal, then \eqref{app:fb-rate-comparison} holds.
\end{lemma}
\begin{proof}
Pathwise, $\eta_t\le\eta_0q$. Within a uniform actual/reference
control-logit discrepancy of at most $1/16$, and provided the correct
reference choice is maximal,
\begin{equation}
 e^{-1/8}\eta_0\le\eta_t p_*^{\rm ref}
       \le e^{1/8}\eta_0.
 \label{app:fb-rate-comparison}
\end{equation}
\end{proof}

\begin{lemma}[Finite reference reward improvement]\label[lemma]{sgd:lem:reward-progress}\label[lemma]{app:fb-population}
Use \cref{sgd:def:reward-reference} with the fixed rewards of \cref{sgd:lem:executor-reward} and feature mean of \cref{sgd:lem:choice-mean}. If the actual/reference discrepancy remains at most $1/16$, the correct control stays maximal and nondecreasing from its uniform start. The predetermined $T$ in \eqref{app:fb-horizon} gives $p_*^T\ge15/16$, and cumulative coefficient displacement is at most $B_0$ in \eqref{app:fb-displacement}.
\end{lemma}
\begin{proof}
Put
\[
 \Delta_{\rm gap}=1-\frac{9c}{8}\left(1+\frac1q\right)
       \ge\frac{31}{32},\qquad
 L_q=\log(15(q-1)).
\]
At every reference step with $p_*\ge1/q$,
\[
 J\ge\frac{r_*}{q}>r_{\bar\eta},\qquad
 z_a^+-z_a=\eta_t k_\mu p_a(r_a-J).
\]
The correct logit strictly increases and every wrong logit decreases.
Consequently each correct-versus-wrong gap increases, and $p_*$ remains
maximal and nondecreasing. Until $p_*\ge15/16$,
$r_*-J\ge\Delta_{\rm gap}/16$, so every such gap increases by at least
$k_\mu\eta_0e^{-1/8}\Delta_{\rm gap}/16$. Therefore the predetermined
number of updates
\begin{equation}
 T=\left\lceil640K L_q\right\rceil
 \label{app:fb-horizon}
\end{equation}
suffices for $p_*^T\ge15/16$ for every $k_\mu\in[2,K]$. Indeed, a sharper
sufficient bound is
\[
 \left\lceil\frac{16e^{1/8}L_q}
 {k_\mu\eta_0\Delta_{\rm gap}}\right\rceil.
\]
The threshold persists after crossing. Only the correct gradient
coefficient is positive, and all coefficients sum to zero. Hence
\[
 \norm{g}_F\le\sqrt{2k_\mu}\,p_*(r_*-J)
       \le\sqrt{2K}\,p_*.
\]
Using \eqref{app:fb-rate-comparison}, a conservative bound on cumulative
reference displacement is
\begin{equation}
 B_0=2\sqrt K\,\eta_0T.
 \label{app:fb-displacement}
\end{equation}
It is $O(\log q)$ because $K=12$ is fixed.
\end{proof}

\begin{lemma}[Local reward stability for the realized rate]\label[lemma]{sgd:lem:reward-stability}\label[lemma]{app:fb-stability}
Under \cref{sgd:lem:executor-reward,sgd:lem:choice-mean,sgd:lem:rate-comparison}, fix the realized predictable $\eta_t$. In the convex tube with control-logit distance at most $1/16$ from the reference, the coefficient reward Hessian obeys $\|\nabla_U^2 F\|\le7Kp_*^{\rm ref}$ and the fixed-rate update has Lipschitz factor less than $9/8$, as in \eqref{app:fb-local-lipschitz}.
\end{lemma}
\begin{proof}
For an arbitrary coefficient array $U$, define
\[
 F(U)=\E_+\sum_a p_a(U\Phi(x))r_a.
\]
The source control-class logits are equal and cancel in this conditional
softmax. With $v_a=p_a(r_a-J)$, the exact logit Hessian is
\[
 \nabla_z^2J=\diag(v)-vp^\top-pv^\top.
\]
Nonnegative rewards imply $\norm{v}_1\le2J$ and
$\norm{\nabla_z^2J}\le6J$. In the convex tube where every control-logit
vector differs from $\Delta_t\Phi$ by at most $1/16$,
\[
 J(U,x)\le p_*(U,x)+r_{\bar\eta}
 \le\left(e^{1/8}+\frac{9c}{8}\right)p_*^{\rm ref},
 \qquad
 \norm{\nabla_U^2F}\le7Kp_*^{\rm ref}.
\]
For the \emph{same fixed} $\eta_t$, the update map is therefore locally
Lipschitz with factor at most
\begin{equation}
 1+7K\eta_t p_*^{\rm ref}
 \le1+\frac{7e^{1/8}}{64}<\frac98.
 \label{app:fb-local-lipschitz}
\end{equation}
No derivative of $p_{\max}$ is taken. The bound allows arbitrary
prompt-dependent coefficient perturbations. The reference path may itself
depend on previous actual batches through the predictable rate sequence;
the displayed inequalities hold pathwise for the realized sequence.
Concentration below conditions only on the pre-batch filtration.
Conditioning on the entire future adaptive sequence would be invalid.
The fixed bounds $K=12$ and $k_\mu\ge2$ keep the horizon $O(\log q)$
and the amplification polynomial in $q$.
\end{proof}

\begin{definition}[Feedback tolerances]\label[definition]{sgd:def:feedback-budget}\label[definition]{app:fb-interface}
Set
\begin{equation}
\begin{aligned}
 s&=T\eta_0q,\qquad \Lambda=(9/8)^T,\qquad
 \nu=\frac1{4096\sqrt K(1+s)\Lambda},\\
 h&=\frac{\nu}{128n^2A^2},\qquad
 B_{\exp}=2\sqrt K(B_0+1)+1.
\end{aligned}
 \label{app:fb-budgets}
\end{equation}
\end{definition}

\begin{corollary}[Source entrance bounds for feedback]\label[corollary]{sgd:cor:feedback-entrance}
The event in \cref{sgd:prop:source-event}, with the accuracy and widths of \cref{sgd:def:source-budget,sgd:def:source-widths}, provides the five entrance bounds below for either architecture. Here $R_{\rm tot}=R_{\rm src}+\sqrt2nAs$ and $s,h,B_{\exp}$ are defined in \cref{sgd:def:feedback-budget}.
\end{corollary}
\begin{proof}
The source analysis, starting from initialization and raw minibatch SGD,
supplies the following actual source event:
\begin{itemize}
 \item An invariant reference $f_{\rm src}$ with
 $|b_j-(1-c/H)|\le c/(8H)$ for $j=1,2$.
 \item Correct-class leakage at every valid source/generation mask at
 most $he^{-B_{\exp}}/2$.
 \item Uniform actual-source/reference logit discrepancy at most $h$
 on all legal-prefix strings, including incorrect state histories.
 \item A representation
 $h_{\rm source}=C\Phi+e_{\rm source}$ with
 $\norm{C^\top C-I}\le\gamma\le h$ and
 $\sup\norm{e_{\rm source}}\le h/[4(R_{\rm tot}+1)]$.
 \item $\norm{W_{\rm source}}\le R_{\rm src}$, a feature/backbone
 derivative envelope $D$ on the remaining radius $r_{\rm ball}/2$, and
 actual feature norm at most $A$ throughout the initialization ball.
\end{itemize}
Here $C$ is the initialized hard-feature analysis operator $C_{\rm all}$,
not the bilinear feed-forward factor with the same architectural name.
It is never supplied to the learner. The deterministic radii satisfy
\[
 R_{\rm tot}=R_{\rm src}+\sqrt2\,nAs.
\]
The source proof derives these bounds from initialization in the order
specified in \cref{app:source-budgets}.
\end{proof}

\begin{definition}[All-parameter feedback schedule]\label[definition]{sgd:def:feedback-schedule}
Actual REINFORCE uses fresh independent training-suffix prompts and
rollouts, the single binary terminal reward, baseline zero, and the complete
generated-token score sum, including EOS. Control-output readout rows
have rate $\eta_t$. Every other readout row and every backbone/gain
parameter have the strictly positive rate $\eta_t\rho_R$. All rows and
blocks update simultaneously. Let
\[
 G_{\rm slow}=\sqrt2\,n(A+DR_{\rm tot})
\]
and prescribe
\begin{equation}
 \rho_R=\min\left\{1,
 \frac{r_{\rm ball}}{2(1+s)G_{\rm slow}},
 \frac{h}{4(1+s)G_{\rm slow}(A+DR_{\rm tot})},
 \frac{h}{4D(R_{\rm tot}+1)(1+s)G_{\rm slow}}\right\}.
 \label{app:fb-slow-rate}
\end{equation}
\end{definition}

\begin{lemma}[Feedback slow-block and projection bounds]\label[lemma]{sgd:lem:feedback-motion}
Starting from \cref{sgd:cor:feedback-entrance}, the actual schedule of \cref{sgd:def:feedback-schedule} stays within the initialization ball. Throughout feedback, $\|W_t\|\le R_{\rm tot}$, slow-block displacement is at most $\rho_RsG_{\rm slow}$, and its direct noncontrol/backbone logit effect is at most $h/4$. With projected fast increment $U_t$ defined below, the complete logit error relative to $f_{\rm src}+\operatorname{control\_rows}(U_t\Phi)$ is at most $3h$ and $\|h_t^\top C-\Phi^\top\|\le6h$.
\end{lemma}
\begin{proof}
The source has already used at most $r_{\rm ball}/2$ of the backbone ball.
A stopped pathwise induction bounds target slow-block displacement by
$\rho_RsG_{\rm slow}$, using at most the remaining half-radius. Its
direct noncontrol-logit and backbone-logit effects total at most $h/4$;
feature displacement is at most $h/[4(R_{\rm tot}+1)]$. Every head update
has norm at most $\eta_t\sqrt2\,nA$, because all block rates are at most
$\eta_t$. Consequently $\norm{W_t}\le R_{\rm tot}$ even without a
favorable reward batch.

Define the projected actual fast-row increment
\[
 U_t=(W_{{\rm control},t}-W_{{\rm control},{\rm source}})C.
\]
On each legal prefix the actual logits differ from
$f_{\rm src}+\operatorname{control\_rows}(U_t\Phi)$ by at most $3h$.
This pays the source discrepancy, the fast increment times
$e_{\rm source}$, and the total slow/backbone effect. Moreover,
\[
 \norm{h_t^\top C-\Phi^\top}
 \le\gamma\sqrt K+2h\le6h.
\]
The projection does not discard head nullspace: its contribution is paid
by the ambient head norm times the feature error.
\end{proof}

\begin{lemma}[Transport of the complete terminal score]\label[lemma]{sgd:lem:score-transport}\label[lemma]{app:fb-transport}
Under \cref{sgd:cor:feedback-entrance,sgd:lem:feedback-motion}, suppose $\|U_t-\Delta_t\|\le1/(128\sqrt K)$. For the complete-trajectory terminal-reward score, conditional on the pre-batch history, the projected expected fast gradient differs from $\nabla F(U_t)$ by at most $32n^2A^2h\le\nu/4$, as in \eqref{app:fb-transport-bound}.
\end{lemma}
\begin{proof}
Conditional on the past, the expected-gradient comparison is
\begin{equation}
 \norm{\E[g_{{\rm actual},{\rm fast}}C\mid\mathrm{past}]
                  -\nabla F(U_t)}
 \le32n^2A^2h\le\frac\nu4.
 \label{app:fb-transport-bound}
\end{equation}
We give the full transport argument. As long as
$\norm{U_t-\Delta_t}\le1/(128\sqrt K)$, put
$B=\sqrt K(B_0+1)$. Every added control logit has magnitude at most $B$.
With source class leakage $\ell_0\le he^{-B_{\exp}}/2$ and the paid
$3h$ logit discrepancy, the actual wrong-class probability satisfies
\begin{align*}
 \ell_{\rm actual}
 &\le\frac{\ell_0}{1-\ell_0}\exp(2B+6h)\\
 &\le\frac{h\exp(-1+6h)}{1-he^{-B_{\exp}}}\le h,
\end{align*}
using $h\le1/12$ and $B_{\exp}=2B+1$. Thus both the odds denominator and
all source/slow logit errors remain in the calculation.

Couple actual generation to the conditional control softmax $U_t\Phi$,
the fixed $b_1,b_2$ conditional state executor, and forced EOS. The
conditional-softmax error is at most $3h$ per decision, and class exits
add at most $h$, so sequential coupling fails with probability at most
$4nh$. The projected complete actual fast score has norm at most $2nA$;
the ideal choice score has norm at most $\sqrt{2K}\le2A$. On a matched
correct-class path the choice-score error is at most
$(6\sqrt2+4\sqrt K)h$. Feature-projection error supplies its first term;
conditional probability and the control-class deficit supply its second.
Every later state/EOS fast score is at most $2Ah$. The expected-gradient
difference is therefore at most
\begin{align*}
 &[8n(n+1)A+6\sqrt2+4\sqrt K+2(n-1)A]h\\
 &\hspace{2cm}\le20An^2h\le32n^2A^2h,
\end{align*}
which proves \eqref{app:fb-transport-bound}. Matched complete trajectories
receive the same actual terminal bit, including repaired hidden-state
histories. On unmatched trajectories the score envelopes pay the reward
change. No hidden correct-trace constraint is inserted. In particular,
the nullspace term is explicitly bounded by
\[
 \norm{\Delta W}\le2R_{\rm tot},\qquad
 \norm{\Delta W e_{\rm source}}\le h/2;
\]
source and slow effects then total less than $3h$. An ambient head norm
is never identified with its projection.
\end{proof}

\begin{lemma}[Uniform fresh-feedback batch concentration]\label[lemma]{sgd:lem:feedback-sampling}
Under the same source event and the stopped feedback path in \cref{sgd:lem:score-transport}, use independent fresh batches of size \eqref{app:fb-batch-size}. With probability at least $1-\delta_R$, the projected fast-score batch noise is at most $\nu$ simultaneously for all $T$ predictable iterates. The original batch denominator is retained, including for zero-reward batches.
\end{lemma}
\begin{proof}
For a target failure allowance $\delta_R$, take
\begin{equation}
 B_R=\left\lceil\frac{256n^2A^2\log(2T/\delta_R)}{\nu^2}\right\rceil.
 \label{app:fb-batch-size}
\end{equation}
Conditional on the pre-batch filtration, the independent reward-weighted
scores have norm at most $M_{\rm score}=2nA$, and their batch mean obeys
\[
 \E\norm{\mathrm{batch\ mean}-\E[\mathrm{score}]}
 \le\frac{M_{\rm score}}{\sqrt{B_R}}.
\]
Bounded differences gives deviation at most
\[
 \frac{M_{\rm score}[1+\sqrt{2\log(T/\delta_R)}]}{\sqrt{B_R}}
 <\nu
\]
with failure probability at most $\delta_R/T$. A union bound gives
simultaneous batch noise at most $\nu$ with probability at least
$1-\delta_R$. Batches with zero rewards are included. The denominator
is the original $B_R$, without reward filtering. No independence between
an iterate or its rate and earlier batches is asserted.
\end{proof}

\begin{lemma}[The actual feedback path stays in the reference tube]\label[lemma]{sgd:lem:stopped-comparison}\label[lemma]{app:fb-stopped}
On the source event and the batch event of \cref{sgd:lem:feedback-sampling}, use \cref{sgd:def:feedback-budget,sgd:def:feedback-schedule}. The actual and reference projected increments satisfy \eqref{app:fb-stopped-error} for every $0\le t\le T$. Neither the coefficient nor parameter stopping boundary is reached, and the rate comparison holds throughout.
\end{lemma}
\begin{proof}
Compare actual and population updates using the same $\eta_t$ and put
$E_t=\norm{U_t-\Delta_t}$. Inside the stopped tube,
\eqref{app:fb-local-lipschitz}, \eqref{app:fb-transport-bound}, and the
batch event imply conservatively
\begin{equation}
 E_{t+1}\le\frac98E_t+8\eta_t\nu,\qquad E_0=0,
 \qquad E_t\le8s\Lambda\nu<\frac1{512\sqrt K}.
 \label{app:fb-stopped-error}
\end{equation}
This is stricter than the stopping threshold $1/(128\sqrt K)$. Together
with the $3h$ source/soft/slow discrepancy, it makes the actual/reference
conditional choice-logit difference less than $1/16$, closing the rate
comparison used in \eqref{app:fb-rate-comparison} and
\eqref{app:fb-local-lipschitz}. The same induction closes coefficient
and parameter exits. Rates are predictable before each batch.
Since $K=12$ and $T=O(\log q)$, $\Lambda$ is a fixed power of $q$ up to
absolute constants.
\end{proof}

\begin{proposition}[Final SGD metrics and source-to-reward gain]\label[proposition]{sgd:prop:final-metrics}
On the events in \cref{sgd:prop:source-event,sgd:lem:feedback-sampling}, at the predetermined source and reward endpoints, simultaneously on every test prompt, $\min\{E(w_f,X),M(w_f,X),S(w_f,X)\}>7/8$, $\max\{E(w_s,X),M(w_s,X),S(w_s,X)\}\le1/q+nh$, each final-minus-source gain exceeds $3/4$, and $V(w_f,X)\le9c/8+4nh$.
\end{proposition}
\begin{proof}
At the predetermined final iterate the actual conditional correct-choice
probability differs by at most $\sqrt K E_T+3h$ from a reference
probability of at least $15/16$, on every valid choice prompt, including
held-out suffixes. In particular the uniform choice-logit discrepancy is
less than $1/256$, so the comparison choice probability is at least
$239/256$. The fixed executor has expected reward conditional on the correct initial control at least
$1-9c/8\ge503/512$ and correct first-state probability at least the same
lower bound. Final trajectory coupling costs at most
$4nh<1/1024$. Thus task utility and first-state memory success exceed
\begin{equation}
 \frac{239}{256}\frac{503}{512}-\frac1{1024}>\frac78.
 \label{app:fb-final-utility}
\end{equation}
Endpoint success dominates task utility.

In the source ancestral sampler, uniform choice makes the first state
uniform. Every subsequent transition kernel preserves uniformity, while
class masses depend only on position. Therefore the reference has
$E_{\rm source}\le1/q$, $S_{\rm source}\le E_{\rm source}$, and
$M_{\rm source}\le1/q$. Actual source coupling bounds all three by
$1/q+nh$. Since $q\ge16$ and $nh<1/4096$, each final-minus-source gain
exceeds $3/4$. The fixed executor satisfies the public-coordinate checks with probability
$D_{\rm exec}$ independently of its choice, so actual grammar-or-public-coordinate violation risk is at most
\[
 1-D_{\rm exec}+4nh\le\frac{9c}{8}+4nh.
\]
\end{proof}

\paragraph{Arbitrary source-only decoders.}
\begin{proposition}[Source-only decoding with an independent binding]\label[proposition]{sgd:prop:source-only}\label[proposition]{app:fb-information}
If the binding $\eta$ is uniform over the $q$ controls and independent of source observations, initialization, learner randomness and evaluation prompts, every source-only decoder has binding-averaged endpoint success at most $1/q$.
\end{proposition}
\begin{proof}
Let $\eta$ be uniform over the $q$ controls, independently of all source
observations, initialization, learner randomness, and evaluation prompts.
For each fixed prompt the map
\[
 \eta\longmapsto
 \phi_{U_H}\cdots\phi_{U_2}\phi_\eta(S_0)
\]
is bijective. Condition on every source-only observation and algorithmic
random choice. Any endpoint equals exactly one of these $q$ targets, so
averaged endpoint success is at most $1/q$. Malformed output only reduces
success. This includes arbitrary source-only computation, decoding, and
calibration. It is an average-over-binding or minimax statement, not a
pointwise assertion for every binding.
\end{proof}

\paragraph{Indistinguishable direct-feedback worlds.}
\begin{proposition}[Paired direct-feedback indistinguishability]\label[proposition]{sgd:prop:direct-feedback}
With no source observations and only the prescribed terminal training bit, the involution below pairs worlds whose adaptive training transcripts have identical laws for any number of queries. On a common tag-one test prompt, their average endpoint success is at most $1/2$.
\end{proposition}
\begin{proof}
Let $\ell(u)$ be the public tag bit. Pair the worlds by
\[
 \theta'_{u,z}=\theta_{u,z}\mathbin\oplus\ell(u),\qquad
 \eta'=(b_\eta\mathbin\oplus\ell(u_\eta),u_\eta).
\]
The adjusted first transition is identical:
$\phi'_{\eta'}=\phi_\eta$. The central state-bit flip $B$ commutes with
every transition map. Each suffix operation adds $B^{\ell(u)}$, so paired
target endpoints differ by $B$ raised to the suffix tag parity. They
agree on every tag-zero training suffix and are opposite on every
tag-one test suffix. Grammar and the public $z$-consistency checks are identical
for every candidate trace in both worlds, including hidden errors that
repair later. Hence their entire terminal training oracles are pointwise
identical.

Without source observations, couple the learner's random tape. Every
adaptive training transcript remains identical, even with unlimited
training queries. On a common test prompt the identical output law has
summed endpoint success at most one against the two opposite targets;
average paired success is at most $1/2$. The pairing is an involution and
preserves uniform $\eta$. It does not imply a first-state memory lower
bound, because the paired first targets agree. Hidden-mechanism
intermediate verification would invalidate this pointwise oracle
equivalence and is outside the information model.
\end{proof}

\paragraph{Singleton-query resource bound.}
\begin{proposition}[Singleton-query lower bound]\label[proposition]{sgd:prop:singleton}
Let the binding be uniform and independent of learner randomness and test prompts. Even if $\theta$ is given, a learner with $Q$ adaptive pre-test candidate-verification queries and no binding-dependent prompt information obeys \eqref{app:fb-singleton}. Hence binding-averaged endpoint success at least $7/8$ requires $Q\ge7q/8-1$.
\end{proposition}
\begin{proof}
Even grant $\theta$ for free. Endpoint bijectivity implies that any fixed
prompt/candidate terminal query accepts for at most one $\eta$;
malformed or public-inconsistent candidates accept for none. Fix learner
randomness and follow the all-zero transcript. At most $Q$ adaptive
queries name at most $Q$ distinct bindings. Every world producing a
positive response belongs to these at most $Q$ values, irrespective of
subsequent behavior. All remaining bindings yield the same all-zero
transcript. On each independent test prompt a final output succeeds for
at most one remaining value. Therefore
\begin{equation}
 \E_\eta[\text{test endpoint success}]
 \le\min\left\{1,\frac{Q+1}{q}\right\}.
 \label{app:fb-singleton}
\end{equation}
In particular, averaged success at least $7/8$ requires
$Q\ge7q/8-1$. Each sampled candidate verification counts separately.
Expected-reward oracles, intermediate labels, binding-dependent prompt
side information, and test-time verification are excluded. This is a
membership-query law, not a logarithmic entropy argument; the neural
upper need not match this bound.
\end{proof}

\paragraph{Packed-mechanism information bound.}
The separate proof in \cref{app:packing} establishes, for
$m\ge512$, $H\ge5$, independent reset source records, and at most $Q$
adaptive binary pre-test replies,
\[
 \sup_{\theta,\eta}\E[\text{endpoint error}]
 \ge\frac7{32}\left[1-\frac{128(n_{\rm rec}H+Q+1)}{m^2}\right]_+.
\]
It permits a fixed known binding and strengthens query permissions to
adaptive prompt choice. The quantity $n_{\rm rec}H+Q$ counts information
bits, not heterogeneous monetary or computation costs. It supplements
the independent-binding singleton bound and the source-free paired-world
ambiguity; its proof is not duplicated here.

\begin{proposition}[Opaque transition-table recovery]\label[proposition]{sgd:prop:table-recovery}\label[proposition]{app:fb-table}
Under the sequential raw source law, let $0<\delta_{\rm tab}<1$ and use the complete-record budget \eqref{app:fb-table-budget}. The rowwise empirical-mode estimator recovers all $q^2$ opaque successors with probability at least $1-\delta_{\rm tab}$, using at most $6M_0+2H+2$ raw tokens and no downstream goal labels.
\end{proposition}
\begin{proof}
Under the raw source law, execution controls $U_1,\ldots,U_{H-1}$ are
jointly iid uniform: $U_1$ is independent, and $U_2,\ldots,U_{H-1}$ form
a strict subset of the tag-conditioned suffix. Retain odd execution
transitions through $H-1$ and average the intervening even controls.
Their inputs become independent uniform states because
$q^{-1}\sum_{U\in\mathcal U} T_U=\Pi$. Thus a complete record supplies
$\lfloor H/2\rfloor$ iid triples in this marginal retained experiment.
Take
\begin{equation}
 M_0=\left\lceil8q^2a^{-2}\log\frac{2q^3}{\delta_{\rm tab}}\right\rceil,
 \qquad
 n_{\rm rec}=\left\lceil\frac{M_0}{\lfloor H/2\rfloor}\right\rceil.
 \label{app:fb-table-budget}
\end{equation}
The charged raw-token count satisfies
\[
 Nn_{\rm rec}=(2H+2)n_{\rm rec}\le6M_0+2H+2.
\]
The empirical mode for every opaque state/control row recovers all $q^2$
successors with probability at least $1-\delta_{\rm tab}$. This estimator
requires neither a coordinate chart nor downstream goal labels.

For completeness, let $M=n_{\rm rec}\lfloor H/2\rfloor$ be the number
of retained iid triples and put
$t=\log(2q^3/\delta_{\rm tab})$. Each row count has mean
$\mu=M/q^2\ge8a^{-2}t$. Binomial lower-tail concentration bounds the
probability of fewer than $\mu/2$ observations by $e^{-\mu/8}$.
Conditional on $n$ observations of a row, true-minus-competitor indicator
differences lie in $[-1,1]$ and have mean $a$. Their centered log moment
generating function is at most $\lambda^2/2$, so an incorrect empirical
mode, including a tie, has probability at most $e^{-na^2/2}$. Union
bounding the $q^2$ row counts and $q^2(q-1)$ comparisons gives failure
probability at most $\delta_{\rm tab}$. This proves recovery with full-
record rounding. It does not assume independence of all $H$ transitions
within a record.
\end{proof}

\begin{corollary}[Binding selection from a recovered table]\label[corollary]{sgd:cor:table-binding}
On the recovery event of \cref{sgd:prop:table-recovery}, at most $q-1$ exact-path terminal verifications on fresh uniform training prompts identify the binding and give perfect test utility and public-coordinate compliance. Additional costs, storage and computation are as stated below.
\end{corollary}
\begin{proof}
On recovery, enumerate candidate bindings on separately drawn fresh
uniform training prompts. Generate each candidate's exact path using the
recovered table. The binary terminal reward equals one exactly when that
candidate is $\eta$: suffix invertibility and sharp transitivity of the
first transition prove the equivalence. Test at most $q-1$ candidates,
stopping at success or inferring the last binding if all reject. No
canonical-prompt query selection is needed.

The additional costs are at most $q-1$ verifier calls,
$(H+2)(q-1)$ generated symbols, and $H$ table lookups per candidate or
deployment. There are zero gradient updates and no test-time oracle.
Final storage is $q^2$ successor indices and one binding index; training
may use $q^3$ counts. Deployment has perfect utility and public-coordinate
compliance on the recovery event. This comparator uses the same observations with a table-based computation, separating information requirements from the neural training route. The $2H+2$ granularity term and every
unused raw token remain charged.
\end{proof}

\begin{proposition}[Greedy source execution with an external binding controller]\label[proposition]{sgd:prop:greedy-controller}\label[proposition]{app:fb-controller}
On \cref{sgd:prop:source-event}, externally choosing any valid first control and greedily decoding thereafter yields its exact trajectory and EOS. At most $q-1$ fresh-prompt verifications identify the binding, after which every test prompt has the correct first state and endpoint. This holds with probability at least $1-7\delta/20$ for each fixed world and uses the additional resources enumerated below.
\end{proposition}
\begin{proof}
On the proved source event, the conditional transition parameters obey $b_j\ge1-9c/(8H)$, the wrong-class mass is at most $\ell_0=he^{-B_{\exp}}/2\le h/2$, and actual source logits differ from the reference by at most $h$ uniformly over legal-pattern prefixes. An individual softmax probability changes by at most $h$ under this logit discrepancy: integrate its coordinate gradient, whose $\ell_1$ norm is $2p(1-p)\le1/2$. The correct state token therefore has actual probability at least
\[
 (1-\ell_0)b_j-h\ge1-\frac{9c}{8H}-\frac{3h}{2}>\frac12.
\]
The correct EOS probability is at least $1-\ell_0-h>1/2$. The strict inequalities follow from $c\le1/64$, $H\ge5$ and $nh<1/4096$. After externally selecting any valid first control, greedy decoding consequently follows its unique exact state trajectory and then EOS, by induction over legal-pattern prefixes. This uses the uniform source function bound, not an assumption about teacher-forced visits.

That candidate satisfies the public-coordinate consistency checks. Suffix invertibility and sharp transitivity make its endpoint equal the target if and only if its first control equals $\eta$. At most $q-1$ fresh-prompt verifications therefore identify the binding, with the last candidate inferred if all others reject. The same corrected choice gives the target first state and exact execution on every test prompt. Success is deterministic on the source event, whose probability is at least $1-7\delta/20$ for each fixed world.

The comparator pays the original source tokens and updates. Additional costs are at most $q-1$ verifier queries, $n(q-1)$ generated symbols including the externally supplied control, $H(q-1)$ prompt tokens and $(H+1)(q-1)$ model forward evaluations, each retaining $N$ context tokens. There are no additional optimizer updates. The binding index and enumeration counter are additional persistent controller state; model parameters, inference context and forward workspace are separate. Each verification still performs $H$ trusted transitions and public checks. Deployment uses $H+1$ forwards and $n$ output symbols, with no verification. This comparator combines the frozen source model with an external binding search.
\end{proof}

\begin{proposition}[Complete SGD resource accounting]\label[proposition]{sgd:prop:feedback-resources}\label[proposition]{app:fb-resources}
For \cref{sgd:def:source-budget,sgd:def:source-widths,sgd:def:feedback-budget,sgd:def:feedback-schedule}, there are $Q=B_RT$ feedback queries, $nQ$ generated decisions, $T_{\rm src}+T$ optimizer updates, and $T$ additional inference-only rate inspections. Padding, context, prompt, parameter-storage and verifier costs are the counts below; no test score chooses a checkpoint.
\end{proposition}
\begin{proof}
No model updates occur after the fixed source and target budgets, and
no test score selects a checkpoint. A canonical rate-inspection prompt
may use a fixed state and $H-1$ copies of the publicly tag-positive zero
control. Before each fresh rollout batch, read its conditional control
probabilities without generation or verification. This charges $T$
inference-only evaluations, $HT$ prompt-token appearances, $NT$ processed
context-token appearances, and $nT$ PAD appearances, separately from
zero data-selection queries and from optimizer updates.

Each source record prepends $N$ PADs and trains on $N$ non-PAD targets.
If predictions are separately evaluated, it processes $N^2$ context-token
occurrences with $N(N+1)/2$ PAD appearances. Each feedback rollout starts
its $N$-token context with $n=N-H$ PADs and $H$ real prompt tokens. Its
$n$ decisions process $nN$ context-token occurrences, including
$n(n+1)/2$ PAD appearances; no real token is truncated. Across $Q=B_RT$
rollouts, charge
\[
\begin{aligned}
 &nQ\ \text{prepended PADs},\qquad
 nNQ\ \text{context-token occurrences},\\
 &\frac{n(n+1)Q}{2}\ \text{attended PAD occurrences},\\
 &HQ\ \text{prompt tokens},\qquad nQ\ \text{generated decisions}.
\end{aligned}
\]
Canonical inspections are additional. Optimization uses real arithmetic and stores $P$ real parameter scalars. Each verifier
call evaluates $H$ unknown transition-table entries for the target
endpoint and $H$ public-coordinate comparisons. The verifier reveals the prescribed bit; its private $\theta$ table remains part of the verifier's resources.
\end{proof}

\begin{proposition}[Pathwise equivalence to on-policy reward-weighted likelihood]\label[proposition]{sgd:prop:likelihood-equivalence}
For one fixed complete sampled batch and current model, baseline-zero terminal REINFORCE equals the negative gradient of \eqref{app:fb-weighted-likelihood}, holding sampled tokens and rewards fixed. With the same simultaneous block rates, inspection rule and fresh coupled batches, the two procedures have identical parameter paths; independently sampled fresh batches give the same law.
\end{proposition}
\begin{proof}
For a fixed complete sampled batch and current model, baseline-zero
REINFORCE ascent is exactly the negative gradient of
\begin{equation}
 -\frac1{B_R}\sum_b R_b\sum_{j=1}^{n}
 \log\pi_w(Y_{b,j}\mid\mathrm{prefix}_{b,j}).
 \label{app:fb-weighted-likelihood}
\end{equation}
Hold the realized rewards and trajectory tokens fixed for this single
differentiation. With identical simultaneous block rates and the same
canonical inspection rule, reward-weighted likelihood makes exactly the
same update. Fresh coupled batches preserve pathwise equality;
independently generated fresh batches give equality in law. This does
not identify arbitrary expert SFT, filtered-denominator training, or
repeated off-policy replay with this algorithm. Source-only continuation
remains independent of the binding, so the arbitrary-decoder average
bound still applies. Test-time best-of-$N$ scored by the task verifier adds post-test-prompt oracle
calls outside the pre-test lower bounds and cannot be counted as free.
\end{proof}

\phantomsection\label{sgd:proof:main}
\begin{proof}[Proof of \cref{thm:neural-sgd}]
Choose the feedback constants in \cref{sgd:def:reward-reference,sgd:def:feedback-budget} first, and then the horizon, accuracy, widths, temperature and positive rates in \cref{sgd:def:source-budget,sgd:def:source-widths}. The nonempty horizon and its error inequalities follow from \cref{sgd:lem:source-budget}; \cref{sgd:lem:source-applicability} verifies the shared-foundation and moving-feature premises for each architecture.

By \cref{sgd:prop:source-event}, the prescribed actual source endpoint satisfies all five feedback entrance conditions in \cref{sgd:cor:feedback-entrance} except on an event of probability at most $7\delta/20$. Set $\delta_R=\delta/2$ in \eqref{app:fb-batch-size}. Conditional on the source event, \cref{sgd:lem:feedback-sampling} gives the fresh-batch event with failure probability at most $\delta/2$. Their union failure is $17\delta/20<\delta$ by \cref{sgd:lem:joint-event}; no independence between these events is required.

On their intersection, \cref{sgd:lem:feedback-motion,sgd:lem:score-transport,sgd:lem:stopped-comparison} compare the same-parameter actual training path with the fixed-executor reference, and \cref{sgd:lem:reward-progress} supplies its final correct-choice probability. \Cref{sgd:prop:final-metrics} then gives all four displayed metric conclusions, simultaneously for every test prompt. In particular it combines $239/256$ correct-choice probability, executor success at least $503/512$, and trajectory-coupling loss below $1/1024$ to exceed $7/8$; the source bound $1/q+nh$, with $q\ge16$ and $nh<1/4096$, gives the $3/4$ gain. The same coupling yields the violation bound $9c/8+4nh$.

Finally \cref{sgd:prop:source-resources,sgd:prop:feedback-resources} give $NB_{\rm src}T_{\rm src}$ source targets, $B_RT$ reward queries, $nB_RT$ generated decisions and $T_{\rm src}+T$ updates, plus the stated rate-inspection and context costs. The formulas in the cited definitions establish the finite polynomial resource dependence under real arithmetic. The endpoints are predetermined, so neither checkpoint nor test selection adds a query.
\end{proof}

\end{document}